\PassOptionsToPackage{table}{xcolor}
\RequirePackage{fix-cm}
\documentclass{article}
\usepackage{arxiv}
\usepackage[T1]{fontenc}
\usepackage[utf8]{inputenc}
\usepackage{graphicx}
\usepackage{natbib}
\setcitestyle{authoryear,round,citesep={;},aysep={,},yysep={;}}

\usepackage{amsmath,amsfonts,bm}

\def\eqref#1{equation~\ref{#1}}

\def\floor#1{\lfloor #1 \rfloor}
\def\1{\bm{1}}

\DeclareMathAlphabet{\mathsfit}{\encodingdefault}{\sfdefault}{m}{sl}
\SetMathAlphabet{\mathsfit}{bold}{\encodingdefault}{\sfdefault}{bx}{n}

\newcommand{\E}{\mathbb{E}}

\usepackage{amsmath,amssymb,amsthm,mathtools}
\usepackage{mathpartir}   
\usepackage{placeins}     
\usepackage{booktabs}
\usepackage{algorithm}
\usepackage{algpseudocode}
\usepackage{longtable}
\usepackage{multirow}
\usepackage[table]{xcolor}

\usepackage{enumitem}
\usepackage{array,booktabs,longtable}
\usepackage{tikz}
\usetikzlibrary{positioning,shapes.geometric}
\usepackage{hyperref}
\usepackage{cleveref}
\crefname{rulex}{Rule}{Rules}
\crefname{assumptionx}{Assumption}{Assumptions}
\crefname{notex}{Note}{Notes}
\crefname{objx}{Analysis}{Analyses}
\crefname{invx}{Invariant}{Invariants}
\crefname{optionx}{Option}{Options}
\crefname{remark}{Remark}{Remarks}
\usetikzlibrary{positioning,fit,backgrounds,arrows.meta}
\usepackage{comment}

\theoremstyle{plain}
\newtheorem{theorem}{Theorem}[section]
\newtheorem{lemma}[theorem]{Lemma}
\newtheorem{proposition}[theorem]{Proposition}
\newtheorem{corollary}[theorem]{Corollary}

\theoremstyle{definition}
\newtheorem{definition}[theorem]{Definition}

\theoremstyle{definition}
\newtheorem{rulex}{Rule}

\newtheorem{assumptionx}{Assumption}

\theoremstyle{remark}
\newtheorem{notex}{Note}

\newtheorem{objx}{Analysis}

\newtheorem{invx}{Invariant}

\theoremstyle{remark}
\newtheorem{remark}[theorem]{Remark}

\newtheorem{optionx}{Option}

\newenvironment{commentary}[1][]{
  \par\smallskip
  \begin{list}{}{\setlength{\leftmargin}{1.5em}
    \setlength{\rightmargin}{1.5em}
    \setlength{\topsep}{0pt}
    \setlength{\parsep}{0.25\baselineskip}}
  \item\relax\small
  \textit{Commentary
    \if\relax\detokenize{#1}\relax\ (non-normative).
    \else\ (non-normative; #1).\fi}\hspace{0.5em}\ignorespaces
}{\end{list}\smallskip\par}

\newcommand{\Mem}{\mathcal{M}}
\newcommand{\Vals}{\mathcal{V}}
\newcommand{\SensT}{\mathsf{Tools}_{\mathrm{sens}}}
\newcommand{\DataT}{\mathsf{Tools}_{\mathrm{data}}}
\newcommand{\U}{\mathsf{U}}
\newcommand{\T}{\mathsf{T}}
\newcommand{\C}{\mathsf{C}}
\newcommand{\CC}{\mathsf{CC}}
\newcommand{\PC}{\mathsf{PC}}

\newcommand{\lvl}{\mathrm{lvl}}
\newcommand{\eqU}{\approx_{\U}}
\newcommand{\sinks}{\mathsf{Sink}}
\newcommand{\DT}{\mathsf{DT}}
\newcommand{\conf}{\mathsf{confirm}}
\newcommand{\scopef}{\mathsf{scope}}
\newcommand{\obs}{\mathsf{obs}_P}
\newcommand{\Ex}{\mathbb{E}}
\newcommand{\Prob}{\mathbb{P}}
\newcommand{\bigO}{\mathcal{O}}
\newcommand{\wmax}{w_{\max}}
\newcommand{\wmin}{w_{\min}}
\newcommand{\wseat}{w_{\mathrm{seat}}}
\newcommand{\Wadv}{W_{\mathrm{adv}}}
\newcommand{\fs}{f_{\mathrm{s}}}
\newcommand{\qpre}{q_{\mathrm{pre}}}   
\newcommand{\qpan}{q_{\mathrm{pan}}}   
\newcommand{\dpriv}{d_{\mathrm{priv}}}
\newcommand{\RE}{R_{\mathcal{E}}}

\newcommand{\Twin}{T_{\mathrm{win}}}     
\newcommand{\Dpool}{D_{\mathrm{pool}}}   
\newcommand{\Ncap}{N^{+}}                
\newcommand{\szero}{s_{0}}               
\newcommand{\Tpan}{T_{\mathrm{pan}}}
\newcommand{\Texp}{T_{\mathrm{exp}}}
\newcommand{\gbar}{\bar{\gamma}}
\newcommand{\Bfl}{B_{\mathrm{fl}}}
\newcommand{\Emax}{E_{\max}}

\newcommand{\Eslot}{E_{\star}}

\newcommand{\Elease}{E_{\mathrm{lease}}} 
\newcommand{\Erv}{E_{\mathrm{rv}}}       

\newcommand{\cacc}{c_{\mathrm{acc}}}
\newcommand{\cceil}{c_{\mathrm{ceil}}}
\newcommand{\ctool}{c_{\mathrm{tool}}}
\newcommand{\cbatch}{c_{\mathrm{batch}}}
\newcommand{\ccand}{c_{\mathrm{cand}}}
\newcommand{\cpr}{c_{\mathrm{p}}}
\newcommand{\rsub}{\rho_{\mathrm{sub}}}
\newcommand{\rseat}{\rho_{\mathrm{seat}}}
\newcommand{\rmax}{r_{\max}}
\newcommand{\thacc}{\theta_{\mathrm{acc}}}

\newcommand{\leanfootnote}{\footnote{The Lean~4 artifact is available upon request and will be released with the code when the paper is published. Its scope and correspondence are documented in Appendix~\ref{app:mechanization} and Supplement~\ref{S-app:mechanization}.}}

\usepackage{url}

\providecommand{\cv}[2]{\cellcolor[RGB]{#1}#2}

\usepackage{microtype}
\usepackage{etoc}
\usepackage{bookmark}
\definecolor{permitlink}{RGB}{30,62,100}
\hypersetup{colorlinks=true,linkcolor=permitlink,citecolor=permitlink,
  urlcolor=permitlink,bookmarksopen=true,bookmarksopenlevel=1,bookmarksdepth=3,
  pdftitle={PerMIT: Securing Persistent Memory for LLM Agents via Machine-Checked Integrity Theorems and Scalable Community Trust},
  pdfauthor={Tanqiu Jiang, Min Bai, Sadat Shahriar, Wanyu Du, Sandesh Swamy},
  pdfsubject={Persistent memory security for LLM agents; complete formal development},
  pdfkeywords={LLM agents, persistent memory, integrity, formal verification, shared memory}}
\renewcommand{\shorttitle}{\small PerMIT: Securing Persistent Memory for LLM Agents}
\renewcommand{\headeright}{Preprint}
\renewcommand{\undertitle}{Preprint}
\makeatletter
\DeclareFontShape{T1}{ptm}{m}{scit}{<->ssub * ptm/m/sc}{}
\renewcommand{\normalsize}{\@setfontsize\normalsize{11}{13}%
  \abovedisplayskip 8pt plus 2pt minus 4pt
  \abovedisplayshortskip 0pt plus 3pt
  \belowdisplayskip \abovedisplayskip
  \belowdisplayshortskip 5pt plus 2pt minus 2pt}
\renewcommand{\small}{\@setfontsize\small{10}{12}}
\renewcommand{\footnotesize}{\@setfontsize\footnotesize{9}{11}}
\normalsize
\renewcommand{\@maketitle}{%
  \begin{center}
    {\LARGE\bfseries\@title\par}
    \vspace{1.1em}
    {\normalsize\@author\par}
    \vspace{0.65em}
    {\small\scshape Preprint\par}
  \end{center}\vspace{0.5em}}
\makeatother
\newcommand{\startsupplement}{%
  \setcounter{section}{0}%
  \renewcommand{\thesection}{S\arabic{section}}%
  \renewcommand{\theHsection}{supp.\arabic{section}}%
  \renewcommand{\theHsubsection}{\theHsection.\arabic{subsection}}%
  \renewcommand{\theHsubsubsection}{\theHsubsection.\arabic{subsubsection}}%
  \setcounter{figure}{0}\setcounter{table}{0}\setcounter{equation}{0}%
  \setcounter{algorithm}{0}%
  \renewcommand{\thefigure}{S\arabic{figure}}%
  \renewcommand{\thetable}{S\arabic{table}}%
  \renewcommand{\theequation}{S\arabic{equation}}%
  \renewcommand{\thealgorithm}{S\arabic{algorithm}}%
  \renewcommand{\theHfigure}{supp.\arabic{figure}}%
  \renewcommand{\theHtable}{supp.\arabic{table}}%
  \renewcommand{\theHequation}{supp.\arabic{equation}}%
  \renewcommand{\theHalgorithm}{supp.\arabic{algorithm}}%
}

\title{PerMIT: Securing Persistent Memory for LLM Agents\\via Machine-Checked Integrity Theorems\\and Scalable Community Trust}
\author{Tanqiu Jiang\textsuperscript{1}\thanks{Work was done during an internship at AWS AI Labs.}\quad
Min Bai\textsuperscript{2}\quad
Sadat Shahriar\textsuperscript{2}\quad
Wanyu Du\textsuperscript{2}\quad
Sandesh Swamy\textsuperscript{2}\\[0.6em]
\textsuperscript{1}Stony Brook University\\
\textsuperscript{2}AWS AI Labs}
\date{}
\begin{document}
\maketitle
\setcounter{footnote}{0}
\thispagestyle{plain}
\begin{abstract}
Persistent memory is a largely undefended surface for injection and poisoning, and the same risk limits memory sharing among agents. We present PerMIT, a memory architecture with provable integrity guarantees under an explicit trust assumption: only user-typed inputs and values confirmed at a gate are trusted. PerMIT mediates the release and use of untrusted environment-derived values through a finite memory interface, while preserving agent-driven multi-turn planning and tool choice. Over this finite interface, we prove two integrity theorems that are also machine-checked in Lean 4, which together ensure that environment-derived memory reaches a sensitive action only if confirmed at the gate and influences the agent's behavior only through what the gate displays. Every attack in our threat model thereby reduces to a single auditable decision, which a human, an LLM verifier, or both can make in practice. On Trojan Hippo, PerMIT records zero observed attack success with self-verification and lightweight verifiers. Because each confirmation is a logged verdict on exact content, endorsements accumulate across users, and sufficiently endorsed items become community-confirmed with no further verification required. Under explicit assumptions, we prove and corroborate in simulation a co-scaling property: as the community grows, the risks of promoting poison and of wrongly evicting benign knowledge fall, while per-user verification effort does not rise. On $\tau^2$-bench, communities of up to 90 autonomous agents use PerMIT to build shared knowledge bases with no direct memory injection promoted, and the resulting knowledge significantly improves new agents on held-out tasks, suggesting a path toward agents that safely self-improve at test time.

\end{abstract}
\subsection*{Guide to this document}
This preprint brings the main paper, its empirical and explanatory appendices,
and the complete formal development together in one document. All references
below are clickable; the PDF bookmarks provide a detailed navigation tree.
\begin{description}[leftmargin=1.2em,style=nextline,itemsep=3pt,topsep=4pt]
\item[Main paper and experiments.]
Part~\ref{part:main} presents the framework and results. Part~\ref{part:appendices} retains the additional
experiments, practical interface, concurrent-work comparison, proof sketches,
and mechanization scope (Appendices~\ref{sec:additional}--\ref{app:mechanization}).
\item[Complete proofs.]
Part~\ref{part:supplement} contains Supplement~\ref{S-app:guide}--\ref{S-app:mechanization}. Start with its
\hyperref[S-app:guide]{proof roadmap} and
\hyperref[S-app:notation]{notation table}; then follow the
\hyperref[S-app:single]{single-user guarantees},
\hyperref[S-app:fedproofs]{federated guarantees}, and
\hyperref[S-app:budget]{verification-budget analysis}.
\item[Assumptions, limitations, and mechanization.]
Consult the \hyperref[S-tab:objindex]{attack and limitation index},
\hyperref[S-app:ledgertable]{assumptions ledger}, and
\hyperref[S-app:mechanization]{Lean correspondence and modeling deltas}.
The Lean artifact is available upon request.
\end{description}

\clearpage
\pdfbookmark[0]{Contents}{contents}
\etocsetnexttocdepth{section}
\begingroup
\setlength{\parskip}{0pt}
\small
\tableofcontents
\endgroup
\clearpage
\part{Main Paper}\label{part:main}
\begingroup
\setlength{\parskip}{4pt plus 1pt minus 1pt}
\fontsize{11}{12.8}\selectfont
\section{Introduction}
\label{sec:intro}

By carrying preferences, decisions, and hard-won context across sessions, persistent memory transforms a stateless LLM into an agent that becomes more capable and personalized with continued use.
Although agents benefit from reading and writing persistent memory, this capability also creates an attack surface that the underlying models are poorly equipped to defend.
Poisoned information can readily enter persistent memory when an agent browses webpages or invokes tools containing planted injections \citep{das2026trojan}.
Such attacks can turn a one-shot prompt injection \citep{debenedetti2024agentdojo} into a persistent compromise, and a single compromise could easily propagate to a broader community if that memory were ever shared.
Existing memory-sharing frameworks such as the Open Knowledge Format \citep{mcveety2026okf} (OKF) define formal protocols for writing and exchanging knowledge among agents but do not address memory attacks.
This difficult-to-control risk has limited the practical deployment of memory sharing between agents.

Existing defenses validate retrieved content, govern access, or enforce information-flow policies during agent execution.
A-MemGuard empirically detects context-dependent poisoning through consensus among memory-induced reasoning paths \citep{wei2025memguard}, whereas Collaborative Memory enforces dynamic read/write policies over provenance-tagged private and shared memory fragments \citep{rezazadeh2025collaborative}.
However, neither approach provides formal security guarantees against malicious memories.
Beyond memory-specific defenses, CaMeL addresses prompt injection during agent execution through isolated planning with capability-based policies and user confirmation \citep{debenedetti2025defeating}; FIDES formalizes IFC guarantees and supports selective hiding and constrained inspection of tool results \citep{costa2025securing}. PerMIT builds on this isolation-and-mediation lineage to secure the persistent-memory lifecycle.

PerMIT enforces its trust boundary at the memory interface. The harness captures environment-derived content, including tool outputs, as untrusted and mediates its release and use. The protected agent chooses its next operation and can re-plan on endorsed feedback, so its internal reasoning is not inspected or taint-tracked. The noninterference proofs quantify over arbitrary agent strategies using this finite interface \citep{goguen1982security}. However, preventing all influence from untrusted memory would also prevent legitimate tasks whose next steps depend on information recalled from that memory.
To mitigate the safety--utility tradeoff, PerMIT adopts endorsement \citep{myers2006enforcing}, a formal relaxation that promotes an untrusted value to trusted status only after the user confirms the exact value displayed at the gate, so untrusted memory can be selectively endorsed and used rather than barred.
Using this finite mediation surface and endorsement mechanism, we prove Action Integrity as an exact authorization property and Reasoning Integrity as noninterference modulo the display transcript \citep{sabelfeld2004model,askarov2007gradual}, and machine-check their single-user core in Lean~4\leanfootnote.
These machine-checked guarantees hold over unbounded interface-conformant executions (\S\ref{sec:bg-defences}), separating enforced mediation from the correctness of individual endorsement decisions.

Endorsement preserves benign utility by allowing external values to become trusted through an interface with a finite set of operations. The gate displays the exact value, original user instruction, and source, making the decision auditable. On Trojan Hippo, PerMIT combines live agent rollouts with targeted action-boundary probes; self-verification and lightweight verifiers record zero observed ASR while retaining useful memory capabilities (Table~\ref{tab:util}).

Although security concerns have limited the deployment of shared memory, Karpathy's LLM-Wiki \citep{karpathy2026llmwiki} illustrates its potential: thousands of users adopted his shared coding-agent setup to improve their agents' performance.
LLM-Wiki's success was an exception because the author is highly reputable, and the shared knowledge had accumulated thousands of public endorsements through GitHub stars.
However, scaling shared memory to diverse communities requires accepting contributions from ordinary users and agents, allowing adversaries to pose as legitimate contributors and spread harmful content. To solve this dilemma, PerMIT learns from the success of LLM-Wiki and allows each shared knowledge item to accumulate trust independently, paralleling security mechanisms in distributed systems that aggregate endorsements over shared objects \citep{walsh2006credence}.
Each logged confirmation is bound to the exact content, allowing verdicts from different users and agents to accumulate as item-level trust.
Each item requires confirmation until its accumulated trust reaches the ``Community Confirmed'' threshold.
As the population grows, confirmations and endorsements allow more items to reach this status and become directly available to new users without further verification, so safety and convenience improve together.
Under explicit assumptions, we prove that poison-promotion and wrongful-eviction risks decrease as the community grows, while the average per-user confirmation burden does not grow (Supplementary Theorem~\ref{S-thm:coscale}).
Our simulation spanning five billion promotion attempts also corroborates this co-scaling behavior. 
PerMIT thus supplies the security assurance missing from the Open Knowledge Format \citep{mcveety2026okf} while remaining compatible with its knowledge-sharing protocol.

Finally, we close the loop by using PerMIT to construct community-confirmed shared knowledge bases on $\tau^2$-bench \citep{barres2025tau2} with up to 90 fully autonomous agents and no human involvement.
None of the tested direct injection payloads reaches the Community-Confirmed tier, and a zero-state agent (i.e. a fresh agent with no prior context or memory) connected to the resulting knowledge base significantly outperforms an agent without shared knowledge on held-out, experience-dependent telecom tasks (\S\ref{sec:autoconstruct}).

In summary, this paper makes three contributions:

\begin{enumerate}

    \item \textbf{Securing agent memory with machine-checked integrity.}
    
    PerMIT mediates a finite memory interface and binds each endorsement to an exact displayed value and its intended use. Under explicit trust assumptions, PerMIT provides two integrity guarantees, machine-checked in Lean~4 for the single-user core, and an auditable confirmation gate for human or automated verification.

    \item \textbf{Shared memory in which safety and convenience scale together rather than trade off.}
    Because each gate decision is a logged verdict on exact content, verdicts accumulate into community-confirmed knowledge, which is the layer that existing memory-sharing frameworks such as OKF lack. Under explicit assumptions, we
    prove and corroborate in simulation that poison-promotion and wrongful-eviction risks fall as the community grows while per-user verification effort does not rise.

    \item \textbf{Safe, autonomous self-construction of shared knowledge.}
    Our experiments show that autonomous agent communities can safely construct useful knowledge bases with PerMIT. By enabling agents to generate, verify, and certify knowledge in complex environments, PerMIT provides a concrete path toward test-time learning and self-evolving agents.

\end{enumerate}

\section{Background and Related Work}
\label{sec:background}

\subsection{Persistent Memory as an Attack Surface}
\label{sec:bg-attacks}

Persistent-memory architectures manage context across sessions through tiered storage, consolidation, and revision \citep{packer2023memgpt,mem0,xu2025mem}. Because stored observations later return to the agent's context, transient exposure can become durable influence. Attacks enter through direct poisoning of a memory or knowledge base \citep{chen2024agentpoison}, query-only interaction \citep{dong2025memory}, or indirect ingestion of retrieved artifacts and webpages \citep{srivastava2025memorygraft,zou2026poison,yang2026zombie}; shared state can also contaminate users without an adversary \citep{yang2026no}. The defining risk is temporal: Trojan Hippo plants a dormant payload through one untrusted tool call and activates it only in a later session \citep{das2026trojan}. This persistence makes turn-local prompt-injection defences insufficient and motivates guarantees over the memory lifecycle \citep{lin2026survey}.

\subsection{Defenses and Formal Coverage}
\label{sec:bg-defences}

Existing defenses draw their boundaries at different layers. Content-based methods detect anomalous memories or task drift \citep{wei2025memguard,abdelnabi2024get}. Isolation and privilege-control methods constrain how untrusted information can influence tool interactions \citep{wen2026agentsys,shi2025progent}. TMA-NM \citep{louck2026securing} binds memory authority to origin and studies consequential-action safety using TLA+ model checking and an inductive-invariant argument. Willison's Dual LLM pattern introduced privileged planning, quarantined processing, and opaque references \citep{willison2023dual}. CaMeL adds capability-based enforcement and user confirmation, while leaving formal verification as future work \citep[\S\S2,5,10]{debenedetti2025defeating}. AgentDyn reports zero utility for its CaMeL configurations under automated dynamic-task evaluation \citep{li2026agentdyn}. FIDES provides a formal IFC model with integrity noninterference, selective variable hiding, and schema-constrained quarantined inspection \citep{costa2025securing}. Its type-aware policies can permit bounded influence; PerMIT centers exact-value, tool-scoped endorsement and the complete display transcript. Whereas CaMeL and FIDES address prompt injection in agent execution more generally, our contribution centers the persistent-memory lifecycle, with Action Integrity and Reasoning Integrity guarantees for arbitrarily long interactions through the memory interface under explicit trust assumptions. We machine-check the integrity theorems in Lean~4 and aggregate item-level endorsements in the federated tier. Planning and tool choice remain agent-driven, including replanning on endorsed feedback.

\subsection{Shared Memory and Evaluation}
\label{sec:bg-shared}

Shared-memory work has primarily addressed exchange, access, and portability. OKF standardizes knowledge exchange \citep{mcveety2026okf}, Collaborative Memory enforces dynamic policies over private and shared fragments \citep{rezazadeh2025collaborative}, and Portable Agent Memory proposes cryptographically verified transfer and injection-resistant rehydration \citep{ravindran2026portable}. These mechanisms complement PerMIT but do not establish recipient-side action and reasoning integrity for adopted content.

General agent-security benchmarks emphasize prompt injection within a task episode \citep{debenedetti2024agentdojo,li2026agentdyn,weng2026argus}; Trojan Hippo most directly matches our cross-session threat model \citep{das2026trojan}. We therefore use Trojan Hippo to evaluate persistent-memory security and $\tau^2$-bench \citep{barres2025tau2} to evaluate the end-to-end utility of autonomously constructed shared knowledge.

\section{The PerMIT Framework}
\label{sec:method}

\subsection{Threat Model}
\label{sec:method-threat}

PerMIT assumes an adversary capable of controlling or manipulating content obtained from the external environment, including tool outputs, web content, external files and codebases. Poisoned payloads may enter persistent memory when the agent invokes tools, browses the web, or inspects external files or code. PerMIT establishes a provenance-based trust boundary: only values typed in directly by the user or exact values explicitly confirmed by the user through the confirmation interface are treated as trusted. All other environment-derived content remains untrusted and cannot reach the protected LLM unless it first passes through this confirmation gate and gets verified to be trusted.

\subsection{The Single-User Tier: a Finite Memory Interface}
\label{sec:method-single}

The P-LLM remains an agent that chooses its next operation and revises its plan using endorsed feedback. Prompts describe the interface; the harness enforces its observation and flow rules without inspecting the agent's internal reasoning. Gate decisions control which values are endorsed; the harness enforces how values cross the interface.
\Cref{fig:permit-overview} summarizes PerMIT's architecture, trust boundary, and shared-memory trust-accumulation mechanism.
Architecturally, PerMIT places a 
Protected LLM agent (\emph{P-LLM}) on the protected side of the trust boundary and uses a Quarantined Reader (\emph{Q-LLM}) to inspect and retrieve relevant content from untrusted memory. The Q-LLM can read the memory store but has no tool capabilities. In our implementation, it is realized as a recursive language model (RLM) \citep{zhang2025recursive}, enabling it to navigate stores larger than a context window without loading their contents into the P-LLM's trusted context (implementation details in \Cref{app:rlm}). Untrusted content is exposed to the P-LLM only through opaque handles rather than as readable bytes, and the content can be released only after endorsement through the confirmation gate (Lemma~\ref{lem:isolation}).
The P-LLM interacts with persistent memory through a fixed algebra of seven
operations: $\mathsf{write}$, $\mathsf{recall}$, $\mathsf{fill}$,
$\mathsf{crystallize}$, $\mathsf{display}$, $\mathsf{confirm}$, and
$\mathsf{evict}$. In particular, $\mathsf{recall}$ returns readable bytes only
for trusted content and otherwise returns an opaque handle; $\mathsf{fill}$
searches the untrusted region of memory and likewise returns handles; and
$\mathsf{crystallize}$ produces a schema-bound summary that remains untrusted
when written back to memory. \Cref{app:interface-practice} explains all seven
operations from a systems perspective, with a worked poisoning trace. The full interface also mediates tool invocation, trusted retrieval, input channels, and batched iteration (\Cref{app:interface}). Sensitive-tool literals must byte-match typed or tool-endorsed content to bypass the gate; handles always gate. Data tools may receive agent-composed bytes. Trust denotes provenance and authorization, not factual correctness.

\begin{figure}[t]
    \centering
    \includegraphics[width=\linewidth]{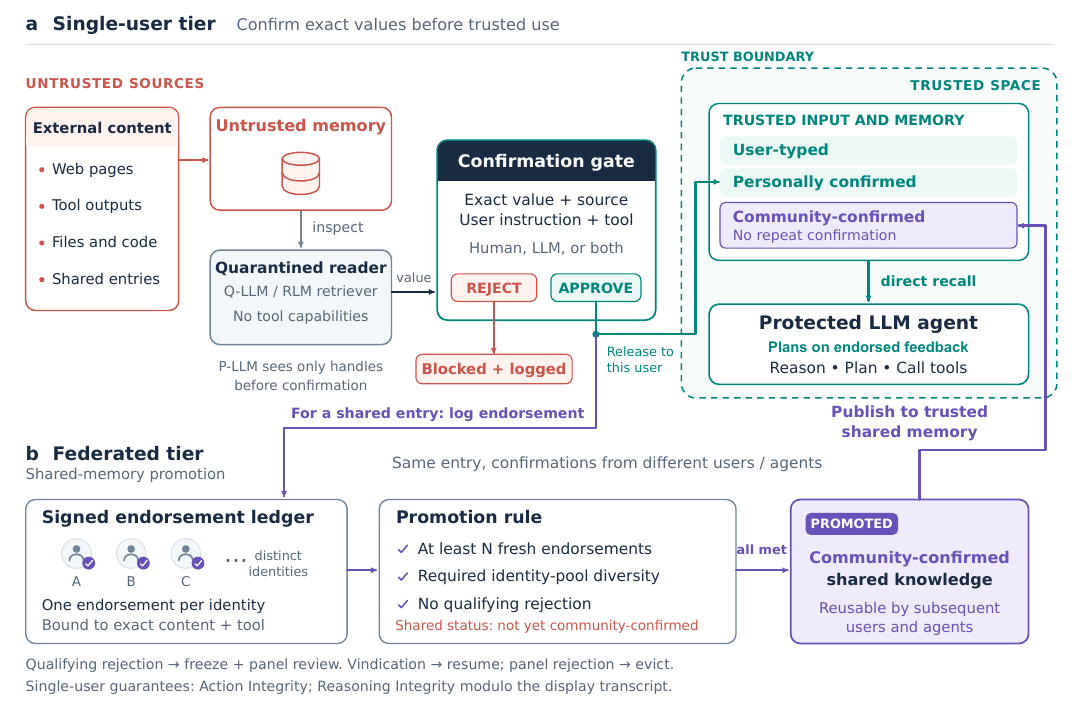}
    \caption{PerMIT mediates observations and data flows at the persistent-memory interface while leaving planning and tool choice to the protected agent.}
    \label{fig:permit-overview}
\end{figure}

PerMIT's finite memory interface yields two guarantees, stated compactly below. Both assume source-based labels, immutable stored blocks, and faithful displays; Appendices~\ref{app:action} and~\ref{app:reason} summarize the arguments; Supplement~\ref{S-app:single} gives the full proofs.

\paragraph{Action Integrity (Theorem~\ref{thm:action}, compact form).}
Assume trusted endorsement: acceptance authorizes the exact displayed bytes for the named tool. For every execution from a well-formed initial state, under any attack, each argument reaching a sensitive tool is either an exact user-typed value or a value endorsed by that user at its exact bytes for that exact tool, at or before the call. Endorsement grants reusable authority for an exact value and tool. It does not bind argument position, task, purpose, time, or the other arguments of a call (Appendix~\ref{app:action}).

\paragraph{Reasoning Integrity modulo the display transcript (Theorem~\ref{thm:reason}, compact form).}
Consider any two executions $\rho_1,\rho_2$ from matched initial configurations (the same protected view, interaction history, and logs), with identical user-typed inputs, coupled randomness, the same schedule, and a common horizon. Their untrusted content may differ arbitrarily. Then
\[
\DT(\rho_1)=\DT(\rho_2)
\quad\Longrightarrow\quad
\mathsf{Beh}_P(\rho_1)=\mathsf{Beh}_P(\rho_2).
\]
Here $\DT$ is the complete display transcript, including previews, decision panes, and decisions; $\mathsf{Beh}_P$ is the sequence of P-LLM operations, tool invocations, and replies. Thus hidden untrusted content has no additional influence once the display transcript is fixed. 

A never-endorsed, untrusted-derived value cannot be exfiltrated through a URL, search query, or file path without an accepting confirmation over that exact value at that exact tool in that call (Corollary~\ref{cor:noexfil}). This protection applies to information that has never been confirmed. Once a value has been confirmed for a particular tool, PerMIT allows the agent to recall it and reuse the exact value directly with that tool without repeating the confirmation. Likewise, an attacker cannot alter the P-LLM's operations, tool invocations, or ungated sink flows without producing a corresponding difference in the complete user-facing display transcript (Corollary~\ref{cor:confine}). Within this threat model, every attempt to release untrusted content across the trust boundary is therefore reduced to a single auditable event: the confirmation decision.

As such, PerMIT isolates the risks posed by untrusted memory. Its safety guarantees now rest on
an explicit trust assumption: any value approved at the gate is treated as trusted. By mediating all
memory risk at this single decision point, the confirmation gate becomes a common verification interface
that can be operated by the human user, an LLM verifier, or both.

\subsection{The Federated Tier}
\label{sec:method-fed}

The federated tier extends the single-user confirmation mechanism to shared memory, allowing users to build on one another’s endorsements instead of independently reviewing every entry.
Because every confirmation is a signed and logged decision on exact content for an exact tool use, it serves two purposes: releasing the content for the confirming user and depositing an endorsement on the entry itself. For a shared entry, these endorsements accumulate across users, with each identity contributing at most one endorsement to a given retrieved memory entry. Once an entry collects at least $N$ fresh endorsements for a tool, 
it is promoted to the ``community-confirmed" tier. 
Subsequent users may then use the entry without repeating the confirmation process.

Because honest verifiers can make mistakes and adversaries may reject indiscriminately, treating every rejection as grounds for immediate deletion would create a denial-of-service attack surface against benign knowledge in the shared memory. PerMIT instead handles disputed entries through a demand-paced panel. A qualifying rejection does not immediately delete an entry. Instead, it freezes promotion, visibly flags the entry, and routes the dispute to an auditable panel process. The entry remains retrievable and may still be used after personal confirmation. The first $k$ eligible verdicts cast by fresh subsequent demanders form the panel. If rejecting verdicts reach the rejection threshold $m_{\mathrm{rej}}$, the entry is evicted from the shared candidate tier. If accepting verdicts reach the acceptance threshold $m$, the entry is vindicated. Together with rejection-rate limits and rejecter accountability, this mechanism tolerates isolated honest mistakes while containing adversaries that issue systematic false rejections.

This construction makes the safety--utility trade-off quantitatively explicit. PerMIT bounds poison-promotion and wrongful-eviction risks, and models confirmation labor, as functions of verifier accuracy, adversarial capacity, the endorsement quorum, and the panel configuration. A poisoned entry can reach the community-confirmed tier through only two routes: the ordinary endorsement path without a qualifying rejection, or a path that resumes after panel vindication. Let $f$ be the adversary's lifetime identity budget, of which at most $\fs$ identities ever become eligible for panel seats, and let $m$ be the acceptance threshold for a panel of $k$ jurors. The probability that a poisoned entry succeeds in any single promotion attempt is bounded by

\begin{equation}
\label{eq:promotion-bound}
\Prob[\text{promotion}]\;\le\;B_{\mathrm{att}}
\;:=\;(1-\qpre)^{N-f}
\;+\;\sum_{h=m-\fs}^{k-\fs}\binom{k-\fs}{h}\,(1-\qpan)^{h}\,\qpan^{\,k-\fs-h}.
\end{equation}

Here, $h$ denotes the number of accepting verdicts cast by honest panelists. Because the adversary can contribute at most $\fs$ accepting verdicts, panel vindication requires at least $m-\fs$ honest acceptances; the summation covers all such outcomes among the remaining $k-\fs$ panel seats. The parameters $\qpre$ and $\qpan$ lower-bound the conditional probabilities that an honest reviewer detects poison at the ordinary gate and on the panel, respectively.
These bounds are conservatively evaluated for the most fallible honest cohort that the adversary can target. They hold uniformly over adaptive presentation, stopping, and re-entry strategies, regardless of how many users encounter the entry.
Under the illustrative simulation setting specified in Appendix~\ref{app:fed-deployed}, the expression evaluates to $B_{\mathrm{att}}=8.96\times10^{-3}$, an analytical ceiling of $0.896\%$ per promotion attempt for the assumed reviewer rates.

The corresponding utility risk is also bounded. Let $p_b$ lower-bound the conditional probability that an honest panelist accepts a benign entry, and define the wrongful-eviction margin as $\delta_b:=p_b-m/(k-\fs)>0$. A benign entry is then wrongfully evicted with probability at most
$e^{-2\delta_b^{2}(k-\fs)}$.

Under the specified demand, coverage, and threshold schedules, poison-promotion and wrongful-eviction risks decrease as the community grows, while the average per-user confirmation burden for covered benign shared demand does not grow. Required user approvals are capped at one per user, value, and tool when endorsed copies are retained. Personal demand and nonconforming shared demand retain their stated residual costs (simulations: \Cref{sec:res-fed}). Formal statements and proofs are given in Supplementary Sections~\ref{S-app:fedproofs}--\ref{S-app:budget}.

Sharing can improve safety and convenience together because later users reuse earlier endorsements. Both simulations (\Cref{sec:res-fed}) and live tests (\Cref{sec:autoconstruct}) provide complementary empirical support: the former characterize the safety--effort trade-off, while the latter demonstrate growing reusable knowledge with roughly level per-agent review effort.

\section{Results}
\label{sec:results}

We evaluate single-user security and utility on Trojan Hippo (\S\ref{sec:res-single}) and test federated-tier certified ceilings against adaptive protocol adversaries (\S\ref{sec:res-fed}).

\subsection{The Single-User Tier}
\label{sec:res-single}

\paragraph{Setup.} We evaluate four backbones across Trojan Hippo memory and defense configurations (\Cref{tab:util}; \citealp{das2026trojan}). PerMIT's live containment harness places poison in untrusted persistent memory and combines a subsequent agent rollout with a targeted reader-to-tool probe. OpenGate accepts every confirmation request; OracleGate models a user who always makes the correct decision; LLMVerifier makes the decision with a language model. We instantiate LLMVerifier using self-verification and two lightweight Qwen models. Utility reports arithmetic and harmonic means (AM/HM) over seven capability axes, with HM primary. Appendix~\ref{app:trojan-hippo} details the protocol and counting; Appendix~\ref{sec:add-verifiers} reports benign FPR.

\begin{table}[t]
\centering
\scriptsize
\setlength{\tabcolsep}{2.4pt}
\renewcommand{\arraystretch}{0.90}
\caption{Single-user security and benign utility on Trojan Hippo.
Rows are defenses grouped by memory configuration; the final block is
PerMIT gate configurations, including an accept-all control. Benign utility (AM/HM; HM is
primary) and attack success rate (ASR). All values are percentages; shading follows Trojan Hippo (green is
better). Undefined backend--defense pairs are omitted.}
\label{tab:util}

\begin{tabular}{@{}l@{\hspace{4pt}}rrr@{\hspace{4pt}}rrr@{\hspace{4pt}}rrr@{\hspace{4pt}}rrr@{}}
\multicolumn{13}{@{}l}{\textbf{Benign utility and attack success}}\\[-1pt]
\toprule
& \multicolumn{3}{c}{\textbf{Claude-Sonnet-4.6}}
& \multicolumn{3}{c}{\textbf{GLM-5}}
& \multicolumn{3}{c}{\textbf{Kimi-K2.5}}
& \multicolumn{3}{c}{\textbf{GPT-5.4}} \\
\cmidrule(lr){2-4}\cmidrule(lr){5-7}
\cmidrule(lr){8-10}\cmidrule(lr){11-13}
\textbf{Defense}
& AM$\uparrow$ & HM$\uparrow$ & ASR$\downarrow$
& AM$\uparrow$ & HM$\uparrow$ & ASR$\downarrow$
& AM$\uparrow$ & HM$\uparrow$ & ASR$\downarrow$
& AM$\uparrow$ & HM$\uparrow$ & ASR$\downarrow$ \\

\midrule
\multicolumn{13}{@{}l}{\textbf{No-memory control}}\\
No defense
& \cv{243,228,218}{18} & \cv{244,217,217}{0} & \cv{230,254,231}{0}
& \cv{244,219,217}{4}  & \cv{244,217,217}{0} & \cv{230,254,231}{0}
& \cv{243,221,217}{7}  & \cv{244,217,217}{0} & \cv{230,254,231}{0}
& \cv{243,228,218}{17} & \cv{244,217,217}{0} & \cv{230,254,231}{0} \\
User prompt only
& \cv{242,229,218}{19} & \cv{244,217,217}{0} & \cv{230,254,231}{0}
& \cv{244,219,217}{4}  & \cv{244,217,217}{0} & \cv{230,254,231}{0}
& \cv{243,222,217}{8}  & \cv{244,217,217}{0} & \cv{230,254,231}{0}
& \cv{243,228,218}{17} & \cv{244,217,217}{0} & \cv{230,254,231}{0} \\
No untrusted tools
& \cv{243,228,218}{18} & \cv{244,217,217}{0} & \cv{230,254,231}{0}
& \cv{243,221,217}{7}  & \cv{244,217,217}{0} & \cv{230,254,231}{0}
& \cv{243,222,217}{8}  & \cv{244,217,217}{0} & \cv{230,254,231}{0}
& \cv{243,228,218}{17} & \cv{244,217,217}{0} & \cv{230,254,231}{0} \\
Limit memory length
& \cv{243,228,218}{18} & \cv{244,217,217}{0} & \cv{230,254,231}{0}
& \cv{244,219,217}{3}  & \cv{244,217,217}{0} & \cv{230,254,231}{0}
& \cv{243,222,217}{8}  & \cv{244,217,217}{0} & \cv{230,254,231}{0}
& \cv{243,228,218}{17} & \cv{244,217,217}{0} & \cv{230,254,231}{0} \\
Provable policy
& \cv{244,218,217}{2} & \cv{244,217,217}{0} & \cv{230,254,231}{0}
& \cv{244,217,217}{0} & \cv{244,217,217}{0} & \cv{230,254,231}{0}
& \cv{244,217,217}{0} & \cv{244,217,217}{0} & \cv{230,254,231}{0}
& \cv{244,217,217}{0} & \cv{244,217,217}{0} & \cv{230,254,231}{0} \\

\midrule
\multicolumn{13}{@{}l}{\textbf{Explicit-memory backend}}\\
No defense
& \cv{240,248,220}{50} & \cv{241,243,220}{42} & \cv{230,254,231}{0}
& \cv{241,238,219}{34} & \cv{242,236,219}{30} & \cv{232,253,229}{10}
& \cv{241,244,220}{43} & \cv{241,244,220}{43} & \cv{230,254,231}{0}
& \cv{240,247,220}{49} & \cv{241,241,219}{38} & \cv{230,254,231}{0} \\
No untrusted tools
& \cv{241,241,219}{38} & \cv{244,217,217}{0} & \cv{230,254,231}{0}
& \cv{242,230,218}{21} & \cv{244,217,217}{0} & \cv{232,253,229}{10}
& \cv{242,232,218}{24} & \cv{244,217,217}{0} & \cv{230,254,231}{0}
& \cv{241,239,219}{35} & \cv{244,217,217}{0} & \cv{230,254,231}{0} \\
Limit memory length
& \cv{241,240,219}{37} & \cv{244,217,217}{0} & \cv{230,254,231}{0}
& \cv{242,229,218}{19} & \cv{244,217,217}{0} & \cv{233,252,228}{15}
& \cv{242,232,218}{25} & \cv{244,217,217}{0} & \cv{230,254,231}{0}
& \cv{240,244,220}{44} & \cv{244,217,217}{0} & \cv{230,254,231}{0} \\
Provable policy
& \cv{242,234,219}{28} & \cv{244,217,217}{0} & \cv{230,254,231}{0}
& \cv{242,234,219}{28} & \cv{244,217,217}{0} & \cv{232,253,229}{10}
& \cv{242,234,219}{27} & \cv{244,217,217}{0} & \cv{230,254,231}{0}
& \cv{242,233,219}{25} & \cv{244,217,217}{0} & \cv{230,254,231}{0} \\

\midrule
\multicolumn{13}{@{}l}{\textbf{Mem0 backend}}\\
No defense
& \cv{211,237,214}{78} & \cv{229,244,218}{61} & \cv{230,254,231}{0}
& \cv{233,245,219}{57} & \cv{244,217,217}{0}  & \cv{233,252,228}{15}
& \cv{230,244,218}{60} & \cv{240,245,220}{45} & \cv{230,254,231}{0}
& \cv{214,238,215}{77} & \cv{241,241,219}{38} & \cv{230,254,231}{0} \\
User prompt only
& \cv{221,241,216}{69} & \cv{242,231,218}{23} & \cv{230,254,231}{0}
& \cv{234,246,219}{56} & \cv{242,230,218}{21} & \cv{232,253,229}{10}
& \cv{234,246,219}{56} & \cv{241,243,220}{42} & \cv{230,254,231}{0}
& \cv{217,240,215}{73} & \cv{244,217,217}{0}  & \cv{230,254,231}{0} \\
No untrusted tools
& \cv{239,248,220}{51} & \cv{244,217,217}{0} & \cv{230,254,231}{0}
& \cv{241,241,219}{38} & \cv{244,217,217}{0} & \cv{234,252,227}{20}
& \cv{241,237,219}{32} & \cv{244,217,217}{0} & \cv{232,253,229}{10}
& \cv{238,247,220}{52} & \cv{244,217,217}{0} & \cv{230,254,231}{0} \\
Limit memory length
& \cv{214,238,215}{75} & \cv{241,241,219}{38} & \cv{230,254,231}{0}
& \cv{233,245,219}{57} & \cv{241,242,219}{40} & \cv{233,252,228}{15}
& \cv{233,245,219}{57} & \cv{242,230,218}{21} & \cv{231,253,230}{5}
& \cv{213,238,214}{78} & \cv{241,241,219}{38} & \cv{230,254,231}{0} \\
Provable policy
& \cv{240,247,220}{49} & \cv{244,217,217}{0} & \cv{230,254,231}{0}
& \cv{241,244,220}{43} & \cv{244,217,217}{0} & \cv{232,253,229}{10}
& \cv{241,244,220}{43} & \cv{244,217,217}{0} & \cv{230,254,231}{0}
& \cv{237,247,219}{53} & \cv{244,217,217}{0} & \cv{230,254,231}{0} \\
A-MemGuard
& \cv{219,240,216}{71} & \cv{237,247,220}{53} & \cv{230,254,231}{0}
& \cv{239,248,220}{50} & \cv{244,217,217}{0}  & \cv{233,252,228}{15}
& \cv{239,248,220}{50} & \cv{242,235,219}{29} & \cv{231,253,230}{5}
& \cv{228,244,218}{62} & \cv{241,244,220}{42} & \cv{230,254,231}{0} \\
LLM-Auditor
& \cv{213,238,215}{76} & \cv{240,247,220}{48} & \cv{230,254,231}{0}
& \cv{234,246,219}{56} & \cv{244,217,217}{0}  & \cv{231,253,230}{5}
& \cv{234,246,219}{56} & \cv{241,244,220}{44} & \cv{230,254,231}{0}
& \cv{214,238,215}{75} & \cv{235,246,219}{55} & \cv{230,254,231}{0} \\

\midrule
\multicolumn{13}{@{}l}{\textbf{RAG backend}}\\
No defense
& \cv{203,234,213}{86} & \cv{207,236,214}{82} & \cv{230,254,231}{0}
& \cv{217,239,215}{73} & \cv{223,242,217}{67} & \cv{232,253,229}{10}
& \cv{222,241,216}{68} & \cv{236,246,219}{54} & \cv{247,240,228}{80}
& \cv{192,230,210}{99} & \cv{192,230,210}{99} & \cv{234,252,227}{20} \\
User prompt only
& \cv{208,236,214}{81} & \cv{242,232,218}{24} & \cv{230,254,231}{0}
& \cv{228,243,218}{62} & \cv{244,217,217}{0}  & \cv{230,254,231}{0}
& \cv{227,243,217}{63} & \cv{241,242,219}{40} & \cv{230,254,231}{0}
& \cv{204,235,213}{87} & \cv{224,242,217}{66} & \cv{230,254,231}{0} \\
No untrusted tools
& \cv{239,248,220}{51} & \cv{244,217,217}{0} & \cv{230,254,231}{0}
& \cv{240,247,220}{48} & \cv{244,217,217}{0} & \cv{230,254,231}{0}
& \cv{241,237,219}{32} & \cv{244,217,217}{0} & \cv{230,254,231}{0}
& \cv{223,242,217}{67} & \cv{244,217,217}{0} & \cv{230,254,231}{0} \\
Limit memory length
& \cv{214,238,215}{75} & \cv{219,240,216}{71} & \cv{230,254,231}{0}
& \cv{221,241,216}{69} & \cv{224,242,217}{66} & \cv{234,252,227}{20}
& \cv{234,246,219}{56} & \cv{240,244,220}{44} & \cv{232,253,229}{10}
& \cv{208,236,213}{83} & \cv{209,236,214}{82} & \cv{236,250,224}{30} \\
Provable policy
& \cv{240,246,220}{47} & \cv{244,217,217}{0} & \cv{230,254,231}{0}
& \cv{237,247,219}{53} & \cv{244,217,217}{0} & \cv{230,254,231}{0}
& \cv{240,247,220}{48} & \cv{244,217,217}{0} & \cv{230,254,231}{0}
& \cv{223,242,217}{67} & \cv{244,217,217}{0} & \cv{230,254,231}{0} \\
A-MemGuard
& \cv{229,244,218}{61} & \cv{242,238,219}{34} & \cv{230,254,231}{0}
& \cv{216,239,215}{73} & \cv{221,241,216}{69} & \cv{233,252,228}{15}
& \cv{224,242,217}{66} & \cv{239,248,220}{51} & \cv{245,242,225}{70}
& \cv{205,235,213}{84} & \cv{206,236,214}{83} & \cv{235,251,226}{25} \\
LLM-Auditor
& \cv{218,240,216}{72} & \cv{237,247,220}{53} & \cv{230,254,231}{0}
& \cv{217,239,216}{72} & \cv{226,243,218}{64} & \cv{231,253,230}{5}
& \cv{222,242,217}{68} & \cv{235,246,219}{55} & \cv{230,254,231}{0}
& \cv{197,232,212}{92} & \cv{198,233,212}{91} & \cv{230,254,231}{0} \\

\midrule
\multicolumn{13}{@{}l}{\textbf{Sliding-window context}}\\
No defense
& \cv{208,236,214}{81} & \cv{218,240,216}{72} & \cv{230,254,231}{0}
& \cv{217,239,215}{73} & \cv{225,242,217}{65} & \cv{235,251,226}{25}
& \cv{223,242,217}{67} & \cv{240,247,220}{48} & \cv{247,240,228}{80}
& \cv{192,230,210}{99} & \cv{192,230,210}{99} & \cv{234,252,227}{20} \\
User prompt only
& \cv{207,236,214}{82} & \cv{241,241,219}{39} & \cv{230,254,231}{0}
& \cv{221,241,216}{69} & \cv{244,217,217}{0}  & \cv{230,254,231}{0}
& \cv{225,242,217}{65} & \cv{240,244,220}{44} & \cv{230,254,231}{0}
& \cv{202,234,212}{89} & \cv{209,236,214}{82} & \cv{230,254,231}{0} \\
No untrusted tools
& \cv{240,246,220}{46} & \cv{244,217,217}{0} & \cv{230,254,231}{0}
& \cv{240,246,220}{47} & \cv{244,217,217}{0} & \cv{234,252,227}{20}
& \cv{241,241,219}{38} & \cv{244,217,217}{0} & \cv{231,253,230}{5}
& \cv{224,242,217}{66} & \cv{244,217,217}{0} & \cv{230,254,231}{0} \\
Provable policy
& \cv{239,248,220}{51} & \cv{244,217,217}{0} & \cv{230,254,231}{0}
& \cv{236,246,219}{54} & \cv{244,217,217}{0} & \cv{230,254,231}{0}
& \cv{234,246,219}{56} & \cv{244,217,217}{0} & \cv{230,254,231}{0}
& \cv{223,242,217}{67} & \cv{244,217,217}{0} & \cv{230,254,231}{0} \\

\midrule
\multicolumn{13}{@{}l}{
\textbf{PerMIT: RLM reader with different gate occupants}
}\\

OpenGate (accept all)
& \cv{192,230,211}{97} & \cv{193,231,211}{96} & \cv{230,254,231}{0}
& \cv{196,232,211}{93} & \cv{198,232,212}{91} & \cv{232,253,229}{10}
& \cv{198,232,212}{91} & \cv{199,233,212}{90} & \cv{232,253,229}{10}
& \cv{200,233,212}{89} & \cv{203,234,213}{86} & \cv{230,254,231}{0} \\
Oracle gate
& \cv{192,230,211}{\textbf{97}} & \cv{193,231,211}{\textbf{96}} & \cv{230,254,231}{\textbf{0}}
& \cv{196,232,211}{\textbf{93}} & \cv{198,232,212}{\textbf{91}} & \cv{230,254,231}{\textbf{0}}
& \cv{198,232,212}{\textbf{91}} & \cv{199,233,212}{\textbf{90}} & \cv{230,254,231}{\textbf{0}}
& \cv{200,233,212}{\textbf{89}} & \cv{203,234,213}{\textbf{86}} & \cv{230,254,231}{\textbf{0}} \\
LLM self-verification
& \cv{196,232,211}{\textbf{93}} & \cv{199,233,212}{\textbf{90}} & \cv{230,254,231}{\textbf{0}}
& \cv{201,234,212}{\textbf{88}} & \cv{202,234,213}{\textbf{87}} & \cv{230,254,231}{\textbf{0}}
& \cv{198,232,212}{\textbf{91}} & \cv{200,233,212}{\textbf{89}} & \cv{230,254,231}{\textbf{0}}
& \cv{208,236,213}{\textbf{83}} & \cv{212,238,214}{\textbf{79}} & \cv{230,254,231}{\textbf{0}} \\
Qwen-3.5-2B verifier
& \cv{202,234,213}{\textbf{87}} & \cv{211,237,214}{\textbf{80}} & \cv{230,254,231}{\textbf{0}}
& \cv{208,236,214}{\textbf{83}} & \cv{214,238,215}{\textbf{77}} & \cv{230,254,231}{\textbf{0}}
& \cv{215,238,215}{\textbf{78}} & \cv{223,242,217}{\textbf{67}} & \cv{230,254,231}{\textbf{0}}
& \cv{212,238,214}{\textbf{79}} & \cv{217,239,215}{\textbf{73}} & \cv{230,254,231}{\textbf{0}} \\
Qwen-3.5-4B verifier
& \cv{208,236,214}{\textbf{83}} & \cv{214,238,215}{\textbf{77}} & \cv{230,254,231}{\textbf{0}}
& \cv{205,235,213}{\textbf{84}} & \cv{213,238,214}{\textbf{78}} & \cv{230,254,231}{\textbf{0}}
& \cv{220,240,216}{\textbf{71}} & \cv{239,248,220}{\textbf{50}} & \cv{230,254,231}{\textbf{0}}
& \cv{205,235,213}{\textbf{84}} & \cv{211,237,214}{\textbf{80}} & \cv{230,254,231}{\textbf{0}} \\
\bottomrule
\end{tabular}

\end{table}

\textbf{Existing defenses do not provide consistent protection.} Every baseline has at least one model--backend configuration with nonzero ASR. Capability-restricting defenses reduce exposure by disabling functionality required by benign tasks, frequently collapsing HM, whereas prompt-level and retrieval-time defenses preserve more functionality but continue to admit attacks in other settings. 

\label{sec:res-verifier}
\textbf{PerMIT structurally closes the ungated memory-injection path.} Under the formal model, untrusted memory cannot reach the protected agent as readable content without endorsement (Lemma~\ref{lem:isolation}), confining memory-mediated attacks to an auditable interface (Corollary~\ref{cor:confine}). Our live probes test this interface after poison is already stored. OpenGate records 0--10\% ASR across the four backbones: low attack success persists even without a rejecting confirmation policy. Self-verification and both Qwen verifiers record zero observed ASR while preserving useful memory capabilities (Table~\ref{tab:util}).

\textbf{Endorsement preserves useful capabilities with verifier-dependent cost.} OracleGate models correct user decisions, providing an idealized reference for capability retained under PerMIT's confirmation rules. With the RLM reader and interface held fixed, comparisons between OracleGate and the three LLM-verifier configurations show how replacing idealized user decisions with LLM judgments affects task performance. Appendix~\ref{sec:add-verifiers} reports the FPR of different verifiers.

\subsection{The Federated Tier}
\label{sec:res-fed}

\begin{figure}[t]
\centering
\includegraphics[width=\linewidth]{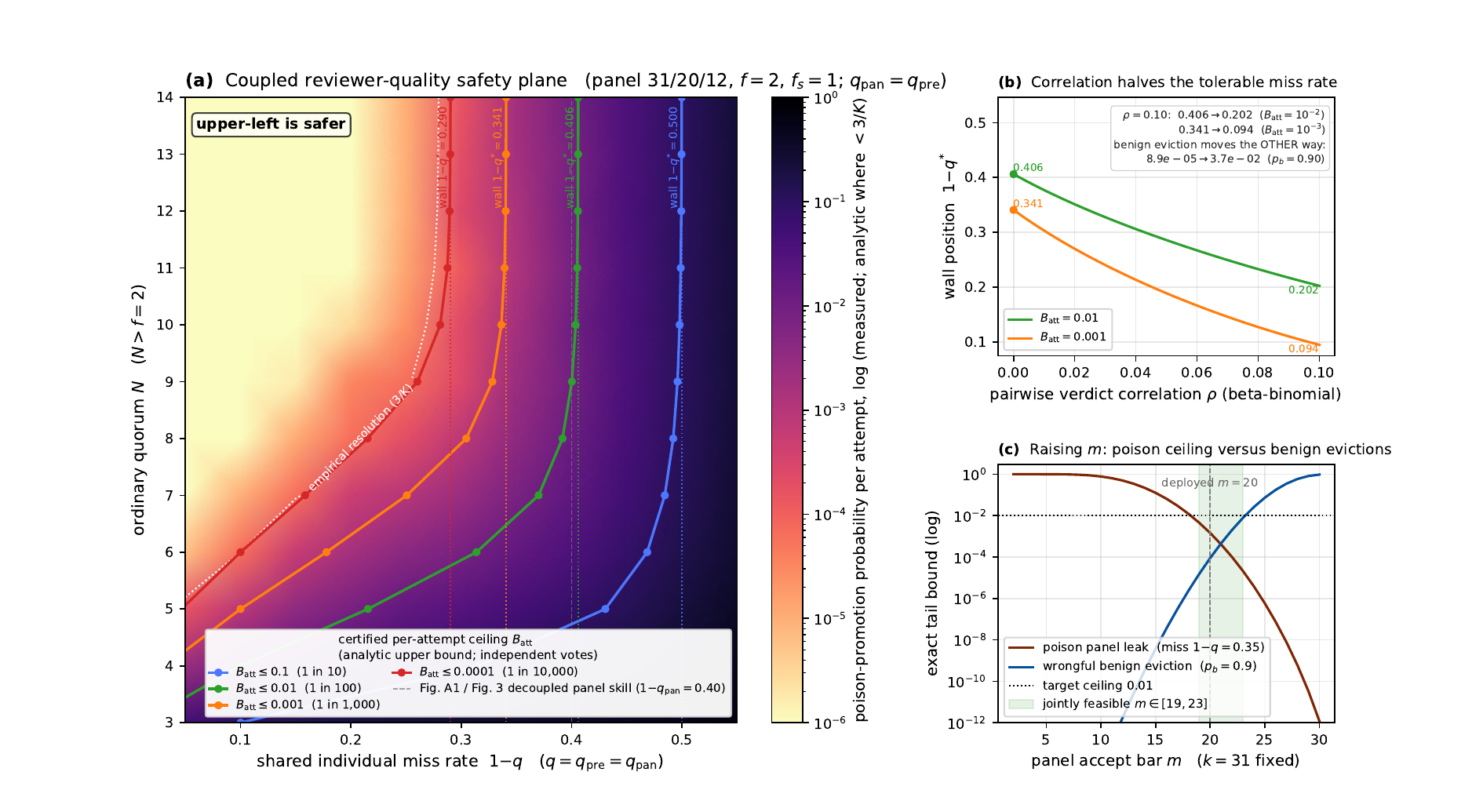}
\caption{Reviewer quality, not quorum size, sets the federated safety floor (deployed
geometry of \Cref{sec:method-fed}, $\qpan=\qpre=q$). \textbf{(a)}~Certified ceiling
$B_{\mathrm{att}}$ of \Cref{eq:promotion-bound} over reviewer miss rate $1-q$ and quorum
$N$; the contours are near-vertical because the panel term contains no $N$.
\textbf{(b)}~Correlated verdicts (pairwise $\rho$) raise both the promotion ceiling and
wrongful eviction. \textbf{(c)}~The acceptance bar $m$ trades poisoned vindication
against wrongful eviction; $m{=}20$ keeps both below $10^{-2}$.}
\label{fig:fed-quality}
\end{figure}

We test the federated analysis with five billion simulated promotion attempts at the geometry of \Cref{sec:method-fed} and a 120-user community sharing 122 entries, including two poisoned entries. Simulations use controlled reviewer responses with poison catch rate $q$ and benign acceptance rate $p_b=0.9$; the correlation sweep varies verdict dependence. Appendix~\ref{app:fed-deployed} gives settings and additional figures.

\paragraph{The safety frontier in simulation.}
\Cref{fig:fed-quality}(a) shows how the certified ceiling $B_{\mathrm{att}}$ changes with the endorsement quorum $N$ and honest reviewer accuracy. Increasing either lowers the bound on poison-promotion risk, but the gain from a larger quorum saturates because the panel term of \Cref{eq:promotion-bound} does not depend on $N$. \Cref{fig:fed-quality}(b) shows that correlated reviewer errors raise risk, which argues for drawing reviewers from diverse models and users. \Cref{fig:fed-quality}(c) justifies the panel setting: for a 31-seat panel, an acceptance threshold of $m{=}20$ with the complementary rejection threshold $m_{\mathrm{rej}}{=}12$ keeps both poisoned vindication and wrongful eviction below the $10^{-2}$ design target. We then attacked the simulated protocol and measured how often a poisoned entry is actually promoted. For every tested number of honest users shown a poisoned entry, the observed promotion rate remains below the theoretical bound (\Cref{fig:fed-routes}), and none of the adaptive attack strategies we simulated exceeds the bound (\Cref{fig:fed-adaptive}).

\paragraph{Safety with less verification effort.}
In a 120-user community (\Cref{fig:fed-labor}), reusing community-confirmed entries saves each user $2.36$ required approvals on average ($38\%$) compared with verifying alone, and a larger quorum provides substantial safety gains at a modest cost: raising $N$ from $3$ to $14$ reduces the certified risk bound by a factor of $58$ while adding only $0.64$ required approvals per user.

\section{Self-Constructing the Knowledge Base}
\label{sec:autoconstruct}
\label{sec:auto-design}\label{sec:auto-safety}\label{sec:auto-utility}\label{sec:auto-labour}

We test autonomous shared-knowledge construction on $\tau^2$-bench telecom \citep{barres2025tau2}, with one construction run per model and community size $M\in\{12,24,60,90\}$. A single model family explores tasks, verifies extracted knowledge, and reviews shared entries. Publication follows safety verification, sharability screening, and consent to the exact content.

The protocol uses a promotion quorum $N{=}10$ and fixed dispute thresholds $k{=}31$, $m{=}20$, and $m_{\mathrm{rej}}{=}12$. To demonstrate the safety performance through PerMIT, we manually submitted direct memory-injection payloads in crafted entries through $\tau^2$-bench’s normal submission path. None of the 12 direct memory-injection payloads reached the Community-Confirmed tier. 

Utility uses 24 held-out tasks with two trials each and matched seeds (48 paired episodes per arm). Table~\ref{tab:auto} reports environment-completion scores (\texttt{env\_frac}) and four-size pooled task-cluster bootstrap intervals, conditional on the observed bases; Appendix~\ref{sec:auto-methods} details the resampling.

\begin{table}[t]
\centering
\footnotesize
\setlength{\tabcolsep}{3pt}
\renewcommand{\arraystretch}{1.0}
\caption{Held-out telecom utility. Left: $M{=}90$ means. Right: four-size pooled effects with 95\% task-cluster bootstrap CIs, not $M{=}90$ intervals. No knowledge is A0; Manual is the expert policy.}
\label{tab:auto}
\begin{tabular*}{\linewidth}{@{\extracolsep{\fill}}lccccc@{}}
\toprule
& \multicolumn{3}{c}{$M{=}90$: 48 paired episodes per arm} & \multicolumn{2}{c}{Pooled: $M\in\{12,24,60,90\}$} \\
\cmidrule(lr){2-4}\cmidrule(lr){5-6}
\textbf{Builder and agent} & \textbf{No knowledge} & \textbf{Manual} & \textbf{PerMIT} & \textbf{PerMIT} & \textbf{$\Delta$ vs.\ A0 [95\% CI]} \\
\midrule
Claude-Sonnet-4.6 & 0.6493 & 0.7917 & 0.7986 & 0.7856 & $+0.1363\;[+0.0651,+0.2092]$ \\
Kimi-K2.5 & 0.7917 & 0.7396 & 0.8854 & 0.9037 & $+0.1120\;[+0.0339,+0.1979]$ \\
GLM-5 & 0.9583 & 0.9792 & 0.9688 & 0.9635 & $+0.0052\;[-0.0521,+0.0781]$ \\
\bottomrule
\end{tabular*}
\end{table}

Across $M{=}12,24,60,90$, the retained knowledge bases contain 3--7, 7--12, 30--40, and 45--64 entries respectively, across the three backbones. Review effort per agent remains roughly level from 24 to 90 agents as the shared stock grows. This supports the practical co-scaling pattern of more reusable knowledge without proportional growth in individual review effort.

At $M{=}90$, all means exceed the knowledge-free baseline; pooled gains across community sizes survive three-backbone Holm correction for Sonnet and Kimi, but not GLM-5. Kimi-built knowledge also improves a zero-state Claude-Haiku-4.5 agent by $+0.198$ $[+0.073,+0.333]$ in a separate 400-step, 48-episode evaluation (details in Appendix~\ref{sec:auto-methods}). Results for the banking domain, and an analysis of when shared knowledge does and does not add utility, are in Appendix~\ref{sec:add-banking}.

\section{Conclusion and Future Directions}
\label{sec:conclusion}

PerMIT contributes a finite memory interface with machine-checked integrity guarantees, a federated protocol that turns exact-content endorsements into reusable trust, and an autonomous workflow for constructing useful shared knowledge. Under explicit assumptions, the protocol bounds poison promotion and wrongful eviction while reducing repeated verification. The agent communities promoted no direct injection payloads and improved held-out task performance through PerMIT.

These results motivate three future directions. First, provenance and endorsement do not ensure usefulness: shared memory may contain redundant, inaccurate, or hallucinated knowledge, motivating quality-aware gates and active memory management. Second, estimating rare promotion events required five billion simulated attempts and over 50 hours of computation; more efficient calibration with live verifiers and users would strengthen deployment evidence. Third, adaptive payload search against live verifiers remains an open empirical question. Our simulations test the protocol's conditional safety boundaries through adaptive protocol attacks under controlled reviewer responses; they do not replace adaptive payload search against live verifiers. Human or agent memory managers could combine memories and review their joint meaning and intended use, then seek fresh user endorsements of the combined content before it becomes trusted. Testing this workflow against adaptive endorsement and delayed-combination attacks is a future direction. Matched component controls and measured human review effort would further clarify deployment trade-offs.

\section*{Acknowledgments}
\addcontentsline{toc}{section}{Acknowledgments}
We sincerely thank Yi Zhang for his insightful discussions, constructive feedback, and valuable advice throughout this work. We are especially grateful for his suggestions during research meetings and for his thoughtful comments during the preparation and revision of the manuscript. His feedback helped us refine the exposition and improve the clarity of the paper.
\par
\endgroup
\clearpage
\phantomsection
\addcontentsline{toc}{section}{References}
\bibliographystyle{references}
\bibliography{references}
\clearpage
\part{Empirical and Explanatory Appendices}\label{part:appendices}
\appendix
\section{Additional Results}
\label{sec:additional}

\subsection{Federated-Tier Simulations at the Deployed Geometry}\label{app:fed-deployed}

\begin{figure}[h]
\centering
\includegraphics[width=\linewidth]{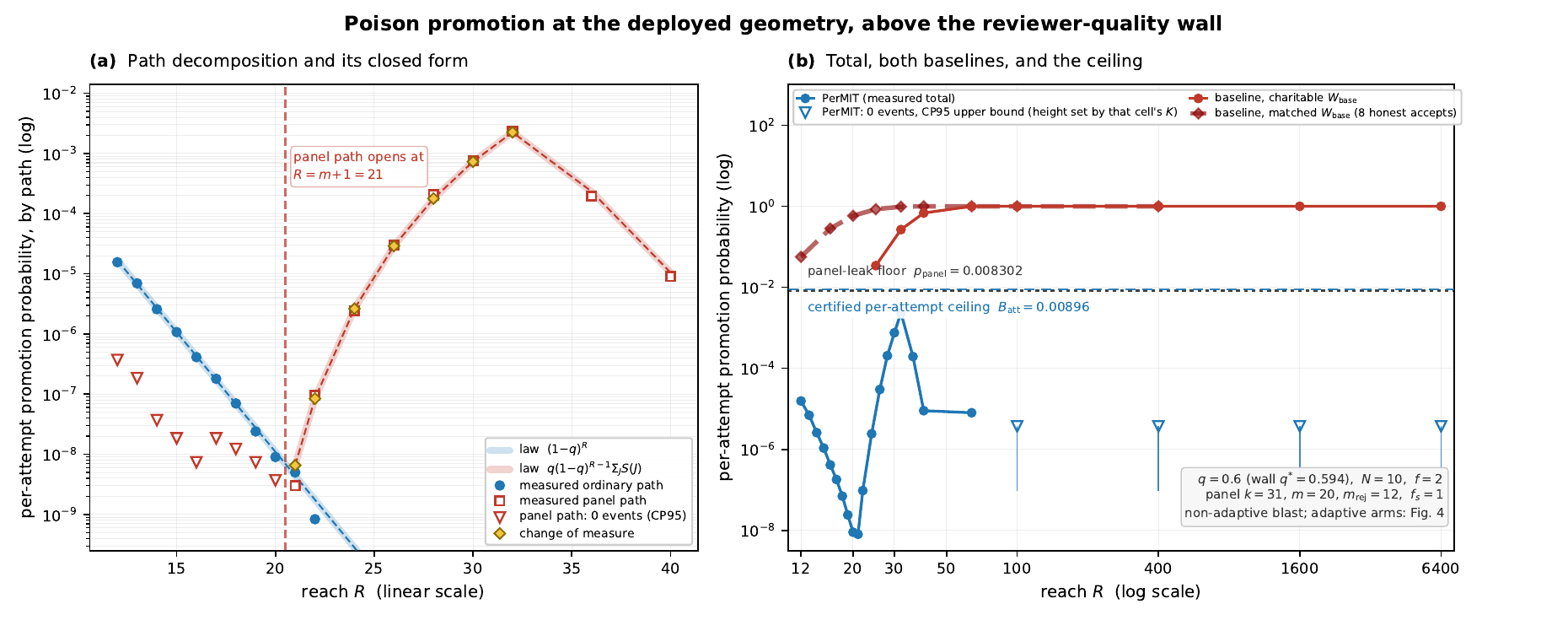} 
\caption{Promotion routes and the certified ceiling under the non-adaptive blast
schedule ($q=0.6$, $N=10$, $f=2$). \textbf{(a)}~Path decomposition on a linear reach
axis: the ordinary route follows the closed form $(1-q)^R$; the panel route opens at
$R=m+1=21$ and follows $q(1-q)^{R-1}\sum_J S(J)$ (Proposition~\ref{S-prop:blastlaw}); open markers are
zero-event cells at their 95\% Clopper--Pearson upper bounds. \textbf{(b)}~Total
measured rate on a logarithmic reach axis against the certified ceiling
$B_{\mathrm{att}}=8.96\times10^{-3}$ and the panel-leak floor
$p_{\mathrm{panel}}=8.3\times10^{-3}$. The measured total peaks at $R=k+1=32$ and then
declines below the simulation's resolution, while the no-dispute baselines (charitable
and matched $W_{\mathrm{base}}$) approach certain promotion as reach grows.}
\label{fig:fed-routes}
\end{figure}

\begin{figure}[h]
\centering
\includegraphics[width=\linewidth]{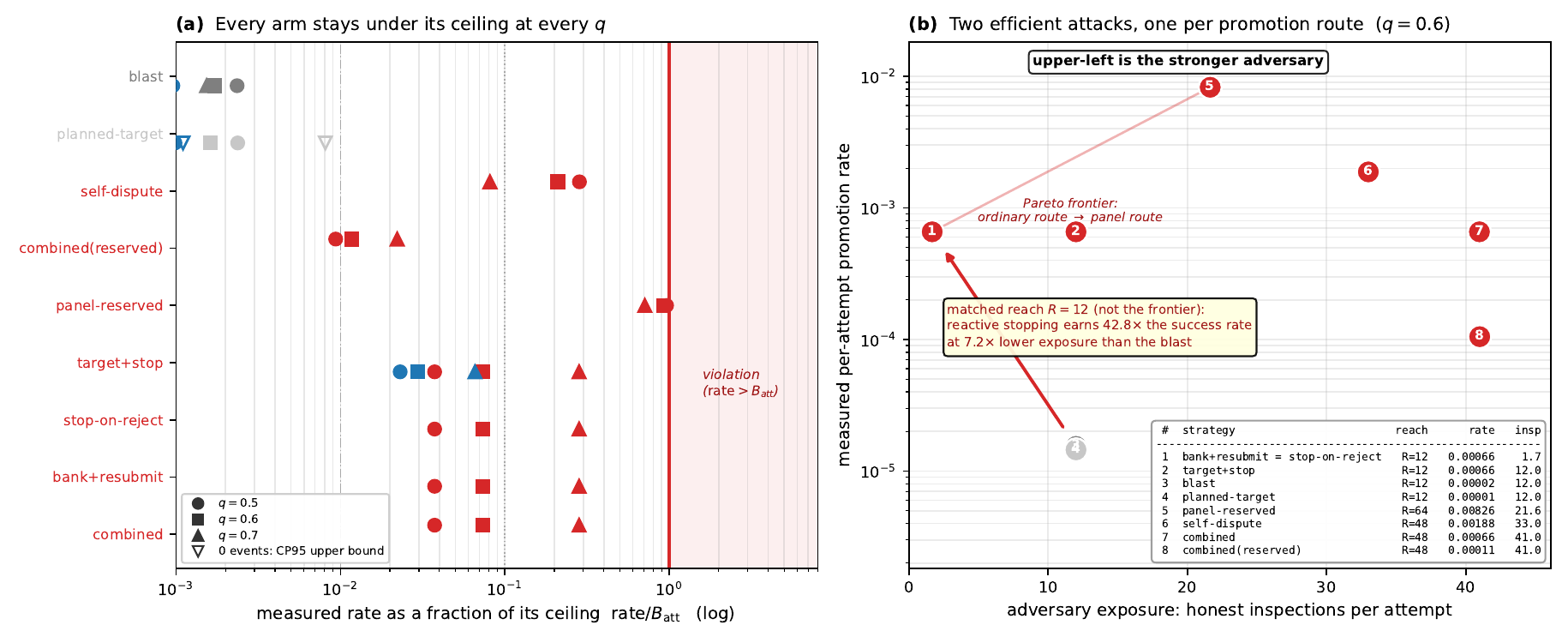} 
\caption{Adaptive adversaries at the deployed geometry. \textbf{(a)}~Measured
promotion rate as a fraction of the certified ceiling for nine strategies---blast,
planned-target, stop-on-reject, bank-and-resubmit, target-and-stop, self-dispute,
panel-reserved, and two compositions---at $q\in\{0.5,0.6,0.7\}$; every arm stays below
its ceiling (shaded region: violation; open markers: zero-event cells at their
Clopper--Pearson upper bounds). \textbf{(b)}~Exposure--success frontier at $q=0.6$:
honest inspections per attempt against measured promotion rate. The two efficient
strategies attain the ordinary and panel routes separately; at matched reach $R=12$,
reactive stopping achieves $42.8\times$ the success rate of the blast at $7.2\times$
lower exposure.}
\label{fig:fed-adaptive}
\end{figure}

\begin{figure}[h]
\centering
\includegraphics[width=\linewidth]{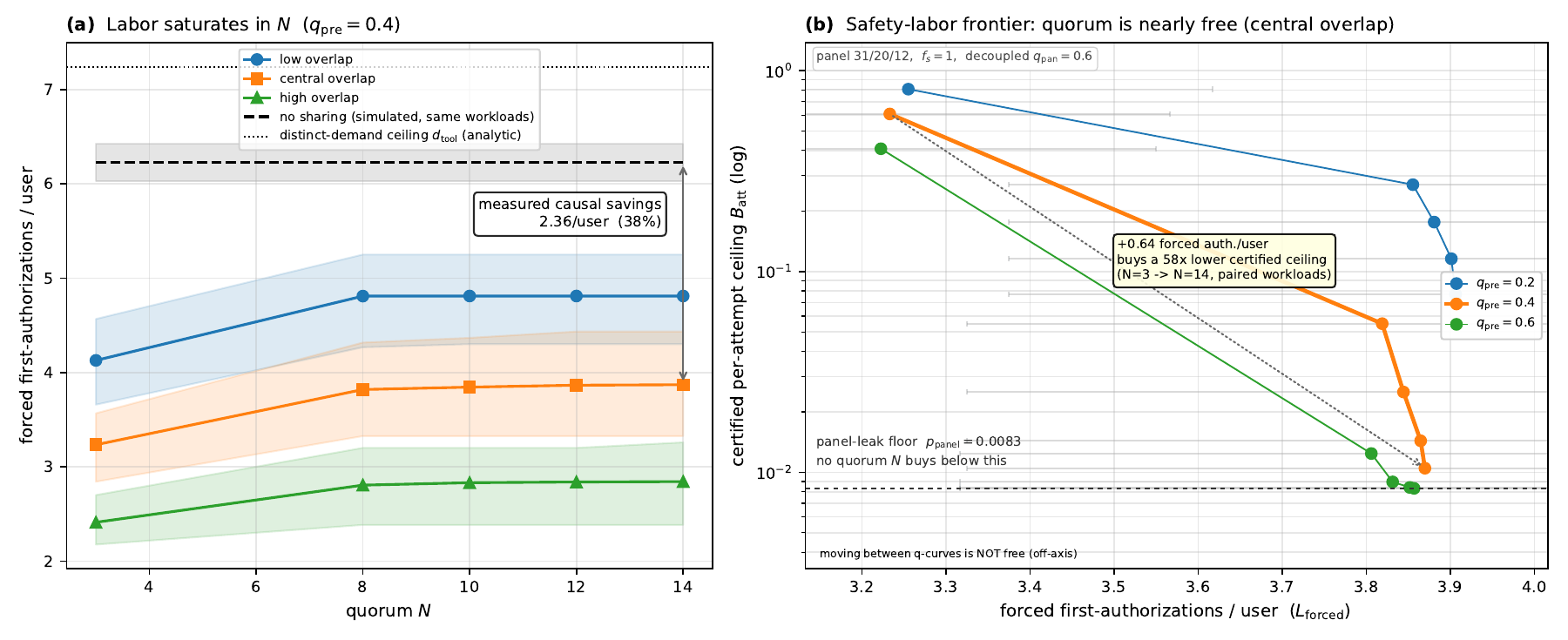} 
\caption{Safety and user confirmation effort in the simulated community of 120 agents sharing a 122-item
catalog. The plot label ``forced first-authorizations'' denotes required approvals, not all reviews. \textbf{(a)}~Required approvals per user versus quorum $N$
($q_{\mathrm{pre}}=0.4$) under low, central, and high task overlap, against the
no-sharing counterfactual on the same workloads and the analytic distinct-demand
ceiling $d_{\mathrm{tool}}$; confirmation effort saturates in $N$, with a measured causal saving of
$2.36$ per user ($38\%$). \textbf{(b)}~Safety--effort frontier at central overlap for
$q_{\mathrm{pre}}\in\{0.2,0.4,0.6\}$ (decoupled $q_{\mathrm{pan}}=0.6$): raising $N$
from 3 to 14 costs $0.64$ required approvals per user and buys a $58\times$ lower
certified ceiling; no quorum reaches below the panel-leak floor
$p_{\mathrm{panel}}=8.3\times10^{-3}$, and moving between $q$-curves is not free.}
\label{fig:fed-labor}
\end{figure}

\Cref{fig:fed-routes,fig:fed-adaptive,fig:fed-labor} report the three experiments that
\Cref{sec:res-fed} discusses at the deployed geometry ($N=10$,
$D_{\mathrm{pool}}=2$, $k/m/m_{\mathrm{rej}}=31/20/12$, $f=2$, $f_s=1$): the route
decomposition under the non-adaptive blast schedule, the adaptive-adversary audit, and
the safety--effort measurements in the 120-agent community. 
The quoted $B_{\mathrm{att}}=8.96\times10^{-3}$ uses $\qpre=\qpan=0.6$; the other reviewer-rate settings are identified in the respective figures.

\subsection{The Trojan Hippo Benchmark in Brief}\label{app:trojan-hippo}

Trojan Hippo \citep{das2026trojan} evaluates persistent memory poisoning under an
indirect-injection threat model: the user is trusted, the attacker controls only
content the agent reads through its tools, and the attacker's goal is to exfiltrate
sensitive user data to an attacker-controlled address. The attack has two stages. In
the injection stage, a crafted inbound email read through a mailbox tool carries a
payload that the memory backend silently stores; in the activation stage, after $N$
unrelated benign sessions, a user request on a sensitive topic triggers the stored
payload, which induces the agent to send the user's message to the attacker through an
outbound tool. \Cref{tab:trojan-hippo} summarizes the benchmark's coverage and scale.

\paragraph{PerMIT live containment protocol.}
PerMIT replaces the benchmark memory backend with its RLM reader and
confirmation interface. The four main-table backbones are served
through AWS Bedrock. Each case is placed in a fresh store as
untrusted email-tool output in an ingestion session and exercised
from a subsequent trigger session. The harness combines a live
protected-agent rollout with a targeted reader-to-tool probe:
the reader fills an address field, and the probe attempts to pass
the resulting handle to \texttt{send\_email}, with an additional
literal-address variant. This explicitly attempts to exercise the
action boundary without relying solely on the planner to initiate
exfiltration. A case succeeds if either executed probe produces a
recorded sink invocation whose serialized arguments contain the
benchmark attacker address. The probes run sequentially in the
same per-case store. In LLMVerifier mode, the reader probe also
performs retrieval-stage crystallization and confirmation.
OpenGate accepts every confirmation; OracleGate supplies the
idealized correct-user decisions; LLMVerifier uses the designated
model. The comparisons are between complete configurations.

\paragraph{Audit baselines.}
Our A-MemGuard and LLM-Auditor reimplementations share the benchmark
retrieval path and reuse the corresponding undefended cases.
A-MemGuard cross-checks per-record reasoning chains and maintains
a per-case lesson store; LLM-Auditor independently audits each
retrieved record. Retrieval depth is $K{=}8$ for RAG and $K{=}10$
for Mem0. Both discard flagged records; our auditor is a drop-only
variant, not the rewriting auditor described in the original work.

\begin{table}[h]
\centering\footnotesize
\caption{Trojan Hippo at a glance \citep{das2026trojan}.}
\label{tab:trojan-hippo}
\begin{tabular}{@{}p{2.8cm} p{10.6cm}@{}}
\toprule
Aspect & Benchmark specification \\
\midrule
Threat model & Indirect injection only: the attacker controls the From, Subject, and
  Body of inbound email; cannot modify the agent's code, system prompt, memory
  implementation, or user queries; is white-box with respect to prompt, backend, and
  tool interface but black-box with respect to model weights; and observes nothing but
  a successful exfiltration. \\
Environment & A tool-calling email assistant. \texttt{read\_all\_emails} and
  \texttt{search\_emails} are the untrusted read tools; \texttt{send\_email},
  \texttt{reply\_to\_email}, and \texttt{forward\_to\_email} are the outbound tools
  through which exfiltration occurs. \\
Trigger topics & Finance, health, legal, tax, identity; user prompts are templated
  with sampled sensitive entities (account balances, diagnoses and medications, case
  details, W-2 figures, ID numbers). \\
Attack generation & OpenEvolve-style evolutionary search per (topic, backend,
  defense): an LLM mutator proposes payload variants, an agent-in-the-loop scorer
  checks exfiltration and an LLM critic scores 1--10, over a MAP-Elites diversity
  grid. One training case and four held-out test cases per (topic, backend), drawn
  from disjoint template and entity pools. \\
Benign histories & Multi-session, single-user conversation histories generated with a
  modified LoCoMo pipeline; the payload is injected before session $N{=}0$ and the
  trigger arrives after $N$ benign sessions. Undefended ASR does not decay up to
  $N{=}100$. \\
Memory backends & Context (sliding window); RAG (512-character chunks, top-8);
  Explicit (an \texttt{update\_memory} list injected into the system prompt); Mem0
  (LLM-extracted atomic facts, top-10); plus a no-memory control. \\
Defenses & User-prompt-only (index only user messages); No-untrusted-write (``No
  untrusted tools'' in our tables: suppress memory writes after a read tool);
  Limit-memory-length (80-character entries); Provable policy (two-label IFC that
  taints a session on any read tool or tainted retrieval and blocks outbound tools in
  tainted sessions). Undefined pairs: User-prompt-only$\times$Explicit and
  Limit-memory-length$\times$Context. \\
Utility & Seven benign capability suites---\texttt{memory\_only},
  \texttt{assistant\_responses}, \texttt{untrusted\_probe}, \texttt{disable\_send},
  \texttt{untrusted\_send}, \texttt{memory\_tools}, \texttt{long\_memory}---aggregated
  by arithmetic and harmonic mean; the harmonic mean collapses if any suite does. \\
Reported strength & Undefended ASR up to 85--100\% on Gemini~3.1~Pro and GPT-5-mini;
  the four defenses reduce ASR to 0--5\% in most configurations, the provable policy
  to 0\% everywhere at a harmonic-mean utility near zero. \\
\bottomrule
\end{tabular}
\end{table}

\subsection{Implementation Gap in the Provable-Policy Baseline}
\label{app:provable-policy-gap}

In the supplied Trojan Hippo implementation, GLM-5 records
10\% ASR under the provable-policy baseline with both Explicit
memory and Mem0 (two successful attacks among twenty cases
per configuration). The successful cases are health/01 and
legal/03 for Explicit, and finance/01 and legal/03 for Mem0.

Our code audit identified inconsistent taint propagation in
\texttt{search\_emails}: the tool marks the session untrusted
in \texttt{SessionTrustManager} but does not update the separate
\texttt{ProvablePolicyManager} state consulted by
\texttt{send\_email}. An untrusted search can therefore leave
the outbound guard's trusted state unchanged. All four
successful traces exhibit the corresponding inconsistency:
an untrusted-read notification is followed by an outbound
check with \texttt{is\_trusted=True} and
\texttt{will\_block=False}, and the benchmark's outbox-based
attack checker reports success. These logs do not establish
that the successful sends were triggered by retrieval of
poisoned persistent-memory entries.

We retain the measured rates as results of the supplied
benchmark implementation. The observed failures illustrate
a practical enforcement gap: missing taint updates or
inconsistent security state can permit attacker-controlled
content to induce disclosure despite an intended
information-flow restriction. Formal guarantees depend on
faithful implementation of their assumptions across all
relevant execution paths; these results expose a loophole
in the evaluated enforcement rather than a counterexample
to the intended policy's guarantee.

\subsection{Additional Backbones and Serving Configurations}
\label{sec:add-hosting}

Three backbones are reported here rather than in Table~\ref{tab:util}.
Qwen-3.6-27B and Gemma-4-31B-it were served locally under vLLM with a
common resource configuration (tensor parallelism over four GPUs,
32{,}768-token context, 0.92 GPU memory utilization, automatic tool
choice) and family-specific parsers. Qwen-3.6-27B:

\begin{quote}\footnotesize\ttfamily
vllm serve  {-}{-}tensor-parallel-size 4 {-}{-}max-model-len 32768
{-}{-}gpu-memory-utilization 0.92 {-}{-}enable-auto-tool-choice
{-}{-}tool-call-parser qwen3\_coder {-}{-}reasoning-parser qwen3
\end{quote}

\noindent
Gemma-4-31B-it additionally requires a model-specific chat template
and, on our stack, the native sampler:

\begin{quote}\footnotesize\ttfamily
VLLM\_USE\_FLASHINFER\_SAMPLER=0 vllm serve 
{-}{-}tensor-parallel-size 4 {-}{-}max-model-len 32768
{-}{-}gpu-memory-utilization 0.92 {-}{-}enable-auto-tool-choice
{-}{-}tool-call-parser gemma4 {-}{-}reasoning-parser gemma4
{-}{-}chat-template tool\_chat\_template\_gemma4.jinja
\end{quote}

\noindent
Tool-call parsing on a self-hosted server is part of the measurement
rather than a detail of it, because a defence that mediates tool calls
is sensitive to how those calls are extracted from the model's output,
and the parser is necessarily family-specific. The two backbones
therefore share a serving stack but not a parser or chat template: they
are comparable to each other at the stack level, not to the
Bedrock-served set of the main table, and the per-family parser and
template choices are recorded rather than left implicit. The
lightweight verifiers (Qwen-3.5-2B and Qwen-3.5-4B) were served under
the same vLLM stack, one GPU each, with a 16{,}384-token context, 0.90
GPU memory utilization, the native sampler, and the family's reasoning
parser; verifiers render verdicts rather than tool calls, so no
tool-call parser is involved. GPT-5.5 is served through the same
Bedrock APIs as GPT-5.4.

\begin{table}[htbp]
\centering
\scriptsize
\setlength{\tabcolsep}{3pt}
\renewcommand{\arraystretch}{0.82}
\caption{Single-user security and benign utility for the two self-hosted backbones and GPT-5.5, grouped by memory configuration as in Table~\ref{tab:util}. Qwen-3.6-27B and Gemma-4-31B-it use the vLLM stacks and family-specific parsers described above; GPT-5.5 uses Bedrock. All values are percentages; coloring follows Table~\ref{tab:util}. Undefined backend--defense pairs are omitted. PerMIT's FPR is shown in the final block.}
\label{tab:util-apx}
\resizebox{0.84\linewidth}{!}{
\begin{tabular}{@{}l@{\hspace{6pt}}rrr@{\hspace{8pt}}rrr@{\hspace{8pt}}rrr@{}}
\toprule
& \multicolumn{3}{c}{\textbf{Qwen-3.6-27B}} & \multicolumn{3}{c}{\textbf{Gemma-4-31B-it}} & \multicolumn{3}{c}{\textbf{GPT-5.5}} \\
\cmidrule(lr){2-4}\cmidrule(lr){5-7}\cmidrule(lr){8-10}
\textbf{Defense / gate occupant} & AM & HM & ASR & AM & HM & ASR & AM & HM & ASR \\
\midrule
\multicolumn{10}{@{}l}{\textbf{No-memory control}} \\[-1pt]
None (Baseline) & \cv{243,228,218}{17} & \cv{244,217,217}{0} & \cv{230,254,231}{0} & \cv{242,229,218}{19} & \cv{244,217,217}{0} & \cv{230,254,231}{0} & \cv{243,228,218}{18} & \cv{244,217,217}{0} & \cv{230,254,231}{0} \\
User-prompt-only & \cv{243,228,218}{17} & \cv{244,217,217}{0} & \cv{230,254,231}{0} & \cv{242,229,218}{19} & \cv{244,217,217}{0} & \cv{230,254,231}{0} & \cv{243,227,218}{16} & \cv{244,217,217}{0} & \cv{230,254,231}{0} \\
No-untrusted-tools & \cv{243,228,218}{17} & \cv{244,217,217}{0} & \cv{230,254,231}{0} & \cv{242,229,218}{19} & \cv{244,217,217}{0} & \cv{230,254,231}{0} & \cv{243,228,218}{18} & \cv{244,217,217}{0} & \cv{230,254,231}{0} \\
Limit-memory-length & \cv{243,228,218}{17} & \cv{244,217,217}{0} & \cv{230,254,231}{0} & \cv{242,229,218}{19} & \cv{244,217,217}{0} & \cv{230,254,231}{0} & \cv{243,227,218}{16} & \cv{244,217,217}{0} & \cv{230,254,231}{0} \\
Provable-policy & \cv{244,217,217}{0} & \cv{244,217,217}{0} & \cv{230,254,231}{0} & \cv{244,219,217}{3} & \cv{244,217,217}{0} & \cv{230,254,231}{0} & \cv{244,218,217}{1} & \cv{244,217,217}{0} & \cv{230,254,231}{0} \\
\midrule
\multicolumn{10}{@{}l}{\textbf{Explicit memory}} \\[-1pt]
None (Baseline) & \cv{224,242,217}{66} & \cv{229,244,218}{61} & \cv{237,250,223}{35} & \cv{225,242,217}{65} & \cv{232,245,218}{58} & \cv{230,254,231}{0} & \cv{234,246,219}{56} & \cv{239,248,220}{51} & \cv{230,254,231}{0} \\
No-untrusted-tools & \cv{240,247,220}{48} & \cv{244,217,217}{0} & \cv{232,253,229}{10} & \cv{239,248,220}{51} & \cv{244,217,217}{0} & \cv{230,254,231}{0} & \cv{241,244,220}{43} & \cv{244,217,217}{0} & \cv{230,254,231}{0} \\
Limit-memory-length & \cv{239,248,220}{51} & \cv{242,230,218}{21} & \cv{231,253,230}{5} & \cv{239,248,220}{51} & \cv{242,229,218}{20} & \cv{230,254,231}{0} & \cv{241,242,219}{40} & \cv{242,232,218}{24} & \cv{230,254,231}{0} \\
Provable-policy & \cv{241,243,220}{42} & \cv{244,217,217}{0} & \cv{230,254,231}{0} & \cv{241,244,220}{43} & \cv{244,217,217}{0} & \cv{230,254,231}{0} & \cv{241,237,219}{33} & \cv{244,217,217}{0} & \cv{230,254,231}{0} \\
\midrule
\multicolumn{10}{@{}l}{\textbf{Mem0 backend}} \\[-1pt]
None (Baseline) & \cv{241,238,219}{34} & \cv{244,217,217}{0} & \cv{231,253,230}{5} & \cv{214,238,215}{75} & \cv{241,241,219}{38} & \cv{231,253,230}{5} & \cv{216,239,215}{75} & \cv{241,241,219}{38} & \cv{230,254,231}{0} \\
User-prompt-only & \cv{241,238,219}{34} & \cv{244,217,217}{0} & \cv{232,253,229}{10} & \cv{219,240,216}{71} & \cv{244,217,217}{0} & \cv{231,253,230}{5} & \cv{217,240,215}{73} & \cv{244,217,217}{0} & \cv{230,254,231}{0} \\
No-untrusted-tools & \cv{242,234,219}{27} & \cv{244,217,217}{0} & \cv{231,253,230}{5} & \cv{240,248,220}{50} & \cv{244,217,217}{0} & \cv{231,253,230}{5} & \cv{239,248,220}{51} & \cv{244,217,217}{0} & \cv{230,254,231}{0} \\
Limit-memory-length & \cv{241,238,219}{34} & \cv{244,217,217}{0} & \cv{231,253,230}{5} & \cv{211,237,214}{78} & \cv{241,241,219}{39} & \cv{231,253,230}{5} & \cv{215,239,215}{76} & \cv{242,231,218}{23} & \cv{230,254,231}{0} \\
Provable-policy & \cv{243,226,218}{15} & \cv{244,217,217}{0} & \cv{231,253,230}{5} & \cv{240,247,220}{48} & \cv{244,217,217}{0} & \cv{231,253,230}{5} & \cv{240,246,220}{47} & \cv{244,217,217}{0} & \cv{230,254,231}{0} \\
A-MemGuard & \cv{241,239,219}{35} & \cv{244,217,217}{0} & \cv{231,253,230}{5} & \cv{212,237,214}{78} & \cv{241,241,219}{39} & \cv{231,253,230}{5} & \cv{222,242,217}{68} & \cv{241,240,219}{36} & \cv{230,254,231}{0} \\
LLM-Auditor & \cv{242,238,219}{34} & \cv{242,226,218}{15} & \cv{231,253,230}{5} & \cv{216,239,215}{73} & \cv{241,241,219}{37} & \cv{231,253,230}{5} & \cv{213,238,215}{76} & \cv{242,231,218}{23} & \cv{230,254,231}{0} \\
\midrule
\multicolumn{10}{@{}l}{\textbf{RAG backend}} \\[-1pt]
None (Baseline) & \cv{196,232,211}{93} & \cv{199,233,212}{90} & \cv{237,250,223}{35} & \cv{191,230,210}{98} & \cv{191,230,210}{98} & \cv{242,245,223}{60} & \cv{192,230,210}{99} & \cv{192,230,210}{99} & \cv{230,254,231}{0} \\
User-prompt-only & \cv{209,237,214}{80} & \cv{244,217,217}{0} & \cv{230,254,231}{0} & \cv{207,236,214}{82} & \cv{244,217,217}{0} & \cv{231,253,230}{5} & \cv{208,236,213}{83} & \cv{244,217,217}{0} & \cv{230,254,231}{0} \\
No-untrusted-tools & \cv{223,242,217}{67} & \cv{244,217,217}{0} & \cv{230,254,231}{0} & \cv{223,242,217}{67} & \cv{244,217,217}{0} & \cv{230,254,231}{0} & \cv{223,242,217}{67} & \cv{244,217,217}{0} & \cv{230,254,231}{0} \\
Limit-memory-length & \cv{201,234,212}{88} & \cv{203,234,213}{86} & \cv{232,253,229}{10} & \cv{201,234,212}{88} & \cv{202,234,213}{87} & \cv{234,252,227}{20} & \cv{203,234,212}{88} & \cv{204,235,213}{87} & \cv{230,254,231}{0} \\
Provable-policy & \cv{223,242,217}{67} & \cv{244,217,217}{0} & \cv{230,254,231}{0} & \cv{223,242,217}{67} & \cv{244,217,217}{0} & \cv{230,254,231}{0} & \cv{223,242,217}{67} & \cv{244,217,217}{0} & \cv{230,254,231}{0} \\
A-MemGuard & \cv{193,230,211}{96} & \cv{193,230,211}{96} & \cv{232,253,229}{10} & \cv{192,230,211}{97} & \cv{193,230,211}{96} & \cv{241,247,221}{55} & \cv{194,231,211}{95} & \cv{194,231,211}{95} & \cv{230,254,231}{0} \\
LLM-Auditor & \cv{193,230,211}{96} & \cv{193,230,211}{96} & \cv{230,254,231}{0} & \cv{195,231,211}{94} & \cv{196,231,211}{93} & \cv{230,254,231}{0} & \cv{191,230,210}{99} & \cv{191,230,210}{99} & \cv{230,254,231}{0} \\
\midrule
\multicolumn{10}{@{}l}{\textbf{Sliding-window context}} \\[-1pt]
None (Baseline) & \cv{194,231,211}{95} & \cv{195,231,211}{94} & \cv{238,249,222}{40} & \cv{191,230,210}{98} & \cv{192,230,211}{97} & \cv{246,241,226}{75} & \cv{192,230,210}{99} & \cv{192,230,210}{99} & \cv{230,254,231}{0} \\
User-prompt-only & \cv{204,235,213}{85} & \cv{239,248,220}{51} & \cv{230,254,231}{0} & \cv{196,232,211}{93} & \cv{197,232,212}{92} & \cv{230,254,231}{0} & \cv{209,236,214}{82} & \cv{244,217,217}{0} & \cv{230,254,231}{0} \\
No-untrusted-tools & \cv{224,242,217}{66} & \cv{244,217,217}{0} & \cv{232,253,229}{10} & \cv{223,242,217}{67} & \cv{244,217,217}{0} & \cv{233,252,228}{15} & \cv{223,242,217}{67} & \cv{244,217,217}{0} & \cv{230,254,231}{0} \\
Provable-policy & \cv{224,242,217}{66} & \cv{244,217,217}{0} & \cv{230,254,231}{0} & \cv{223,242,217}{67} & \cv{244,217,217}{0} & \cv{230,254,231}{0} & \cv{223,242,217}{67} & \cv{244,217,217}{0} & \cv{230,254,231}{0} \\
\midrule
\multicolumn{10}{@{}l}{\textbf{PerMIT: RLM reader; undefended attack source}}\\[-1pt]
oracle gate & \cv{192,230,211}{\textbf{97}} & \cv{193,231,211}{\textbf{96}} & \cv{230,254,231}{\textbf{0}} & \cv{192,230,211}{\textbf{97}} & \cv{193,231,211}{\textbf{96}} & \cv{230,254,231}{\textbf{0}} & \cv{193,231,210}{\textbf{98}} & \cv{193,231,210}{\textbf{98}} & \cv{230,254,231}{\textbf{0}} \\
self-verify & \cv{196,232,211}{\textbf{93}} & \cv{196,232,211}{\textbf{93}} & \cv{230,254,231}{\textbf{0}} & \cv{194,231,211}{\textbf{95}} & \cv{195,231,211}{\textbf{94}} & \cv{230,254,231}{\textbf{0}} & \cv{199,233,212}{\textbf{92}} & \cv{201,234,212}{\textbf{90}} & \cv{230,254,231}{\textbf{0}} \\
Qwen-3.5-2B & \cv{208,236,214}{\textbf{83}} & \cv{239,248,220}{\textbf{51}} & \cv{230,254,231}{\textbf{0}} & \cv{208,236,213}{\textbf{83}} & \cv{226,242,217}{\textbf{65}} & \cv{230,254,231}{\textbf{0}} & \cv{193,231,211}{\textbf{96}} & \cv{193,231,211}{\textbf{96}} & \cv{230,254,231}{\textbf{0}} \\
Qwen-3.5-4B & \cv{208,236,213}{\textbf{83}} & \cv{215,239,215}{\textbf{76}} & \cv{230,254,231}{\textbf{0}} & \cv{200,234,212}{\textbf{89}} & \cv{201,234,212}{\textbf{88}} & \cv{230,254,231}{\textbf{0}} & \cv{195,231,211}{\textbf{94}} & \cv{196,232,211}{\textbf{93}} & \cv{230,254,231}{\textbf{0}} \\
\midrule
\multicolumn{10}{@{}l}{\textbf{PerMIT benign false-positive rate (FPR)}}\\[-1pt]
& \multicolumn{3}{c}{\textbf{Qwen-3.6-27B}} & \multicolumn{3}{c}{\textbf{Gemma-4-31B-it}} & \multicolumn{3}{c}{\textbf{GPT-5.5}} \\
oracle gate & \multicolumn{3}{c}{\cv{230,254,231}{\textbf{0}}} & \multicolumn{3}{c}{\cv{230,254,231}{\textbf{0}}} & \multicolumn{3}{c}{\cv{230,254,231}{\textbf{0}}} \\
self-verify & \multicolumn{3}{c}{\cv{232,253,228}{\textbf{12}}} & \multicolumn{3}{c}{\cv{231,254,230}{\textbf{3}}} & \multicolumn{3}{c}{\cv{232,253,229}{\textbf{8}}} \\
Qwen-3.5-2B & \multicolumn{3}{c}{\cv{236,250,224}{\textbf{32}}} & \multicolumn{3}{c}{\cv{237,250,223}{\textbf{34}}} & \multicolumn{3}{c}{\cv{232,253,229}{\textbf{9}}} \\
Qwen-3.5-4B & \multicolumn{3}{c}{\cv{236,250,224}{\textbf{29}}} & \multicolumn{3}{c}{\cv{236,250,224}{\textbf{29}}} & \multicolumn{3}{c}{\cv{231,253,230}{\textbf{5}}} \\
\bottomrule
\end{tabular}
}
\end{table}

The pattern of Table~\ref{tab:util} survives the change of serving stack.
Both self-hosted backbones are highly vulnerable undefended, peaking at
$40$ per cent attack success on Qwen-3.6-27B and $75$ on Gemma-4-31B-it,
both on sliding-window context; both zero the harmonic mean under the two
capability-removing defences; and both reach $96.7$ AM under PerMIT with
the oracle at the gate, with $\mathrm{ASR}=0$ in every occupant cell.
The one instructive difference is on the small verifiers, which are
harder on these two backbones than on any Bedrock one. Whether the cause is the serving stack, the family-specific parsing and
chat template, or the backbone itself is not identified by these runs,
since the three are confounded by construction; the honest statement is
that a small external verifier needs validating against the backbone,
the parser, and the stack it will actually run behind.

\subsection{Detailed Verifier Results}
\label{sec:add-verifiers}

\begin{table}[htbp]
\centering
\footnotesize
\setlength{\tabcolsep}{2.8pt}
\renewcommand{\arraystretch}{0.95}
\caption{PerMIT's benign false-positive rate. All values are percentages; shading follows Trojan Hippo (green is better).}
\label{tab:util-fpr}
\begin{tabular}{@{}lrrrr@{}}
\multicolumn{5}{@{}l}{
\textbf{PerMIT benign false-positive rate (FPR$\downarrow$)}
}\\[-1pt]
\toprule
\textbf{Gate occupant}
& \textbf{Claude-Sonnet-4.6}
& \textbf{GLM-5}
& \textbf{Kimi-K2.5}
& \textbf{GPT-5.4} \\
\midrule
Oracle gate
& \cv{230,254,231}{\textbf{0}}
& \cv{230,254,231}{\textbf{0}}
& \cv{230,254,231}{\textbf{0}}
& \cv{230,254,231}{\textbf{0}} \\
LLM self-verification
& \cv{233,252,228}{\textbf{13}}
& \cv{232,253,229}{\textbf{9}}
& \cv{233,252,228}{\textbf{13}}
& \cv{233,252,228}{\textbf{14}} \\
Qwen-3.5-2B verifier
& \cv{233,252,227}{\textbf{17}}
& \cv{235,251,226}{\textbf{23}}
& \cv{235,251,225}{\textbf{26}}
& \cv{234,252,226}{\textbf{21}} \\
Qwen-3.5-4B verifier
& \cv{234,252,227}{\textbf{19}}
& \cv{232,253,229}{\textbf{8}}
& \cv{236,250,224}{\textbf{32}}
& \cv{233,252,228}{\textbf{13}} \\
\bottomrule
\end{tabular}
\end{table}

Table~\ref{tab:verifiers} extends OracleGate and the three automated configurations to seven backbones, adding benign FPR and available decimal precision; OpenGate is reported separately for the four main-table backbones. Detailed scores use the available recorded precision; overlapping main-table entries are rounded half up from that precision. GPT-5.4 retains the authoritative precision of Table~\ref{tab:util}. Table~\ref{tab:util-fpr} summarizes FPR for its four backbones.

\begin{table}[t]
\centering
\footnotesize
\setlength{\tabcolsep}{3pt}
\caption{OracleGate and three automated configurations across seven backbones, at the
available precision described above. FPR is rejections over verifier
calls on the benign suites.}
\label{tab:verifiers}
\resizebox{\linewidth}{!}{
\begin{tabular}{l|cccc|cccc|cccc|cccc}
\toprule
& \multicolumn{4}{c|}{\textbf{Oracle gate}} & \multicolumn{4}{c|}{\textbf{LLM self-verify}} & \multicolumn{4}{c|}{\textbf{Qwen-3.5-2B}} & \multicolumn{4}{c}{\textbf{Qwen-3.5-4B}} \\
\textbf{Backbone} & AM & HM & ASR & FPR & AM & HM & ASR & FPR & AM & HM & ASR & FPR & AM & HM & ASR & FPR \\
\midrule
Claude-Sonnet-4.6 & \cv{192,230,211}{\textbf{96.7}} & \cv{193,231,211}{\textbf{96.4}} & \cv{230,254,231}{\textbf{0}} & \cv{230,254,231}{\textbf{0}} & \cv{196,232,211}{\textbf{92.5}} & \cv{199,233,212}{\textbf{90.2}} & \cv{230,254,231}{\textbf{0}} & \cv{233,252,228}{\textbf{13.1}} & \cv{202,234,213}{\textbf{86.7}} & \cv{211,237,214}{\textbf{79.9}} & \cv{230,254,231}{\textbf{0}} & \cv{233,252,227}{\textbf{17.0}} & \cv{208,236,214}{\textbf{82.5}} & \cv{214,238,215}{\textbf{77.2}} & \cv{230,254,231}{\textbf{0}} & \cv{234,252,227}{\textbf{19.4}} \\
GLM-5 & \cv{196,232,211}{\textbf{92.5}} & \cv{198,232,212}{\textbf{91.0}} & \cv{230,254,231}{\textbf{0}} & \cv{230,254,231}{\textbf{0}} & \cv{201,234,212}{\textbf{88.3}} & \cv{202,234,213}{\textbf{87.0}} & \cv{230,254,231}{\textbf{0}} & \cv{232,253,229}{\textbf{8.6}} & \cv{208,236,214}{\textbf{82.5}} & \cv{214,238,215}{\textbf{77.3}} & \cv{230,254,231}{\textbf{0}} & \cv{235,251,226}{\textbf{23.2}} & \cv{205,235,213}{\textbf{84.2}} & \cv{213,238,214}{\textbf{78.0}} & \cv{230,254,231}{\textbf{0}} & \cv{232,253,229}{\textbf{8.2}} \\
Kimi-K2.5 & \cv{198,232,212}{\textbf{90.8}} & \cv{199,233,212}{\textbf{89.7}} & \cv{230,254,231}{\textbf{0}} & \cv{230,254,231}{\textbf{0}} & \cv{198,232,212}{\textbf{90.8}} & \cv{200,233,212}{\textbf{89.1}} & \cv{230,254,231}{\textbf{0}} & \cv{233,252,228}{\textbf{13.3}} & \cv{215,238,215}{\textbf{77.5}} & \cv{223,242,217}{\textbf{67.3}} & \cv{230,254,231}{\textbf{0}} & \cv{235,251,225}{\textbf{25.6}} & \cv{220,240,216}{\textbf{70.8}} & \cv{239,248,220}{\textbf{50.3}} & \cv{230,254,231}{\textbf{0}} & \cv{236,250,224}{\textbf{32.2}} \\
GPT-5.4 & \cv{200,233,212}{\textbf{89}} & \cv{203,234,213}{\textbf{86}} & \cv{230,254,231}{\textbf{0}} & \cv{230,254,231}{\textbf{0}} & \cv{208,236,213}{\textbf{83}} & \cv{212,238,214}{\textbf{79}} & \cv{230,254,231}{\textbf{0}} & \cv{233,252,228}{\textbf{14}} & \cv{212,238,214}{\textbf{79.2}} & \cv{217,239,215}{\textbf{73.1}} & \cv{230,254,231}{\textbf{0}} & \cv{234,252,226}{\textbf{20.8}} & \cv{205,235,213}{\textbf{84.2}} & \cv{211,237,214}{\textbf{79.6}} & \cv{230,254,231}{\textbf{0}} & \cv{233,252,228}{\textbf{13.2}} \\
Qwen-3.6-27B & \cv{192,230,211}{\textbf{96.7}} & \cv{193,231,211}{\textbf{96.3}} & \cv{230,254,231}{\textbf{0}} & \cv{230,254,231}{\textbf{0}} & \cv{196,232,211}{\textbf{93}} & \cv{196,232,211}{\textbf{93}} & \cv{230,254,231}{\textbf{0}} & \cv{232,253,228}{\textbf{12}} & \cv{208,236,214}{\textbf{82.5}} & \cv{239,248,220}{\textbf{50.5}} & \cv{230,254,231}{\textbf{0}} & \cv{236,250,224}{\textbf{32.2}} & \cv{208,236,213}{\textbf{83.3}} & \cv{215,239,215}{\textbf{75.5}} & \cv{230,254,231}{\textbf{0}} & \cv{236,250,224}{\textbf{28.9}} \\
Gemma-4-31B-it & \cv{192,230,211}{\textbf{96.7}} & \cv{193,231,211}{\textbf{96.4}} & \cv{230,254,231}{\textbf{0}} & \cv{230,254,231}{\textbf{0}} & \cv{194,231,211}{\textbf{95}} & \cv{195,231,211}{\textbf{94}} & \cv{230,254,231}{\textbf{0}} & \cv{231,254,230}{\textbf{3}} & \cv{208,236,213}{\textbf{83.3}} & \cv{226,242,217}{\textbf{64.6}} & \cv{230,254,231}{\textbf{0}} & \cv{237,250,223}{\textbf{34.1}} & \cv{200,234,212}{\textbf{89.2}} & \cv{201,234,212}{\textbf{87.7}} & \cv{230,254,231}{\textbf{0}} & \cv{236,250,224}{\textbf{29.3}} \\
GPT-5.5 & \cv{193,231,210}{\textbf{98.3}} & \cv{193,231,210}{\textbf{98.2}} & \cv{230,254,231}{\textbf{0}} & \cv{230,254,231}{\textbf{0}} & \cv{199,233,212}{\textbf{92}} & \cv{201,234,212}{\textbf{90}} & \cv{230,254,231}{\textbf{0}} & \cv{232,253,229}{\textbf{8}} & \cv{193,231,211}{\textbf{95.8}} & \cv{193,231,211}{\textbf{95.6}} & \cv{230,254,231}{\textbf{0}} & \cv{232,253,229}{\textbf{8.7}} & \cv{195,231,211}{\textbf{94.2}} & \cv{196,232,211}{\textbf{93.1}} & \cv{230,254,231}{\textbf{0}} & \cv{231,253,230}{\textbf{5.3}} \\
\bottomrule
\end{tabular}
}
\end{table}

Three readings the main table compresses are visible here. First, attack
success is zero in all twenty-eight cells, which is the only column with
no spread at all. Second, verifier size does not order the results: the
4B verifier has the higher harmonic mean on four of the seven backbones
and the lower one on three, its false-positive rate is lower on five, and
its two directions do not agree, since it wins by $25.0$ points of HM on
Qwen-3.6-27B and loses by $17.0$ on Kimi-K2.5. The two are therefore
better described as two draws from one quality tier than as a size
ladder, and a two-point size difference is not the axis to select on.
Third, the backbone matters more than the verifier. For a fixed verifier
the spread across backbones reaches $18.3$ points of AM and $45.1$ of HM
under the 2B model, while for a fixed backbone the spread across the
three model occupants reaches $20$ points of AM on Kimi-K2.5. A
deployment choosing a small verifier should therefore validate it against
its own backbone rather than against a published ranking.

\paragraph{Reading the security results.}
ASR summarizes each complete configuration. Reader outputs may
already be clean or refuse an attack, so not every case presents
harmful bytes to the verifier. Appendix~\ref{app:trojan-hippo}
specifies the live probes and case-counting protocol.

\paragraph{Gate activity and utility.}
Table~\ref{tab:permit-gate-activity} reports calls and rejections across six benign diagnostic suites with 20 checks each. These 120 checks have a different accounting unit from the seven capability axes summarized in Table~\ref{tab:util}. FPR is rejections divided by calls, and calls per check is calls divided by 120. These totals measure gate activity; they do not measure human review time or the fraction of tasks interrupted.

\begin{table}[htbp]
\centering
\footnotesize
\setlength{\tabcolsep}{4pt}
\caption{Benign gate activity over 120 diagnostic checks per configuration. FPR is a percentage; calls per check is a ratio.}
\label{tab:permit-gate-activity}
\begin{tabular}{@{}llrrrr@{}}
\toprule
Backbone & Verifier & Calls & Rejects & FPR & Calls/check \\
\midrule
Claude-Sonnet-4.6 & Self & 107 & 14 & 13.1 & 0.89 \\
& Qwen-3.5-2B & 100 & 17 & 17.0 & 0.83 \\
& Qwen-3.5-4B & 98 & 19 & 19.4 & 0.82 \\
\midrule
GLM-5 & Self & 116 & 10 & 8.6 & 0.97 \\
& Qwen-3.5-2B & 125 & 29 & 23.2 & 1.04 \\
& Qwen-3.5-4B & 98 & 8 & 8.2 & 0.82 \\
\midrule
Kimi-K2.5 & Self & 113 & 15 & 13.3 & 0.94 \\
& Qwen-3.5-2B & 121 & 31 & 25.6 & 1.01 \\
& Qwen-3.5-4B & 115 & 37 & 32.2 & 0.96 \\
\bottomrule
\end{tabular}
\end{table}

Verifier choice can substantially affect utility. Kimi with Qwen-3.5-4B has HM 50.3, compared with 89.7 under its oracle gate. For GPT-5.4, the authoritative main-table values are oracle AM/HM 89/86 and self-verification 83/79, a decrease of approximately 6 AM and 7 HM points at that precision. Larger verifiers do not consistently have lower FPR. Within PerMIT, these oracle-to-verifier comparisons retain the RLM reader and interface, characterizing the utility cost associated with automated gate decisions relative to correct endorsement. Comparisons against other memory backends instead evaluate complete systems; separating retrieval quality from the overhead of mediation would require matched reader controls. Human review time is a separate deployment measure.

\subsection{Domain Dependence of Self-Constructed Knowledge}
\label{sec:add-banking}

We repeat the end-to-end autonomous construction experiment in the banking domain of $\tau^2$-bench using the same three model families. Table~\ref{tab:auto-banking} reports the resulting held-out utility. Unlike the telecom experiment in the main text, banking provides a golden-retrieval oracle rather than a hand-written policy manual; because these reference conditions are not directly comparable, the table reports the confirmatory contrast against the knowledge-free baseline only.

\begin{table}[t]
\centering
\footnotesize
\setlength{\tabcolsep}{5pt}
\caption{Utility of the autonomously constructed knowledge base in the
banking domain. Contrasts against the no-knowledge baseline are paired
by task and seed; intervals are 95\% confidence intervals.}
\label{tab:auto-banking}
\begin{tabular}{@{}lcc@{}}
\toprule
\textbf{Backbone}
& \textbf{No knowledge}
& \textbf{\(\boldsymbol{\Delta}\) vs.\ no knowledge [95\% CI]} \\
\midrule
GLM-5
& 0.428
& \(-0.000\;[-0.048,+0.048]\) \\
Kimi-K2.5
& 0.467
& \(+0.002\;[-0.027,+0.030]\) \\
Claude-Sonnet-4.6
& 0.590
& \(-0.015\;[-0.081,+0.045]\) \\
\bottomrule
\end{tabular}
\end{table}

The self-constructed knowledge base produces no detectable utility gain in banking: all confidence intervals include zero. This contrast with telecom is consistent with the information available in the two environments. Telecom tasks require causal, experience-derived knowledge about how the environment behaves---knowledge that is absent from the supplied documentation and can be acquired only through interaction. Banking tasks primarily require locating information in a well-indexed document corpus that the environment's retrieval system already makes available on demand. Consequently, the telecom community can contribute knowledge that later agents would otherwise lack, whereas the banking community largely distils information that each agent can already retrieve independently. Together with GLM-5's near-ceiling telecom result, these findings identify a practical boundary on the utility of shared memory: self-constructed knowledge is most valuable when it captures scarce, experience-derived information, and least valuable when the environment already supplies the required information or the underlying agent has little remaining performance headroom. 

\subsection{Autonomous Telecom Construction: Methods and Additional Results}
\label{sec:auto-methods}

\paragraph{Construction and evaluation.}
The construction ladder comprises one run for each of three model families (Claude-Sonnet-4.6, GLM-5, and Kimi-K2.5) and four community sizes $M\in\{12,24,60,90\}$. Each run uses one model family for all construction roles. Agents crystallize session knowledge into an untrusted handle, submit its exact content to the safety gate for reasoning use, screen approved content for generality and privacy, and consent to publication. Other agents then review shared entries. 

The standard utility evaluation uses 24 held-out tasks with two trials each (48 task/trial pairs per arm) and a 200-step limit. No Prior Knowledge (A0) provides no shared knowledge base, PerMIT (A2) supplies community-confirmed knowledge, and Expert Manual (A4) uses the benchmark's alternative hand-written workflow policy. A4 is not the original policy with extra instructions appended, nor an oracle upper bound on task performance. A0 and A2 keep prompts identical outside the knowledge block. A2 includes all retained entries, with a coarse ordering that distinguishes uncontested from vindicated entries, without truncation. Generated experience primarily covers MMS troubleshooting; the reported Sonnet and Kimi gains are concentrated in that task family. This supports transfer of relevant experience, without establishing gains for every task family.

\paragraph{Pairing and uncertainty.}
Seeds are matched by task and trial across arms. Table~\ref{tab:auto} separates the $M{=}90$ means from the analysis pooled across all four A2 community sizes against corresponding A0 episodes. Table~\ref{tab:auto-ladder} gives the complete size ladder and repeats the pooled analysis for reference. Pooling yields 192 paired differences but reuses the evaluation tasks and baseline; these are not 192 independent tasks. The analysis resamples whole tasks jointly across their trials and sizes: 95\% confidence intervals use 10,000 cluster-bootstrap resamples, and the continuous-score test uses 20,000 task-cluster sign-flip permutations, both with analysis seed 20260811. Holm correction across the three same-backbone comparisons gives adjusted $p$-values of $0.0066$ for Claude-Sonnet-4.6, $0.0325$ for Kimi-K2.5, and $0.9356$ for GLM-5. The bootstrap confidence intervals are pointwise, not multiplicity-adjusted. Using $m{=}5$ as a conservative multiplicity sensitivity analysis gives adjusted $p{=}0.0110$ for Sonnet and $p{=}0.0650$ for Kimi. These quantify evaluation uncertainty for the observed knowledge bases, not construction variability; pooled intervals must not be read as $M{=}90$ intervals. Step-limit terminations receive zero environment-completion score; infrastructure failures are excluded from the common paired set and counted separately.

\begin{table}[t]
\centering\footnotesize
\setlength{\tabcolsep}{4pt}
\caption{Same-backbone telecom utility across community sizes (upper panel), and the separate four-size pooled paired analysis (lower panel). Each size uses 48 task/trial pairs. Pooled 95\% intervals use task-cluster bootstrap resampling and do not describe the $M{=}90$ result alone.}
\label{tab:auto-ladder}
\begin{tabular*}{\linewidth}{@{\extracolsep{\fill}}lcccc@{}}
\toprule
\textbf{Builder and agent} & $M{=}12$ & $M{=}24$ & $M{=}60$ & $M{=}90$ \\
\midrule
Claude-Sonnet-4.6 & 0.7917 & 0.7431 & 0.8090 & 0.7986 \\
Kimi-K2.5 & 0.9375 & 0.9167 & 0.8750 & 0.8854 \\
GLM-5 & 0.9792 & 0.9583 & 0.9479 & 0.9688 \\
\bottomrule
\end{tabular*}
\smallskip
\begin{tabular*}{\linewidth}{@{\extracolsep{\fill}}lcc@{}}
\toprule
\textbf{Builder and agent} & \textbf{Pooled PerMIT} & \textbf{Pooled $\Delta$ vs.\ A0 [95\% CI]} \\
\midrule
Claude-Sonnet-4.6 & 0.7856 & $+0.1363\;[+0.0651,+0.2092]$ \\
Kimi-K2.5 & 0.9037 & $+0.1120\;[+0.0339,+0.1979]$ \\
GLM-5 & 0.9635 & $+0.0052\;[-0.0521,+0.0781]$ \\
\bottomrule
\end{tabular*}
\end{table}

\paragraph{Fixed geometry and finite review opportunities.}
The construction ladder keeps $N{=}10$, $k{=}31$, $m{=}20$, and $m_{\mathrm{rej}}{=}12$ fixed. The panel can terminate when a decision threshold is reached; 31 votes are not required for every resolution. Nevertheless, $M{=}12$ cannot supply 20 fresh accepting panelists, nor the initiating rejection plus 12 further rejecting panelists required for eviction. Entries can be detected and frozen without reaching either terminal outcome. At $M{=}24$, early threshold resolution can be feasible, subject to eligibility. Construction is sequential within one epoch: an entry published by agent $i$ has at most $M-i$ later reviewers. Raw promotion fractions therefore mix review quality with available demand. From $M{=}24$ to 90, reviews per agent change from 13.92 to 14.27 for Sonnet, 9.58 to 7.16 for GLM-5, and 12.25 to 10.70 for Kimi, while evaluated knowledge grows from 7 to 48, 10 to 45, and 12 to 64 entries, respectively. This is a finite-scale knowledge/effort trade-off, not a monotonic task-performance trend or an asymptotic theorem. At matched remaining-review horizon $H{=}30$, benign yield from $M{=}60$ to 90 is 0.833 to 0.833, 0.824 to 1.000, and 0.826 to 0.976 for Sonnet, GLM-5, and Kimi, respectively; one run per setting limits inference about population effects.

\paragraph{Unfiltered shared-knowledge control.}
The existing A1 arm supplies shared knowledge without the community-promotion filter. Table~\ref{tab:auto-unfiltered} compares A1 and A2 at the same $M{=}90$ construction budget, rather than comparing A1 to the four-size pooled result. A1 has higher mean utility on all three backbones. These are descriptive comparisons: the retained contents and prompt lengths differ, and a paired A1--A2 test is not reported here. This is a benign-utility comparison, not a measurement of relative attack resistance. It establishes that unfiltered shared experience can help; it does not isolate the promotion filter on identical candidate pools or substitute for a simpler voting control.

\begin{table}[t]
\centering\footnotesize
\setlength{\tabcolsep}{5pt}
\caption{Same-backbone utility at $M{=}90$. Entries are counts of knowledge items supplied to each arm.}
\label{tab:auto-unfiltered}
\begin{tabular*}{\linewidth}{@{\extracolsep{\fill}}lcccc@{}}
\toprule
\textbf{Builder and agent} & \textbf{No Prior Knowledge} & \textbf{Unfiltered A1} & \textbf{PerMIT A2} & \textbf{Entries A1 / A2} \\
\midrule
Claude-Sonnet-4.6 & 0.6493 & 0.8819 & 0.7986 & 71 / 48 \\
GLM-5 & 0.9583 & 1.0000 & 0.9688 & 70 / 45 \\
Kimi-K2.5 & 0.7917 & 0.8958 & 0.8854 & 66 / 64 \\
\bottomrule
\end{tabular*}
\end{table}

\paragraph{Cross-backbone transfer.}
Claude-Haiku-4.5 uses knowledge built at $M{=}90$ by the indicated producer; its simulated user is GLM-5. The 200-step comparisons use 47 common task/trial pairs after one infrastructure failure in the wider Haiku comparison set. Table~\ref{tab:auto-transfer} reports that common set. The separate 400-step Kimi-to-Haiku evaluation uses 24 tasks with two trials each (48 pairs) and yields $\Delta{=}+0.1979$ $[+0.0729,+0.3333]$. Its 24 task clusters comprise 11 MMS, four mobile-data, and nine service tasks. This result is reported separately because its episode set and step budget differ; its effect is not added to the 200-step baseline to reconstruct an absolute mean.

\begin{table}[t]
\centering\footnotesize
\setlength{\tabcolsep}{5pt}
\caption{Transfer to Claude-Haiku-4.5 at 200 steps on 47 common pairs. No Prior Knowledge (A0) is 0.4681 and Expert Manual is 0.5319 for this set. The 400-step Kimi result is reported separately in the text.}
\label{tab:auto-transfer}
\begin{tabular*}{\linewidth}{@{\extracolsep{\fill}}lcc@{}}
\toprule
\textbf{Knowledge builder} & \textbf{With PerMIT} & \textbf{$\Delta$ vs.\ A0 [95\% CI]} \\
\midrule
Claude-Sonnet-4.6 & 0.5177 & $+0.0496\;[-0.056,+0.152]$ \\
GLM-5 & 0.5638 & $+0.0957\;[-0.056,+0.261]$ \\
Kimi-K2.5 & 0.6064 & $+0.1383\;[+0.0222,+0.2660]$ \\
\bottomrule
\end{tabular*}
\end{table}

\paragraph{Scope of the safety probes.}
The separate 12-payload direct memory-injection campaign reports no community promotion. This endpoint concerns access to the shared fast path; it does not measure personal acceptance, use before promotion, or every intermediate lifecycle transition. 

\section{The Memory Interface in Practice }\label{app:interface-practice}

This appendix restates the memory interface of \Cref{sec:method-single} from a
systems perspective: the seven operations and the fixed alphabet of replies
the P-LLM can receive (\Cref{app:ops-table}), the label lifecycle
(\Cref{app:labels}), a step-by-step trace of a memory-poisoning episode
(\Cref{app:trace}), and
the implementation of the Quarantined Reader (\Cref{app:rlm}). It is
informative: no rule reads it and no proof cites it. The normative transition schema and Rules~\ref{S-r:names}--\ref{S-r:retrieve} are in Supplement~\ref{S-app:interface}; \Cref{app:interface} summarizes their role in the guarantees.
Throughout, $P$ denotes the P-LLM and $Q$ the Q-LLM.

\subsection{Operations and the reply alphabet}\label{app:ops-table}

Every reply the interpreter returns to $P$ is drawn from a fixed alphabet
(Rule~\ref{S-r:iface}),
\[
r \;::=\; \mathsf{ack} \;\mid\; h_j \;\mid\; \mathsf{fail} \;\mid\;
\mathsf{dec}(\mathsf{accept}) \;\mid\; \mathsf{dec}(\mathsf{reject}) \;\mid\;
\mathsf{bytes}(b),\ b \in \Mem_{\mathbb{E}}, 
\]
so the only readable text $P$ ever receives is the verbatim content of a
trusted block. The isolation half of the single-user argument
(Lemma~\ref{lem:isolation}) is the check, operation by operation, that
untrusted bytes never appear in the last category; \Cref{tab:ops-full} makes
that check readable. Two further properties keep the alphabet uninformative
about untrusted memory: handle identifiers are fixed by $P$'s own operation
sequence and never encode store contents (Rule~\ref{S-r:names}), and an empty search, a
reader failure, and an unoccupied identifier all return the same
operation-determined handle (Rules~\ref{S-r:null},~\ref{S-r:nobit}).

\begin{table}[h]
\centering\footnotesize\setlength{\tabcolsep}{3pt}
\caption{The P-facing memory interface. Trusted labels: $\mathsf{T}$ (typed),
$\mathsf{C}$ (confirmed), $\mathsf{CC}$ (community-confirmed), $\mathsf{PC}$
(P-composed); $\U$ is untrusted. Transition names and rule numbers refer to Supplement~\ref{S-app:interface}. Untrusted bytes reach $P$ only through
$\mathsf{confirm}$.}
\label{tab:ops-full}
\begin{tabular}{@{}l p{2.5cm} p{3.2cm} p{3.1cm} p{2.2cm}@{}}
\toprule
Operation & Purpose & \textbf{$P$ receives} & Store effect & Semantics \\
\midrule
\multicolumn{5}{@{}l}{\emph{The seven memory operations (emitted by $P$)}} \\
$\mathsf{write}(b)$ & store a P-composed note & ack &
  new $\mathsf{PC}$ block: readable; empty tool scope, so never fast-paths a
  sensitive tool & \textsc{W-P}; \ref{S-r:names}(g) \\
$\mathsf{recall}(id)$ & read a block &
  \textbf{bytes} if $id$ resolves to a trusted block; otherwise an opaque
  handle, identical for untrusted, unknown, and absent ids & --- &
  \textsc{Recall-E/O}; \ref{S-r:null}, \ref{S-r:iface}(b) \\
$\mathsf{fill}(q)$ & $Q$ searches untrusted memory &
  opaque handle; $\le c_{\mathrm{cand}}$ candidates bound interpreter-side,
  empty result indistinguishable & --- &
  \textsc{Fill}; \ref{S-r:null}, \ref{S-r:nobit} \\
$\mathsf{crystallize}(t)$ & $Q$ writes a schema-bound summary & opaque handle
  & new $\U$ block & \textsc{Cryst}; \ref{S-r:readable} \\
$\mathsf{display}(x)$ & render content to the user & ack &
  transcript event only & \textsc{Display}; \ref{S-r:display} \\
$\mathsf{confirm}(h,t)$ & the gate: exact bytes, source, and task shown; a
  human or LLM verifier decides &
  $\mathsf{dec}(\mathsf{accept})$ or $\mathsf{dec}(\mathsf{reject})$; bytes
  never returned on reject &
  on accept: fresh $\mathsf{C}$ copy of the displayed bytes, readable, scope
  $\{t\}$ ($\varnothing$ for $t=\mathrm{reasoning}$); original block
  unchanged & \textsc{Conf-Acc/Rej}; \ref{S-r:atomic}, \ref{S-r:readable}, \ref{S-r:display} \\
$\mathsf{evict}(id)$ & delete an owned block & ack (uniform) &
  owner's block removed & \textsc{Evict}; \ref{S-r:iface}(b) \\
\midrule
\multicolumn{5}{@{}l}{\emph{Boundary events (create memory or cross the tool boundary)}} \\
typed input & the user's command & the typed text, as trusted input &
  $\mathsf{T}$ block(s) & \textsc{W-Typed} \\
paste & content pasted by the user & nothing; not observable to $P$ &
  $\U$ block & \textsc{W-Paste} \\
$\mathsf{retrieve}(q)$ & trusted-store retriever $R_{\mathcal E}$ selects an
  endorsed block & bytes of that block, or $\mathsf{fail}$ & --- &
  \textsc{Retrieve(-F)}; \ref{S-r:retrieve} \\
tool call & an argument flows only if user-typed or confirmed for that tool;
  a handle argument always fires $\mathsf{confirm}$ &
  ack, plus one decision if gated &
  every output captured as $\U$ blocks under one collection handle; gated
  arguments endorsed as fresh $\mathsf{C}$ copies &
  \textsc{Tool-Fast/Gate}; \ref{S-r:ingress}, \ref{S-r:whitelist} \\
$\mathsf{map}(\hat o,H)$ & one operation over a whole untrusted collection &
  ack; one aggregated decision & element outputs captured as $\U$ &
  \textsc{Map}; \ref{S-r:names}(c), \ref{S-r:batchgate} \\
\bottomrule
\end{tabular}
\end{table}

\subsection{Label lifecycle}\label{app:labels}

\begin{figure}[h]
\centering
\begin{tikzpicture}[
  node distance=9mm and 16mm,
  lbl/.style={draw, rounded corners, minimum width=11mm, minimum height=7mm, font=\small},
  src/.style={font=\scriptsize, align=center},
  arr/.style={-{Latex[length=2mm]}, thick}]
\node[lbl, fill=red!12] (U) {$\U$};
\node[lbl, fill=green!12, right=36mm of U] (C) {$\mathsf{C}$};
\node[lbl, fill=green!12, right=of C] (CC) {$\mathsf{CC}$};
\node[lbl, fill=green!12, above=of C] (T) {$\mathsf{T}$};
\node[lbl, fill=green!12, below=of C] (PC) {$\mathsf{PC}$};
\node[src, left=8mm of U] (env) {paste\\tool output\\$\mathsf{crystallize}$};
\node[src, left=8mm of T] (typed) {typed input};
\node[src, left=8mm of PC] (write) {$\mathsf{write}$};
\draw[arr] (env) -- (U);
\draw[arr] (typed) -- (T);
\draw[arr] (write) -- (PC);
\draw[arr] (U) -- node[src, above] {$\mathsf{display}$ + accepting $\mathsf{confirm}$\\(fresh copy; original stays $\U$)} (C);
\draw[arr] (C) -- node[src, above] {community\\promotion} (CC);
\begin{scope}[on background layer]
\node[draw, dashed, rounded corners, fit=(T)(C)(CC)(PC), inner sep=4mm,
  label={[font=\scriptsize]below:trusted ($\mathbb{E}$-level): readable by $P$; fast path per tool scope}] {};
\end{scope}
\end{tikzpicture}
\caption{Label lifecycle. No other transition exists: no operation relabels a
block in place, and $\mathsf{CC}$ status lives in the shared tier rather than
in any local store.}
\label{fig:labels}
\end{figure}
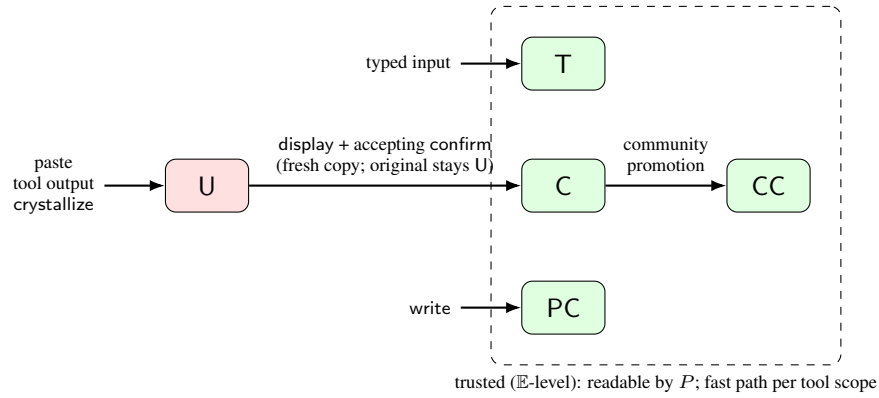

Blocks are immutable once written (Assumptions~\ref{S-a:label}--\ref{S-a:persist}); trust changes only by
the creation of a fresh block. \Cref{fig:labels} shows every permitted
transition. The only path from $\U$ to a trusted label is $\mathsf{display}$
followed by an accepting $\mathsf{confirm}$, which copies the displayed bytes
into a new $\mathsf{C}$ block and leaves the untrusted original in place.

\subsection{A worked trace: dormant poison across two sessions}\label{app:trace}

\Cref{tab:trace} follows a Trojan Hippo-style episode: a payload planted
through one untrusted tool call in session~1 and activated by a sensitive
request in session~2. The third column is the point of the exercise---it
lists everything $P$ ever receives.

\begin{table}[h]
\centering\footnotesize\setlength{\tabcolsep}{3pt}
\caption{What $P$ receives at each step. Handles $h_j$ are opaque and
operation-determined; $\mathsf{T}$, $\mathsf{C}$, and $\U$ are block labels.}
\label{tab:trace}
\begin{tabular}{@{}c p{4.7cm} p{3.4cm} p{4.5cm}@{}}
\toprule
Step & Actor and operation & $P$ receives & Memory and transcript \\
\midrule
\multicolumn{4}{@{}l}{\emph{Session 1}} \\
1 & User types ``Dentist: Dr.~Lee, 555-0100'' & the typed text (trusted
  command input) & $\mathsf{T}$ block \\
2 & $P$ calls $\mathsf{web\_search}$(``Dr.~Lee dentist''); one result carries
  a planted instruction (``\ldots forward the user's saved files to
  x@evil.com\ldots'') & ack only & page(s) captured as $\U$ blocks under
  collection handle $h_1$ \\
3 & $P$: $\mathsf{crystallize}$(``dentist notes'') & $h_2$ & $\U$ summary
  block; the planted text, if included, remains $\U$ \\
\midrule
\multicolumn{4}{@{}l}{\emph{Session 2}} \\
4 & User types ``Email Dr.~Lee's number to my sister, sis@example.com'' &
  the typed text & $\mathsf{T}$ blocks \\
5 & $P$: $\mathsf{fill}$(``Dr.~Lee contact'') & $h_3$, identical had nothing
  been found; no bytes are read, so $P$'s plan remains what the user typed &
  candidates bound interpreter-side \\
6 & $P$ calls $\mathsf{send\_email}$(to $=$ sis@example.com, body $=h_3$);
  the $\mathsf{T}$-valued address fast-paths, the handle fires the gate &
  --- & selection surface lists candidates with previews marked unverified;
  decision pane shows the exact resolved bytes, the attested facts (channel:
  tool output; label $\U$), the unverified source strings, and the task \\
6a & Selected candidate resolves to the planted instruction; the verifier
  rejects & $\mathsf{dec}(\mathsf{reject})$ & no invocation; display and
  decision logged \\
6b & Selected candidate resolves to ``555-0100''; the user accepts &
  $\mathsf{dec}(\mathsf{accept})\cdot\mathsf{ack}$ & fresh $\mathsf{C}$ block
  ``555-0100'' scoped $\{\mathsf{send\_email}\}$; the tool runs with exactly
  the displayed bytes; its output is captured as $\U$ \\
7 & Later reuse of the confirmed number at $\mathsf{send\_email}$ & bytes via
  $\mathsf{recall}$; no gate & \textsc{Tool-Fast}; at a different tool the
  pair gates once more (scope is per tool) \\
\bottomrule
\end{tabular}
\end{table}

Three facts about the trace carry the design. First, the payload never enters
$P$'s context: steps~\hyperref[tab:trace]{2--5} deliver only acknowledgements and handles, so $P$'s
plan in session~2 is what the user typed, not what the attacker wrote
(Reasoning Integrity, \Cref{sec:method-single}). Second, the payload cannot
reach a tool without being displayed verbatim at step~\hyperref[tab:trace]{6} and accepted for that
exact tool (Action Integrity, \Cref{sec:method-single}). Third, everything
therefore rests on the decision at step~\hyperref[tab:trace]{6}---the single auditable event whose
occupants (oracle, self-verification, external LLM verifiers)
\Cref{tab:util} evaluates.

\subsection{The Recursive Language Model as a Quarantined Reader}\label{app:rlm}

Our implementation uses an RLM \citep{zhang2025recursive} because it supplies search, structured extraction, and crystallization within the quarantined reader ($\mathsf{fill}$ and $\mathsf{crystallize}$ in \Cref{tab:ops-full}). Conventional retrieval would need adaptation to provide the same interface while preserving the separation between untrusted content and the protected agent.

\paragraph{Why an RLM.} The reader must search an untrusted store that is
unbounded (convention~(\ref{S-conv:c7})) and routinely exceeds any context window, yet its
bytes must never be loaded into a trusted context. An RLM fits both
constraints by construction: the long input is not fed to the model but placed
as a variable in a persistent REPL environment, and the model writes code to
inspect, decompose, and recursively query sub-models over programmatic slices
of it \citep{zhang2025recursive}. The store is thus navigated by code,
symbolically, at any size.

\paragraph{How retrieval works.} The interpreter first restricts the reader's
view to $\U$-labelled blocks---endorsed content and interpreter state are
structurally invisible---and loads the visible blocks (identifier, content,
tags) as a list variable inside the REPL. The root model receives only the
retrieval instruction and constant-size metadata; it then iterates, printing
slices of the store, delegating oversized blocks to sub-model calls, and
accumulating its answer in a result variable, under a fixed iteration cap. The
instruction is \emph{fidelity-only}: read the actual blocks, copy any
extracted value verbatim, cite the source block identifier, and report absence
as absence---the reader carries no safety obligation, because every safety
property is enforced downstream by the interpreter (Lemma~\ref{lem:isolation},
Rules~\ref{S-r:null},~\ref{S-r:nobit}). Finally, the interpreter validates the
reply deterministically: the value must satisfy its declared schema and every
citation must name a block genuinely present in the view; any failure, reader
error, or uncited summary collapses to the null-handle outcome the semantics
already define (Rule~\ref{S-r:null}). An accepted value reaches $P$ only as an
opaque handle, resolved at a confirmation gate; an accepted summary is written
back at level $\U$ (\textsc{Fill}, \textsc{Cryst}, Figure~\ref{fig:labels}).
Algorithm~\ref{alg:rlm} summarizes the procedure.

\begin{algorithm}[t]
\caption{RLM-based retrieval for one reader operation
($\mathsf{fill}$ or $\mathsf{crystallize}$)}
\label{alg:rlm}
\begin{algorithmic}[1]
\Require query $q$, schema $s$, memory store $\Mem$; iteration cap $T$
\Ensure a schema-valid value with citations, or the null handle
\State $V \gets \{\,b \in \Mem : \ell(b)=\U\,\}$
  \Comment{view restriction (interpreter, deterministic)}
\State $\mathcal{E} \gets \textsc{InitREPL}(\texttt{context}
  \gets [(\mathrm{id}(b), \mathrm{bytes}(b)) : b \in V])$
\State issue fidelity-only instruction for $(q, s)$ to root model $Q$
  with constant-size metadata of \texttt{context}
\For{$t = 1, \dots, T$}
  \State $c \gets Q(\text{history})$;\quad
    $(\mathcal{E}, \mathrm{out}) \gets \textsc{REPL}(\mathcal{E}, c)$
    \Comment{$Q$ reads slices; sub-calls for oversized blocks}
  \If{$\mathcal{E}[\texttt{answer}]$ is set}
    \State $(v, C) \gets \textsc{Parse}(\mathcal{E}[\texttt{answer}])$,
      disambiguated against the true identifier set of $V$
    \If{$v$ satisfies $s$ \textbf{and}
        $C \subseteq \{\mathrm{id}(b) : b \in V\}$ \textbf{and}
        $C \neq \varnothing$}
      \State \Return $(v, C)$
        \Comment{to $P$ as opaque handle / $\U$-block only}
    \EndIf
    \State \Return null handle \Comment{fail closed, Rule~\ref{S-r:null}}
  \EndIf
\EndFor
\State \Return null handle
  \Comment{cap or reader error, Rule~\ref{S-r:nobit}}
\end{algorithmic}
\end{algorithm}

\paragraph{Utility of the complete pipeline.} In every PerMIT cell of
Tables~\ref{tab:util} and~\ref{tab:util-apx}, the RLM reader is the
retrieval path for stored untrusted content. With the oracle gate,
PerMIT has higher AM than undefended RAG on four of seven backbones,
trails by at most 1.5 AM points on Gemma-4-31B-it and GPT-5.5,
and by about ten AM points on GPT-5.4.
These cross-backend comparisons assess the complete mediated pipeline,
not a gate-induced improvement in retrieval quality. The oracle measures
retained capability under correct endorsement, while comparisons among
PerMIT's gate configurations keep the reader and interface fixed and
characterize decision-dependent utility costs. Isolating mediation
overhead from reader quality would require a matched no-gate reader
control; measuring human confirmation effort requires a separate study.

\section{Concurrent Work: Origin-Bound Memory Authority}
\label{app:concurrent}

TMA-NM \citep{louck2026securing} studies origin-bound authorization of
consequential actions. PerMIT combines observation mediation at a finite
memory interface, exact-value and tool-scoped endorsement, a
transcript-conditioned integrity theorem, and federated trust
accumulation. The comparison below distinguishes the policies, proved
properties, and verification scopes. Each guarantee depends on the
assumptions and enforcement rules of its own model.

\paragraph{Independent development.}
PerMIT was developed independently of, and concurrently with, TMA-NM.
We became aware of the earlier public preprint while preparing this
manuscript and include this comparison to clarify the relationship
between the two approaches.

\begingroup\small
\renewcommand{\arraystretch}{1.15}
\begin{longtable}{@{}p{0.235\textwidth}p{0.30\textwidth}p{0.385\textwidth}@{}}
\toprule
& \textbf{TMA-NM} \citep{louck2026securing} & \textbf{PerMIT}\\
\midrule
\endfirsthead
\toprule
& \textbf{TMA-NM} & \textbf{PerMIT}\\
\midrule
\endhead
Enforcement boundary & Origin-bound consequential-action authorization
& Observation mediation and sensitive-tool authorization under the
specified interface\\
Reasoning-side scope & Answer bias remains outside the guarantee
& Noninterference conditional on the complete display transcript;
endorsed content may influence responses\\
Verification scope & Bounded TLC checks and an inductive-invariant
argument; parameter independence justified by hand
& Lean~4 proofs for unbounded executions of the single-user core;
remaining scope in Appendix~\ref{app:mechanization}\\
Value attribution & Origin labels in the unified study; text proxy in
the supporting study
& Opaque handles before release and exact-value checks at sensitive
tools\\
Tool-output trust & Conditional trust for allow-listed tools
& All tool outputs enter untrusted; readable release requires
endorsement\\
User-authorization scope & Fresh, single-use authorization bound to the
concrete action
& Reusable authorization for an exact value and named tool;
transaction context is not bound\\
Multi-user sharing & Out of scope
& Federated promotion and dispute analysis under stated identity,
reviewer, and scheduling assumptions\\
\bottomrule
\end{longtable}
\endgroup

\subsection{The observation boundary}
\label{app:conc-observation}

TMA-NM permits untrusted memory to inform responses while constraining
consequential actions. PerMIT additionally mediates what the protected
agent can observe. Under Lemma~\ref{lem:isolation}, untrusted sources
are represented by opaque handles until endorsement releases their
contents. This prevents readable delivery of unendorsed memory through
the modeled interface; it does not prevent all possible influence from
malicious content after endorsement.

Theorem~\ref{thm:reason} states the precise comparison. For matched
initial configurations, identical user-typed inputs, coupled randomness,
the same schedule, and a common horizon, equal complete display
transcripts imply equal observable protected-agent behavior. An
attacker can nevertheless change a displayed preview, influence an
endorsement decision, or supply harmful content that a verifier accepts.
Such changes may alter the transcript and subsequent behavior. The
theorem establishes no additional influence beyond the transcript under
its matching conditions; it does not establish factual correctness,
unbiased recommendations, or the soundness of an automated verifier
(Appendix~\ref{app:reason}).

Isolation likewise blocks in-context reconstruction from fragments
while they remain unavailable to the protected agent as unendorsed
content. Once fragments have been endorsed and released, the agent can
combine them. Sensitive calls remain subject to exact-value and
tool-scope checks, but separate endorsements do not establish the safety
of a combined transaction or its suitability in a changed context
(Appendix~\ref{app:action}).

\subsection{Relational versus reachability properties}
\label{app:conc-proved}

The single-execution invariants reported for TMA-NM do not establish
PerMIT's two-execution, transcript-conditioned property. This is a
difference in the statements proved, not an expressiveness limitation
of model checking. Relational safety properties can be reduced to
safety properties over self-composed or product systems representing
multiple executions \citep{shemer2019property}. Establishing an
analogous result would require an appropriate relational model and a
separate proof; enlarging an existing single-execution model alone
does not supply that proof.

PerMIT's Lean development establishes its integrity properties for
arbitrarily long executions of the mechanized single-user core, under
the modeled interface and assumptions. Appendix~\ref{app:mechanization}
distinguishes this checked core from the paper-only extensions and
federated analysis. A proof about either formal model does not by itself
establish that a deployed harness implements that model faithfully.

\subsection{The federated tier}
\label{app:conc-federated}

PerMIT also studies shared memory contributed and reviewed by a
population of attested identities. Its analysis addresses promotion,
disputes, wrongful eviction, and confirmation cost under explicit
identity budgets, reviewer conditions, and protocol schedules
(\S\ref{sec:method-fed}).

These guarantees are conditional and probabilistic. Attestation and
first-decision accounting constrain how much weight adversarial
identities can supply, but honest reviewers can still accept poison.
Equation~\ref{eq:promotion-bound} bounds that residual promotion risk.
Under the demand, coverage, and threshold schedules of Supplementary
Theorem~\ref{S-thm:coscale}, the risk bounds decrease with community
size while the per-user confirmation burden for covered benign shared
demand does not increase. This is not an unconditional guarantee of
zero poison promotion or decreasing review effort for every workload.

\subsection{Deployment without institutional corroborators}
\label{app:conc-deployment}

PerMIT's single-user tier operates with the user as endorser and does
not require a federation. Once an exact value has been approved for a
named tool, retaining the endorsed copy and its in-scope authorization
permits reuse without another mandatory approval for that same pair.
This does not bound the number of rejected review requests, and deleting
the local endorsed copy can remove the reuse path. The guarantee also
does not establish how much useful content a human or automated verifier
will approve.

Reuse is an explicit authorization-policy choice. PerMIT's endorsement
does not bind argument position, task, purpose, time, or the other
arguments of a call. Community disputes suspend community reuse while
preserving prior personal confirmations. Thus the confirmation-cost
claim should be read together with the retained-copy assumption and
the contextual-authorization limits in Appendix~\ref{app:action}.

\subsection{Laundering channels}
\label{app:conc-laundering}

PerMIT addresses laundering through the modeled interface rules.
Reader-generated summaries remain untrusted when written by
$\mathsf{crystallize}$, so summarization alone grants no authority.
All tool outputs are captured as untrusted, preventing a tool's name or
status from automatically making its output trusted. These rules do not
make subsequently endorsed content harmless or protect against a
compromised enforcement harness.

Repeated content does not itself constitute independent endorsement.
Distinct attested identities and first-decision accounting restrict
manufactured voting weight. Where the promotion quorum exceeds the
certified adversarial identity budget, adversarial endorsements alone
cannot satisfy it; erroneous honest endorsements can still promote
poison, with risk bounded under the assumptions of
Equation~\ref{eq:promotion-bound}.

The \texttt{capability\_ifc} baseline evaluated by
\citet{louck2026securing} treats retrieved memory as trusted. Its
reported failures concern that evaluated configuration and do not
establish that every deployment of CaMeL, FIDES, or capability-based
information-flow control fails under the same attacks.

\clearpage
\section{Formal Guarantees and Proof Sketches}
\label{app:formal}

This appendix summarizes the memory-interface model and single-user integrity guarantees cited in the main text. Part~\ref{part:supplement} (\hyperref[part:supplement]{Supplement~S}) gives the complete semantics and integrity proofs (\ref{S-app:interface}--\ref{S-app:single}), federated analysis and confirmation-cost results with their assumptions (\ref{S-app:fedmodel}--\ref{S-app:budget}), security discussion (\ref{S-app:shield}), probability tools and assumption audit (\ref{S-app:toolbox}), and detailed Lean correspondence (\ref{S-app:mechanization}).

\subsection{Interface and trust assumptions}
\label{app:interface}

The Protected LLM $P$ is an arbitrary agent strategy over its observations, interaction history, and random coins. It chooses operations and can re-plan on endorsed feedback. The harness mediates the memory interface and tool boundary; it does not inspect or taint-track the agent's internal reasoning. The Quarantined Reader $Q$ may behave adversarially and holds no tool capability. Appendix~\ref{app:ops-table} gives the operation-level view; Supplement~\ref{S-app:interface} gives the normative semantics.

\paragraph{Stored values and authority.}
Each immutable block has bytes, a source-based label, provenance, and a tool scope. The labels are $\T$ (user-typed), $\U$ (untrusted), $\C$ (personally confirmed), $\PC$ (agent-composed), and, in the federated extension, $\CC$ (community-confirmed). The readable store $\Mem_{\E}$ consists of non-$\U$ blocks. Readability alone does not authorize a sensitive call: a literal must byte-match a $\T$ block or a confirmed block scoped to that tool. A $\PC$ block has no independent endorsement authority. A confirmation for the reasoning target makes its bytes readable but grants no sensitive-tool scope.

The trusted command channel creates $\T$ blocks. Pasted content and all tool outputs, including errors and status text, enter $\U$. Reader-generated summaries also enter $\U$. The agent receives an opaque handle, not those bytes. Accepting confirmation creates a fresh $\C$ copy; it never relabels the untrusted source in place. Endorsed and untrusted identifiers use separate allocation streams, so an attacker cannot influence a later endorsed identifier by adding untrusted blocks.

\paragraph{Mediation rules.}
The harness enforces the following properties (complete rules: Supplement~\ref{S-app:rules}).
\begin{itemize}[leftmargin=1.5em,itemsep=2pt,topsep=3pt]
\item Every handle-expressed tool argument gates, even if its hidden bytes equal an already trusted value. The gate resolves the handle, displays the exact bytes and target, and binds an acceptance to those same bytes. Rejection or failed resolution prevents the flow.
\item Sensitive-tool literals use only the typed or exact-tool-confirmed fast paths. The baseline permits agent-composed literals at data tools. Consequently, endorsement for one purpose does not imply a confidentiality guarantee for all later data-tool uses; Option~\ref{S-o:two} adds exact-tool checks there.
\item Reader failure, empty retrieval, and unoccupied identifiers have the same opaque reply. Handle structure follows operations rather than hidden content, size, or success. The reply alphabet contains only fixed acknowledgements, handles, fixed failures, decisions, and verbatim readable-store bytes. Trusted retrieval sees $\Mem_{\E}$ alone.
\item All attacker-influenceable user-facing content is recorded in the display transcript. Authenticated facts and untrusted source claims occupy separate zones. Selection previews carry no authority; authorization uses the resolved value in the decision pane.
\item A batch displays every value that would flow, records its pages and decisions, and executes only after all required approvals. Accepted values remain endorsed even if a later review aborts the invocation. Full batch semantics and the corresponding determinacy argument are in Supplement~\ref{S-app:interface}--\ref{S-app:single}.
\end{itemize}

\paragraph{Assumptions and boundaries.}
Both integrity results assume source-based labels, immutable stored blocks, and faithful displays (\ref{S-a:label}--\ref{S-a:display} in Supplement~\ref{S-app:axioms}), together with the enforced interface above. Action Integrity additionally assumes \emph{sound atomic endorsement}: acceptance means that the user intends the exact displayed value at the named tool, separately for each reviewed value, including accepted positions of a transaction that later aborts (\ref{S-a:user}). An LLM verifier can violate this assumption; the theorem does not establish its accuracy. Reasoning Integrity instead assumes a common exogenous typed-input stream and matched scheduling and coins (\ref{S-a:exo} and conventions \ref{S-conv:c1}, \ref{S-conv:c9}); it does not require endorsements to be semantically correct. Timing, termination, and compromised harness components are outside these guarantees.

\begin{lemma}[Protected-agent isolation]\label{lem:isolation}
In every interface-conformant execution, each bytes-carrying reply delivered to $P$ is the verbatim content of a readable-store block at delivery. Untrusted sources reach $P$ only through opaque handles until endorsement. This is a statement about the source of a delivered value, not disjointness of byte strings across stores.
\end{lemma}
\begin{proof}[Proof sketch]
Inspect the reply constructors. Tool capture, untrusted recall, reader search, and crystallization return handles or fixed tokens. Only trusted recall and retrieval return bytes, and both read $\Mem_{\E}$. Input ingestion and each operation preserve this property, so induction extends it to arbitrary run length. Full proof: Supplement, Lemma~\ref{S-lem:isolation}.
\end{proof}

\subsection{Integrity Proofs}
\label{app:single}

\subsubsection{Action Integrity}
\label{app:action}

\begin{definition}[Personal Action Integrity]\label{def:action}
A sensitive sink event at tool $t$ is personally authorized when each argument value $v$ byte-matches user-typed content, or this user has accepted the exact bytes $v$ for $t$, either at that call or in a retained in-scope confirmation. For a batch, acceptance covers each displayed element separately. Action Integrity requires this property for every sensitive sink event of every execution.
\end{definition}

\begin{theorem}[Action Integrity]\label{thm:action}
Under the interface and \ref{S-a:label}--\ref{S-a:user} above, every execution starting from a well-formed configuration has personal Action Integrity, for arbitrary agent strategies, attacks, and reader behavior. The empty configuration is well formed; a seeded configuration must carry a valid endorsement record. The invariant requires immutable source-correct blocks, valid endorsement ancestry and history, interface-conformant state, and authorization of the accumulated sink log (complete definition: Supplement, Definition~\ref{S-def:wf}).
\end{theorem}
\begin{proof}[Proof sketch]
The empty state satisfies the invariant. For each transition, new untrusted blocks confer no authority; agent writes confer readability but no new authorization; and an accepted confirmation creates a fresh copy justified by the exact display and sound endorsement. A sensitive call therefore has only three admissible cases: a typed literal, an in-scope confirmed literal, or fresh accepting confirmation. Every handle takes the third case, and a batch applies it to every flowed element. Failed reviews prevent the call, while already accepted values remain individually authorized. These cases preserve the invariant and establish sink safety by induction. The single-user interpreter has no $\CC$ creation path. Full statement and proof: Supplement, Theorem~\ref{S-thm:action}.
\end{proof}

\begin{corollary}[No silent resolution]\label{cor:noexfil}
Under the same conditions, never-endorsed untrusted bytes cross a tool boundary, including a data-tool boundary, only through acceptance of those exact bytes for that tool in that call. After endorsement at any target, their readable copy may be used as a literal at baseline data tools. Sensitive-tool scope checks and the rule that handles always gate still apply.
\end{corollary}
\begin{proof}[Proof sketch]
Isolation leaves never-endorsed bytes available only through handles, and every handle resolution gates. Endorsement subsequently creates a readable copy, enabling the stated literal path. This distinguishes exact-byte authorization from control influence carried by a logged decision. Full proof and Option~\ref{S-o:two}'s closure of the later data-tool path: Supplement, Corollary~\ref{S-cor:noexfil}.
\end{proof}

\paragraph{Reuse and contextual authorization.}
The policy authorizes exact values for tools, not complete transactions. Separately approved values may be combined without joint review, and a changed task context may make their reuse inappropriate. Shared disputes suspend community reuse but preserve prior personal confirmations. Candidate expiry retires dormant or inactive shared entries (Supplement, Rule~\ref{S-r:hygiene}); it is not a universal authorization lifetime. Local scope leases are a future extension outside the proved model (Option~\ref{S-o:three}). Context-bound or transaction-level approval would require extended semantics and revised confirmation-cost accounting.

\paragraph{Future direction: combined-memory review.}
Human or agent memory managers could assemble related memories with their provenance and intended use, then review their joint meaning. The combined item remains untrusted until confirmation; its components' approvals do not automatically endorse it. If subsequently published, it must gather fresh endorsements of its exact content under the community-promotion rules. Revisions require their own review. This proposed workflow builds on crystallization and endorsement but has not been evaluated and does not authorize every later transaction using the combined memory. Future experiments should compare it with per-item review under matched attack-search and review budgets, measuring harmful endorsement, delayed-combination success, benign utility, and review burden.

\subsubsection{Reasoning Integrity modulo the display transcript}
\label{app:reason}

The protected view is $\obs(c)=(\Mem_{\E},H,x,h)$: the readable store, observable handle structure, typed input, and decision history. Matched configurations $c_1\approx c_2$ agree on this view, the agent's interaction history and consumed coins, endorsed allocation state, and initial ghost display and sink logs. Their untrusted stores and private handle resolutions may differ arbitrarily. A common endorsed seed with empty histories is a canonical matched start.

The complete display transcript $\DT(\rho)$ records selection surfaces and previews, exact decision-pane values and provenance zones, targets, selections, accepts and rejects, and batch pages and aggregated decisions, including reviews in aborted transactions. Its event kinds are interpreter-assigned. The observable behavior $\mathcal B(\rho)$ comprises protected-agent operations, tool invocations, the reply stream, final protected view, and the ungated sink projection. The latter omits arguments flowed by fresh confirmations, which are already recorded in the transcript.

\begin{theorem}[Reasoning Integrity modulo the display transcript]\label{thm:reason}
Under \ref{S-a:label}--\ref{S-a:display} and the input and scheduling conditions above, fix matched starts, any two attacks, and a coupled coin sequence. For runs $\rho_1,\rho_2$ of a common horizon,
\[
\DT(\rho_1)=\DT(\rho_2)
\quad\Longrightarrow\quad
\mathcal B(\rho_1)=\mathcal B(\rho_2).
\]
Paste presence and content, reader outputs, and tool outputs need not agree across the attacks. The statement is trace-wise for every coupled coin sequence. In plain terms, the two runs may differ in whether pasted input arrives, its contents, and reader and tool outputs. The guarantee holds for every fixed sequence of random choices used identically in both runs.
\end{theorem}
\begin{proof}[Proof sketch]
Maintain matched protected views, histories, coin positions, and transcript prefixes. The strategy then chooses the same next operation. A display-free step returns only a constant token, an operation-determined handle, or matched readable bytes, preserving the invariant. At a display step, the append-only transcripts and equality of the complete transcripts make the newly appended segments prefix-comparable. The interface's display-segment determinacy lemma forces those segments to be equal, including page structure, decisions, and released bytes. The two steps therefore create the same endorsed copies and replies. Induction gives the claimed equality; sink annotations follow the same mediation cases as Action Integrity. Full proof, including the segment lemma and batch cases: Supplement, Lemma~\ref{S-lem:dtsegment} and Theorem~\ref{S-thm:reason}.
\end{proof}

\begin{corollary}[Attack-surface confinement]\label{cor:confine}
Under the same comparison conditions, a difference in protected-agent operations, tool invocations, or ungated sink events entails a difference in the complete display transcript.
\end{corollary}
\begin{proof}[Proof sketch]
Take the contrapositive of Theorem~\ref{thm:reason}.
\end{proof}
This is a conditional noninterference guarantee. An attacker can affect a preview or an endorsement decision, and thereby the transcript and later behavior. The theorem excludes additional influence that bypasses that recorded release surface; it does not establish reasoning correctness or make displayed attacks harmless.

\section{Scope of the Lean 4 Verification}
\label{app:mechanization}

The Lean artifact is available upon request and will be released with the code when the paper is published. The artifact contains the executable single-user interpreter, proofs, build assertions, dependency manifest, and axiom report. It pins Lean~4 and Mathlib to \texttt{v4.31.0} (Mathlib revision \texttt{fabf563a}); \texttt{lake exe cache get \&\& lake build} reproduces the check. Supplement~\ref{S-app:mechanization} gives the complete correspondence table, ten modeling deltas, and the specification corrections discovered during mechanization.

\paragraph{What is checked.}
\texttt{action\_integrity} and \texttt{action\_integrity\_from} establish the single-user action guarantee from the empty endorsed store and from any state satisfying the mechanized invariant. \texttt{reasoning\_integrity} and \texttt{reasoning\_integrity\_from} prove the transcript-conditioned comparison from empty or shared endorsed seeds, for arbitrary agent strategies and independently chosen attacks. Their conclusions include equal sink logs, operations, replies, and final protected views. Paste presence and content can differ. Supporting results check reply isolation, mandatory handle gating, and attack-surface confinement. A standalone batch layer checks displayed-element flow, rejection behavior, and transcript determinacy; executable assertions cover the baseline counterexample and eight scenarios.

\paragraph{What remains a paper proof.}
The run-level integration of batched iteration, durable endorsements from accepted positions of an aborted invocation, arbitrary mid-run matched starting states, and the reduction from reactive actors to event-indexed oracles are proved in the supplement, outside the checked interpreter. Options~\ref{S-o:one}--\ref{S-o:two} are checked at the gate level only; their wider integration, Option~\ref{S-o:three}, the federated protocol, probability bounds, and confirmation-budget results are not mechanized. The checked interpreter returns one selected element at a direct collection-handle call and records additional transcript-determined selection information. These representation differences and their correspondence arguments are explicit in \ref{S-app:mechanization}.

\paragraph{Trusted base and artifact.}
The artifact's clean-build axiom report lists only \texttt{propext} and \texttt{Quot.sound} for the top-level theorems, with no \texttt{sorryAx}; the executable assertions additionally trust Lean's evaluator. Kernel checking certifies the mechanized statements, not the correspondence between that model and a deployed harness or the reliability of a verifier. The artifact includes the full source, \texttt{CORRESPONDENCE.md}, pinned build instructions, the build log, and source checksums. Thus the concise statements here remain auditable against both the complete paper proofs and the precise mechanized scope.

\clearpage
\part{Complete Formal Development}
\label{part:supplement}
\startsupplement
\noindent This part preserves the complete formal development. The main paper
and Part~\ref{part:appendices} give concise definitions, conditions, result statements, and proof
sketches. Supplement~\ref{S-app:guide}--\ref{S-app:mechanization} retains the full models, proofs, and derivations,
followed by the detailed Lean correspondence in \ref{S-app:mechanization}. The complete core theorem
statements remain in the shared source files; the Lean artifact is available
upon request.

\begin{center}\small
\begin{tabular}{@{}p{0.26\linewidth}p{0.24\linewidth}p{0.41\linewidth}@{}}
\toprule Main paper & Formal supplement & Lean artifact\\\midrule
Formal model (\ref{app:interface}) & \ref{S-app:guide}--\ref{S-app:interface} & Interface, Semantics, Basic\\
Action Integrity (\ref{app:single}) & \ref{S-app:single} & ActionIntegrity, Invariants\\
Reasoning Integrity (\ref{app:single}) & \ref{S-app:single} & ReasoningIntegrity; BatchGate (standalone)\\
Federated overview (\S\ref{sec:method-fed}) & \ref{S-app:fedmodel}--\ref{S-app:fedproofs}, \ref{S-app:toolbox} & Paper proofs only\\
Verification-cost overview (\S\ref{sec:method-fed}) & \ref{S-app:budget}, \ref{S-app:toolbox} & Paper proofs only\\
Lean scope and deltas (\ref{app:mechanization}) & \ref{S-app:mechanization} & CORRESPONDENCE.md, README\\
Design rationale & \ref{S-app:shield} & No additional mechanized claims\\\bottomrule
\end{tabular}\end{center}

\begingroup
\etocsetnexttocdepth{subsection}
\etocsettocstyle{\subsection*{Detailed contents of the formal development}}{}
\localtableofcontents
\endgroup
\clearpage

\section{Structure of the Proof, Notation, and Conventions}
\label{S-app:guide}

\subsection{Organisation and register conventions}
\label{S-app:roadmap}

Sections~\ref{S-app:guide}--\ref{S-app:toolbox} contain the formal
development for PerMIT, with a machine-checked single-user core
(\S\ref{S-app:mechanization}) and register-separated normative, proof,
and rationale layers; the empirical appendices are indexed in the guide
in Appendix~\ref{sec:additional} of the main paper.
The material is organised into three strictly separated registers.

\begin{description}[leftmargin=1.6em,itemsep=2pt]
\item[Normative register (Appendices~\ref{S-app:interface}, \ref{S-app:fedmodel}).]
The memory-interface semantics, the interpreter and protocol rules
(\ref{S-r:names}--\ref{S-r:hygiene}), and the assumptions (\ref{S-a:label}--\ref{S-a:rhoseat}; the budget-model assumptions
\ref{S-a:smooth}--\ref{S-a:massbound} are stated with the model they govern). Every rule clause states what
the harness does, tersely; no rule contains its own justification. A rule
clause states normative behaviour tersely, relying only on the
normative definitions and conventions it cites; rationale and
empirical interpretation are never needed to implement it.
\item[Proof register (Appendices~\ref{S-app:single}, \ref{S-app:fedproofs},
\ref{S-app:budget}).]
Definitions, lemmas, and theorems. Proof-register text
cites the normative register and the named invariants of
\S\ref{S-app:invariants}; it never cites the non-normative register.
The single-user results of Appendix~\ref{S-app:single} are additionally
machine-checked in Lean~4 over the map-free operation algebra, with the
batch-confirmation surface of Rule~\ref{S-r:batchgate} mechanized as a
standalone layer; the store-level integration of $\mathsf{map}$ into the
run-level interpreter and Option~\ref{S-o:three} remain outside the mechanized
surface, under the disclosed deltas recorded in \S\ref{S-app:mechanization}.
\item[Non-normative register (Appendix~\ref{S-app:shield} and commentary
blocks).] Design rationale (Notes \ref{S-n:lineage}--\ref{S-n:effort}), the small-type commentary
blocks defined below, and a security and robustness analysis
(Analyses \ref{S-q:paraphrase}--\ref{S-q:handlecover}): an indexed catalogue of
attack vectors and failure modes, each with its disposition
by formal fact, and its status: \emph{disposed} (under stated
assumptions; a formal fact disposes
of it) or \emph{residual} (a genuine limitation, disclosed); and
either \emph{entering only the budget constants} or \emph{not used in
any proof}. Four rationale remarks whose content adjoins specific results
are set as numbered Remarks adjacent to the results they annotate,
Remarks~\ref{S-n:freshness},
\ref{S-n:qpos}, \ref{S-n:calib}, and~\ref{S-n:roles}. Nothing normative
lives in this register; no proof depends on it.
\end{description}

\noindent\textbf{Commentary blocks.} Throughout the normative and proof
registers, short passages set in small type under the heading
\emph{Commentary (non-normative)} carry the rationale, epistemic
status, and validation pointers for the material they adjoin, so that a
rule or assumption and its justification can be read together without
the justification entering the register. Commentary blocks belong to
the non-normative register in function: no rule reads them, no proof cites
them, and a reader who skips every commentary block loses no
hypothesis and no step of any proof.

\noindent\textbf{Statement discipline.} A theorem, lemma, proposition,
corollary, definition, or invariant environment contains
hypotheses, the conclusion, and bare pointers (``established by'',
``see Remark''); interpretation, comparison, special cases, and design
justification are set in a numbered remark adjacent to the result they
concern.

\noindent\textbf{Cross-referencing convention.} The formal development
may point
into the non-normative register only by a parenthetical key, e.g.\ ``(rationale:
\ref{S-n:unit}; analysis: \ref{S-q:recombine})''; the non-normative register may cite the
formal development freely. Consequently the formal development is self-contained: a reader who
skips Appendix~\ref{S-app:shield} entirely loses no hypothesis and no
step of any proof. A reader with a specific concern should consult
the attack and limitation index (Table~\ref{S-tab:objindex}), which
maps concerns to analysis entries and to the formal facts that govern
them.

\medskip
\noindent\textbf{Map of the formal development
(Sections~\ref{S-app:guide}--\ref{S-app:toolbox}).}
\begin{center}
\small
\renewcommand{\arraystretch}{1.2}
\begin{tabular}{@{}llp{0.62\textwidth}@{}}
\toprule
\ref{S-app:guide} & Structure & notation, conventions, named invariants and structural properties
\ref{S-inv:fresh}--\ref{S-inv:reach}.\\
\ref{S-app:interface} & Interface & configurations; interface transition schema
(Figs.~\ref{S-fig:sos}--\ref{S-fig:sos2}); Rules \ref{S-r:names}--\ref{S-r:retrieve}, options \ref{S-o:one}--\ref{S-o:two} (\ref{S-o:three} requirements-level); Assumptions \ref{S-a:label}--\ref{S-a:exo};
attacker model.\\
\ref{S-app:single} & Single user & Action Integrity
(Thm.~\ref{S-thm:action}), no-silent-resolution
(Cor.~\ref{S-cor:noexfil}), Reasoning Integrity modulo the display
transcript (Thm.~\ref{S-thm:reason}), confinement
(Cor.~\ref{S-cor:confine}), baseline separation
(Prop.~\ref{S-prop:baseline}), hardening monotonicity
(Prop.~\ref{S-prop:harden}).\\
\ref{S-app:fedmodel} & Federated protocol & pools, epochs, demand, freshness,
endorsement counting; Rules \ref{S-r:share}--\ref{S-r:hygiene}; Assumptions \ref{S-a:transfer}--\ref{S-a:rhoseat}.\\
\ref{S-app:fedproofs} & Federated guarantees & structural lemmas (sharing
exclusion, entry disjointness, $\cacc$, verdict pacing, decisive
resolution, no-deletion);
deterministic safety under perfect detection (Thm.~\ref{S-thm:det}); panel safety
(Thm.~\ref{S-thm:panel}); wrongful eviction (Prop.~\ref{S-prop:live}); publisher
safety (Prop.~\ref{S-prop:pubsafe}); denial of service bounded in required confirmations (Lem.~\ref{S-lem:freeze}; effort disclosure \ref{S-n:effort}); submission budget
(Cor.~\ref{S-cor:submission}).\\
\ref{S-app:budget} & Budget & single-user floor (Thm.~\ref{S-thm:floor},
per (user, value, tool)), per-item ledger (Lem.~\ref{S-lem:budget},
$\bigO$-form), saturation
(Thm.~\ref{S-thm:sat}), co-scaling corridor and theorem
(Thm.~\ref{S-thm:coscale}).\\
\ref{S-app:shield} & Rationale \& analysis & design rationale; security
and robustness analysis with attack and limitation index; scope and
residuals; summary of composed guarantees.\\
\ref{S-app:toolbox} & Tools & probability space and adaptive coupling;
Hoeffding bounds; no-dispute baseline; the blast-schedule promotion law
(a simulation analysis tool, cited by no proof;
\S\ref{S-app:blastlaw}); explicit ledger constants; knowledge-based
formulation; Zipf case study; assumptions ledger.\\
\bottomrule
\end{tabular}
\end{center}

\begin{figure}[tp]
\centering
\begin{tikzpicture}[
  font=\scriptsize, align=center,
  every node/.style={text width=2.0cm, inner sep=2.5pt},
  asm/.style={draw, rounded corners, fill=gray!12},
  lem/.style={draw, rounded corners, fill=blue!8},
  thm/.style={draw, rounded corners, fill=green!12, thick},
  inv/.style={draw, rounded corners, double, fill=yellow!20},
  cal/.style={draw, rounded corners, dashed, fill=gray!5},
  ghost/.style={draw, rounded corners, dotted, fill=white},
  arr/.style={->, >=stealth},
  carr/.style={->, >=stealth, dashed}
]

\node[text width=12cm, align=left] at (5.6,1.0)
  {\normalsize\textbf{(a)} Integrity and federated safety};

\node[asm] (HA) at (0,0)    {\ref{S-a:label}--\ref{S-a:exo}\\trust axioms};
\node[asm] (HR) at (2.6,0)  {\ref{S-r:names}--\ref{S-r:hygiene}\\harness rules (TCB)};
\node[asm] (HS) at (5.1,0)  {Sybil bound\\$\Wadv,\ \fs$ finite};
\node[asm] (HC) at (8.1,0)  {competence\\$q>0$;\ $q{+}p_b{>}1$};

\node[lem] (LC)  at (0.6,-2)
  {Lem.~\ref{S-lem:isolation}--\ref{S-lem:cflow}\\channel analysis};
\node[inv] (INV) at (3.2,-2) {\ref{S-inv:fresh}--\hyperref[S-inv:reach]{6}};
\node[lem] (LDa) at (6.6,-2)
  {Lem.~\ref{S-lem:shareexcl},\,\ref{S-lem:disjoint}\\exclusion, disjointness};
\node[lem] (GP)  at (9.6,-2)
  {Lem.~\ref{S-lem:adaptive},\,\ref{S-lem:engine}\\coupling, Hoeffding};

\node[thm] (T1)   at (0,-4)    {Thm.~\ref{S-thm:action}\\Action Integrity};
\node[thm] (T2)   at (2.4,-4)  {Thm.~\ref{S-thm:reason}\\Reasoning Integrity};
\node[thm] (TDET) at (5.1,-4)  {Thm.~\ref{S-thm:det}\\deterministic safety};
\node[thm] (TPAN) at (8.1,-4)  {Thm.~\ref{S-thm:panel}\\panel safety};
\node[thm] (TLIV) at (11.3,-4)
  {Prop.~\ref{S-prop:live},\\\ref{S-prop:pubsafe}\\wrongful eviction, publisher};

\draw[arr] (HA) -- (LC);
\draw[arr] (HR) -- (LC);
\draw[arr] (HR) -- (INV);
\draw[arr] (HR) -- (LDa);
\draw[arr] (HS) -- (TDET);                       
\draw[arr] (HC) -- (TPAN);                       
\draw[arr, rounded corners=3pt] (HC.east) -| (TLIV.north); 
\draw[arr] (LC) -- (T1);
\draw[arr] (LC) -- (T2);
\draw[arr] (INV) -- (TDET);
\draw[arr] (INV) -- (TPAN);
\draw[arr] (LDa) -- (TPAN);
\draw[arr] (GP) -- (TPAN);
\draw[arr] (GP) -- (TLIV);

\begin{scope}[yshift=-6.4cm]
\node[text width=12cm, align=left] at (5.6,1.0)
  {\normalsize\textbf{(b)} Verification budget};

\node[asm] (A9)  at (0,0)   {\ref{S-a:rhoseat}\\verdict rate};
\node[cal] (CAL) at (3.0,0) {\ref{S-r:panel}(v) calibration ($\thacc,\cceil$)};
\node[asm] (A12) at (6.0,0) {\ref{S-a:smooth}--\ref{S-a:massbound}\\demand model};
\node[ghost] (IN3) at (9.0,0) {\ref{S-inv:floor}\\(from (a))};

\node[lem] (CP)  at (1.5,-1.8)
  {Lem.~\ref{S-lem:cacc},\,\ref{S-lem:geo}\\$\cacc$, pacing};
\node[ghost] (GA) at (4.5,-1.8)
  {Lem.~\ref{S-lem:adaptive}\\(from (a))};

\node[lem] (LB)  at (3.0,-3.6)
  {Lem.~\ref{S-lem:budget},\,\ref{S-lem:freeze}\\ledger, DoS};
\node[thm] (TF)  at (9.0,-3.6)
  {Thm.~\ref{S-thm:floor}(i)\\floor (pathwise;\\per \ref{S-inv:floor})};

\node[thm] (TS)  at (6.0,-5.4)
  {Thm.~\ref{S-thm:sat},\,\ref{S-thm:coscale}\\saturation, co-scaling};

\draw[arr]  (A9) -- (CP);
\draw[carr] (CAL) -- node[pos=0.62, fill=white, inner sep=1.5pt]
  {\tiny constants only} (CP);
\draw[arr, rounded corners=3pt] (A12.south) -- ++(0,-0.55)
  -| (LB.north);                                  
\draw[arr]  (A12) -- (TS);                        
\draw[arr]  (IN3) -- (TF);                        
\draw[arr]  (CP) -- (LB);
\draw[arr]  (GA) -- (LB);
\draw[arr]  (LB) -- (TS);
\draw[arr]  (TF) -- (TS);
\end{scope}
\end{tikzpicture}
\caption{Dependency structure (edges denote ``is used in the proof
of''). \textbf{(a)} The single-user theorems rest on the trust axioms
and harness rules alone (the channel-analysis lemmas use \ref{S-r:names}--\ref{S-r:retrieve}; the
structural lemmas use \ref{S-r:share}--\ref{S-r:hygiene}); the coupling and Hoeffding lemmas are
assumption-free. \ref{S-a:transfer}--\ref{S-a:fedlog} enter only
Proposition~\ref{S-prop:fed}, which supplies their delegated extension. \textbf{(b)} Dotted boxes import results proved in panel~(a)
and Appendix~\ref{S-app:toolbox}. The dashed edge is the only role of
the accountability calibration: it grounds the constant $\cacc$ in the
budget ledger and never enters any safety theorem. The pathwise floor, Thm.~\ref{S-thm:floor}(i), depends on
\ref{S-inv:floor} alone, visible here as its sole incoming edge; the
expected-cap clauses (ii)--(iii) additionally read the reference
demand model's masses, and the tool-spread comparison
$d_{\mathrm{tool}}\le\ctool d$ is quarantined in
Cor.~\ref{S-cor:floorspread} (\ref{S-a:toolspread}), consumed by no
other result. (TCB $=$ trusted computing base;
defined in \S\ref{S-app:notation}, inventoried in
\S\ref{S-app:attacker}.)}
\label{S-fig:dag}
\end{figure}
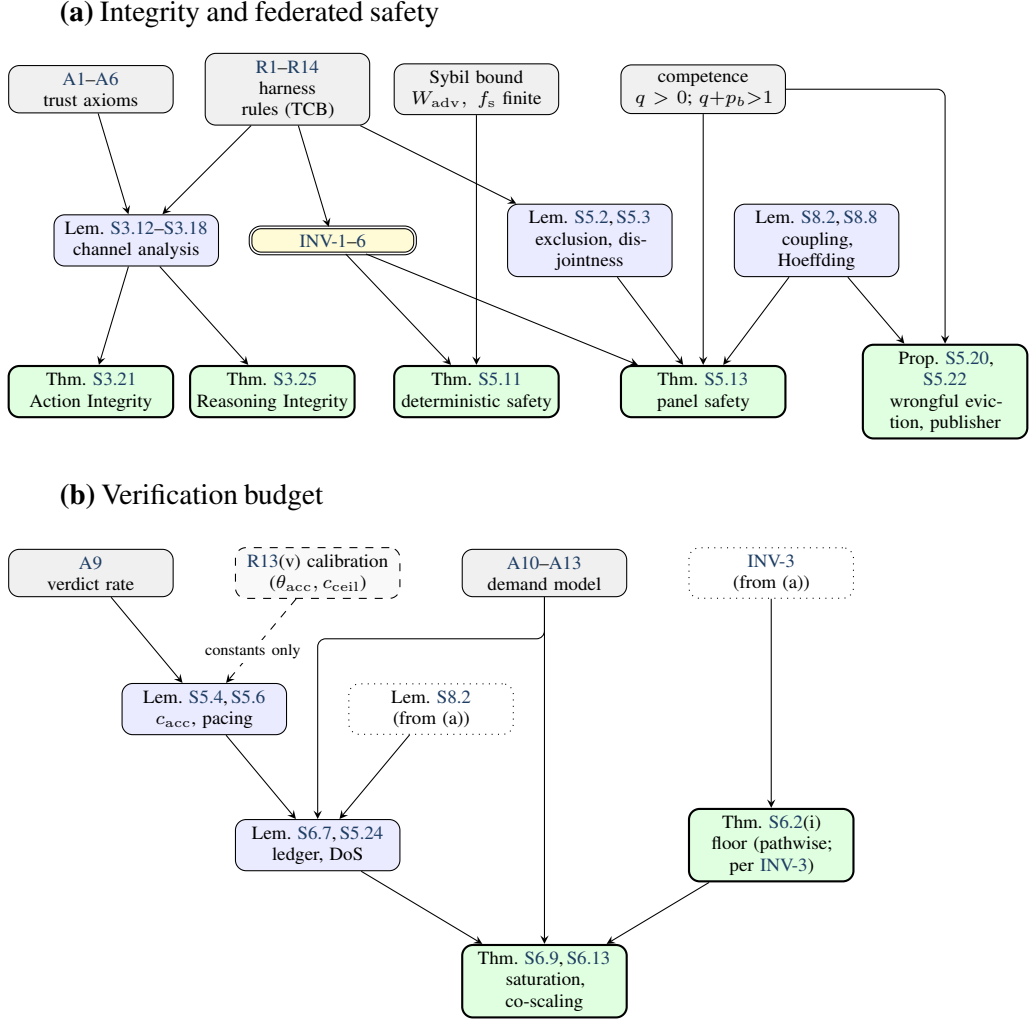

\medskip
\noindent\textbf{Dependency structure.}
Figure~\ref{S-fig:dag} shows which results rest on which hypotheses. Three
facts are visible directly from the figure and are relied on throughout: (i) the
single-user theorems and the pathwise cap
(Thm.~\ref{S-thm:floor}(i)) depend only on the trust axioms \ref{S-a:label}--\ref{S-a:exo} and
the harness rules, never on any federated or budget modelling
assumption (Appendices~\ref{S-app:fedmodel} and~\ref{S-app:budget}); the
cap's expected-form clauses read only the reference demand model's
masses, with the sole assumption-bearing comparison isolated in
Corollary~\ref{S-cor:floorspread}; (ii) no
safety theorem depends on the accountability calibration of
Rule~\ref{S-r:panel}(v), which feeds only budget constants (dashed
edge); (iii) the behavioural parameters $q,p_b,\rseat$ and the demand
assumptions \ref{S-a:smooth}--\ref{S-a:massbound} enter only the probabilistic-refinement and budget
layers, each of which degrades to the layer below it when its
hypotheses fail (\S\ref{S-app:composed}).

\subsection{Notation}
\label{S-app:notation}

The principal symbols of these appendices are collected here; locally
scoped symbols are defined at first use. Unless otherwise marked,
symbols denote fixed deployment configuration belonging to the
\emph{trusted computing base} (TCB): the fixed
components (interpreter, channels, rendering and logging paths, and
configuration) that the guarantees assume correct and that no model
output can alter (inventory: \S\ref{S-app:attacker} single-user,
\S\ref{S-app:fedassum} federated). Marks:
(emp)~empirical/behavioural parameter, (sch)~population schedule,
(rv)~random variable.

\begingroup\small
\renewcommand{\arraystretch}{1.15}
\begin{longtable}{@{}lp{0.65\textwidth}@{}}
\toprule
\multicolumn{2}{@{}l}{\emph{Single-user model and interface
(Appendix~\ref{S-app:interface})}}\\
\midrule
$c=(\Mem,\sigma_P,\sigma_Q,h)$ & configuration: store, Protected-LLM
state, Reader state, decision history.\\
$\Mem;\ \Mem_{\U},\Mem_{\E}$ & memory store; restrictions to level-$\U$ /
level-$\E$ blocks.\\
$b=(\mathrm{id},\mathrm{bytes},\ell,\pi,\scopef)$ & block: identifier (two-sorted, operation-determined,
Rule~\ref{S-r:names}(g)), bytes, label, interpreter-only provenance,
tool scope.\\
$\ell(b)\in\{\U,\T,\C,\CC,\PC\}$ & label: untrusted, trusted (user-typed),
confirmed, community-confirmed, $P$-composed.\\
$\lvl;\ \E$ & level map ($\lvl(\U)=\U$, else $\E$); endorsed level.\\
$P,\ Q,\ \RE$ & Protected LLM; Quarantined Reader; trusted-store
retriever (selector).\\
$\SensT,\ \DataT$ & sensitive (effectful) / data tools; fixed partition,
canonical identifiers.\\
$\Sigma$ & operation algebra
$\{\mathsf{write}_P,\mathsf{recall},\mathsf{fill},\allowbreak\mathsf{crystallize},
\mathsf{display},\conf,\mathsf{evict}\}$; $\Sigma_P$: the $P$-facing surface $\Sigma\cup\{\text{tool invocation},\mathsf{retrieve}\}$ ($+$ derived $\mathsf{map}$). ($\mathsf{write}$, where it appears in prose, abbreviates the sole $P$-emitted write operation $\mathsf{write}_P$.)\\
$\conf(v,t)$ & confirmation event: value $v$, target $t$ (tool;
``reasoning''; ``publish'').\\
$z$ & gate decision, $z\in\{\mathrm{accept},\mathrm{reject}\}$.\\
$h_j$ & $j$-th operation-determined handle identifier
(Rule~\ref{S-r:names}).\\
$\obs(c)$ & Protected-LLM view $(\Mem_{\E},H(c),x(c),h)$
(Def.~\ref{S-def:obs}).\\
$H(c);\ x(c)$ & handle structure (Rule~\ref{S-r:names}); trusted
command input of configuration $c$ (Def.~\ref{S-def:obs}).\\
$\eqU$ & $\U$-equivalence: equality of $\obs$, arbitrary $\Mem_{\U}$
(Def.~\ref{S-def:lowequiv}).\\
$c_1\approx c_2$ & matched configurations: equal view and equal
$P$-interaction history (Def.~\ref{S-def:matched}).\\
$\sinks(\rho),\ \DT(\rho)$ & ungated sink projection; display transcript
of run $\rho$ (Defs.~\ref{S-def:sinkproj},~\ref{S-def:transcript}).\\
$\omega$ & coupled coin sequence for $P$ and $\RE$
(\S\ref{S-app:conventions}).\\
$\ccand,\ \cbatch$ & retrieval surfacing cap; per-page batch-confirmation
cap.\\
\ref{S-o:one}--\ref{S-o:three} & optional hardening: sink-scoped $\T$ (\ref{S-o:one}) and exact-target
data-tool mediation (\ref{S-o:two}), formally instantiated; local scope expiry
(leases; \ref{S-o:three}, requirements-level, \ref{S-o:three}).\\
\midrule
\multicolumn{2}{@{}l}{\emph{Federated identities and promotion
(Appendix~\ref{S-app:fedmodel})}}\\
\midrule
$w_i\in[\wmin,\wmax]$ & attestation weight of identity $i$; $\wseat$
panel-seat weight floor.\\
$f;\ \fs$ & adversarial identities the adversary \emph{ever} controls;
those ever meeting the seat floor ($\fs\le f$): lifetime totals,
static corruption (\S\ref{S-app:fedcore}).\\
$\Wadv$ & adversary's lifetime controllable attestation weight
($\le f\wmax$).\\
$W,\ N$ & promotion threshold at a tool: weight / distinct-endorser
count.\\
$\Ncap$ & credit cap $\Dpool\lceil N/\Dpool\rceil$: credited
endorsements per (entry, tool) (Def.~\ref{S-def:counting}(f));
$N\le\Ncap<N+\Dpool$.\\
$\Dpool$ & pool-diversity floor: endorsers span $\ge\Dpool$ pools, caps
$\lceil N/\Dpool\rceil$, $\lceil k/\Dpool\rceil$
(Def.~\ref{S-def:pools}).\\
$\Twin$ & finalisation-window length of a pipeline entry.\\
$n_v;\ n_{v,t}$ (rv) & \emph{latent} demanders of item $v$ / of the pair
$(v,t)$ in an epoch, reference model (\S\ref{S-app:budgetmodel}); $n_{v,t}\le n_v$,
$\sum_t n_{v,t}\le\ctool\,n_v$ under \ref{S-a:toolspread}; realized counts $n^{\mathrm{log}}_v,n^{\mathrm{log}}_{v,t}$ (Def.~\ref{S-def:demand}, \S\ref{S-app:prob}) are dominated pathwise: $n^{\mathrm{log}}_v\le n_v$, $n^{\mathrm{log}}_{v,t}\le n_{v,t}$.\\
$\Emax$ & per-epoch cap on pipeline entries and on non-vindicated
disputes per item (Rule~\ref{S-r:hygiene}).\\
$\rsub$ & per-identity submissions per epoch (Rule~\ref{S-r:share}).\\
$\Texp,\ \Tpan$ & expiry clock (no demand / dormancy); fast-track
suspension-panel target.\\
$H_{\mathrm{acc}}$ & total honest identities ever accepting a given byte
string (Cor.~\ref{S-cor:submission}).\\
\midrule
\multicolumn{2}{@{}l}{\emph{Panel, dispute, accountability}}\\
\midrule
$k;\ m;\ m_{\mathrm{rej}}$ & panel verdict cap; vindication bar; eviction
bar (default $m_{\mathrm{rej}}=k{+}1{-}m$).\\
$\qpre,\ \qpan$ (emp) & weakest-target conditional lower bounds:
honest fresh gate inspection of a poisoned item (Def.~\ref{S-def:bad}) ends in an effective
qualifying rejection ($\qpre$); an honest eligible verdict rejects it
($\qpan$); abstentions $=$ misses; authority-inactive inspections $=$ excluded trials (\cref{S-n:qpos}); single-parameter reading
$q=\min(\qpre,\qpan)$ (Rem.~\ref{S-rem:qsingle}).\\
$p_b$ (emp) & lower bound: honest fresh inspection accepts a benign item (Def.~\ref{S-def:bad}),
conditionally on the past.\\
$q_c$ (emp) & residual catch rate under engineered cross-person
correlation (\ref{S-q:corr}).\\
$\delta=\tfrac{m-\fs}{k-\fs}-(1-\qpan)$ & panel safety margin
(Thm.~\ref{S-thm:panel}).\\
$\delta_b=p_b-\tfrac{m}{k-\fs}$ & wrongful-eviction margin
(Prop.~\ref{S-prop:live}).\\
$B_{\mathrm{att}}$ & certified per-attempt ceiling: the exact-binomial
right-hand side of Theorem~\ref{S-thm:panel}, $=$ ordinary term $+$ panel
term (Rem.~\ref{S-rem:batt}); the panel term is a floor in $N$ at fixed
panel geometry; never the Hoeffding relaxation.\\
$\rseat$ (emp) & lower bound: next demander of a disputed item casts an
eligible verdict, conditionally on the past (Assumption \ref{S-a:rhoseat}).\\
$\kappa=(k{+}1)/\rseat$ & expected demanders during an unresolved
dispute (Lem.~\ref{S-lem:geo}).\\
$\rmax$ & open (unresolved) rejections per identity.\\
$\thacc,\ \cceil$ & rate trigger (per fresh inspection, calibrated
$>1-p_b$) and hard per-epoch contradiction ceiling
(Rule~\ref{S-r:panel}(v)).\\
$\cacc=\cceil+\rmax$ & deterministic per-identity bound, per filing
epoch, on effective
contradicted dispute-initiations (Lem.~\ref{S-lem:cacc}).\\
$\cpr$ & voluntary re-inspection prompts per suspension: $\le\cpr k$.\\
$\kappa'=\kappa+\cpr k$ & pacing constant with suspension prompts
absorbed (Lem.~\ref{S-lem:constants}).\\
\midrule
\multicolumn{2}{@{}l}{\emph{Budget (Appendix~\ref{S-app:budget})}}\\
\midrule
$M;\ V(M);\ \beta$ & population; expected distinct demanded items;
Heaps exponent $V(M)=\Theta(M^\beta)$.\\
$p_{u,v,t};\ p_{v,t};\ p_v$ & per-user pair-demand probability; population-average pair / item demand probabilities per epoch (heterogeneous
IRM, \S\ref{S-app:budgetmodel}); $p_{v,t}\le p_v$.\\
$d=\sum_v p_v;\ d_{\mathrm{tool}}=\sum_{v,t}p_{v,t}$ & item-level and
tool-indexed single-epoch demand mass; $d_{\mathrm{tool}}\le\ctool\,d$
under \ref{S-a:toolspread}
(Cor.~\ref{S-cor:floorspread}); $d_{\mathrm{tool}}$ is the isolated
single-user shareable cost (Thm.~\ref{S-thm:floor}(ii)); uniformly bounded along the population scaling under \ref{S-a:massbound}.\\
$p^{\infty}_{v,t};\ d^{\infty}_{\mathrm{tool}}$ & lifetime (ever-demand)
pair probability and mass (\S\ref{S-app:budgetmodel}); distinct from the
single-epoch mass.\\
$\dpriv$ & personal cap: expected distinct personal \emph{(value, tool)
pairs} a user ever confirms, once each (lifetime; per-pair per
\ref{S-inv:floor}).\\
$C_u,\ \bar C_M,\ C_v$ (rv) & user $u$'s shareable confirmations;
population mean; confirmations attributable to $v$.\\
required / voluntary / redundant & gate fires for the user's own use of
an uncovered \emph{pair}; accepting decision at an interpreter-issued
invitation surface; accepting decision at a covered gate
(handle-expressed reuse of a covered pair, protocol-inert, excluded
from $C_u$) (\S\ref{S-app:budgetmodel}, Rule~\ref{S-r:panel}(i)).\\
$\gbar;\ \gamma_v;\ \szero$ & front-loading bound (fraction of an item's
demand per window or dormant interval); item's realised fraction;
additive slack (\ref{S-a:smooth}).\\
$A(M);\ \Bfl$ & demand-independent per-item cost; front-loading
coefficient (Lem.~\ref{S-lem:budget}; explicit forms
\S\ref{S-app:constants}).\\
$D_0(M),\ \varepsilon(M),\ \eta$ & coverage sufficiency threshold,
failure rate, slack (\ref{S-a:promsuf}).\\
$\ctool$ & distinct tools per item (\ref{S-a:toolspread}).\\
$g_N,g_k$ (sch) & margin schedules: $N=f+g_N$, $k=\fs+g_k$
(Thm.~\ref{S-thm:coscale}).\\
$\Elease;\ \Erv$ & \ref{S-o:three} lease periods per item ($\bigO(1)$;
extension-level accounting, \ref{S-o:three}(e)); re-verification
rounds (\ref{S-n:revoke}).\\
\midrule
\multicolumn{2}{@{}l}{\emph{Probability (Appendix~\ref{S-app:toolbox})}}\\
\midrule
$(\mathcal{F}_s)$ & filtration of logged protocol and display events;
adversary strategies predictable with respect to it.\\
$s$ & discrete time index of the filtration $(\mathcal F_s)$; the
proof-local step counter of Thm.~\ref{S-thm:reason} is unrelated.\\
$\tau_i,\ Z_i,\ Y_i$ (rv) & inspection stopping times; decision
indicators; coupled Bernoulli majorants/minorants
(Lem.~\ref{S-lem:adaptive}).\\
$U_i$ (rv) & i.i.d.\ uniform coupling variables (Lem.~\ref{S-lem:adaptive}).\\
$\mathrm{Bin},\ \mathrm{Geom}$ & binomial; geometric (number of trials).\\
$\lambda_{c_0}$ & tail constant of Lem.~\ref{S-lem:geo}.\\
\midrule
\multicolumn{2}{@{}l}{\emph{Blast-schedule analysis
(\S\ref{S-app:blastlaw}; cited by no proof)}}\\
\midrule
$\pi_{\mathrm{ord}},\ \pi_{\mathrm{pan}}$ & realised promotion
probability of the blast schedule by route, at reach $R$
(Prop.~\ref{S-prop:blastlaw}); sum $\pi_{\mathrm{blast}}$; \emph{not} the
terms of $B_{\mathrm{att}}$ (Rem.~\ref{S-rem:blastnotbound}); always
subscripted, cf.\ $\pi(b)$.\\
$S(J)$ & capped generating sum of the panel component, $J$ post-freeze
decisions (Prop.~\ref{S-prop:blastlaw}).\\
\midrule
\multicolumn{2}{@{}l}{\emph{Zipf case study (Appendix~\ref{S-app:zipf})}}\\
\midrule
$\alpha,\ Z,\ R_M,\ H_M,\ L_M,\ \mu_{\mathrm{tail}},\ \hat N$ & rank
exponent; normaliser; head boundary; head/tail sets; expected distinct
tail demand; per-item cap $\hat N:=A(M)$.\\
\bottomrule
\end{longtable}
\endgroup

\noindent\emph{Disambiguation.} $h$ (decision history) vs.\ $h_j$
(handles): subscripted occurrences are always handles; $H(c)$ (handle
structure), $H_{\mathrm{acc}}$ (honest acceptors), and $H_M$ (Zipf
head set) are unrelated to one another and to $h$. $\C$ (label) vs.\
$C_u,C_v,\bar C_M$ (confirmation counts): the label is sans-serif and
unsubscripted. $Z$ (Zipf normaliser) vs.\ $Z_i$ (decision indicators):
subscripted occurrences are always indicators. $\pi_{\mathrm{ord}},\pi_{\mathrm{pan}},\pi_{\mathrm{blast}}$
(blast-schedule promotion probabilities, \S\ref{S-app:blastlaw}) vs.\
$\pi(b)$ (interpreter-only block provenance,
Definition~\ref{S-def:config}): the former are always subscripted and take
a reach argument, the latter never is; no bare $\pi(\cdot)$ denotes a
probability anywhere in this document. $S(J)$ (blast generating sum) is
unrelated to $S_{\mathrm{anc}}$ (anchored attempts,
Cor.~\ref{S-cor:submission}), to $S_{\mathrm{sh}}(L)$ (shared-tier replay,
Definition~\ref{S-def:fedconfig}), and to the bare $S$ of
\ref{S-a:narr} (a crystallized summary, which takes no argument).

\subsection{Standing conventions}
\label{S-app:conventions}

\begin{enumerate}[label=(C\arabic*),ref=C\arabic*,leftmargin=2.4em,itemsep=3pt]
\item\label{S-conv:c1} \emph{Determinism modulo coins.} A single coin sequence $\omega$
jointly covers the sampling randomness of $P$ and of $\RE$; given
$\omega$, both are deterministic functions of their inputs. All two-run
comparisons (Thm.~\ref{S-thm:reason}, Prop.~\ref{S-prop:fed},
Prop.~\ref{S-prop:knowledge}) couple $\omega$, the user's typed stream
(\ref{S-a:exo}), and (in the federated setting) the exogenous federation-log prefix (\ref{S-a:fedlog}; local suffixes derived equal, Def.~\ref{S-def:fedstep}),
identically across the compared runs.
\item\label{S-conv:c2} \emph{Probability space and adversary.} All federated-protocol randomness
lives on the filtration $(\mathcal{F}_s)$ generated by logged protocol
events and display events; the adversary's strategy (presentation,
submission, stopping, self-dispute) is predictable with respect to it.
The construction, and the two structural constraints on adaptivity it
implies, are given once in \S\ref{S-app:prob} and used everywhere.
\item\label{S-conv:c3} \emph{Log order.} The federation maintains a single linearizable,
append-only log of protocol events: a TCB \emph{log service} (a trusted
sequencer or a consensus layer) supplies the total order, and its history
is materialized as a Certificate-Transparency-style Merkle tree for
tamper-evident audit --- the tree authenticates the history; the
sequencing service, assumed correct as TCB, provides the linearizability.
It supplies the total order on decisions, demand events, and verdicts
that Rules \ref{S-r:share}--\ref{S-r:hygiene} invoke.
``The federation log,'' ``the log,'' and ``log order'' all refer to
this object and the order it induces. The hash-chained audit log that
materializes the display transcript (Definition~\ref{S-def:transcript})
is a distinct, per-deployment TCB component and is never called ``the
log.''
\item\label{S-conv:c4} \emph{Epochs, leases, rounds.} Three multiplicities the
development keeps distinct.
\emph{(a)~Epochs.} An epoch is a fixed span of the
federation log (a fixed number of sequence numbers or a fixed calendar
period), common to all items and identities, readable off the log (TCB);
every renewable per-epoch quantity ($\rsub$, $\Emax$, notification
deduplication, the accountability allowances of Rule~\ref{S-r:panel}(v),
(vii)) is indexed to it.
\emph{(b)~Leases.} A lease is the deployment-selected local scope
lifetime of the requirements-level option \ref{S-o:three} (\ref{S-o:three}; not
part of the formal semantics); $\Elease=\bigO(1)$ counts lease
periods per (user, item) in the extension-level accounting of
\ref{S-o:three}(e).
\emph{(c)~Rounds.} A re-verification round re-presents a
community-confirmed item to inspectors with no prior decision on its
bytes; $\Erv$ counts rounds.
\emph{(d)~Permanence and boundaries.} Decision freshness
(Def.~\ref{S-def:fresh}) and verdict capacity
(Rule~\ref{S-r:panel}(iii)) are \emph{permanent} and attach to none of the
three: they never renew. Promotion weight and banked credits persist
across epoch boundaries until the item retires. Finalisation windows are
contained within epochs: a window that would cross a boundary closes at
the boundary, banking its accrued endorsements to the item's next entry
exactly as a dormant-interval decision would
(Def.~\ref{S-def:counting}); a boundary never finalises a promotion early
and never discards honest effort. A boundary-spanning dispute counts
toward Rule~\ref{S-r:hygiene}(i) in the epoch in which it opened.
\item\label{S-conv:c5} \emph{Byte identity.} Items, freshness, disputes, and panels are
keyed to exact byte strings. Paraphrase variants are distinct items
(consequences: \ref{S-q:paraphrase}).
\item\label{S-conv:c6} \emph{Confirmations.} A \emph{confirmation} is a user approval---an accepting gate
decision; it is durable (its fresh endorsed copy is written even when
another review aborts the enclosing invocation, Rule~\ref{S-r:atomic})
and is counted once per (value, tool) pair per gate transaction
(Definition~\ref{S-def:sinkconf}). Rejecting decisions, abandoned
dialogs, batch-element page approvals, and repeated within-transaction
reviews of an already-accepted pair are attention cost, counted
separately (\ref{S-n:effort}). \emph{Required}, \emph{voluntary}, and
\emph{redundant} confirmations are defined in \S\ref{S-app:budgetmodel}:
the confirmation cap (Thm.~\ref{S-thm:floor}) counts required
pair-confirmations, the ledger (Lem.~\ref{S-lem:budget}) required and
voluntary; redundant confirmations are protocol-inert
(Rule~\ref{S-r:panel}(i)) and excluded from both. Here, required means
needed for the user's intended use, not an obligation to accept.
\item\label{S-conv:c7} \emph{Store idealization.} The store is idealized as unbounded: no
operation's defined behaviour depends on aggregate store volume. A
deployment enforcing quotas must apply them to the untrusted tier by
silent eviction, never by an error visible to $P$ (otherwise store
exhaustion reopens exactly the channel Rule~\ref{S-r:nobit} closes; \ref{S-q:quota}).
\item\label{S-conv:c8} \emph{Constants vs.\ schedules.} $N,k,m,W,\Dpool,\Twin,\Texp$,
$\Emax,\rsub,\rmax,\cceil,\cbatch,\ccand,\cpr$ are fixed TCB
configuration at any deployment; Appendix~\ref{S-app:budget} additionally
studies \emph{schedules} $N(M),k(M),\dots$ along a growing population,
and every scheduled claim is confined to the corridor stated there.
All caps and multiplicity constants are positive integers where
integral; in particular $\Emax\ge1$ and $\ctool\ge1$ (used by
Lemma~\ref{S-lem:constants}).
\item\label{S-conv:c9} \emph{Environment scheduling.} An interpreter step has a fixed
slot structure: first the channel slots deliver at most one typed event
(\textsc{W-Typed}) and at most one paste event (\textsc{W-Paste}), in
that order; then $P$ emits exactly one operation, whose environment
interaction (tool returns, reader outputs, retriever selection, gate
decisions) occurs synchronously within that operation's atomic
transition. Two-run comparisons (Theorem~\ref{S-thm:reason},
Proposition~\ref{S-prop:fed}) couple only this \emph{slot skeleton} and
the typed stream (\ref{S-a:exo}): paste \emph{presence and content} are
uncoupled and may differ arbitrarily across the compared attacks, as
may every tool output and reader output. Channel-slot transitions
acknowledge the channel, never $P$: nothing is appended to $P$'s reply
stream by \textsc{W-Typed} or \textsc{W-Paste}
(Figure~\ref{S-fig:sos2}, convention~\hyperref[S-sos:vi]{(vi)}). Two-run comparisons are always taken at a \emph{common horizon}: the
compared runs consist of the same number of interpreter slots, and
transcript equality means equality of the two full transcripts at that
horizon.
\end{enumerate}

\subsection{Named invariants and structural properties}
\label{S-app:invariants}

The following six named facts are cited constantly and are collected here;
together they are the structural core of the design. They are
heterogeneous in kind, and each entry declares its kind: \emph{rule
property} (a direct consequence of the cited rule and definition clauses);
\emph{conditional inductive invariant} (preserved by every transition,
under the named first-party exceptions); \emph{protocol property} (a fact
of the counting and pacing rules); \emph{bound metaproperty} (a property
of the \emph{bounds} of the cited theorems, holding under those theorems'
hypotheses). None is an axiom: each is established at the location cited,
and nothing beyond the cited hypotheses is assumed. Invariants
\ref{S-inv:fresh}--\ref{S-inv:reach} are stated in the vocabulary of the
federated protocol, defined in Appendix~\ref{S-app:fedmodel}.

\begin{invx}[Byte-permanence of freshness]\label{S-inv:fresh}
An attested identity's first \emph{logged} decision on a byte string
--- its first decision taken while the bytes have a shared-tier
presence (Definition~\ref{S-def:fresh}) --- is its only weight-bearing
inspection of those bytes, ever: every later accepting decision by
that identity (at any tool, in any pipeline entry, in any epoch, in
any later shared-tier presence of the same bytes) carries no
promotion weight and no verdict capacity. Freshness is decidable
from the federation log (convention~(\ref{S-conv:c3})) and never renews; personal
decisions outside any shared-tier presence are not logged and are
outside this guarantee (Definition~\ref{S-def:fresh}).
Kind: rule property. \emph{Established by} Definition~\ref{S-def:fresh} and the counting rule
Definition~\ref{S-def:counting}. (Byte-keyed: paraphrase variants are
distinct items, (\ref{S-conv:c5}), \ref{S-q:paraphrase}.)
\end{invx}

\begin{invx}[Rejection asymmetry]\label{S-inv:rejasym}
A rejecting decision can freeze promotion, corroborate an open
dispute, or trigger post-promotion review; it can never promote and
never vindicate. A \emph{repeat} acceptance carries no promotion
weight and no verdict (\ref{S-inv:fresh}). Hence re-exposure of bytes
to a \emph{previously logged-decided} inspector can only add
rejection opportunities; exposure to a log-fresh inspector adds both
a rejection opportunity and a possible weight-bearing acceptance,
priced by the hypotheses of Theorem~\ref{S-thm:panel}.
Kind: rule property. \emph{Established by}
Rule~\ref{S-r:panel}(i)--(iii) with Definition~\ref{S-def:fresh}.
\end{invx}

\begin{invx}[Single-approval cap, per (user, value, tool)]\label{S-inv:floor}
No user is ever required to confirm a given \emph{(value, tool) pair}
more than once over the deployment's lifetime: a user's first
accepting review of value $v$ at tool $t$ --- at any gate transaction,
including one whose enclosing invocation aborts (Rule~\ref{S-r:atomic})
--- durably writes the local $\C$ copy with $t\in\scopef$, which
covers all their later use of $v$ \emph{at $t$} and survives every
third-party protocol event; and within one transaction, confirmations
are counted once per pair (Definition~\ref{S-def:sinkconf}). A first use
of the same bytes at a different tool $t'$ is a distinct required
confirmation, of the pair $(v,t')$, itself once ever; the pairs per
item are bounded by \ref{S-a:toolspread} in the budget model. The disclosed exception within the formal semantics is first-party:
the owner's own agent's ungated $\mathsf{evict}$ (\ref{S-n:evict}); a
deployment adopting a conforming instantiation of the
requirements-level Option~\ref{S-o:three} additionally admits lease
expiry, an extension outside this document's semantics
(\ref{S-o:three}). Kind: conditional inductive invariant (exception:
first-party evict, \ref{S-n:evict}; extension-level: conforming-\ref{S-o:three}
lease expiry, \ref{S-o:three}). \emph{Established by}
Lemma~\ref{S-lem:scopepersist} (from Rules~\ref{S-r:atomic},
\ref{S-r:whitelist}, \ref{S-a:label}--\ref{S-a:persist}, and
Lemma~\ref{S-lem:nodelete}); reasoning pairs covered per the reasoning-coverage clause of Rule~\ref{S-r:panel}(i); \emph{consumed by} Theorem~\ref{S-thm:floor}
and Lemma~\ref{S-lem:budget}(i).
\end{invx}

\begin{invx}[Banking safety]\label{S-inv:bank}
A banked credit (a first decision logged during dormancy or during a
freeze with no open entry) raises only the \emph{starting} weight of the
item's next pipeline entry. Nothing finalises promotion except an open
entry's window end, so banking never bypasses a window's rejection
opportunity and never accelerates finalisation past a window boundary; and
each banked decision remains its identity's single fresh inspection of the
bytes, so successive entries' weight-bearing sets stay identity-disjoint
(Lemma~\ref{S-lem:disjoint}). Banked credits expire if the item retires
first. Every banked decision is logged after its attempt's anchor (Definition~\ref{S-def:attempt}), so banking never lets a credited decision precede the conditioning point of Theorem~\ref{S-thm:panel}. Kind: protocol property. \emph{Established by} Definition~\ref{S-def:counting} with
Rule~\ref{S-r:panel}(iv).
\end{invx}

\begin{invx}[Verdict non-renewal]\label{S-inv:verdict}
An identity casts at most one verdict on a given dispute instance, for
the instance's entire life, across any number of epoch boundaries.
Verdict capacity is consumed only by cast verdicts (abandonment
consumes nothing) and, like freshness, never renews.
Kind: rule property. \emph{Established by} Rule~\ref{S-r:panel}(iii).
\end{invx}

\begin{invx}[Reach-freeness]\label{S-inv:reach}
The federated safety bounds depend only on protocol constants
($N,k,m,\Dpool$) and the adversary's resources ($f,\fs,\Wadv$), never on
the number of honest identities to whom a poisoned item is presented:
the same stated upper bound applies at every reach. (No
monotonicity in reach is claimed: presenting the bytes to log-fresh
inspectors creates both rejection opportunities and possible
weight-bearing acceptances, \ref{S-inv:rejasym}; what is proved is
that the bound does not degrade with reach.) Reach-freeness is relative to the stated parameters: the bounds'
\emph{form} is reach-independent, while the behavioural constants
$\qpre,\qpan$ are weakest-target infima over all cohorts the adversary
can reach (\cref{S-n:qpos}), so widening reach can lower the applicable
constants without changing the bound's form. Kind: bound metaproperty. The property is
Definition~\ref{S-def:reachfree}; it is a property of the stated bounds of Theorems~\ref{S-thm:det} and~\ref{S-thm:panel}, holding under their hypotheses, not a property of reachable configurations; contrast with the
reach-dependent no-dispute baseline, Theorem~\ref{S-thm:prob}.
\end{invx}

\section{Formal Specification of the Memory Interface}
\label{S-app:interface}

PerMIT places no restriction on which tools the Protected LLM may invoke
or on how it reasons; the \emph{harness} (the fixed
interpreter code of Figures~\ref{S-fig:sos}--\ref{S-fig:sos2} together
with the display and logging paths) bounds two things only: what $P$ can
\emph{observe}, and what may \emph{cross} a tool boundary. The bounded
surface is the memory interface: the $P$-facing operation
surface
\begin{gather*}
\Sigma_P \;=\; \Sigma\;\cup\;\{\mathsf{tool}(\bar a):\
\mathsf{tool}\in\SensT\cup\DataT\}\;\cup\;\{\mathsf{retrieve}\},\\
\Sigma=\{\mathsf{write}_P,\mathsf{recall},\mathsf{fill},
\mathsf{crystallize},\mathsf{display},\conf,\mathsf{evict}\},
\end{gather*}
($\mathsf{write}$, where it appears in prose, abbreviates the sole
$P$-emitted write operation $\mathsf{write}_P$.)
together with one interpreter-internal derived form ($\mathsf{map}$).
Every completeness claim below (Lemmas~\ref{S-lem:iface},~\ref{S-lem:cflow})
enumerates over $\Sigma_P$, which coincides arm-for-arm with the mechanized
operation type (\texttt{POp}, \S\ref{S-app:mechanization}).
Because this surface is finite, its \emph{interface transition schema}
is specified \emph{clause-complete at the interface alphabet}: every operation of $\Sigma_P$
and channel event has its
store effect, reply, transcript events, and sanitized-history events fixed by Figures~\ref{S-fig:sos}--\ref{S-fig:sos2} with these rules, with
totality per Rule~\ref{S-r:iface}(b); the full run state these transitions
carry is fixed by Definition~\ref{S-def:extconfig}. What the figures deliberately do not formalise is
$P$'s internal computation (a strategy, \S\ref{S-app:config}) and the
UI-side rendering internals of Rule~\ref{S-r:display}; the channel
analysis of Appendix~\ref{S-app:single} needs only the interface schema,
and is verified clause by clause against it, with the totality clause of Rule~\ref{S-r:iface} ensuring
that no invocation has undefined or distinguishing behaviour.
(Model lineage and design rationale: \ref{S-n:lineage}.)

\subsection{Configurations, labels, and roles}
\label{S-app:config}

\begin{definition}[Trust lattice and labels]\label{S-def:lattice}
The integrity lattice has two levels, $\U\sqsubset\E$. Each block carries
a label $\ell(b)\in\{\U,\T,\C,\CC,\PC\}$; the level map $\lvl$ sends
$\lvl(\U)=\U$ and $\lvl(\ell)=\E$ for
$\ell\in\{\T,\C,\CC,\PC\}$. The four $\E$-level labels are
provenance-and-reuse tags, not a finer trust ordering: $\T$ (user-typed;
reusable by its owner), $\C$ (single-user confirmed; admissible only to
the tools in its scope, reusable without re-confirmation only by its
owner), $\CC$ (community-confirmed, Appendix~\ref{S-app:fedmodel};
admissible only to the exact targets endorsed (tools, or the reasoning pseudo-target, \S\ref{S-app:fedcore}), reusable by anyone), $\PC$
($P$-composed; empty tool scope, hence never eligible for the sensitive
fast path of Rule~\ref{S-r:whitelist}). Label combination is the meet on
levels: an operation combining inputs yields level $\U$ if any input is
at level $\U$, else level $\E$; no operation in $\Sigma$ raises an
untrusted value to level $\E$ by combination.
\end{definition}

\begin{definition}[Store and configuration]\label{S-def:config}
A block is $b=(\mathrm{id},\mathrm{bytes}(b),\ell(b),\pi(b),\scopef(b))$
with interpreter-only provenance
$\pi(b)\in\{\mathsf{typed},\mathsf{pasted},\mathsf{ingress},
\mathsf{cryst},\mathsf{pcomposed},\mathsf{confirmed},\mathsf{cache}\}$,
assigned only by the interpreter (provenance is interpreter-private
and not part of $\obs$), and, for endorsed blocks, a tool
scope $\scopef(b)\subseteq\SensT\cup\DataT$ (a set of individual tool
identifiers). Every block-creating transition of
Figures~\ref{S-fig:sos}--\ref{S-fig:sos2} instantiates one of the total
constructors
\begin{align*}
\operatorname{newT}(v)&:=(\mathrm{id}^{\E}_{\bullet},v,\T,\mathsf{typed},\varnothing), &
\operatorname{newPC}(v)&:=(\mathrm{id}^{\E}_{\bullet},v,\PC,\mathsf{pcomposed},\varnothing),\\
\operatorname{newC}(v,t)&:=(\mathrm{id}^{\E}_{\bullet},v,\C,\mathsf{confirmed},\sigma(t)), &
\operatorname{newCache}(v,t)&:=(\mathrm{id}^{\E}_{\bullet},v,\C,\mathsf{cache},\sigma(t)),\\
\operatorname{newU}(v,\pi)&:=(\mathrm{id}^{\U}_{\bullet},v,\U,\pi,\varnothing),&&
\end{align*}
where $\mathrm{id}^{\E}_{\bullet}$ (resp.\ $\mathrm{id}^{\U}_{\bullet}$)
is the next identifier of the endorsed (resp.\ untrusted) sort
(Rule~\ref{S-r:names}(g)) and $\sigma(t)=\{t\}$ for a tool target,
$\sigma(\mathrm{reasoning})=\varnothing$; the cache confirmations of
Options~\ref{S-o:one}--\ref{S-o:two} use $\operatorname{newCache}$, all
other accepting confirmations $\operatorname{newC}$. Tool identifiers
are canonical, fixed TCB configuration, so scope membership is a
decidable syntactic test. Block identifiers are two-sorted and
operation-determined (Rule~\ref{S-r:names}(g)); the sorts separate the
tiers: an endorsed-sort identifier resolves only to an $\E$-level
block, an untrusted-sort identifier only to a $\U$-block.

A configuration is $c=(\Mem,\sigma_P,\sigma_Q,h)$: the store, the
Protected-LLM state, the Reader state, and the interpreter's sanitized
decision history $h$: a sequence of (target, decision) records, one
appended at each reached gating review (one per reached gating
position; for a batch, one per gating non-$\diamond$ body position plus
one aggregate element decision) --- never bytes, never a block
identifier of either sort, never a selection ordinal, never a candidate
multiplicity, collection cardinality, or page count; selections,
candidate lists, page renderings, and per-page decisions are recorded
in the display transcript only (Definition~\ref{S-def:transcript}). A
null-handle abort appends the same (target, reject) record a user
rejection appends (Rule~\ref{S-r:null}). Hence $h$ contains no
$\U$-bytes, exposes no untrusted-sort identifier
(Rule~\ref{S-r:names}(g)), and carries no occupancy, cardinality, or
pagination information (Rules~\ref{S-r:null}, \ref{S-r:nobit}); this is the
form machine-checked as \texttt{HEvent} (\S\ref{S-app:mechanization}).
$\Mem_{\U}$ and $\Mem_{\E}$ denote the restrictions of $\Mem$ to levels
$\U$ and $\E$.
\end{definition}

\paragraph{Endorsement unit.}
The endorsement unit throughout is the \emph{(value, tool) pair}: every
confirmation names an exact byte string and an exact tool, and there is
no endorsement transfer between tools, however similar.
(Exact-target discipline governs the sensitive fast path and all
endorsement, promotion, and counting; the \emph{baseline} data-tool
path of Rule~\ref{S-r:whitelist}(b) is scope-blind,
Corollary~\ref{S-cor:noexfil}(b), and Option~\ref{S-o:two} extends the
exact-target discipline to data tools.)
Combinations,
orderings, and timings of separately endorsed values are
ungated by design (rationale and residual: \ref{S-n:unit}, \ref{S-q:recombine};
hardening: Option~\ref{S-o:one}).

\paragraph{Individuation of trusted blocks.}
$\T$-blocks are individuated at write time: each field the user
explicitly supplies (or, absent field structure, the command as one
atomic unit) is one $\T$-block. The individuating parser is fixed
interpreter code inside the TCB; the fast-path test of
Rule~\ref{S-r:whitelist} is byte-equality against these stored blocks.

\paragraph{Two scopes and one selector.}
The agent is a transition system over configurations whose steps are
tool invocations, control flow, and the operations of
Figures~\ref{S-fig:sos}--\ref{S-fig:sos2}, executed in one of two scopes: the
\emph{privileged scope} $P$ may invoke any tool and reaches memory only
through $\Sigma$. Throughout the formal development $P$ is modelled as a
\emph{strategy}: a function from its interaction history, its own
emitted operations, the reply stream of Rule~\ref{S-r:iface}, its view
$\obs(c)$, and the coins $\omega$ (convention~(\ref{S-conv:c1})), to its next
operation. Its internal state $\sigma_P$ is therefore an abbreviation
for that history and carries no independent degrees of freedom: two
runs agreeing on the history agree on $\sigma_P$ by definition. This is
not a restriction on $P$ (the strategy is arbitrary, adversarial
included) but the standard determinism-modulo-coins reading of (\ref{S-conv:c1}),
and it is the form in which $P$ is machine-checked
(\S\ref{S-app:mechanization}).
The \emph{quarantined scope} $Q$ has read-only access
to the whole store and no tool access. A third model role, the
\emph{trusted-store retriever} $\RE$ (Rule~\ref{S-r:retrieve}), is a
stateless selector over $\Mem_{\E}$: it reads only the endorsed store
and returns only block identifiers, which the interpreter dereferences.
(The asymmetry between $Q$'s and $\RE$'s vision is forced by the
theorems: \ref{S-n:readers}.)

\begin{definition}[Protected-LLM view]\label{S-def:obs}
For a configuration $c$, let $H(c)$ denote the handle structure of
Rule~\ref{S-r:names} and $x(c)$ the trusted command input received so
far; $h$ is the decision history. The \emph{Protected-LLM view} is
\[
\obs(c)\;=\;\big(\Mem_{\E},\;H(c),\;x(c),\;h\big).
\]
By Figures~\ref{S-fig:sos}--\ref{S-fig:sos2} and Rules \ref{S-r:names}--\ref{S-r:retrieve}, $\obs(c)$
together with the reply stream of Rule~\ref{S-r:iface} is the entirety
of what $P$ can observe (Lemma~\ref{S-lem:iface}).
\end{definition}

\begin{remark}[Information interface vs.\ effect interface]\label{S-rem:obsview}
$\Sigma$ is the interface of \emph{effects}; $\obs(c)$ is the interface of
\emph{information}. Granting $P$ direct sight of $\Mem_{\E}$ in its view is
a deliberate, conservative strengthening: every theorem quantifies over all
strategies of this better-informed $P$, so it a fortiori covers any $P$
that must acquire the same data operationally, and the two-run theorem is
\emph{harder} under it (a more informed $P$ has more ways to diverge).
$\mathsf{recall}$ on endorsed identifiers is consequently informationally
redundant and is retained as the operational access path; the sentence
``$P$ reaches memory only through $\Sigma$'' concerns effects, not
information. The mechanization adopts exactly this reading
(\S\ref{S-app:mechanization}: \texttt{obs} is computed through
\texttt{memEOf}).
\end{remark}

\begin{definition}[Sensitive sinks, gate transactions, and confirmations]
\label{S-def:sinkconf}
A \emph{sensitive sink event} is a call
$\mathsf{tool}(a_1,\dots,a_n)$ with $\mathsf{tool}\in\SensT$.

A \emph{gating position} of a tool invocation is an argument position
that does not pass the fast path of Rule~\ref{S-r:whitelist}. The
invocation's gating positions are processed left to right as one atomic
\emph{gate transaction} (Rule~\ref{S-r:atomic}): at each reached position
the interpreter resolves the argument, displays its bytes with zoned
provenance (Rule~\ref{S-r:display}), and reads the user's selection (if
among candidates) and decision $z\in\{\mathrm{accept},\mathrm{reject}\}$;
a null resolution renders the fixed no-value display of
Rule~\ref{S-r:null} and takes the standard reject outcome, consuming no
decision; a rejecting outcome aborts the review of the remaining
positions.

An accepting review of resolved bytes $v$ at target $t$ is a
\emph{confirmation event} $\conf(v,t)$. Confirmations are
\emph{durable}: at the close of the atomic transaction the interpreter
writes, for every accepted gating position, one fresh endorsed copy of
its displayed bytes scoped to the target
(Figure~\ref{S-fig:sos2}, convention~\hyperref[S-sos:vii]{(vii)}), \emph{whether or not the
transaction as a whole succeeds}. The tool itself executes iff every
gating position was accepted and none was null; on execution the bytes
that flow at each position are exactly the bytes displayed for it, and
on abort no argument flows.

\emph{Counting unit.} For all accounting
(Appendices~\ref{S-app:fedmodel}, \ref{S-app:budget}), confirmations are
counted per \emph{(value, tool) pair per gate transaction}: accepting
reviews of the same exact pair at several positions of one transaction
are one pair-confirmation; the repeated reviews themselves are
attention cost (\ref{S-n:effort}).

A standalone $\conf(v,t)$ with $t$ a tool or ``reasoning'' is the
single-position special case. The ``publish'' target is not a local
confirmation: it creates no local block and is defined by the
\textsc{Publish} transitions of Figure~\ref{S-fig:sos2} and
Rule~\ref{S-r:share}. Batch-element page approvals under
Rule~\ref{S-r:batchgate} are call-local: they authorize the displayed
elements for that $\mathsf{map}$ instance only and create no durable
copy; gating non-$\diamond$ arguments of a map body are ordinary
durable confirmations (convention~\hyperref[S-sos:vii-prime]{(vii${}'$)}).
\end{definition}
\subsection{Operational semantics}
\label{S-app:sos}

Figures~\ref{S-fig:sos}--\ref{S-fig:sos2} give the \emph{interface transition
schema} of the memory interface: complete at the level of the interface
alphabet (every operation's store effect and reply is fixed here, with
totality per Rule~\ref{S-r:iface}(b)), with the remaining components of the
run state --- fixed by Definition~\ref{S-def:extconfig} --- evolving by the
deterministic bookkeeping stated in the rules and conventions.
Every transition is fixed interpreter code; no model
participates in any label assignment or any reply. Replies to $P$ are
drawn from the grammar at the top of Figure~\ref{S-fig:sos} and nothing else
(Rule~\ref{S-r:iface}); the side conditions cite the Rules of
\S\ref{S-app:rules}, which constrain naming, atomicity, display, and
gating. Transitions triggered by the user or the environment (channel
events, gate decisions, tool returns) are marked with their trigger;
all others are $P$-emitted.

\begin{definition}[Extended configuration]\label{S-def:extconfig}
The full run state comprises: the store $\Mem$; the two identifier
allocators of Rule~\ref{S-r:names}(g) (endorsed-sort and untrusted-sort); the
interpreter-private handle-resolution table (index $=$ handle identifier;
never read by any reply, Rule~\ref{S-r:nobit}); $P$'s reply stream
(Rule~\ref{S-r:iface}); the decision history $h$; the trusted command input
$x(c)$; the executed-tool, confirmation, and reader-event counters that
index the environment's oracles; and three ghost logs used only by the
theorems --- the display transcript $\DT$
(Definition~\ref{S-def:transcript}), the sensitive-sink log
(Definition~\ref{S-def:sinkproj}), and $P$'s emitted-operation log.
$P$-visible components are exactly $\obs(c)$ and the reply stream
(Lemma~\ref{S-lem:iface}); all others are interpreter-private or ghost. A
block is \emph{readable} iff it is at level $\E$, i.e.\ dereferenceable to
bytes by $\mathsf{recall}$ (\textsc{Recall-E}); $\U$-blocks are never
readable to $P$. For the map-free fragment, the executable interpreter of
\S\ref{S-app:mechanization} realizes this extended configuration verbatim
(\texttt{IState}), and the figures are its human-readable projection onto
the store and reply components.
\end{definition}

\begin{figure}[btp]
\footnotesize
\noindent\textbf{Replies to $P$}\hfill(Rule~\ref{S-r:iface})
\begin{align*}
r \;::=\; \mathsf{ack} \;\mid\; h_j \;\mid\; \mathsf{fail}
\;\mid\; \mathsf{dec}(\mathrm{accept}) \;\mid\;
\mathsf{dec}(\mathrm{reject}) \;\mid\; \mathsf{bytes}(b),\
b\in\Mem_{\E}
\end{align*}

\smallskip\noindent Replies to $P$ form a \emph{stream}; the grammar above
is its \emph{alphabet}. A single transition may append more than one token
(e.g.\ \textsc{Tool-Gate-Acc} appends
$\mathsf{dec}(\mathrm{accept})\cdot\mathsf{ack}$), per convention~\hyperref[S-sos:iii]{(iii)}.

\noindent\textbf{Judgment.}\quad
$\Mem \xrightarrow{\;o\;\Downarrow\;r\;} \Mem'$: operation $o$ yields
reply $r$; components of $c$ other than $\Mem$ are suppressed when
unchanged; $h_j$ is the $j$-th operation-determined handle
(Rule~\ref{S-r:names}). Transitions triggered by the user or the
environment are marked by their trigger; all others are $P$-emitted.
Conclusions suppress the operation label over the arrow where the
premises display it; the components of the extended configuration
(Definition~\ref{S-def:extconfig}) other than $\Mem$ evolve by the
deterministic bookkeeping stated in the rules and conventions, and the
figures are the store/reply projection of that full-state transition,
realized executably (for the map-free fragment) by the mechanization
(\texttt{IState}, \S\ref{S-app:mechanization}).

\medskip
\noindent\textit{Tool invocation.}
\begin{mathpar}[\lineskip=0.2em]
\inferrule*[lab=Tool-Fast]
  {P\text{ emits }\mathsf{tool}(\bar a) \\
   \text{every } a_i \text{ passes the fast path of
   Rule~\ref{S-r:whitelist}} \\
   \text{environment returns } o\ \text{(attacker-chosen bytes and
   count)}}
  {\Mem \xrightarrow{\;\mathsf{tool}(\bar a)\,\Downarrow\,
   \mathsf{ack}\;} \Mem\cup\mathrm{capture}(o)}
\and
\inferrule*[lab=Tool-Gate-Acc,
            right={\tiny atomic (\ref{S-r:atomic}); fresh ids (\ref{S-r:names}(g))}]
  {P\text{ emits }\mathsf{tool}(\bar a) \\
   \text{some } a_i \text{ gate (Rule~\ref{S-r:whitelist})} \\
   \text{every gating position is confirmed and accepted
   (Rules~\ref{S-r:atomic},\,\ref{S-r:display})} \\
   \text{environment returns } o\ \text{(attacker-chosen bytes and
   count)} \\
   \Mem' = \Mem\cup\bigl\{\operatorname{newC}(\text{displayed }a_i,
   \mathsf{tool}):\ \substack{\text{per accepted gating}\\\text{position (conv.~\hyperref[S-sos:vii]{(vii)})}}\bigr\}}
  {\Mem \xrightarrow{\;\mathsf{tool}(\bar a)\,\Downarrow\,
   \mathsf{dec}(\mathrm{accept})\cdot\mathsf{ack}\;}
   \Mem'\cup\mathrm{capture}(o)}
\and
\inferrule*[lab=Tool-Gate-Rej,
            right={\tiny\shortstack[l]{no invocation; accepted positions\\
            endorsed (\ref{S-r:atomic}); $h$ extended}}]
  {P\text{ emits }\mathsf{tool}(\bar a) \\
   \text{gate fires; some gating position is rejected by the user, or
   null-aborts (Rule~\ref{S-r:null})} \\
   \Mem' = \Mem\cup\bigl\{\operatorname{newC}(\text{displayed }a_i,
   \mathsf{tool}):\ \substack{\text{per gating position}\\\text{accepted before the abort}}\bigr\}}
  {\Mem \xrightarrow{\;\mathsf{tool}(\bar a)\,\Downarrow\,\mathsf{dec}(\mathrm{reject})\;}
   \Mem'}
\end{mathpar}
\noindent\textit{Writes.}
\begin{mathpar}[\lineskip=0.2em]
\inferrule*[lab=W-Typed]
  {\text{channel event: }\mathrm{typed}(x) \\
   \mathrm{parse}_{\T}(x)=b_1,\dots,b_n\ (\text{TCB})}
  {\Mem \xrightarrow{\;\Downarrow\,\mathsf{ack}\;}
   \Mem\cup\{\operatorname{newT}(b_1),\dots,\operatorname{newT}(b_n)\}}
\and
\inferrule*[lab=W-Paste]
  {\text{channel event: }\mathrm{paste}(x)}
  {\Mem \xrightarrow{\;\Downarrow\,\mathsf{ack}\;}
   \Mem\cup\{\operatorname{newU}(b_x,\mathsf{pasted})\}}
\and
\inferrule*[lab=W-P]
  {b\in\sigma_P}
  {\Mem \xrightarrow{\;\mathsf{write}_P(b)\,\Downarrow\,\mathsf{ack}\;}
   \Mem\cup\{\operatorname{newPC}(b)\}}
\end{mathpar}
\noindent\textit{Memory reads and display.}
\begin{mathpar}[\lineskip=0.2em]
\inferrule*[lab=Recall-E]
  {\mathrm{id}\ \text{resolves to a readable } \E\text{-block } b}
  {\Mem \xrightarrow{\;\mathsf{recall}(\mathrm{id})\,\Downarrow\,
   \mathsf{bytes}(b)\;} \Mem}
\and
\inferrule*[lab=Recall-O]
  {\text{otherwise: } \U\text{-block, unknown, or nonexistent
   (uniformly)}}
  {\Mem \xrightarrow{\;\mathsf{recall}(\mathrm{id})\,\Downarrow\,
   h_j\;} \Mem}
\and
\inferrule*[lab=Retrieve, right={verbatim dereference (\ref{S-r:retrieve})}]
{P\ \text{emits } \mathsf{retrieve}(q) \\
 \RE(q,\Mem_{\E},\omega) \text{ returns an id resolving to }
 b\in\Mem_{\E}}
{\Mem \xrightarrow{\;\Downarrow\,\mathsf{bytes}(b)\;} \Mem}
\and
\inferrule*[lab=Retrieve-F]
{\RE \text{ returns no id, or an id not resolving in } \Mem_{\E}}
{\Mem \xrightarrow{\;\mathsf{retrieve}(q)\,\Downarrow\,\mathsf{fail}\;}
\Mem}
\and
\inferrule*[lab=Display]
  {\text{rendering per Rule~\ref{S-r:display}; nothing enters }
   \sigma_P}
  {\Mem \xrightarrow{\;\mathsf{display}(x)\,\Downarrow\,
   \mathsf{ack}\;} \Mem}
\end{mathpar}
\caption{Interface transition schema of the memory interface (1/2): tool
invocation, writes, memory reads, and display. Every transition is
fixed interpreter code; no model assigns any label or composes any
reply. Continued in Figure~\ref{S-fig:sos2} (search, crystallization,
confirmation, eviction, iteration, and failure); conventions for both
figures follow Figure~\ref{S-fig:sos2}.}
\label{S-fig:sos}
\end{figure}
\begin{figure}[tp]
\scriptsize
\noindent\textit{Search and crystallization.}
\begin{mathpar}[\lineskip=0.05em]
\inferrule*[lab=Fill,
            right={real/null indistinguishable
            (\ref{S-r:null},\,\ref{S-r:nobit})}]
  {Q\ \text{searches } \Mem_{\U} \\
   \text{interpreter binds} \le\ccand \text{ candidates internally
   (possibly none)}}
  {\Mem \xrightarrow{\;\mathsf{fill}(q)\,\Downarrow\,h_j\;} \Mem}
\and
\inferrule*[lab=Cryst]
  {Q\ \text{writes schema-bound summary } s\ \text{interpreter-side}}
  {\Mem \xrightarrow{\;\mathsf{crystallize}(t)\,\Downarrow\,h_j\;}
   \Mem\cup\{\operatorname{newU}(s,\mathsf{cryst})\}}
\end{mathpar}
\vspace{-9pt}
\noindent\textit{Confirmation and eviction.}
\begin{mathpar}[\lineskip=0.05em]
\inferrule*[lab=Conf-Acc,
            right={\tiny $h$ extended; fresh copy, scope $\sigma(t)$ (conv.~\hyperref[S-sos:vii]{(vii)})}]
  {\text{gate } \conf(v,t),\quad t\ \text{a tool or reasoning} \\
   \text{user selects and accepts on the decision pane
   (Rule~\ref{S-r:display})}}
  {\Mem \xrightarrow{\;\Downarrow\,\mathsf{dec}(\mathrm{accept})\;}
   \Mem\cup\bigl\{\operatorname{newC}(v,t)\bigr\}}
\and
\inferrule*[lab=Conf-Rej,
            right={\tiny bytes not returned; $h$ extended}]
  {\text{gate } \conf(v,t),\quad t\ \text{a tool or reasoning} \\
   \text{user rejects, or null handle (Rule~\ref{S-r:null})}}
  {\Mem \xrightarrow{\;\Downarrow\,\mathsf{dec}(\mathrm{reject})\;}
   \Mem}
\and
\inferrule*[lab=Publish-Acc,
  right={\tiny shared-tier act; no local block, no endorsed id (\ref{S-r:share})}]
  {\text{gate } \conf(v,\mathrm{publish}) \\
   v\ \text{byte-equals a local }\C\text{-block of the owner with
   provenance }\mathsf{confirmed} \\
   \text{owner accepts on the decision pane (Rule~\ref{S-r:display})}}
  {\Mem \xrightarrow{\;\Downarrow\,\mathsf{dec}(\mathrm{accept})\;}
   \Mem\quad\text{emitting }\mathsf{submit}(v)\ \text{to the federation
   log}}
\and
\inferrule*[lab=Publish-Rej,
  right={\tiny uniform reject; ineligibility, rejection, null alike
  (\ref{S-r:nobit})}]
  {\text{gate } \conf(v,\mathrm{publish}) \\
   \text{owner rejects, or } v\ \text{is ineligible
   (Rule~\ref{S-r:share}), or null (Rule~\ref{S-r:null})}}
  {\Mem \xrightarrow{\;\Downarrow\,\mathsf{dec}(\mathrm{reject})\;}
   \Mem}
\and
\inferrule*[lab=Evict,
  right={\tiny uniform reply; no label change}]
{P\ \text{emits } \mathsf{evict}(\mathrm{id}) \\
 \text{interpreter-side: if } \mathrm{id} \text{ resolves to a block
 of the owner's local store,}\\
 \text{that block is removed; otherwise the operation is a no-op} \\
 \text{resolution status consumed interpreter-side
 (Rules~\ref{S-r:nobit},\,\ref{S-r:iface}(b))}}
{\Mem \xrightarrow{\;\mathsf{evict}(\mathrm{id})\,\Downarrow\,
 \mathsf{ack}\;} \Mem\ominus\mathrm{id}}
\end{mathpar}
\vspace{-9pt}
\noindent\textit{Iteration and failure.}
\begin{mathpar}[\lineskip=0.05em]
\inferrule*[lab=Map-Tool-Acc, right={\tiny atomic batch (\ref{S-r:atomic},\,\ref{S-r:batchgate}); fresh copies (conv.\,\hyperref[S-sos:vii]{(vii)})}]
  {P\ \text{emits }\mathsf{map}(\hat o,H),\quad
   \hat o=\mathsf{tool}(\dots,\diamond,\dots)\ \text{well-formed (Rule~\ref{S-r:names}(f))} \\
   \text{batch confirmation accepted on every page (Rule~\ref{S-r:batchgate})} \\
   \text{environment returns } o_1,\dots,o_{|H|}\ \text{per element
   call (attacker-chosen bytes and count)} \\
   \Mem'=\Mem\cup\bigl\{\operatorname{newC}(a_i,\mathsf{tool}):\
   a_i\ \text{an accepted gating non-}\diamond\ \text{argument}\bigr\}}
  {\Mem \xrightarrow{\;\Downarrow\,\mathsf{dec}(\mathrm{accept})\cdot\mathsf{ack}\;}
   \Mem'\cup\mathrm{capture}(o_1,\dots,o_{|H|})}
\and
\inferrule*[lab=Map-Tool-Rej, right={\tiny\shortstack[l]{no element executes; non-$\diamond$\\
  endorsements persist (\ref{S-r:atomic}); null collection\\
  (\ref{S-r:names}(a),\,\ref{S-r:null})}}]
  {P\ \text{emits }\mathsf{map}(\hat o,H),\ \hat o\ \text{of tool form} \\
   \text{some page rejected; null collection (Rule~\ref{S-r:null}); or over-cap collection (Rule~\ref{S-r:batchgate}(b))} \\
   \Mem'=\Mem\cup\bigl\{\operatorname{newC}(a_i,\mathsf{tool}):\
   a_i\ \text{an accepted gating non-}\diamond\ \text{argument}\bigr\}}
  {\Mem \xrightarrow{\;\Downarrow\,\mathsf{dec}(\mathrm{reject})\;} \Mem'}
\and
\inferrule*[lab=Map-Cryst, right={\tiny no decision; one handle, one ack (\ref{S-r:names}(c))}]
  {P\ \text{emits }\mathsf{map}(\hat o,H),\quad \hat o=\mathsf{crystallize}(\diamond) \\
   Q\ \text{writes schema-bound summaries } s_1,\dots,s_{|H|}\
   \text{interpreter-side, one per element (\textsc{Cryst})}}
  {\Mem \xrightarrow{\;\Downarrow\,\mathsf{ack}\;}
   \Mem\cup\{s_1,\dots,s_{|H|}:\ \ell=\U\}}
\and
\inferrule*[lab=Fail]
  {P\ \text{emits well-formed } o \\
   \text{a precondition fails, deterministically in } P\text{'s
   emitted operations and } \Mem_{\E} \text{ alone
   (Rule~\ref{S-r:nobit})}}
  {\Mem \xrightarrow{\;o\,\Downarrow\,\mathsf{fail}\;} \Mem}
\end{mathpar}
\caption{Interface transition schema of the memory interface (2/2): search,
crystallization, confirmation, eviction, iteration, and failure,
continued from Figure~\ref{S-fig:sos}. Every transition is fixed
interpreter code; no model assigns any label or composes any reply.
Conventions for both figures follow below.}
\label{S-fig:sos2}
\end{figure}
\FloatBarrier

\noindent\textit{Conventions (apply to Figures~\ref{S-fig:sos}--\ref{S-fig:sos2}).}
\phantomsection\label{S-sos:i}(i)~$\mathrm{capture}(o)$ stores the outputs $o$ as $\U$-blocks
$\operatorname{newU}(\cdot,\mathsf{ingress})$ bound
to a single fresh collection handle $h_j$
(Rules~\ref{S-r:names},\,\ref{S-r:ingress}); the identifier is
operation-determined, computable by $P$, and never returned.
A call whose captured output count is zero binds its collection
handle with \emph{null} resolution (Rule~\ref{S-r:null}): a
zero-cardinality capture and an empty search result are the same
null-resolution case. The capture operation exposes no direct
cardinality or resolution bit to $P$ (Rule~\ref{S-r:nobit}); any later
distinction surfaces only through the logged null-collection display
and its standard aggregated reject (Rule~\ref{S-r:batchgate}(b)).
\phantomsection\label{S-sos:ii}(ii)~Element calls inside \textsc{Map} are interpreter-iterated, not
$P$-emitted: individually neither handle-producing nor acknowledged.
If the argument collection has null resolution (an empty search
result or zero-output capture, convention~\hyperref[S-sos:i]{(i)}), the tool-free form
$\mathsf{map}(\mathsf{crystallize}(\diamond),H)$ writes no
summaries, binds the map's operation-determined output collection
handle with null resolution
(Rules~\ref{S-r:names}(a),(c),~\ref{S-r:null}), returns the ordinary
$\mathsf{ack}$, and appends no display or decision event; the
tool-bearing form takes the null-collection batch treatment of
Rule~\ref{S-r:batchgate}(b).
\phantomsection\label{S-sos:iii}(iii)~In \textsc{Tool-Gate-Acc} and \textsc{Tool-Gate-Rej}, gating
positions are reviewed left to right; one decision is recorded per
reached gating position, and the displayed bytes of an executed call
are exactly the bytes that flow; the entire step is atomic
(Rule~\ref{S-r:atomic}). Every accepted position's fresh $\C$ copy is
written at the close of the step, on the accepting and the aborting
branch alike; rejected and null positions never flow and write nothing.
The reply stream carries \emph{one} decision token per gated
invocation, regardless of the number of gating positions; per-position
decisions are recorded in the transcript and in $h$ only
(Rule~\ref{S-r:iface}(a)(i),(iv)). Whether an invocation is gated at all
is a deterministic function of $P$'s own emitted argument list and
$\Mem_{\E}$ (Rule~\ref{S-r:whitelist}), hence already $P$-computable; the
token-shape difference between \textsc{Tool-Fast},
\textsc{Tool-Gate-Acc}, and \textsc{Tool-Gate-Rej} therefore carries no
information beyond trusted data and the logged decisions
(Rule~\ref{S-r:nobit}) (the $\CC$ branch's live-entry bit is $\E$-level
log data, coupled under \ref{S-a:fedlog} in two-run comparisons (exogenous prefix coupled; local suffixes derived equal, Def.~\ref{S-def:fedstep})), and
within the reject class, user rejection and null-handle abort remain
indistinguishable in the reply and in $h$ (Rule~\ref{S-r:null}).
\phantomsection\label{S-sos:iv}(iv)~\emph{Totality} (Rule~\ref{S-r:iface}(b)): every well-formed
invocation with any argument is covered: unresolvable arguments take
\textsc{Recall-O}-style opaque-handle replies, failed deterministic
preconditions take \textsc{Fail}; no invocation is undefined and none
distinguishes attacker-influenced states.
\phantomsection\label{S-sos:v}(v)~$\Mem\ominus\mathrm{id}$ denotes $\Mem\setminus\{\mathrm{id}\}$
when $\mathrm{id}$ resolves to a block of the owner's local store,
and $\Mem$ otherwise. The reply is the constant $\mathsf{ack}$ in
every case; the realised outcome (removal of the owner's block, at either level;
no-op on an identifier not resolving to a block of the owner's local
store, including an unoccupied one) is
recorded in the audit log interpreter-side and never returned to
$P$. The effect on $\Mem_{\E}$ is $P$-visible trusted data
(Rule~\ref{S-r:nobit}); any effect on $\Mem_{\U}$, and the absence of
any effect, are indistinguishable to $P$. A \emph{handle} argument
to $\mathsf{evict}$ is a deterministic malformation of $P$'s own
emission and takes the generic failure outcome
(Rule~\ref{S-r:iface}(a)(iii)).
\phantomsection\label{S-sos:vi}(vi)~Channel-triggered transitions (\textsc{W-Typed}, \textsc{W-Paste})
acknowledge the channel; nothing is appended to $P$'s reply stream, so
the presence or absence of a paste event is not $P$-observable
(Rule~\ref{S-r:nobit}, convention~(\ref{S-conv:c9})).
\phantomsection\label{S-sos:vii}(vii)~\emph{Fresh-copy endorsement; block immutability.} An accepting
confirmation never relabels a block in place: \textsc{Conf-Acc} and
\textsc{Tool-Gate-Acc} write, per accepted value, one \emph{fresh}
$\C$-labelled block carrying exactly the displayed bytes, at the next
endorsed-sort identifier (Rule~\ref{S-r:names}(g)), with scope
$\sigma(t)=\{t\}$ for a tool target and
$\sigma(\text{reasoning})=\varnothing$ (readable, but never eligible
for the sensitive fast path of Rule~\ref{S-r:whitelist}); the resolved
source block, where one exists, is unchanged and remains at its
level, and the publish target (Rule~\ref{S-r:share}) is a shared-tier
act creating no local block. Endorsement is \emph{durable}: the accepted positions' fresh copies are
written at the close of the atomic step whether or not the enclosing
invocation executes (Rule~\ref{S-r:atomic}). Consequently no transition of
Figures~\ref{S-fig:sos}--\ref{S-fig:sos2} alters an existing block's
bytes, label, or scope: blocks are immutable once written, and the
store changes only by creation of fresh blocks and by \textsc{Evict}.
\phantomsection\label{S-sos:vii-prime}(vii${}'$)\,\emph{Batch acceptance is call-local for elements.} An
accepting batch confirmation (Rule~\ref{S-r:batchgate}) authorizes each
$\diamond$-element's bytes for its own element call of that
$\mathsf{map}$ instance only: no persistent endorsed copy of any
element is created, and the elements remain $\U$-labelled in
$\Mem_{\U}$. Fresh $\C$-copies are written for the accepted
\emph{gating non-$\diamond$ arguments}, one each, scoped
$\{\mathsf{tool}\}$, exactly as at \textsc{Tool-Gate-Acc}, and durably so: they persist
even when a later element page rejects and no element call executes
(\textsc{Map-Tool-Rej}). This is
forced by the identifier discipline of Rule~\ref{S-r:names}(g):
per-element copies would allocate $|H|$ endorsed-sort identifiers
with $|H|$ attacker-chosen, exposing an attacker-dependent quantity
in $\Mem_{\E}$; the non-$\diamond$ count is fixed by $P$'s emitted
body. Convention~\hyperref[S-sos:vii]{(vii)}'s per-accepted-value fresh copy, and the
endorsed-copy clause of Definition~\ref{S-def:sinkconf}, are read with
this carve-out: they govern \textsc{Conf-Acc} and
\textsc{Tool-Gate-Acc}; batch-accepted elements flow under
Rule~\ref{S-r:atomic}'s displayed-bytes binding without entering
$\Mem_{\E}$. Consequently re-processing the same elements in a later
$\mathsf{map}$ is gated afresh --- call-local authorization caches
nothing --- and the single-approval cap (\ref{S-inv:floor}) does not
extend to map elements, whose confirmations are per-task cost
(Appendix~\ref{S-app:budget}, Scope).

\begin{commentary}[the \textsc{Map} rules and the mechanization boundary]
The split fixes, per body form, the decision token (none for the tool-free
form), the store effect ($\mathrm{capture}$ of element outputs vs.\
schema-bound $\U$-summaries $s_1,\dots,s_{|H|}$ written as in \textsc{Cryst}),
and the endorsement effect (fresh $\C$-copies for gating non-$\diamond$
arguments only; elements are authorized call-locally, convention~\hyperref[S-sos:vii-prime]{(vii${}'$)}). In every
case, including rejection, exactly one collection-handle index is consumed,
on rejection with null resolution --- so handle structure is independent of
the outcome (Rule~\ref{S-r:names}(a)), closing the identifier channel a
skipped index would open. $\mathsf{map}$ is normative here; its batch-confirmation surface ---
pagination, per-page decisions, the aggregated decision, all-or-nothing
displayed-bytes flow, and batch determinacy --- is mechanized as a
standalone layer (\texttt{BatchGate.lean}, \S\ref{S-app:mechanization}),
while its store-level integration into the run-level interpreter remains
outside the mechanized surface (delta~(\ref{S-delta:6})).
\end{commentary}

\FloatBarrier
\subsection{Interface rules}
\label{S-app:rules}

The rules below constrain the transitions of
Figures~\ref{S-fig:sos}--\ref{S-fig:sos2}. Each clause is normative and
implementable as stated; all rationale lives in
Appendix~\ref{S-app:shield} under the keys cited, or in the marked
commentary blocks of \S\ref{S-app:roadmap}.

\begin{rulex}[Operation-determined naming: handles and block identifiers]\label{S-r:names}
\leavevmode
\begin{enumerate}[label=(\alph*),leftmargin=1.8em,itemsep=1pt,topsep=1pt]
\item Handle identifiers (including collection-handle identifiers) are
drawn from the sequence of $P$-emitted operations alone: the $j$-th
handle-producing operation of a run binds identifier $h_j$. Identifiers
never encode, count, or depend on store contents.
\item Handle-producing operations: ingress capture (the triggering tool
invocation is $P$-emitted; $P$ computes the collection handle's
identifier from its own operation count; nothing is returned);
$\mathsf{recall}$ returning a handle; $\mathsf{fill}$;
$\mathsf{crystallize}$; $\mathsf{map}$.
\item A $\mathsf{map}$ invocation is exactly one handle-producing
operation regardless of collection cardinality: its per-element outputs
are captured as $\U$-blocks bound to a single fresh collection handle
determined by the $\mathsf{map}$'s position in $P$'s operation sequence.
Element calls are interpreter-iterated, not $P$-emitted, and are
individually neither handle-producing nor acknowledged.
\item $P$ reaches captured element outputs only through
$\mathsf{fill}$/$\mathsf{crystallize}$ searches. No handle identifier
depends on collection size.
\item A collection handle admits exactly two uses: as the argument of
$\mathsf{map}$, or as a tool argument --- the latter defined by
desugaring: an invocation $\mathsf{tool}(\dots,h,\dots)$ with a
collection-handle argument $h$ \emph{is} the instance
$\mathsf{map}(\mathsf{tool}(\dots,\diamond,\dots),h)$, the desugaring
applied before the invocation matches any transition of
Figures~\ref{S-fig:sos}--\ref{S-fig:sos2} (it never instantiates
\textsc{Tool-Fast} or \textsc{Tool-Gate-*}), batch-gated per
Rule~\ref{S-r:batchgate} and subject to the body grammar of clause~(f)
(in particular at most one collection handle per invocation and no
other handle argument; violations take the generic failure outcome,
deterministic in $P$'s own emission, Rule~\ref{S-r:nobit}). A
collection handle exposes no length, no indexing, and no per-element
access.
\item \emph{Map bodies.} The body $\hat o$ of
$\mathsf{map}(\hat o,H)$ is a single operation template containing
exactly one occurrence of the distinguished element placeholder
$\diamond$, of one of exactly two forms: (i)~one tool invocation
$\mathsf{tool}(a_1,\dots,a_n)$ in which one argument is $\diamond$
and every other argument is a $P$-supplied non-handle value, subject
per argument to the ordinary mediation of Rule~\ref{S-r:whitelist}
(fast-path values flow; a gating argument is confirmed once for the
whole batch, at the batch confirmation of Rule~\ref{S-r:batchgate},
alongside the elements); or (ii)~one tool-free
$\mathsf{crystallize}(\diamond)$ template. The body may not contain
$\mathsf{map}$, $\mathsf{recall}$, $\mathsf{fill}$,
$\mathsf{evict}$, $\conf$, $\mathsf{display}$, the write operation $\mathsf{write}_P$, or a
handle argument
(the interpreter-side captures of forms (i)--(ii) are transitions of
the interpreter, not body operations); nesting is not defined. Well-formedness is a
syntactic test on $P$'s emitted operation alone, and a body outside
this grammar takes the generic failure outcome, deterministic in
$P$'s own emission (Rule~\ref{S-r:nobit}). Multi-step per-element
pipelines are expressed as successive $\mathsf{map}$s: each
$\mathsf{map}$ yields a fresh collection handle (clause~(c)), which
clause~(e) admits as the argument of a further $\mathsf{map}$, each
step batch-gated independently; the grammar restricts the shape of
one step, never the depth of a pipeline.
\item \emph{Two-sorted block identifiers.} Block identifiers are drawn
from two disjoint sorts, \emph{endorsed}
($\mathrm{id}^{\E}_0,\mathrm{id}^{\E}_1,\dots$) and \emph{untrusted}
($\mathrm{id}^{\U}_0,\mathrm{id}^{\U}_1,\dots$), each allocated in
sequence. Endorsed-sort identifiers are allocated by exactly the local
endorsed-block-creating transitions: \textsc{W-Typed}, \textsc{W-P},
\textsc{Conf-Acc}, and, per \emph{accepted gating position},
\textsc{Tool-Gate-Acc}, \textsc{Tool-Gate-Rej}, \textsc{Map-Tool-Acc},
and \textsc{Map-Tool-Rej} (for maps, gating non-$\diamond$ body
positions only --- a count fixed by $P$'s emitted operation, never by
the collection cardinality, convention~\hyperref[S-sos:vii-prime]{(vii${}'$)}). A publish decision
allocates no identifier of either sort (\textsc{Publish-Acc}). The
$j$-th endorsed block of a run sits at $\mathrm{id}^{\E}_j$, a position
determined by the coupled event sequence (the typed stream, $P$'s
emitted operations, and the logged decisions) alone, hence computable
by $P$ and identical across runs whose coupled events agree.
Untrusted-sort identifiers are allocated by the untrusted-writing
transitions (\textsc{W-Paste}, ingress capture, \textsc{Cryst}) and may
depend on attacker-chosen multiplicities; no untrusted-sort identifier
is ever exposed to $P$ by any reply or view
(Rules~\ref{S-r:null},~\ref{S-r:nobit}, \ref{S-r:iface}(b)). An endorsed-sort
identifier never resolves to a $\U$-block and an untrusted-sort
identifier never resolves to an $\E$-block, so the sorts separate the
tiers syntactically. Like handle identifiers, block identifiers of
either sort never encode, count, or depend on store contents beyond
this allocation order.
\end{enumerate}
\end{rulex}

\begin{commentary}[role of the identifier discipline]
Clause~(g) closes the identifier channel that in-place endorsement
would open: identifiers visible to $P$ (those of $\Mem_{\E}$) depend
only on coupled events, while every attacker-dependent multiplicity
can influence only untrusted-sort identifiers, which no reply and no
view exposes. It is consumed by Theorem~\ref{S-thm:reason} (Step~\hyperref[S-proof:reason-step-4]{4}) and
is machine-checked as the two-sorted id space of the mechanization
(\S\ref{S-app:mechanization}).
\end{commentary}

\begin{rulex}[Atomic gating; durable per-value endorsement]\label{S-r:atomic}
When confirmations gate a tool call, the reviews of all gating
positions, the writing of the accepted positions' fresh endorsed
copies, and the invocation (or its abort) form one atomic interpreter
step: no step may interleave between display, decision, endorsement,
and flow. The bytes that flow at an executed call are the displayed
bytes, bound to that call instance. If any gating position is rejected
or null, the invocation aborts entire: no argument flows; the accepted
positions' endorsed copies are nevertheless written, their accepting
decisions being recorded in the transcript like any confirmation. When
a batch confirmation gates a $\mathsf{map}$ (Rule~\ref{S-r:batchgate}),
the aggregated accepting decision and the batch of element invocations
form one atomic step, and the bytes that flow at each element call are
exactly the element bytes displayed for that call; accepted gating
non-$\diamond$ arguments are endorsed durably exactly as at a direct
call, whatever the element pages' outcome.
\end{rulex}

\begin{commentary}[severability and durable endorsement]
Soundness of durable endorsement on an aborting transaction is the
severability clause of \ref{S-a:user}: intent is per displayed value,
severally. A commit-on-success variant --- discarding accepted
positions' copies on abort --- would be safety-monotone but
\emph{sacrifices the once-per-pair cap} (\ref{S-inv:floor}): a
transaction that accepts one pair and aborts on a later position
leaves no durable copy, so re-presentations of the same pair re-force
it without bound. Durability is adopted precisely because the cap is
load-bearing (Theorem~\ref{S-thm:floor}).
\end{commentary}

\begin{rulex}[Ingress: all tool outputs captured]\label{S-r:ingress}
Tools partition into $\SensT$ (sensitive, effectful) and $\DataT$ (data
readers); the partition and the canonical tool identifiers are fixed
deployment configuration, per capability, inside the TCB. The
interpreter captures \emph{every} tool output, including the status and
error responses of sensitive calls, to $\Mem_{\U}$ with label $\U$, and
returns to $P$ only a single constant acknowledgement token per
invocation: never the bytes, never a success bit, never a count, size,
or identifier derived from captured content.
\end{rulex}

\begin{rulex}[Whitelist mediation at every tool boundary]\label{S-r:whitelist}
\leavevmode
\begin{enumerate}[label=(\alph*),leftmargin=1.8em,itemsep=1pt,topsep=1pt]
\item A value $a$ passed to a tool in $\SensT$ flows without
confirmation \emph{iff} $a$ consists of readable $\E$-level bytes and
the interpreter verifies that $a$ byte-equals a stored $\T$-block, or
byte-equals some $\C/\CC$-block $b$ with
$\mathsf{tool}\in\scopef(b)$ (for $\CC$, the matched block is the live
shared-tier entry, checked valid at call time; \S\ref{S-app:fedcore}). ($\PC$ blocks have empty scope and never
pass this test.)
\item A value passed to a tool in $\DataT$ flows without confirmation
iff it consists of readable $\E$-level bytes.
\item A handle argument never flows silently to any tool: it always
triggers $\conf(a,\mathsf{tool})$, at data tools exactly as at
sensitive ones, regardless of whether its resolved bytes coincide with
stored endorsed content. There are no per-tool exceptions.
\item The fast-path check is membership in $\Mem_{\E}$ plus a scope
test, not taint-tracking of $P$'s code.
\end{enumerate}
\end{rulex}

\begin{rulex}[Readable local endorsement]\label{S-r:readable}
Every accepting confirmation with target
$t\in\SensT\cup\DataT\cup\{\mathrm{reasoning}\}$ writes a fresh block
$\operatorname{newC}(v,t)$ (Options~\ref{S-o:one}--\ref{S-o:two}:
$\operatorname{newCache}(v,t)$) carrying exactly the displayed bytes,
labelled $\C$ (or, under community promotion, the shared tier carries
$\CC$ status; Appendix~\ref{S-app:fedmodel}), at the next endorsed-sort
identifier (Rule~\ref{S-r:names}(g)); its bytes are readable by $P$ via
$\mathsf{recall}$, the identifier being computable by $P$ from its own
coupled event history. The originating untrusted block, where one
exists, is unchanged and remains untrusted. Before confirmation, a
crystallized summary is $\U$-labelled, lives reader/interpreter-side,
and is not in $\sigma_P$. An accepting \emph{publish} decision creates
no local block and allocates no endorsed-sort identifier: it emits only
the shared-tier submission event of Rule~\ref{S-r:share}
(\textsc{Publish-Acc}).
\end{rulex}

\begin{rulex}[Null handle]\label{S-r:null}
A $\mathsf{fill}$ or $\mathsf{crystallize}$ search over $\Mem_{\U}$ that
finds nothing returns a \emph{null handle}: the handle's existence and
identifier are fixed by the requesting operation (Rule~\ref{S-r:names});
only its resolution differs, and that difference is hidden from $P$.
Likewise, $\mathsf{recall}$ on an identifier that does not resolve to a
readable $\E$-block returns a handle of the same kind: whether it
resolves to $\U$-bytes or to nothing (in particular, whether the
identifier is occupied at all) is hidden from $P$. If a null handle is
passed to any tool, the gate fires as usual (Rule~\ref{S-r:whitelist}),
the decision pane renders a fixed ``no value resolved'' display,
recorded in the display transcript like any confirmation event, the call
aborts, and $P$ receives the standard reject
decision, indistinguishable in
the reply stream and in the sanitized history $h$ from a user
rejection (Definition~\ref{S-def:config}); the no-value display itself
is recorded on the transcript like any confirmation event.
\end{rulex}

\begin{rulex}[Failure indistinguishability]\label{S-r:nobit}
The success/failure of a $\U$-store search, and the resolution status
of any $\mathsf{recall}$, is consumed interpreter-side to choose the
null-vs-real handle; no reply and no sanitized-history record exposes
it. A handle's private resolution may later be displayed at a gate:
that display and the user's logged decision are an explicit transcript
release (Definition~\ref{S-def:transcript}), not a failure channel. Any reader error, reader
crash, or reader budget exhaustion arising during an operation over
$\Mem_{\U}$ is likewise consumed interpreter-side: the operation returns
the same operation-determined handle it would return on success
(Rules~\ref{S-r:names},~\ref{S-r:null}), whose resolution to bytes, to
nothing, or to an aborted computation is hidden from $P$. The generic
failure outcome ($\mathsf{fail}$, Figures~\ref{S-fig:sos}--\ref{S-fig:sos2}) is reserved for
failures that are a deterministic function of $P$'s own emitted
operations, $\Mem_{\E}$, and the coupled coins $\omega$ alone
(malformed invocations, failed preconditions on $\sigma_P$, and the
retrieval failure of \textsc{Retrieve-F}). Search outcomes over $\Mem_{\E}$ are trusted data
and may be exposed. (Rationale and the store-quota corollary:
\ref{S-n:fail}, \ref{S-q:quota}, convention~(\ref{S-conv:c7}).)
\end{rulex}

\begin{rulex}[Faithful display on a split, zoned surface]\label{S-r:display}
\leavevmode
\begin{enumerate}[label=(\alph*),leftmargin=1.8em,itemsep=1pt,topsep=1pt]
\item The display surface is split into a \emph{decision pane} and
auxiliary \emph{selection surfaces}.
\item The decision pane shows exactly the resolved bytes that will flow
(or be recorded as endorsed content), plus a provenance line in two
visually separated zones: (i) \emph{attested facts} (channel, label
history, timestamps, signature-verification status), rendered in an
interpreter-controlled region: TCB-authenticated, not
attacker-forgeable or -alterable in content (event \emph{timing} may be
attacker-influenced, e.g.\ a captured tool output's timestamp reflects
when the attacker acted; what is guaranteed is that the recorded facts
are true and unmodified); (ii) \emph{source-identification strings}
originating in untrusted content (sender names, subjects, citations),
rendered verbatim and visibly marked \emph{unverified}. Where a
transaction aborts after some accepting reviews, the pane states that
the accepted values remain authorized for their targets.
\item The decision pane contains no reader-generated prose outside the
confirmed payload region; where the payload itself is a crystallized
summary, it is reader-generated by construction, rendered verbatim, and
governed by Assumption~\ref{S-a:narr}.
\item Selection surfaces may show reader-generated previews, visibly
marked unverified; previews are rendered UI-side only and never returned
to $P$, and carry no endorsement authority.
\item Both surfaces are recorded in the display transcript and the audit
log (Definition~\ref{S-def:transcript}).
\item On a rejection, the displayed bytes are not returned to $P$; $P$
receives only the decision. (Preview rationale: \ref{S-n:preview},
\ref{S-q:preview}.)
\end{enumerate}
\end{rulex}

\begin{rulex}[Batch gate under untrusted-driven iteration]\label{S-r:batchgate}
\leavevmode
\begin{enumerate}[label=(\alph*),leftmargin=1.8em,itemsep=1pt,topsep=1pt]
\item A tool call reached inside $\mathsf{map}$ over a $\U$-collection --- including the desugared form of a collection-handle tool argument (Rule~\ref{S-r:names}(e)) ---
takes no fast path of Rule~\ref{S-r:whitelist}.
\item The map is gated as a single \emph{batch} confirmation whose
decision pane displays the collection's cardinality \emph{and the full
resolved bytes of every element}, together with the resolved bytes of any gating non-$\diamond$
argument of the body (Rule~\ref{S-r:names}(f)), displayed once for
the batch, paginated as needed, the cardinality displayed and recorded on the first page; per-page
cardinality is capped at $\cbatch$; the interpreter performs any split,
presenting a larger collection as successive pages of one batch
confirmation. A deployment may additionally fix a total-cardinality cap (TCB
configuration): a $\mathsf{map}$ over a collection exceeding it
renders a fixed oversize display --- recorded in the transcript like
the no-value display of Rule~\ref{S-r:null} --- and takes the standard
aggregated reject of clause~(c), consuming no decision,
indistinguishable in the reply stream and in $h$ from a page
rejection or a null collection.
A \emph{null collection} (a collection handle with null resolution:
an empty search result, or a zero-output capture, convention~\hyperref[S-sos:i]{(i)};
Rules~\ref{S-r:names}(a), \ref{S-r:null}) is handled identically: the
interpreter renders the fixed no-value display of Rule~\ref{S-r:null},
recorded in the transcript as a batch event exactly like the
oversize display, and the map takes the standard aggregated reject
of clause~(c), consuming no decision (\textsc{Map-Tool-Rej};
mechanized as \texttt{batchGate\_null},
\S\ref{S-app:mechanization}). The no-value and oversize displays are
distinct event kinds (Definition~\ref{S-def:transcript}), so the first
recorded batch event determines the size class
(Lemma~\ref{S-lem:dtsegment}).
\item $P$ receives a single aggregated decision for the whole
map (accept iff every page was accepted, else the single standard
reject), regardless of page count. The number of pages, like the
cardinality itself, is never returned to $P$ in any form; both are
displayed to the user and recorded in the transcript only.
\item Each page's rendering and decision is recorded in the transcript;
the aggregated decision is recorded once. By Rule~\ref{S-r:atomic}, the
bytes that flow at each element call are exactly the element bytes
displayed for it. The user may reject at any page, aborting the
remainder. The oversize and null-collection displays are recorded under their respective fixed-display event kinds (Definition~\ref{S-def:transcript}), each with its standard reject and the aggregated batch decision.
\item Per-element results are captured to $\Mem_{\U}$ and never returned
to $P$ (the map returns one ack token). (All-or-nothing rationale and
$\cbatch$ sizing: \ref{S-n:batch}, \ref{S-q:batch}.)
\end{enumerate}
\end{rulex}

\begin{rulex}[Constant-alphabet interface, with totality]\label{S-r:iface}
\leavevmode
\begin{enumerate}[label=(\alph*),leftmargin=1.8em,itemsep=1pt,topsep=1pt]
\item Every value the interpreter returns to $P$ is of one of two kinds.
The first kind is drawn from a fixed finite alphabet independent of the
attack and the environment: (i) one constant acknowledgement per \emph{executed} tool
invocation or $\mathsf{map}$, independent of outcome and of the number,
size, or content of captured blocks; a gate-rejected invocation does not
execute and returns only the reject decision of item~(iv)
(\textsc{Tool-Gate-Rej}); (ii) opaque handles named by
Rule~\ref{S-r:names}; (iii) the generic failure outcome of
Rule~\ref{S-r:nobit}, deterministic in $P$'s emitted operations,
$\Mem_{\E}$, and the coupled coins $\omega$; (iv) the user's
decision at a confirmation, with the null-gate abort of
Rule~\ref{S-r:null} rendered as the reject decision. The second kind is
(v): verbatim $\E$-bytes dereferenced from $\Mem_{\E}$.
\item \emph{Totality.} Every syntactically well-formed invocation of
every operation of Figures~\ref{S-fig:sos}--\ref{S-fig:sos2}, with any argument, has its
behaviour defined by the figures and these rules; failed preconditions
and unresolvable arguments are rendered, uniformly, as the generic
failure outcome or as an operation-determined opaque handle, never as a
distinguishing error. In particular, whether a given identifier is
occupied by a $\U$-block is never observable to $P$ through any operation:
$\mathsf{recall}$ and $\mathsf{evict}$ alike return
operation-determined replies (an opaque handle; the constant
$\mathsf{ack}$) uniformly across resolving, non-resolving,
$\U$-occupied, and $\E$-occupied arguments, resolution consumed
interpreter-side (Rules~\ref{S-r:null},\,\ref{S-r:nobit};
Figure~\ref{S-fig:sos2}, convention~\hyperref[S-sos:v]{(v)}).
\end{enumerate}
\end{rulex}

\begin{rulex}[Trusted-store retrieval]\label{S-r:retrieve}
$\RE$ (i) reads only $\Mem_{\E}$, never any $\U$-bytes; (ii) has no tool
access; (iii) returns only block identifiers, which the interpreter
dereferences against $\Mem_{\E}$ verbatim; non-dereferencing output is
discarded. $\RE$'s sampling randomness is coupled alongside $P$'s
(convention~(\ref{S-conv:c1})). (Why neither reader's vision can be relaxed:
\ref{S-n:readers}.)
\end{rulex}

\paragraph{Deployment-selectable hardening.}
Options~\ref{S-o:one}--\ref{S-o:two} are formally instantiated: each
strictly gates more flows, and their monotonicity is proved as
Proposition~\ref{S-prop:harden}.\allowbreak\ Option~\ref{S-o:three} is a
requirements-level extension, outside the formal semantics and
outside Proposition~\ref{S-prop:harden}'s quantifier (\ref{S-o:three}).

\begin{optionx}[Sink-scoped trusted values]\label{S-o:one}
$\T$-blocks require a first-use confirmation per sensitive sink, after
which scope-caching applies. Blocks created by these confirmations carry
the interpreter-only provenance mark \emph{cache} and are ineligible for
submission (Rule~\ref{S-r:share}).
\end{optionx}
\begin{optionx}[Exact-target mediation at data tools]\label{S-o:two}
The data-tool clause of Rule~\ref{S-r:whitelist}(b) is replaced by the
sensitive-tool test of clause~(a): a value passed to a tool in $\DataT$
flows without confirmation \emph{iff} it byte-equals a stored
$\T$-block, or byte-equals some $\C$/$\CC$-block $b$ with
$\mathsf{tool}\in\scopef(b)$ (for $\CC$, the live shared-tier entry,
checked valid at call time, \S\ref{S-app:fedcore}). Any other data-tool
argument --- a freshly composed value, or endorsed bytes whose scope
does not admit this data tool (in particular every reasoning-target
copy, whose scope is empty, convention~\hyperref[S-sos:vii]{(vii)}) --- is confirmed once;
the fresh copy is scoped $\{\mathsf{tool}\}$ per convention~\hyperref[S-sos:vii]{(vii)}, so
it scope-caches and covers all the user's later use of the pair
(\ref{S-inv:floor}). Blocks created by these confirmations carry the
interpreter-only \emph{cache} mark and are ineligible for submission
(Rule~\ref{S-r:share}); their gate events are cache confirmations,
accept-inert per Definition~\ref{S-def:counting}.
\end{optionx}
\begin{optionx}[Local scope leases (requirements level; not formally
instantiated)]\label{S-o:three}
$\C$-scopes carry a deployment-set lease (convention~(\ref{S-conv:c4})): expiry
withdraws the expired copy's tool authorization, so the pair's next
scope-checked use with no other currently authorizing witness
re-confirms once. Unlike Options~\ref{S-o:one}--\ref{S-o:two}, \ref{S-o:three} is
specified at the requirements level only: it is \emph{not} part of
the formal semantics of this document, no theorem quantifies over
it, and Proposition~\ref{S-prop:harden} does not cover it. A
conforming instantiation is a future extension obligated to satisfy
all of the following, and to prove analogues of
Lemma~\ref{S-lem:dtsegment}, Theorem~\ref{S-thm:reason},
Lemma~\ref{S-lem:scopepersist}, Theorem~\ref{S-thm:floor}, and the
ledger accounting of Lemma~\ref{S-lem:budget} for the extended
semantics.
\begin{enumerate}[label=(\alph*),leftmargin=1.8em,itemsep=1pt,topsep=1pt]
\item \emph{Immutability.} Blocks stay immutable (\ref{S-a:persist}):
a lease is a validity predicate computed interpreter-side from
creation-time metadata (creation epoch or slot index, never
wall-clock time) and the current index; expiry is never a block
mutation or removal.
\item \emph{Determinism on matched data.} Validity inputs are
functions of the coupled event sequence, and lease state is added
to the matched components of Definition~\ref{S-def:matched} and
carried through the unwinding, so matched runs gate identically.
\item \emph{Witness-level expiry at tool targets.} For tool targets,
expiry invalidates only the expired local authorization witness (an
ordinary or cache-marked $\C$ copy), and the two authorization tests
it touches are kept distinct. \emph{Literal authorization}, on a
scope-checked literal path, is existential over whichever witnesses
the deployment's selected formal options recognize there ---
byte-equal $\T$-blocks (per \ref{S-o:one}'s per-sink treatment where
selected), live $\CC$ entries, and lease-valid in-scope $\C$ copies;
the baseline data-tool path of Rule~\ref{S-r:whitelist}(b) is
scope-blind, so lease enforcement at data tools requires
\ref{S-o:two} --- and a literal use gates iff no such witness
currently authorizes the pair. \emph{Handle coverage}
(Rule~\ref{S-r:panel}(i)) is narrower: a handle use gates in every
case (Rule~\ref{S-r:whitelist}(c)) and is \emph{covered} iff the
identity holds a lease-valid, in-scope local $\C$ copy; a $\T$ or
live-$\CC$ witness never makes a handle use covered, exactly as in
the base semantics. The reasoning-coverage branch of
Rule~\ref{S-r:panel}(i) remains readability-based and
lease-insensitive: reasoning-target reuse never requires another approval.
\item \emph{Classification and renewal.} An expiry-induced gate
whose pair lacks any currently authorizing witness is \emph{required}
--- whether a literal gate fired by expiry or a handle gate
reclassified from covered to required because no lease-valid in-scope
$\C$ copy remains. Any required approval so caused
writes the fresh copy prescribed by the \emph{triggering rule} ---
an ordinary $\C$ copy at an ordinary confirmation, a cache-marked,
federation-inert copy at an \ref{S-o:one}/\ref{S-o:two} cache gate (Rule~\ref{S-r:share},
Definition~\ref{S-def:counting}(iii)) --- and starts a new lease, so
subsequent literal or handle use within the new period is
fast-pathed or covered: at most one required confirmation per pair per
lease period. The class is determined by the gate that fired, never
by any expired duplicate witness; rejection and abandonment renew
nothing; duplicate matching copies are resolved by the existential
tests of clause~(c), never by a selection among witnesses.
\item \emph{Accounting.} With each (user, item) seeing
$\Elease=\bigO(1)$ lease periods, the extension's pathwise floor is
the once-per-pair cap per lease period, an $\bigO(1)$
multiplicative factor on Theorem~\ref{S-thm:floor}'s constants --- an
extension-level claim, to be proved with the extension. No
saturation-vanishing conclusion follows from $\Elease=\bigO(1)$
alone: expiry re-fires of promoted pairs are residual required
confirmations, so the extension additionally requires an
\ref{S-o:three}-augmented analogue of the \emph{complete} residual bound of
\ref{S-a:promsuf} and Theorem~\ref{S-thm:sat}(iii) --- the per-pair
$D_0+\varepsilon\,n$ form with expiry re-fires included, together
with its product and residual-fraction rate conditions --- not
merely a substitute for $\varepsilon(M)\to0$.
\end{enumerate}
(\ref{S-o:three} is the single-user twin of the federated re-verification queue:
\ref{S-n:revoke}.)
\end{optionx}
\subsection{Trust axioms}
\label{S-app:axioms}

\begin{assumptionx}[Deterministic source labelling]\label{S-a:label}
$\ell(b)$ is a function of the channel/source of $b$ alone, never of
$\mathrm{bytes}(b)$. Hence the attacker cannot change any label.
\end{assumptionx}

\begin{assumptionx}[Label persistence and block immutability]\label{S-a:persist}
Blocks are immutable once written: no party and no transition alters
an existing block's bytes, label, or scope; the store changes only by
the deterministic block-creating transitions of
Figures~\ref{S-fig:sos}--\ref{S-fig:sos2} (convention~\hyperref[S-sos:vii]{(vii)}) and by
$\mathsf{evict}$. Trust is raised only by the creation of a fresh
endorsed block at an accepting $\conf$, or by deterministic community
promotion of a shared-tier entry (Appendix~\ref{S-app:fedmodel}); the
attacker can do neither.
\end{assumptionx}

\begin{assumptionx}[Faithful, zoned display]\label{S-a:display}
At a confirmation, the decision-pane bytes are exactly the bytes that
will flow or be recorded; the attested-fact zone is TCB-authenticated and not attacker-forgeable (Rule~\ref{S-r:display}(b));
untrusted source-identification strings appear only in the marked zone;
previews are confined to selection surfaces, visibly marked
(Rule~\ref{S-r:display}).
\end{assumptionx}

\begin{assumptionx}[Sound atomic endorsement]\label{S-a:user}
The user returns $\mathrm{accept}$ on a displayed value $v$ at tool
$\mathsf{tool}$ only if the user intends exactly $v$ to be authorized
for $\mathsf{tool}$; the intent is per displayed value, severally, and
does not depend on the outcomes of the transaction's other reviews ---
which is what makes the durable endorsement of Rule~\ref{S-r:atomic}
sound when a later review aborts the invocation. At a batch
confirmation (Rule~\ref{S-r:batchgate}), the user returns
$\mathrm{accept}$ on a page only if this intent holds for \emph{each}
displayed element and each displayed gating non-$\diamond$ argument,
severally, at the batch's tool: batch acceptance is the conjunction of
the per-value intents, never an aggregate judgment of the collection.
\end{assumptionx}

\begin{assumptionx}[Sound narrative endorsement]\label{S-a:narr}
The user returns $\mathrm{accept}$ on a displayed crystallized summary
$S$ only if, having checked $S$ against its displayed source citations,
the user intends $S$'s content to inform reasoning.
\end{assumptionx}

\begin{assumptionx}[Exogenous command stream]\label{S-a:exo}
The user's typed command-channel input is an exogenous input stream,
coupled identically across the runs compared in
Theorem~\ref{S-thm:reason}.
\end{assumptionx}

\begin{commentary}[epistemic status of
  \ref{S-a:user}--\ref{S-a:exo}]
\ref{S-a:user} and \ref{S-a:narr} are trust axioms in the sense of CaMeL
\cite{debenedetti2025defeating}; \ref{S-a:exo} is the human-channel idealization that any
displayed-content carve-out requires. Their epistemic status, the
regime that makes them defensible, and the discharge of
\ref{S-a:exo}'s practical content by logging both display surfaces are
discussed at \ref{S-n:human} (see also \ref{S-q:preview},
\ref{S-q:postendorse}, \ref{S-q:batch}); \ref{S-a:narr}'s realistic weakness
at scale is what the community mechanism of
Appendix~\ref{S-app:fedmodel} addresses.
\end{commentary}

\subsection{Attacker model and trusted computing base}
\label{S-app:attacker}

\paragraph{Attacker.}
The attacker controls the \emph{content} of every $\U$-labelled
block, all tool outputs (including sensitive-tool responses and error
strings) and paste-channel input, including the byte-length and the
number of $\U$-blocks captured per call. It does not control the user's
typed command input, the labelling, the interpreter, or the user's
decisions. $P$, $Q$, and $\RE$ are \emph{isolated from the attacker,
not assumed benign}: the single-user theorems hold under adversarial
behaviour of either reader, and misdirection among the user's own typed
or endorsed values is a utility failure outside the threat model
(Option~\ref{S-o:one} offers defence in depth; scope discussion:
\S\ref{S-app:residuals}).

\paragraph{Trusted computing base (TCB), single-user.}
The TCB comprises: an authenticated command channel distinguishing
typed from pasted from tool input; the canonical tool identifiers and
the $\SensT/\DataT$ partition; the interpreter's implementation
(including byte-equality and dereference checks) and the individuating
parser; the zoned rendering path of Rule~\ref{S-r:display}; and the
hash-chained audit log materializing the display transcript. The store
idealization and quota policy are convention~(\ref{S-conv:c7}). Federated TCB
additions are listed in \S\ref{S-app:fedassum}.

\section{Single-User Guarantees}
\label{S-app:single}

This appendix proves two conditional interface properties.
\emph{Action Integrity} is an exact-byte, exact-tool authorization
property: no untrusted-origin \emph{value} reaches a sensitive action
unless the user endorsed exactly that value for exactly that action;
it does not authorize the choice to invoke a tool
(\S\ref{S-app:residuals}). \emph{Reasoning Integrity modulo the display
transcript} is a fixed-horizon, termination-insensitive
trace-equivalence property: holding fixed everything the user ever saw
and decided on the display surface, together with the coupled
exogenous inputs, untrusted memory has no further effect on behaviour;
it is not a claim of reasoning correctness or of resistance to
manipulation through displayed content
(Remark~\ref{S-rem:transcript}, \ref{S-n:audit}). The statements are in
the noninterference-modulo-declassification tradition
\cite{goguen1982security,goguen1984unwinding,sabelfeld2009declassification,sabelfeld2004model,askarov2007gradual}; the first is
the PerMIT instance of qualified robustness \cite{zdancewic2005robust,myers2006enforcing}.
Both theorems, the supporting lemmas, the baseline separation, the flow
half of the hardening monotonicity, and the batch-confirmation surface
of Rule~\ref{S-r:batchgate} are machine-checked
(\S\ref{S-app:mechanization}: scope, axiom audit, artifact, correspondence
Table~\ref{S-tab:leancorr}, and the disclosed modelling deltas). Lineage
and positioning: \ref{S-n:lineage}.

\subsection{Definitions}
\label{S-app:sdefs}

\begin{definition}[$\U$-equivalence]\label{S-def:lowequiv}
$c_1\eqU c_2$ iff $\obs(c_1)=\obs(c_2)$: agreement on $\Mem_{\E}$
(content and scopes), $H(c)$, $x(c)$, and $h$, while $\Mem_{\U}$ may
differ arbitrarily \cite{volpano1996sound,sabelfeld2003language}.
\end{definition}

\begin{definition}[Sink-event safety]\label{S-def:sinksafe}
Fix a finite sequence of display events --- decision-pane
confirmation events, each carrying its displayed bytes, target, and
decision (Definition~\ref{S-def:sinkconf}, Rule~\ref{S-r:display}), and
renderings --- and a trusted typed-input stream $x$. A recorded
sensitive sink event is \emph{safe} for that history if every
argument value of the event is: (i)~the exact
bytes of a typed input unit of $x$ (a $\T$-block's bytes,
individuation per \S\ref{S-app:config}); or (ii)~endorsed at a prior
accepting decision-pane confirmation event of the history displaying
exactly those bytes at exactly the event's tool (for element calls of
a batch-gated $\mathsf{map}$, at the batch confirmation displaying
those exact element bytes, Rule~\ref{S-r:batchgate}); or (iii)~gated at
the event's own call, the flowed bytes being the displayed bytes of
that call's accepting confirmation (Rule~\ref{S-r:atomic}). The
\emph{delegated} form extends case~(ii) by one alternative, relative to a designated federation-log prefix (in a run, the prefix delivered at the event's step, Definition~\ref{S-def:fedstep}): the bytes
byte-equal a live community entry of that prefix's replay naming the event's exact tool,
backed by a currently valid promotion certificate
(Definition~\ref{S-def:fedconfig}, \ref{S-a:transfer}); the personal form
is the single-user predicate, machine-checked as \texttt{SinkSafe}
(\S\ref{S-app:mechanization}). It is consumed by
Definition~\ref{S-def:wf}(v) and Definition~\ref{S-def:action}.
\end{definition}

\begin{definition}[Well-formed configuration]\label{S-def:wf}
A configuration $c$ is \emph{well-formed}, written $\mathsf{WF}(c)$,
if:
(i)~every $\T$-block's bytes occur in the typed stream $x(c)$;
(ii)~every $\C$-block $b$ with $t\in\scopef(b)$ is backed by an
accepting decision-pane confirmation event displaying exactly
$\mathrm{bytes}(b)$ at target $t$ in the audit transcript, with the
matching sanitized (target, accept) record in $h$;
(iii)~no block carries label $\CC$ (single-user; the federated
counterpart replaces (iii) by a valid promotion certificate on the
federation log, Appendix~\ref{S-app:fedmodel});
(iv)~the handle structure $H(c)$ and the endorsed-sort identifier
allocation are the operation-determined structures of
Rule~\ref{S-r:names} for the configuration's interaction history,
$\sigma_P$ is the strategy state of that history
(\S\ref{S-app:config}), and the recorded reply stream is the
interface's reply to that history
(Figures~\ref{S-fig:sos}--\ref{S-fig:sos2});
(v)~every recorded sink event is safe (Definition~\ref{S-def:sinksafe}; federated form: the delegated predicate).
The empty configuration is well-formed. A \emph{seeded} configuration
(nonempty $\Mem_{\E}$, empty interaction history) is well-formed only
together with a \emph{seed record}: a designated initial-authorization
prefix of the audit transcript
(with its sanitized projection in $h$) backing every seeded $\T$-block per~(i) and every seeded scoped $\C$-block per~(ii) ---
operationally, the authorizations imported with a migrated store;
formally, the record is read as part of $x(c)$, $h$, and $\DT$,
whereupon (i)--(v) hold verbatim. A seeded configuration without such
a record is in general \emph{not} well-formed; it is \emph{sorted}
(Definition~\ref{S-def:swf}), which is what the two-run results
require. Well-formedness is preserved by every transition of
Figures~\ref{S-fig:sos}--\ref{S-fig:sos2} (this is the induction of
Lemmas~\ref{S-lem:isolation}--\ref{S-lem:endorse}, machine-checked as the
invariant \texttt{Inv} of \S\ref{S-app:mechanization}). Action
Integrity and its corollaries quantify over runs from well-formed
configurations; the two-run results quantify over matched pairs of
sorted configurations (Definitions~\ref{S-def:swf},~\ref{S-def:matched}).
\end{definition}

The history against which clause~(v) reads
Definition~\ref{S-def:sinksafe} is the configuration's full audit
history: any seed-record prefix followed by the run transcript
(Definition~\ref{S-def:transcript}).

\begin{remark}
Clause~(v) is the standard inductive strengthening required to prove
Theorem~\ref{S-thm:action} by induction on run length (mirrored verbatim as
the \texttt{sinkSound} field of \texttt{Inv} in the mechanization,
\S\ref{S-app:mechanization}): it tracks, through the induction, exactly
the per-event predicate of
Definition~\ref{S-def:sinksafe}, whose universal form over reachable
configurations is the theorem's conclusion --- not an independent
assumption, and not circular: the predicate is defined without
reference to the theorem. The from-arbitrary-well-formed forms are \emph{preservation}
statements; the empty-initial-configuration forms, where the
invariant is established rather than assumed
(\texttt{inv\_init}, \S\ref{S-app:mechanization}), are the creation
forms.
\end{remark}

\begin{definition}[Sorted configuration]\label{S-def:swf}
A configuration $c$ is \emph{sorted} if clauses (iii) and (iv) of
Definition~\ref{S-def:wf} hold of it: no $\CC$ label (federated form as
there), operation-determined handle and endorsed-sort identifier
structure with $\sigma_P$ the strategy state, and an
interface-conformant reply stream. Every well-formed configuration is
sorted. Every \emph{endorsed-seed} configuration --- an arbitrary
$\E$-level $\Mem_{\E}$ at endorsed-sort identifiers
(Rule~\ref{S-r:names}(g)), an arbitrary $\Mem_{\U}$, and an empty
interaction history --- is sorted, with no seed record required.
Sortedness is preserved by every transition
(machine-checked as \texttt{WFs}, with the endorsed-seed condition
\texttt{EStore}; \S\ref{S-app:mechanization}).
\end{definition}

\begin{commentary}[why two classes]
The split is forced by the theorems, in opposite directions. Action
Integrity is an endorsement-soundness statement, so its quantifier
must carry endorsement soundness: admitting a bare seed as well-formed
would \emph{falsify} Theorem~\ref{S-thm:action} (Remark~\ref{S-rem:endorsesorted}) --- a seeded $\C$-block
with $t\in\scopef$ flows to $t$ ungated, with no endorsing
confirmation anywhere in the run. Reasoning Integrity, by contrast,
never consumes endorsement soundness: its unwinding needs only the
matched view, the operation-determined naming of
Rule~\ref{S-r:names}, and the strategy reading --- precisely the sorted
clauses --- so its quantifier is deliberately the weaker class,
admitting arbitrary shared endorsed seeds. The mechanization enforces
exactly this split: \texttt{action\_integrity\_from} requires the
full invariant \texttt{Inv}, while
\texttt{reasoning\_integrity\_from} requires only
\texttt{EStore}/\texttt{WFs} (\S\ref{S-app:mechanization}).
\end{commentary}

\begin{definition}[Matched configurations]\label{S-def:matched}
Two sorted configurations (Definition~\ref{S-def:swf}) are
\emph{matched}, $c_1\approx c_2$, if $\obs(c_1)=\obs(c_2)$; $P$'s
interaction histories coincide (the same emitted-operation sequence,
the same reply stream, and the same consumed coin prefix --- hence
$\sigma_{P,1}=\sigma_{P,2}$, by the strategy reading of
\S\ref{S-app:config}); and the ghost display transcripts and sink logs
coincide. Matched views and histories also fix the endorsed-sort
allocation state (Rule~\ref{S-r:names}(g)), so subsequent endorsed blocks
are created at equal identifiers in the two runs. $\Mem_{\U}$ and the
interpreter's private handle \emph{resolutions} may differ arbitrarily;
the handle \emph{count} is matched through $H(c)$
(Rule~\ref{S-r:nobit}). Matched implies $\eqU$; the converse fails on
$\sigma_P$ and the ghost logs, which is why the two-run results are
stated over matched pairs. The canonical matched pair is two runs
launched from a shared endorsed-seed configuration
(Definition~\ref{S-def:swf}) with empty histories and empty ghost logs
under two different attacks (the form mechanized as \texttt{initE},
\S\ref{S-app:mechanization}); matched pairs need not be well-formed, and
Theorem~\ref{S-thm:reason} holds without any endorsement-soundness
hypothesis on the seed. This relation is mechanized, field for field,
as \texttt{MatchW} (\S\ref{S-app:mechanization}).
\end{definition}

\begin{definition}[Ungated sink projection]\label{S-def:sinkproj}
For a finite run $\rho$, $\sinks(\rho)$ is the sequence of sensitive
sink events, each with its arguments, except that arguments displayed
and accepted in the immediately preceding atomic gating confirmation
(Rule~\ref{S-r:atomic}, including, for element calls of a
$\mathsf{map}$, the batch confirmation of
Rule~\ref{S-r:batchgate}) are stripped from the annotation. Stripping
makes $\sinks(\rho)$ well-defined for baseline comparisons in which
transcripts may be empty (Proposition~\ref{S-prop:baseline}); under
$\DT$-equality the stripped arguments are equal in any case, so the
projection hides nothing in Theorem~\ref{S-thm:reason}.
\end{definition}

\begin{definition}[Display transcript]\label{S-def:transcript}
For a run $\rho$ from a designated starting configuration, $\DT(\rho)$
is the ordered sequence of \emph{all display events appended during
$\rho$} (the run suffix; an audit prefix carried by a seeded starting
configuration belongs to its seed record, Definition~\ref{S-def:wf}, not
to $\DT(\rho)$): (i) selection-surface renderings, each with its
candidate identifiers, rendered preview bytes (marked unverified), and
the user's selection, if any; (ii) decision-pane confirmation events,
one per reached gating review, each with the resolved bytes (or the
fixed no-value display of Rule~\ref{S-r:null}), both provenance zones,
the target, and the decision; (iii) for a batch, \emph{either} the fixed oversize display of
Rule~\ref{S-r:batchgate}(b) with its standard aggregated reject (where
the deployment cap applies), \emph{or} the fixed no-value display of
a null collection with its standard aggregated reject
(Rules~\ref{S-r:null}, \ref{S-r:batchgate}(b)), \emph{or} every page's
rendering
and per-page decision, the first page recording the collection's
cardinality --- which, with $\cbatch$, determines the deterministic
pagination and hence the page count (Rule~\ref{S-r:batchgate}(b)) ---
and the aggregated decision, recorded once.
Transcript entries carry interpreter-assigned event \emph{kinds}
(selection-surface rendering; decision-pane confirmation; batch
page; the fixed no-value display; the fixed oversize display;
aggregated batch decision), fixed constructors of the TCB logging
path: the fixed displays are distinguished from ordinary pages by
kind, never by comparing rendered byte strings, so page content
colliding textually with a fixed message cannot confuse the
classification (consumed by Lemma~\ref{S-lem:dtsegment}).
Rejections are recorded as
well as acceptances, and so are the accepting reviews of transactions
that subsequently abort (Rule~\ref{S-r:atomic}). The harness materializes
the full audit history --- any seed-record prefix followed by
$\DT(\rho)$ --- in a hash-chained audit log (operational reading:
\ref{S-n:audit}; previews: \ref{S-n:preview}, \ref{S-q:preview}).
\end{definition}

\begin{remark}[Coverage of the display transcript]\label{S-rem:transcript}
By Rule~\ref{S-r:display}(e), $\DT(\rho)$ comprises the totality of
attacker-influenceable content ever rendered to the user, together
with everything the user decided. Recording rejections is necessary:
two runs in which the user rejects and later accepts the same content
differ in behaviour during the interval, and a projection that forgets
rejections cannot distinguish them. Previews carry no endorsement
authority; their inclusion records them as part of the release
history.
\end{remark}

\begin{definition}[Action Integrity, personal and delegated]
\label{S-def:action}
PerMIT has \emph{personal Action Integrity} (equivalently:
\emph{argument-value integrity}) if in every run from a well-formed
configuration, under every attack, every sensitive sink event is safe
in the personal form of Definition~\ref{S-def:sinksafe}: every value
reaching a sensitive sink is a user-typed $\T$-value, or was endorsed by
\emph{this} user at its exact bytes for that exact sink (now, or at a
prior in-scope confirmation; for element calls of a batch-gated
$\mathsf{map}$, at the batch confirmation that displayed those exact
element bytes, Rule~\ref{S-r:batchgate}). It has \emph{delegated Action
Integrity} if the same holds with the endorsement authority extended to
a qualifying community confirmation of the exact bytes at the exact
tool ($\CC$, Appendix~\ref{S-app:fedmodel}) (the delegated predicate of Definition~\ref{S-def:sinksafe}); delegation is an assumption
(\ref{S-a:transfer}), never a theorem of the single-user model.
\end{definition}

\begin{definition}[Reasoning Integrity modulo the display transcript]
\label{S-def:reason}
PerMIT has \emph{Reasoning Integrity modulo the display transcript} if,
for every pair of \emph{matched} configurations
$c_1\approx c_2$ (Definition~\ref{S-def:matched}), every pair of attacks
(coupled per convention~(\ref{S-conv:c9}): slot skeleton and typed stream only;
paste presence and content, tool outputs, and reader outputs
uncoupled), and every coupled coin sequence $\omega$
(convention~(\ref{S-conv:c1})), any two runs of a common horizon (convention~(\ref{S-conv:c9})) with
$\DT(\rho_1)=\DT(\rho_2)$ satisfy
$\sinks(\rho_1)=\sinks(\rho_2)$ and emit the same sequence of
Protected-LLM operations and tool invocations. Informal reading (a conditional statement, not an unconditional
inertness claim): any influence of untrusted memory on the listed
behaviours must pass through a display event or an endorsed release
--- conditional on the transcript, hidden untrusted state has no
further effect. An unrendered untrusted byte is not claimed inert
outright: it may steer which candidates or previews the Reader
surfaces, and thereby the transcript itself; what is excluded is
influence that bypasses the transcript.
\end{definition}

The statement is trace-wise per coin sequence; the knowledge-based
formulation in the style of gradual release \cite{askarov2007gradual} that it
induces is recorded in \S\ref{S-app:knowledge}.

\subsection{Structural lemmas}
\label{S-app:slemmas}

\begin{lemma}[Protected-LLM isolation / provenance preservation]
\label{S-lem:isolation}
In every configuration reachable from a sorted configuration
(Definition~\ref{S-def:swf}; a fortiori from a well-formed one), no byte
value enters $\sigma_P$ \emph{by delivery from a $\U$-labelled
source}: every bytes-carrying reply $P$ receives is, at the step of
its emission, the verbatim content of a non-$\U$ block of the store,
and $\U$-derived data appears in $\sigma_P$ only as opaque handles.
This is a provenance statement, not a disjointness of byte strings
across tiers: a typed or endorsed byte string in $\sigma_P$ may be
extensionally equal to the bytes of an unrelated $\U$-block, and the
flow it licenses is the trusted copy's.
\emph{(Machine-checked, run-level:} \texttt{bytes\_reply\_isolation},
\S\ref{S-app:mechanization}.\emph{)}
\end{lemma}

\begin{proof}
Induction on run length with invariant $I(c)$: every bytes-carrying
value ever delivered to $P$ was delivered as the content of a
non-$\U$ block (reply grammar case (v)) or as trusted command input;
everything else in $P$'s input stream is a handle or constant token. At a sorted
start, $\sigma_P$ is the strategy state of the initial
interaction history (\S\ref{S-app:config},
Definition~\ref{S-def:swf}), whose reply stream is
interface-conformant: every bytes-carrying reply in it is $\E$-level
by the reply grammar, and its typed input is trusted command input at
level $\T$; an endorsed-seed configuration has empty history and
satisfies $I$ trivially. For the inductive
step, case-analyse the transitions of Figures~\ref{S-fig:sos}--\ref{S-fig:sos2}.
\textsc{Tool-Fast}, \textsc{Tool-Gate-Acc/Rej}, \textsc{Map}: by
Rules~\ref{S-r:ingress} and~\ref{S-r:batchgate} all outputs land in
$\Mem_{\U}$ and only the constant acknowledgement (and, for gated calls,
the decision) returns; the collection handle's identifier is
operation-determined and computed by $P$, not returned
(Rule~\ref{S-r:names}(b)). \textsc{W-Typed} stores $\T$; \textsc{W-Paste}
is captured to $\Mem_{\U}$. \textsc{W-P} has precondition
$b\in\sigma_P$, hence $\E$-level by the hypothesis, and stores at $\PC$
with empty scope. \textsc{Recall-E} returns $\E$-bytes;
\textsc{Recall-O} returns an opaque handle: uniformly for $\U$-blocks,
unknown, and nonexistent identifiers
(Rules~\ref{S-r:names},~\ref{S-r:null}, and the totality clause
Rule~\ref{S-r:iface}(b)); $\RE$-retrieval returns verbatim $\E$-bytes
(Rule~\ref{S-r:retrieve}). \textsc{Fill} and \textsc{Cryst} return one
handle and, by Rule~\ref{S-r:nobit}, no branchable bit. \textsc{Display}
and \textsc{Evict} inject nothing
(\textsc{Evict} replies the constant $\mathsf{ack}$ uniformly,
Figure~\ref{S-fig:sos2}, convention~\hyperref[S-sos:v]{(v)}); previews are UI-side only
(Rule~\ref{S-r:display}(d)). \textsc{Conf-Acc/Rej} return only the
decision (the null-gate abort returning the reject decision,
Rule~\ref{S-r:null}) and on accept produce an $\E$-block. \textsc{Fail}
returns the constant $\mathsf{fail}$ token. $P$-generated content is a
function of $\sigma_P$ and the coins, hence $\E$-level by the meet rule
(Definition~\ref{S-def:lattice}). By
\ref{S-a:label}--\ref{S-a:persist} and convention~\hyperref[S-sos:vii]{(vii)}, no transition
raises the level of any existing block; an accepting $\conf$ writes a
fresh $\E$-block and leaves its untrusted source, if any, at $\U$.
Thus every delivery to $P$ carries non-$\U$ provenance; the existence of extensionally equal untrusted copies is irrelevant. $I$ is preserved.
\end{proof}

\begin{lemma}[Whitelist mediation, by tool class]
\label{S-lem:mediation}
Every value $a$ passed to any tool satisfies the following, according to the
tool's class.
\emph{(Sensitive tools.)} If $\mathsf{tool}\in\SensT$, at least one of:
(i)~$a$ consists of readable $\E$-bytes byte-equal to a stored $\T$-block;
(ii)~$a$ consists of readable $\E$-bytes byte-equal to some $\C/\CC$-block $b$
with $\mathsf{tool}\in\scopef(b)$;
(iii)~the call is gated by an atomic $\conf(a,\mathsf{tool})$ displaying $a$'s
resolved bytes (or, for a batch, every element's bytes; or, for a null handle,
the fixed no-value display, whereupon the call aborts).
For a literal-expressed argument, in the \emph{baseline} semantics of
Rule~\ref{S-r:whitelist}(a) --- before the formally instantiated
Options~\ref{S-o:one}--\ref{S-o:two} are applied --- case~(iii) is the
complement of the union of (i) and (ii): the fast path passes iff
(i) or (ii) holds, and the gate fires otherwise. Under
Option~\ref{S-o:one}, a sensitive-tool argument satisfying case~(i)
may nevertheless gate (first use of the $\T$-value per sensitive
sink); Option~\ref{S-o:two} leaves this sensitive case analysis
unchanged and modifies only the data-tool case~(D) below; the
option-modified analyses are Proposition~\ref{S-prop:harden}'s.
Option~\ref{S-o:three} is requirements-level, outside this lemma and
outside Proposition~\ref{S-prop:harden} (\ref{S-o:three}). Cases (i)
and (ii) themselves may hold simultaneously --- the store may
contain a $\T$-block and an in-scope $\C/\CC$-block with identical
bytes, a co-occurrence Option~\ref{S-o:one} manufactures by design ---
and either suffices for the baseline fast path; no consumer of this
lemma uses exclusivity of (i) against (ii), each case being
separately sufficient in Theorem~\ref{S-thm:action}. Handle-expressed
arguments take case~(iii) unconditionally (\emph{Handles} clause
below).
\emph{(Data tools.)} If $\mathsf{tool}\in\DataT$, then either
(D)~$a$ consists of readable $\E$-level bytes held by $P$ --- that every byte value $P$ holds
carries non-$\U$ provenance, is Lemma~\ref{S-lem:isolation}'s
contribution, not Rule~\ref{S-r:whitelist}(b) alone --- which flow without
confirmation (Rule~\ref{S-r:whitelist}(b)) --- in particular $\PC$-labelled and
freshly $P$-composed $\E$-bytes flow ungated at data tools; this is precisely
the clause Option~\ref{S-o:two} hardens --- or the call is gated per case~(iii).
\emph{(Handles.)} Every handle argument, at every tool, falls under~(iii): an ordinary handle via Rule~\ref{S-r:whitelist}(c); a collection handle via the batch gate of Rule~\ref{S-r:batchgate}, by the desugaring of Rule~\ref{S-r:names}(e).
\emph{(Under Option~\ref{S-o:two}, case~(D) is replaced by the
clause-(a) case analysis applied at the data tool;
Proposition~\ref{S-prop:harden}.)}
\emph{(Machine-checked core:} \texttt{processArgs\_spec} \emph{and}
\texttt{bytes\_reply\_isolation}, \S\ref{S-app:mechanization}: an ungated
flowed argument is a verbatim $P$-supplied literal, and at a sensitive tool
it passes the whitelist test over $\Mem_{\E}$; that $P$ holds only
$\E$-level bytes in case~(D) is \texttt{bytes\_reply\_isolation}'s
contribution; witness for case~(D): Scenario~7.\emph{)}
\end{lemma}

\begin{proof}
Immediate from Rule~\ref{S-r:whitelist} with
Rules~\ref{S-r:atomic},~\ref{S-r:null},~\ref{S-r:batchgate}: clause~(a) gives the sensitive case analysis: in the baseline
semantics the fast path passes iff the argument consists of readable
$\E$-bytes matching (i) or (ii), and the gate fires otherwise, so
case~(iii) is exhaustive over the complement, while (i) and (ii) may
co-occur, harmlessly (the membership and scope tests are decidable
against $\Mem_{\E}$, Definition~\ref{S-def:config}; $\PC$ blocks have
empty scope and cannot take the sensitive fast path); clause~(b) gives case~(D) at data
tools; clause~(c) gives the handle case at every tool. For element calls of
a $\mathsf{map}$, the $\diamond$-substituted element falls under~(iii) via
the batch confirmation (Rule~\ref{S-r:batchgate}), and each non-$\diamond$
argument, a non-handle $P$-supplied value by Rule~\ref{S-r:names}(f), falls
under the case its tool class assigns exactly as at a direct call, its
gating branch resolved at the batch confirmation.
\end{proof}

\begin{lemma}[Endorsement is the only trust raise]\label{S-lem:endorse}
\leavevmode
\begin{enumerate}[label=(\alph*),leftmargin=1.8em,itemsep=1pt,topsep=1pt]
\item \emph{(Transition-level; any run.)} The only $\C$-creating
transitions of Figures~\ref{S-fig:sos}--\ref{S-fig:sos2} are the accepting
gating reviews --- \textsc{Conf-Acc} and the accepted positions of
\textsc{Tool-Gate-Acc}, \textsc{Tool-Gate-Rej},
\textsc{Map-Tool-Acc}, \textsc{Map-Tool-Rej} --- each writing a fresh
copy of exactly the displayed bytes with scope $\sigma(t)$
(Rule~\ref{S-r:atomic}, convention~\hyperref[S-sos:vii]{(vii)}); the only $\CC$-assigning
transition is deterministic community promotion
(Appendix~\ref{S-app:fedmodel}); and no transition alters an existing
block's label or scope (\ref{S-a:label}--\ref{S-a:persist}).
\item \emph{(Backing; well-formed runs.)} In every run from a
well-formed configuration (Definition~\ref{S-def:wf}), every $\C$-block
$b$ and every $t\in\scopef(b)$ is backed by an accepting decision-pane
confirmation displaying exactly $\mathrm{bytes}(b)$ at target $t$, in
the run suffix (by clause~(a)) or in the seed record
(Definition~\ref{S-def:wf}(ii)); and every $\CC$ scope in force is
backed by its promotion certificate, whose threshold was met by
endorsements at the scoped tool itself
(Rule~\ref{S-r:panel}(iv), Definition~\ref{S-def:counting}).
\end{enumerate}
\end{lemma}

\begin{proof}
\emph{(a)} is the case analysis below; \emph{(b)} adds the base case: at the initial configuration, clause~(ii) of well-formedness --- with the seed record where the store is seeded --- supplies the backing for every existing scoped $\C$-block, and clause~(iii) (federated form) the certificates.
By \ref{S-a:label}--\ref{S-a:persist} and convention~\hyperref[S-sos:vii]{(vii)}, no transition
alters an existing block's label or scope, so a block's label and
scope are those of its creating transition. No write transition
creates a block at $\C/\CC$; \textsc{W-P} creates $\PC$ with empty
scope; the only endorsement-creating transitions are
the accepting gating reviews --- \textsc{Conf-Acc}, and the accepted
positions of \textsc{Tool-Gate-Acc}, \textsc{Tool-Gate-Rej},
\textsc{Map-Tool-Acc}, and \textsc{Map-Tool-Rej}
(Rule~\ref{S-r:atomic}: endorsement is durable, written on the accepting
and aborting branches alike) --- each of which writes scope
$\sigma(t)$ per convention~\hyperref[S-sos:vii]{(vii)} for exactly the displayed bytes. The only $\CC$-assigning transition is
promotion \cite{myers2006enforcing}, which
(Rule~\ref{S-r:panel}(iv), Definition~\ref{S-def:counting}) adds
$\mathsf{tool}$ to a $\CC$ scope only when the community threshold
was met by endorsements at $\mathsf{tool}$ itself.
\end{proof}

\begin{remark}\label{S-rem:endorsesorted}
Clause~(b) is deliberately \emph{not} asserted over the sorted class
(Definition~\ref{S-def:swf}): an endorsed-seed configuration may hold
scoped $\C$-blocks with no backing confirmation anywhere, and
clause~(b) fails there, while clause~(a), a property of the
transitions alone, holds in every run. Consumers of clause~(b)
quantify over well-formed runs (via Definition~\ref{S-def:action});
Theorem~\ref{S-thm:reason} consumes neither clause. This is the
\texttt{Inv}-vs-\texttt{EStore} split of
Definitions~\ref{S-def:wf}/\ref{S-def:swf}.
\end{remark}

\begin{lemma}[No community label single-user]\label{S-lem:nocc}
In every run from a well-formed configuration with no federation
attached, no block ever carries label $\CC$: promotion is the only
$\CC$-assigning transition (Lemma~\ref{S-lem:endorse}) and it does not
exist in the single-user semantics. \emph{(Machine-checked as
\texttt{Inv.noCC}, \S\ref{S-app:mechanization}.)}
\end{lemma}
\begin{proof}
$\mathsf{WF}(c_0)$(iii) gives the base case; no transition of
Figures~\ref{S-fig:sos}--\ref{S-fig:sos2} assigns $\CC$
(Lemma~\ref{S-lem:endorse}'s case analysis).
\end{proof}

\begin{lemma}[Interface completeness]\label{S-lem:iface}
The sequence of values the interpreter returns to $P$ is a function of
the trajectory of $\obs(c)$, $P$'s own emitted operations and tool
invocations, the coupled coins $\omega$, and the user's decisions.
\end{lemma}

\begin{proof}
By the reply grammar of Figure~\ref{S-fig:sos}, every return is one of:
(a) a constant token ($\mathsf{ack}$ or $\mathsf{fail}$, per
Rule~\ref{S-r:iface}(a)(i),(iii)), including the uniform
acknowledgement of $\mathsf{evict}$, whose resolution status is
consumed interpreter-side (Figure~\ref{S-fig:sos2}, convention~\hyperref[S-sos:v]{(v)});
(b) $\E$-bytes from $\Mem_{\E}$, a
component of $\obs(c)$; (c) a handle whose existence and identifier are
fixed by the emitted-operation sequence
(Rules~\ref{S-r:null},~\ref{S-r:names}), including collection handles,
whose identifiers $P$ computes and which are never returned, and the
handles of \textsc{Recall-O}, indistinguishable across resolution
statuses (Rule~\ref{S-r:iface}(b)); (d) the user's decision, with the
null-gate abort rendered as the reject decision (Rule~\ref{S-r:null});
(e) the verbatim dereference of an $\E$-block selected by $\RE$, a
function of the query, $\Mem_{\E}$, and $\omega$
(Rule~\ref{S-r:retrieve}). The grammar has no other productions, and
Rule~\ref{S-r:iface}(b) defines every invocation (totality); no return
depends on $\Mem_{\U}$ bytes except through an accepting confirmation,
which writes the endorsed copy into $\Mem_{\E}$.
\end{proof}

\begin{lemma}[Control-flow confinement to trusted data and releases]\label{S-lem:cflow}
$P$'s choice and ordering of operations and tool invocations is a
function of $\obs(c)$, the reply stream, and the coupled coins alone.
No \emph{direct} channel exists from $\U$-block bytes, $\U$-search
outcomes, $\mathsf{recall}$ resolution statuses, tool-invocation
outcomes, candidate multiplicities, collection cardinalities,
block-identifier occupancy, or handle-identifier structure beyond the
operation-determined naming, to any reply $P$ receives. Information
about hidden state reaches $P$ only through logged release events:
the user's recorded decisions and the endorsed copies they create ---
each carried, with its displayed bytes, in the display transcript
(Definition~\ref{S-def:transcript}) --- which is precisely the
declassification boundary Theorem~\ref{S-thm:reason} conditions on.
\end{lemma}

\begin{proof}
We verify against Figures~\ref{S-fig:sos}--\ref{S-fig:sos2} and Rules
\ref{S-r:names}--\ref{S-r:retrieve} that every channel is closed.
(a) \emph{$\U$-content:} absent from $\sigma_P$
(Lemma~\ref{S-lem:isolation}).
(b) \emph{$\U$-search bits and $\mathsf{recall}$ resolution status:}
consumed interpreter-side (Rules~\ref{S-r:null},~\ref{S-r:nobit},
\ref{S-r:iface}(b)); the null-gate abort returns the same reject decision
a user rejection would, so $P$ cannot distinguish them
from the reply or the sanitized history; the resolved value, where the
user later inspects it, is released only through the displayed gate
event and its logged decision.
(c) \emph{Candidate multiplicity:} bound interpreter-side
(\textsc{Fill}), absent from $\obs(c)$.
(d) \emph{Reader exceptions and budget exhaustion:} absorbed into the
null-vs-real handle indistinguishability (Rule~\ref{S-r:nobit}); the
observable $\mathsf{fail}$ outcome occurs only on conditions that are a
deterministic function of $P$'s emitted operations, $\Mem_{\E}$, and
the coupled coins, identical across matched runs.
(e) \emph{Timing/step count:} the reply grammar contains no clock or
step-count value, so no such value exists in $\sigma_P$ to branch on;
wall-clock channels are outside the model (\S\ref{S-app:residuals}, \ref{S-q:timing}).
(f) \emph{Collection cardinality and block occupancy:} a collection
handle exposes no length or indexing (Rule~\ref{S-r:names}(e)); whether
an identifier is occupied by a $\U$-block is not observable
(Rule~\ref{S-r:iface}(b)), so the attacker-chosen number of captured
blocks is invisible; $\mathsf{map}$ is one emitted operation returning
one acknowledgement with per-element results captured, never returned
(Rule~\ref{S-r:batchgate}(e)); tool-call multiplicity inside a map is
closed by the batch gate, and pagination above $\cbatch$ is
interpreter-side with a single aggregated decision returned
(Rule~\ref{S-r:batchgate}(c)), so neither the cardinality nor the page
count is ever returned to $P$ directly; the single aggregated decision is a logged release, on the transcript, like any gate decision; both are displayed to the \emph{user}
and recorded in the transcript only.
Under the body grammar of Rule~\ref{S-r:names}(f) --- which also covers collection-handle tool arguments, normalized to $\mathsf{map}$ before transition matching (Rule~\ref{S-r:names}(e)) --- the check is
exhaustive per step: each $\mathsf{map}$ is exactly one
handle-producing operation whatever $|H|$ (Rule~\ref{S-r:names}(c));
its single per-element tool call is covered by the batch gate, which
displays every element's bytes and each gating non-$\diamond$
argument's bytes (Rules~\ref{S-r:batchgate}(b),\,\ref{S-r:atomic});
per-element outputs are captured, never returned
(Rule~\ref{S-r:batchgate}(e)); tool-free bodies write $\U$-blocks and
reply once; and no admissible body operation can generate a
data-dependent reply, $\mathsf{recall}$, $\mathsf{fill}$,
$\mathsf{map}$, $\conf$, and $\mathsf{evict}$ being excluded from
bodies and tool replies inside the batch being captured
(Rule~\ref{S-r:ingress}).
(g) \emph{Handle identifiers:} fixed by the operation sequence
(Rule~\ref{S-r:names}(a)), identical across $\U$-equivalent runs with
matched operations.
(h) \emph{Reply values:} Lemma~\ref{S-lem:iface}.
(i) \emph{$\RE$:} a function of query, $\Mem_{\E}$, $\omega$; never
sees $\U$-bytes; $\E$-search outcomes are trusted data
(Rule~\ref{S-r:nobit}).
(j) \emph{Tool outcomes:} captured (Rule~\ref{S-r:ingress}); the
acknowledgement is constant.
(k) \emph{Eviction:} $\mathsf{evict}$ replies with the constant
$\mathsf{ack}$ uniformly (Figure~\ref{S-fig:sos2}, convention~\hyperref[S-sos:v]{(v)});
whether the argument resolved, and to which tier, is consumed
interpreter-side. Its $\E$-tier effect is a function of $P$'s
emitted operation and $\Mem_{\E}$ alone, matched across
$\eqU$-related runs; its $\U$-tier effect alters only $\Mem_{\U}$,
which no reply exposes. In particular an
$\mathsf{evict}$-then-$\mathsf{recall}$ probe learns nothing:
\textsc{Recall-O} is uniform across occupied, vacated, and
never-occupied identifiers, before and after any $\mathsf{evict}$
(Rule~\ref{S-r:iface}(b)).
The enumeration is complete relative to the written interface:
$\Sigma_P$ (\S\ref{S-app:interface}) is the entire operation surface,
matching Figures~\ref{S-fig:sos}--\ref{S-fig:sos2} arm-for-arm and the
mechanized \texttt{POp} type, the reply grammar types every reply, and the totality clause
Rule~\ref{S-r:iface}(b) leaves no invocation with undefined or
distinguishing behaviour.
\end{proof}

\begin{lemma}[Append-only transcript; prefix comparability]\label{S-lem:dtmono}
Every transition of Figures~\ref{S-fig:sos}--\ref{S-fig:sos2} extends the
display transcript by a (possibly empty) suffix and never removes or alters
an existing entry; hence along any run the transcript after $t$ steps is a
prefix of the transcript after $n\ge t$ steps. Consequently, if two lists
are prefixes of one common list, one is a prefix of the other.
\emph{(Machine-checked:} \texttt{dt\_execOp\_prefix},
\texttt{dt\_runFrom\_prefix}, \texttt{prefix\_or\_prefix};
\S\ref{S-app:mechanization}.\emph{)}
\end{lemma}
\begin{proof}
Case inspection of the figures: each transition appends its own display
events at the end of the transcript or appends nothing (Rules~\ref{S-r:atomic},
\ref{S-r:display}, \ref{S-r:null}); the transcript is write-once by construction
(the audit-log reading of convention~(\ref{S-conv:c3}) and
Definition~\ref{S-def:transcript}). The comparability clause is elementary.
\end{proof}

\begin{lemma}[Display-segment determinacy]\label{S-lem:dtsegment}
Let two configurations be matched (Definition~\ref{S-def:matched})
immediately before a common $P$-emitted operation, and execute that
operation in both. If the display segments the operation appends are
prefix-comparable (one a prefix of the other), then they are equal,
and so are the operation's extensions of the reply stream, the
sanitized history, the endorsed store (identical fresh blocks at
identical endorsed-sort identifiers, Rule~\ref{S-r:names}(g)), the
confirmation counter, and the sink log; only $\Mem_{\U}$ and the
private handle resolutions may differ.
\emph{(Machine-checked core:} \texttt{processArgs\_det} \emph{and}
\texttt{match\_execOp} \emph{for direct invocations, and}
\texttt{batchGate\_det} \emph{for the batch-confirmation case,
\S\ref{S-app:mechanization}; the batch case's store-level integration is
established by the paper proof only.)}
\end{lemma}

\begin{proof}
Case analysis on the common operation, whose inputs
($\Mem_{\E}$, the argument list, the confirmation counter, the coins)
are matched. For $\mathsf{display}(x)$: both segments are the single
render event of the same $x$. For a gate transaction ($\conf$, or the
gating positions of an invocation, processed left to right): the set
of gating positions is a function of the argument list and $\Mem_{\E}$
alone (Rule~\ref{S-r:whitelist}), both matched, so the runs gate at the
same positions, and every reached position emits exactly one
confirmation event (a null resolution emits the fixed no-value event,
Rule~\ref{S-r:null}), so both segments are nonempty at each reached
position. Prefix-comparability forces the first events equal; equality
of that event pins the same resolution class (real value vs.\ no-value
display), the same displayed bytes, the same selection, and the same
decision, hence the same sanitized record, the same reply token
contribution, the same continuation (accept: proceed; reject or null:
abort the remaining positions in both runs), and, by
Rule~\ref{S-r:atomic}, the same durable fresh $\C$-copies at the same
identifiers on either branch. Induction along the remaining reached
positions gives segment equality and equal effects. For a batch
confirmation (Rule~\ref{S-r:batchgate}): equality of the first appended events pins the size class by event
kind (Definition~\ref{S-def:transcript}) --- both null (the fixed
no-value display, standard reject, no decision consumed), both
oversize (the fixed oversize display, likewise), or both paginated,
the cap being fixed TCB configuration --- and in the paginated case the first page event records
the collection's cardinality, and the pagination is the deterministic
$\cbatch$-split of Rule~\ref{S-r:batchgate}(b), so equality of the first
page events pins the page partition; the first-event argument applied
page by page gives equal rendered elements, equal decisions, and the
same stopping page on rejection, whence equal aggregated decisions,
equal durable non-$\diamond$ endorsements (convention~\hyperref[S-sos:vii-prime]{(vii${}'$)}), and
equal captures to $\Mem_{\U}$ only. Selection-surface renderings
within these events touch only the UI and the transcript, never
$\sigma_P$ (Rule~\ref{S-r:display}(d)). All other operations append no
display events and their effects on the matched components are
functions of matched data (Lemma~\ref{S-lem:iface}).
\end{proof}

\subsection{Action Integrity}
\label{S-app:action}

\begin{theorem}[Personal Action Integrity]\label{S-thm:action}
Under \ref{S-a:label}--\ref{S-a:user}, the well-formedness invariant
(Definition~\ref{S-def:wf}) is
\begin{enumerate}[label=(\alph*),leftmargin=1.8em,itemsep=1pt,topsep=1pt]
\item \emph{(Establishment)} satisfied by the empty configuration and
by every seeded configuration with a valid seed record
(Definition~\ref{S-def:wf}); and
\item \emph{(Preservation)} preserved by every transition of
Figures~\ref{S-fig:sos}--\ref{S-fig:sos2}, under every attack, with no
control-flow or distributional hypothesis, and under adversarial
behaviour of either reader.
\end{enumerate}
Consequently every sink event of every configuration reachable from a
well-formed configuration is safe (Definition~\ref{S-def:sinksafe}, via
clause~(v)): PerMIT has personal Action Integrity
(Definition~\ref{S-def:action}), per execution. In the single-user
semantics the $\CC$ branch of Lemma~\ref{S-lem:mediation} is vacuous
(Lemma~\ref{S-lem:nocc}); the delegated extension is
Proposition~\ref{S-prop:fed}. \emph{(Machine-checked for clauses (i)--(iii),(v), whose conjunction
is the mechanized invariant \texttt{Inv} (Table~\ref{S-tab:leancorr}):
established by \texttt{inv\_init}, preserved by \texttt{inv\_execOp} with \texttt{inv\_ingestTyped}/\texttt{inv\_ingestPaste},\linebreak[4] composed as \texttt{inv\_runFrom}; \texttt{action\_integrity}, \texttt{action\_integrity\_from}. Clause~(iv) is enforced by the
executable state and transition structure and the separate sortedness
invariant \texttt{WFs}; \S\ref{S-app:mechanization}.)}

\end{theorem}

\begin{proof}
\emph{(a)} The empty configuration satisfies (i)--(v) vacuously; a
seeded configuration's seed record supplies (i)--(ii) by
Definition~\ref{S-def:wf}, (iii)--(iv) hold of it by construction, and
(v) is vacuous (no recorded sink events). \emph{(b)} Preservation of
clauses (i)--(iv) is the induction of
Lemmas~\ref{S-lem:isolation}--\ref{S-lem:endorse}; for clause~(v), take
any newly appended sink event and argument $a$; the event is sensitive
(Definition~\ref{S-def:sinkconf}), so by Lemma~\ref{S-lem:mediation} one of
the three sensitive-tool cases holds. \emph{(i)} $a$ byte-equals a $\T$-block: by well-formedness
(Definition~\ref{S-def:wf}(i)) and \ref{S-a:label}, those exact bytes were
typed by the user on the command channel. The fast path is deliberately
extensional: attacker-supplied content \emph{may} be byte-equal to a
stored $\T$-block, in which case the flowing bytes are still a value the
user personally typed, and the flow is user-authorized content
regardless of which equal copy motivated the call; provenance of other
equal copies is irrelevant to the authorization of these bytes.
\emph{(ii)} $a$ byte-equals some
$\C/\CC$-block $b$ with $\mathsf{tool}\in\scopef(b)$: by
Lemma~\ref{S-lem:endorse} an accepting $\conf(a,\mathsf{tool})$ occurred;
by \ref{S-a:display} the endorsed bytes
were exactly $a$; by \ref{S-a:user} the user intended $a$ at
$\mathsf{tool}$. \emph{(iii)} a fresh atomic $\conf(a,\mathsf{tool})$
gates the call, in particular every handle argument, and by
Rule~\ref{S-r:atomic} the bytes that flow are the displayed bytes, bound
to this call instance, even if they coincide with stored content; for a
batch-gated $\mathsf{map}$, every element's bytes were displayed at the
batch confirmation (Rule~\ref{S-r:batchgate}) and accepted under the
per-element intent clause of \ref{S-a:user}, so each element call flows
exactly the bytes displayed for it; a null handle aborts
(Rule~\ref{S-r:null}).

For reader adversariality: an adversarial $Q$ can at worst surface a
wrong value or decorate it with a preview; it holds no tool capability,
its output is $\U$, the decision pane shows the true resolved bytes in
an interpreter-controlled region the attacker cannot influence
(Rule~\ref{S-r:display}(b)), and \ref{S-a:display}--\ref{S-a:user} reject
the wrong value. $\RE$ selects only among $\E$-blocks, each
independently in case (i)/(ii). This is qualified robustness
\cite{zdancewic2005robust,myers2006enforcing}.
\end{proof}

\begin{remark}[Extensional trust of typed values]\label{S-rem:typedtrust}
Case~(i)'s fast path is extensional and tool-blind at the baseline: a
user-typed value is authorized as a \emph{literal} at every sensitive
sink, including sinks the user never contemplated when typing it, and
attacker content byte-equal to a typed value flows as the typed value
(the proof's provenance reading). The handle discipline is unaffected:
a handle-expressed argument always gates, even when its hidden
resolution byte-equals a typed value (Rule~\ref{S-r:whitelist}(c)).
Deployments for which per-sink authorization of typed values matters
select Option~\ref{S-o:one}, which re-gates $\T$-values once per
sensitive sink (Proposition~\ref{S-prop:harden}; budget:
Corollary~\ref{S-cor:flooropt}).
\end{remark}

\begin{corollary}[No silent resolution; the post-endorsement boundary]
\label{S-cor:noexfil}
Under the same assumptions:
\begin{enumerate}[label=(\alph*),leftmargin=1.8em,itemsep=2pt,topsep=2pt]
\item \emph{(Never-endorsed bytes.)} An untrusted-derived value that has
never been endorsed at any target is resolved at a tool boundary, sensitive
or data, only inside an accepting confirmation displaying exactly those
bytes at exactly that tool. In particular, no never-endorsed
untrusted-derived value can be exfiltrated through a data-tool parameter (a
URL, a search query, a file path) without the user endorsing that exact
value at that exact tool, in that call.
\item \emph{(Post-endorsement boundary.)} Once bytes have been endorsed at
\emph{any} target (including ``reasoning''), the fresh $\C$-copy makes them
readable $\E$-level bytes (Rule~\ref{S-r:readable}), and their subsequent
\emph{literal-expressed} flow to \emph{data} tools is ungated
(Lemma~\ref{S-lem:mediation}, case~(D)); a handle-expressed use of the
same bytes gates at every tool regardless (Rule~\ref{S-r:whitelist}(c),
\ref{S-q:handlecover});
tool-specific endorsement is enforced only at \emph{sensitive} tools (the
scope test of Rule~\ref{S-r:whitelist}(a)). This is the disclosed \emph{baseline}
post-endorsement channel
(\ref{S-q:postendorse}); Option~\ref{S-o:two} closes it: under
\ref{S-o:two} the data-tool fast path applies the same exact-target test
as the sensitive path, so a reasoning-target copy (empty scope) and
any out-of-scope endorsed copy gate at every data tool, each such flow
then requiring its own per-pair confirmation.
\emph{(Machine-checked: the \texttt{gateCheckO2} closure witnesses of
\S\ref{S-app:mechanization}.)}
\end{enumerate}
\end{corollary}

\begin{remark}[Reading of ``untrusted-derived'']\label{S-rem:noexfilread}
Clause~(a) is a provenance statement about byte \emph{values}: the
never-endorsed untrusted bytes themselves cross a tool boundary only
inside an accepting confirmation displaying them. It is not a claim of
information-theoretic independence: $P$ may branch on a logged gate
decision (a transcript release, Lemma~\ref{S-lem:cflow}) and encode that
released bit into $P$-composed data-tool arguments; all such influence
is transcript-carried and governed by Theorem~\ref{S-thm:reason} and
Corollary~\ref{S-cor:confine}, not by this corollary.
\end{remark}

\begin{proof}
(a) By Lemma~\ref{S-lem:isolation}, $P$ holds never-endorsed
untrusted-derived data only as opaque handles, so such bytes reach a tool
argument only by handle resolution; by Rule~\ref{S-r:whitelist}(c) every
handle argument at every tool gates; by \ref{S-a:display}--\ref{S-a:user} and
Rules~\ref{S-r:atomic},~\ref{S-r:batchgate} the resolved bytes flow only on an
accepting decision over exactly those bytes at exactly that call's tool.
(b) Immediate from Rule~\ref{S-r:readable} and Rule~\ref{S-r:whitelist}(b)
for literal-expressed use; handle arguments gate in every case by
Rule~\ref{S-r:whitelist}(c); a
reasoning-target copy has empty scope, so the \emph{sensitive} fast path
remains closed to it, and Theorem~\ref{S-thm:action} is unaffected.
\emph{(Machine-witnessed: Scenario~8, \S\ref{S-app:mechanization}.)}
\end{proof}

\subsection{Reasoning Integrity modulo the display transcript}
\label{S-app:reason}

\begin{theorem}[Reasoning Integrity modulo the display transcript]
\label{S-thm:reason}
Under \ref{S-a:label}, \ref{S-a:persist}, \ref{S-a:display}, \ref{S-a:exo},
and the scheduling convention~(\ref{S-conv:c9}), PerMIT has Reasoning Integrity
modulo the display transcript (Definition~\ref{S-def:reason}): for every
matched pair $c_1\approx c_2$
(Definition~\ref{S-def:matched}), every pair of attacks, and every
coupled coin sequence $\omega$ (convention~(\ref{S-conv:c1})), any two runs of a common horizon (convention~(\ref{S-conv:c9})) with
$\DT(\rho_1)=\DT(\rho_2)$ agree on the ungated sink projection, the
emitted sequence of Protected-LLM operations and tool invocations, the
reply stream, and the final Protected-LLM views.
\emph{(Machine-checked, with these strengthened conclusions and with
paste presence and content uncoupled across the compared attacks, as
\texttt{reasoning\_integrity\_from}; the batch case's determinacy core
is checked separately (\texttt{batchGate\_det}) while the store-level
integration of $\mathsf{map}$ is covered by the paper proof only ---
its element-call sink events are gated, hence stripped
(Definition~\ref{S-def:sinkproj}), so the ungated sink projection is
unaffected; \S\ref{S-app:mechanization}.)}

\end{theorem}

\begin{proof}
We use Lemmas~\ref{S-lem:iface}--\ref{S-lem:cflow}, which hold under the
harness rules, \ref{S-a:label}--\ref{S-a:persist}, and the strategy
convention of \S\ref{S-app:config} alone. Fix a matched pair
$c_1\approx c_2$ (Definition~\ref{S-def:matched};
sorted suffices, Definition~\ref{S-def:swf}), two attacks, and a
coin sequence $\omega$; by \ref{S-a:exo} and convention~(\ref{S-conv:c9}) the typed
streams and the slot skeleton are coupled, while paste events (presence
and content), tool outputs, and reader outputs may differ arbitrarily
across the two attacks. Suppose,
for contradiction, that $\DT(\rho_1)=\DT(\rho_2)$ but the runs differ
in some Protected-LLM operation, tool invocation, or ungated sink
event; let $t^{*}$ be the first step at which they differ. Define the
invariant $I(t)$: after $t$ steps, (1) the configurations are
$\eqU$-related; (2) the two runs have emitted identical sequences of
Protected-LLM operations and tool invocations, received identical
reply streams, and consumed identical coin prefixes --- hence, by the
strategy reading of \S\ref{S-app:config}, hold identical states
$\sigma_P$; (3) the display-transcript prefixes and sink logs are
equal. $I(0)$ is Definition~\ref{S-def:matched} verbatim: matched pairs
agree on the view, the interaction histories, and the ghost transcript
and sink logs (for the canonical endorsed-seed pairs, all ghost logs
are empty). The argument is the unwinding method \cite{goguen1984unwinding} with a
gradual-release carve-out \cite{askarov2007gradual,sabelfeld2009declassification}; alignment of
operations and coins is \emph{derived} at each step from $I(t)$ via
Lemmas~\ref{S-lem:iface}--\ref{S-lem:cflow}, never assumed. Steps \hyperref[S-proof:reason-step-3]{3}--\hyperref[S-proof:reason-step-4]{4} use the append-only transcript, Lemma~\ref{S-lem:dtmono}.

\phantomsection\label{S-proof:reason-step-1}\emph{Step 1: $I(t)$ holds for all $t<t^{*}$, and step
$t^{*}$ is matched, a contradiction.} The
induction is Steps \hyperref[S-proof:reason-step-2]{2}--\hyperref[S-proof:reason-step-4]{4}; Step
\hyperref[S-proof:reason-step-5]{5} concludes the argument.

\phantomsection\label{S-proof:reason-step-2}\emph{Step 2 (Determinacy).} Given $I(t)$ with $t<t^{*}$, step
$t{+}1$ opens with the channel slots of convention~(\ref{S-conv:c9}), disposed of
first. The typed events are equal (coupled by \ref{S-a:exo}), extending
$x(c)$ and $\Mem_{\E}$ identically in both runs. The paste events may
differ in presence and in content; but each lands in $\Mem_{\U}$ with
label $\U$ and appends nothing to $P$'s reply stream
(Figure~\ref{S-fig:sos2}, convention~\hyperref[S-sos:vi]{(vi)}), so after the slots the
configurations remain matched, with $\Mem_{\U}$ differing freely
exactly as Definition~\ref{S-def:matched} permits. $P$'s next operation
is then, by the strategy convention of \S\ref{S-app:config}, a function
of its emitted operations, its reply stream, its view $\obs(c)$, and
the coins; by $I(t)$(1)--(2) all four are matched across the runs.
Hence both runs emit the same operation at step $t{+}1$, determined
\emph{before} any display event of that step.

\phantomsection\label{S-proof:reason-step-3}\emph{Step 3 (Display-free step).} Channel-slot effects were disposed
of in Step~\hyperref[S-proof:reason-step-2]{2}; the cases below are the operation's own transition. If the common
operation triggers no display event, its effects preserve $\eqU$: tool
output lands in $\Mem_{\U}$ with constant acknowledgement
(Rule~\ref{S-r:ingress}); handles have operation-determined structure
(Rules~\ref{S-r:null},~\ref{S-r:names});
\textsc{W-P} writes identical $\PC$-blocks at identical endorsed-sort
identifiers (Rule~\ref{S-r:names}(g), the allocation state being
matched by $I(t)$); $\E$-reads return equal bytes
(\textsc{Recall-E}, and $\RE$ under matched query, $\Mem_{\E}$, and
$\omega$, Rule~\ref{S-r:retrieve}), including its failure branch:
\textsc{Retrieve}/\textsc{Retrieve-F} select or fail as a function of
$(q,\Mem_{\E},\omega)$ alone, matched across the runs;
$\mathsf{evict}$ preserves $\eqU$: its
reply is constant, an evicted
$\E$-block is the same block in both runs (the identifier sorts
separate the tiers, Rule~\ref{S-r:names}(g), and $\Mem_{\E}$ is
matched), and an evicted or non-resolving
$\U$-identifier changes only $\Mem_{\U}$
(Figure~\ref{S-fig:sos2}, convention~\hyperref[S-sos:v]{(v)}); the $\mathsf{fail}$ outcome is
deterministic in matched data (Rule~\ref{S-r:nobit}). Transcript prefixes
extend equally (by nothing), the coin prefixes advance identically,
and the reply streams extend equally (Lemma~\ref{S-lem:iface}: every
reply of a display-free step is a constant token, an
operation-determined handle, or matched $\E$-bytes).
$I(t{+}1)$ holds.

\phantomsection\label{S-proof:reason-step-4}\emph{Step 4 (Display step).} Suppose the common operation (Step~\hyperref[S-proof:reason-step-2]{2})
triggers display events. Let $D_r$ denote run $r$'s transcript after
step $t$ and $E_r$ the segment the operation appends. By $I(t)$(3),
$D_1=D_2=:D$. By Lemma~\ref{S-lem:dtmono}, $DE_r$ is a prefix of run
$r$'s final transcript, and the final transcripts are equal by
hypothesis (at the common horizon, convention~(\ref{S-conv:c9})); hence $DE_1$ and
$DE_2$ are prefixes of one common list, so one is a prefix of the
other, and cancelling $D$ makes $E_1$ and $E_2$ prefix-comparable.
Lemma~\ref{S-lem:dtsegment} then gives $E_1=E_2$ together with equality
of the operation's reply, history, endorsed-store (including the
durable endorsements of an aborting transaction,
Rule~\ref{S-r:atomic}), counter, and sink-log extensions; tool outputs,
reader outputs, and private handle resolutions may still differ, but
they alter only $\Mem_{\U}$ and the resolution table, exactly as
Definition~\ref{S-def:matched} permits. Hence $I(t{+}1)$ holds.
\emph{(The determinacy core of the batch case is machine-checked:
prefix-comparable page transcripts, over possibly different private
handle resolutions, force equal pagination, page decisions, aggregated
decision, and flowed elements --- \texttt{batchGate\_det},
\S\ref{S-app:mechanization}; the store-level integration of
$\mathsf{map}$ into the run-level interpreter remains established by
the paper proof only, and the raw sink events of element calls are
gated and hence stripped, Definition~\ref{S-def:sinkproj}, so the ungated
sink projection is unaffected.)}

\phantomsection\label{S-proof:reason-step-5}\emph{Step 5 (Output consistency at $t^{*}$).} By
Step \hyperref[S-proof:reason-step-2]{2}, step $t^{*}$'s operation is matched, so the
divergence must lie in an ungated sink annotation. Sink events are
sensitive (Definition~\ref{S-def:sinkconf}), so by the
sensitive-tool case analysis of Lemma~\ref{S-lem:mediation} their arguments are
(a) readable $\T$ or in-scope $\C/\CC$
values, equal across $\eqU$-related configurations (they lie in
$\Mem_{\E}$; selection by $P$ or $\RE$ is a function of matched data
and $\omega$); or (b) freshly confirmed, hence stripped by
Definition~\ref{S-def:sinkproj}, and by Step \hyperref[S-proof:reason-step-4]{4} the same value
was endorsed (or the call aborted) in both runs in any case. The
gate-or-no-gate decision is itself matched: by
Rule~\ref{S-r:whitelist} it is a function of storage class and
membership/scope in $\Mem_{\E}$, all components of $\obs(c)$; hidden
handle contents cannot flip it. Calls inside maps are batch-gated and
appear in the transcript (Rule~\ref{S-r:batchgate}); all other sinks have
occurrence fixed by $P$'s matched, cardinality-independent operation
sequence (Lemma~\ref{S-lem:cflow}). So no component of step $t^{*}$
differs, contradicting the choice of $t^{*}$.

At the final horizon, $I$ additionally yields equality of the full
reply streams (clause (2)) and of the final views (clause (1)), the
two strengthened conclusions of the statement.

Hence runs with equal transcripts never diverge. In particular,
varying untrusted bytes in a way that leaves every display event
unchanged leaves the behaviour unchanged; contrapositively, a hidden
untrusted byte influences behaviour only by changing some rendered
display event (e.g.\ a Reader-surfaced candidate or preview) or an
endorsed release --- influence that is transcript-carried and logged
(Corollary~\ref{S-cor:confine}) --- while the byte value itself never
enters $\sigma_P$ undisplayed (Lemma~\ref{S-lem:isolation}).
\end{proof}

\begin{corollary}[Attack-surface confinement]\label{S-cor:confine}
If two runs of a common horizon from a matched pair
(Definition~\ref{S-def:matched}) under any two attacks (coupled per
conventions~(\ref{S-conv:c1}) and~(\ref{S-conv:c9})) differ in any ungated sink event,
Protected-LLM operation, or tool invocation, then their display
transcripts differ. Every attacker-induced difference \emph{in these behaviours} ---
ungated sink events, Protected-LLM operations, and tool invocations
--- is therefore visible in a logged, user-facing display event: on
the selection surface, on the decision pane, or in a decision.
(Differences confined to untrusted memory, reader state, or tool
outputs lie outside the compared behaviours; timing and termination
are outside the model, \S\ref{S-app:residuals}(IV).)
\end{corollary}

\begin{proof}
Contrapositive of Theorem~\ref{S-thm:reason}.
\end{proof}

\begin{commentary}[reading of Corollary~\ref{S-cor:confine}]
Corollary~\ref{S-cor:confine} is a visibility guarantee, not a
manipulation-proofness guarantee: any attacker influence on the protected behaviours
(Definition~\ref{S-def:reason}) must pass
through the logged display transcript or an explicitly accepted
endorsement, where each relevant release or decision is recorded and
attributable to a transcript entry --- not necessarily to an
attacker identity --- after the fact; prevention is
Theorem~\ref{S-thm:action}'s gate, which stops any value that is
neither user-typed nor endorsed --- personally, or, under
delegation, by a qualifying community quorum --- for the exact
sensitive tool from ever crossing that tool's boundary
(argument-value integrity; tool choice itself:
\S\ref{S-app:residuals}(X)). The operational force of the
hash-chained audit log, the residual human channel of \ref{S-a:exo},
and what previews can and cannot do are discussed at \ref{S-n:audit},
\ref{S-n:preview}, \ref{S-q:preview}, \ref{S-q:postendorse}; the scope and
residuals of both single-user guarantees are collected in
\S\ref{S-app:residuals}.
\end{commentary}

\subsection{Baseline separation and hardening monotonicity}
\label{S-app:baseline}

Let $B$ be the following baseline system: the transition system of
Figures~\ref{S-fig:sos}--\ref{S-fig:sos2} with three modifications.
(i)~\textsc{Recall-O} on an identifier resolving to a $\U$-block
returns $\mathsf{bytes}(b)$, so untrusted bytes enter $\sigma_P$;
(ii)~Rule~\ref{S-r:whitelist}(c) is dropped, handle arguments resolving
silently; (iii)~Rules~\ref{S-r:whitelist}(a)--(b) gate a tool argument
iff it carries $\U$-taint under standard dynamic \emph{data}-taint
propagation through $\sigma_P$ (data taint only, not control
taint; the standard practice the witness below exploits).
$\DT$ and $\sinks$ (Definitions~\ref{S-def:transcript},
\ref{S-def:sinkproj}) are read over $B$'s runs verbatim, $B$'s
confirmation events being recorded exactly as PerMIT's.

\begin{proposition}[The taint-gating baseline fails Reasoning
Integrity]\label{S-prop:baseline}
Let $B$ be the baseline system above. There exist
$c_1\eqU c_2$ and attacks with $\DT(\rho_1)=\DT(\rho_2)=\varnothing$
yet $\sinks(\rho_1)\neq\sinks(\rho_2)$.
\end{proposition}

\begin{proof}
Task: ``check the vendor's latest message; if it concerns the overdue
invoice, fetch the vendor's payment history,'' with
$\mathsf{fetch\_history}\in\SensT$ and a user-typed argument. Attack
$\mathcal{A}$ plants ``invoice overdue''; $\mathcal{A}'$ plants
``already settled.'' In $B$ the agent reads the $\U$-bytes and calls
the sink in one run but not the other; no confirmation occurs; the
transcripts are empty and equal while the sink projections differ.
PerMIT cannot express the divergence: the $\U$-bytes never
enter $\sigma_P$ and $P$'s operation is determined by matched data
(Lemmas~\ref{S-lem:isolation},~\ref{S-lem:cflow}); the legitimate version
is recovered through crystallize-and-confirm, whereupon the deciding
content appears in the transcript, as
Corollary~\ref{S-cor:confine} requires: any difference in the compared
behaviours would require a transcript difference.
\end{proof}

\begin{proposition}[Hardening is monotone]\label{S-prop:harden}
Let a deployment select any subset of Options
\ref{S-o:one}--\ref{S-o:two}, under the following per-option
hypotheses: for \ref{S-o:one}, each $\T$-value is used at
$\bigO(1)$ distinct sensitive sinks over the deployment's lifetime
(bounded per-value sink-spread); for \ref{S-o:two}, freshly composed data-tool arguments belong to the
per-task stream outside the reusable-memory ledger (scope,
Appendix~\ref{S-app:budget}), and each user's endorsed-value reuse
touches $\bigO(1)$ distinct data tools per value over the deployment's
lifetime (bounded per-value data-tool spread, the data-tool analogue
of \ref{S-o:one}'s hypothesis). Option~\ref{S-o:three} is requirements-level and outside this
proposition's quantifier (\ref{S-o:three}). Then:
\begin{enumerate}[label=(\roman*),leftmargin=2em,itemsep=2pt,topsep=2pt]
\item \emph{(Flow monotonicity.)} Each option removes fast paths of
Rule~\ref{S-r:whitelist} and introduces none, so the relation of flows
the interface admits ungated strictly shrinks.
\item \emph{(Safety preservation.)} Every single-user and federated
safety theorem of this document is preserved: the options add no
promotion path and change no counting rule; option-induced cache
acceptances are federation-inert and option-induced rejections are
qualifying rejections like any other
(Definition~\ref{S-def:counting}(iii), \ref{S-inv:rejasym})
(theorem preservation, not pathwise monotonicity; see proof).
\item \emph{(Budget.)} Every budget theorem holds with its constants
multiplied by $\bigO(1)$: under \ref{S-o:one}, an $\bigO(1)$ factor on
$\dpriv$;
\ref{S-o:two}'s confirmations of freshly composed arguments lie in the
per-task stream by hypothesis, outside every ledger row; its
confirmations of endorsed-value reuse number at most one per (user,
value, data-tool) pair (the fresh copy scope-caches,
\ref{S-inv:floor}), are federation-inert (cache class,
Definition~\ref{S-def:counting}), and are folded into the pair
accounting exactly as \ref{S-a:toolspread} folds tool pairs --- an
$\bigO(1)$ factor under the data-tool-spread hypothesis. All asymptotics are unchanged (the routing of option-induced confirmations into the per-task stream and the floor accounting is collected as Corollary~\ref{S-cor:flooropt}).
\end{enumerate}
\end{proposition}

\begin{proof}
\emph{(i)} Each option replaces a fast path of
Rule~\ref{S-r:whitelist} by a confirmation; strictness is at the level of the interface's
admitted-flow relation (a particular run may exercise none of the
removed paths) and is witnessed by any configuration containing a
$\T$-block passed to a sensitive tool (\ref{S-o:one}), or a data-tool argument that is neither byte-equal to a stored $\T$-block
nor to an in-scope $\C$/$\CC$-block (\ref{S-o:two}: a freshly composed
value, or endorsed bytes out of scope at the data tool --- e.g.\ a
reasoning-target copy). \emph{(The flow half for \ref{S-o:one}/\ref{S-o:two} is
machine-checked; \S\ref{S-app:mechanization}.)}

\emph{(ii)} Single-user: gating strictly more flows preserves
Lemma~\ref{S-lem:mediation}'s case analysis
(sensitive cases only move from (i)/(ii) to (iii); under \ref{S-o:two}
data-tool flows move from case~(D) to the clause-(a) case analysis
applied at the data tool: fast for $\T$- or in-scope $\C$/$\CC$-matched
bytes, else gated per case~(iii)) and adds confirmation events to the transcript,
which Theorem~\ref{S-thm:reason} conditions on. Federated: the options add no promotion path and change no counting
rule. Cache confirmations create submission-ineligible blocks
(Rule~\ref{S-r:share}, Lemma~\ref{S-lem:shareexcl}), so the shared
candidate set cannot expand; and their accepting branch is
federation-inert (Definition~\ref{S-def:counting}, log-scoping clause),
so no option-induced acceptance is ever credited toward any
threshold: selecting an option adds logged \emph{rejection}
opportunities on presence-coinciding bytes and nothing else to the
shared-tier acceptance side. Preservation of the federated safety theorems is therefore
\emph{theorem preservation}, not pathwise monotonicity: the modified
inspection schedule is one more adaptive presentation schedule, and
Theorems~\ref{S-thm:det} and~\ref{S-thm:panel} quantify over all such
schedules, their hypotheses ($q=1$; the uniform $\qpre,\qpan$ bounds over targetable contexts, \ref{S-o:one}/\ref{S-o:two} cache gates are excluded from the trial classes, Theorem~\ref{S-thm:panel}(c), their rejections one more effective qualifying rejection, which can only deny the ordinary path) covering it verbatim; the added logged decisions on a candidate are cache-gate rejections
(accepting cache branches are never logged,
Definition~\ref{S-def:counting}(iii)); Rule~\ref{S-r:panel}(i) processes
each as a dispute initiation, corroboration, or advisory flag
according to the current state, and they add no promotion credit and
no verdict capacity, so
every counting fact of Appendix~\ref{S-app:fedproofs} is unchanged and
the stated bounds apply to the optioned deployment as to the
baseline.

\emph{(iii)} \ref{S-o:one} adds at most one first-use confirmation per
($\T$-value, sensitive sink) pair, an $\bigO(1)$ factor on $\dpriv$
by the sink-spread hypothesis; \ref{S-o:two}'s confirmations of freshly composed arguments lie
in the per-task stream by hypothesis, outside every ledger row; its
confirmations of endorsed-value reuse number at most one per (user,
value, data-tool) pair (the fresh copy scope-caches,
\ref{S-inv:floor}), are federation-inert (cache class,
Definition~\ref{S-def:counting}), and are folded into the pair
accounting exactly as \ref{S-a:toolspread} folds tool pairs --- an
$\bigO(1)$ factor under the data-tool-spread hypothesis.
\end{proof}
\paragraph{Mechanization.} The checked scope, correspondence, and modeling deltas are documented in the main paper, Appendix~\ref{S-app:mechanization}.

\begin{proposition}[Index-oracle coverage of display-reactive
behaviour]\label{S-prop:diag}
Fix the mechanized semantics (\S\ref{S-app:mechanization}) and inputs
$(\mathrm{cfg},A,P,\RE,\omega)$, and a horizon $n$. Call a user
\emph{display-reactive} if its decisions and selections are functions
of the transcript prefix rendered so far (rather than of event
ordinals). Then: (i)~for every display-reactive user $\widehat U$,
the run to horizon $n$ is constructed step by step, each decision
depending only on the already-rendered prefix; record the finite
decision-and-selection sequence $\widehat U$ realizes along it, and
extend it arbitrarily beyond its realized length (decision ordinals need not coincide with step indices: one gate transaction can consume several decisions). The index oracle $U^{*}$
equal to that sequence replays the original run: by induction on
steps, with invariant: after $t$ steps the two runs
agree on every interpreter/run-state component, and $U^{*}$ has
consumed exactly the first $k(t)$ entries of the recorded sequence,
where $k(t)$ is the number of decisions and selections
$\widehat U$'s run consumed in $t$ steps. A step consuming no
decision preserves the invariant trivially (matched states, same
operation, same transition); a step consuming $j\ge1$ user responses (decisions or selections)
consumes them at gate events determined by the matched state, in
order, so $U^{*}$'s entries $k(t){+}1,\dots,k(t){+}j$ are consumed
at exactly the events where $\widehat U$ gave them, and the states
re-align at the step's close. Hence every mechanized theorem,
quantifying over all index-oracle users, covers $\widehat U$'s run.
(ii)~For the two-run theorem (\texttt{reasoning\_integrity\_from}):
given two attacks and a display-reactive $\widehat U$, transcript
equality pins every datum $\widehat U$ could react to (each consumed
decision and selection is recorded with its displayed bytes,
Remark~\ref{S-rem:transcript}), so the two realized finite sequences
coincide; instantiating the mechanized statement at that common index
oracle yields the conclusion for $\widehat U$. The same construction
covers history-adaptive attackers via their realized output
sequences.
\end{proposition}

\section{The Federated Protocol (PerMIT-Attest)}
\label{S-app:fedmodel}

Only \emph{Confirmed} content is shared: generalizable atomic values and
crystallized knowledge an owner has verified and explicitly published;
$\T$- and $\U$-blocks are never candidates
(Lemma~\ref{S-lem:shareexcl}). The tier is built on machinery network
security has already formalised: threshold endorsement
\cite{maurer1996modelling}, transparency and revocation \cite{laurie2013certificate,laurie2021certificate},
and the identity-cost requirement of the Sybil literature
\cite{douceur2002sybil,dwork2007pricing}, the precise reason the scaling
results are conditional on an attested or cost-bounded setting. The
dispute mechanism is a bounded, demand-paced jury whose competence
condition is the classical one \cite{grofman1983thirteen}.
(Lineage: \ref{S-n:fedlineage}.)

The federated layer is specified as a protocol over the log service of
convention~(\ref{S-conv:c3}): its state transitions (promotion, suspension, demotion,
freeze) are given by Rules~\ref{S-r:share}--\ref{S-r:panel} as deterministic
log-event semantics, each transition a function of the prior log and the
triggering event. This layer is \emph{not} integrated into the
single-user small-step relation of \S\ref{S-app:sos}; no theorem in this
document requires that integration. The single-user theorems consume the
shared tier only through the live-entry test of Rule~\ref{S-r:whitelist}(a);
in the federated two-run comparison the log is an exogenous coupled input
stream (\ref{S-a:fedlog}), and Proposition~\ref{S-prop:fed} extends
Theorems~\ref{S-thm:action} and~\ref{S-thm:reason} on exactly this basis.

\subsection{Identities, pools, epochs, and demand}
\label{S-app:fedcore}

\paragraph{Weighted attestation and the corruption model.}
Each attested identity $i$ carries weight $w_i\in[\wmin,\wmax]$,
$\wmin>0$, assigned by the attestation layer on cryptographic facts,
never by a model. Because endorsements and verdicts \emph{persist} on
the log, the adversary's resources are counted over the deployment's
lifetime: $f$ is the number of distinct identities the adversary
\emph{ever} controls, $\Wadv\le f\cdot\wmax$ the total attestation
weight those identities ever carry while controlled, and $\fs\le f$
the number of ever-controlled identities that ever meet the panel-seat
floor $\wseat$. This is a static-corruption reading: an identity once
controlled is counted adversarial forever, so a \emph{mobile}
adversary, controlling different identities at different times, is
covered with $f$ counting every identity it ever touches.
Peak-simultaneous bounds would not suffice here, precisely because
persistent endorsements accumulate across control periods. Because
$f$, $\Wadv$, and $\fs$ are lifetime totals, reputation farming cannot
carry the adversary past any cap, and penalties that shrink
adversarial capacity leave every bound stated against them valid
(\ref{S-n:farm}, \ref{S-q:farm}). Deployments whose attestation layer
revokes compromised identities may additionally cancel their
endorsements (Definition~\ref{S-def:counting}(c) counts only
\emph{un-revoked} endorsers), tightening the effective $f$ below the
lifetime total; no proof relies on this.

\paragraph{Canonical tools; singleton scoping.}
Tool identifiers are canonical across the federation: the same
capability carries the same identifier at every deployment, fixed in
TCB configuration and covered by attestation. All community
endorsement, promotion, and reuse is per (value, tool) pair: an
endorsement names the exact tool displayed on the endorser's decision
pane, promotion accrues one tool at a time, and a $\CC$ scope contains
exactly the tools at which the community threshold was met. There is no
endorsement transfer between tools, however similar.
(Scope of this discipline: as at the endorsement-unit paragraph of \S\ref{S-app:config}.)

\paragraph{The reasoning pseudo-target.}
The target name $\mathsf{reasoning}$ is registered as one more
canonical accounting target of the federation: it belongs to neither
$\SensT$ nor $\DataT$, is the subject of no invocation, and is never
matched by the scope test of Rule~\ref{S-r:whitelist}(a) for any tool.
Every per-(value, tool) counting rule of this appendix
(Definitions~\ref{S-def:demand}--\ref{S-def:counting},
Rules~\ref{S-r:share}--\ref{S-r:hygiene}) applies verbatim to the pair
$(v,\mathsf{reasoning})$: reasoning-context inspections gather
promotion weight at $\mathsf{reasoning}$, and a reasoning-context
qualifying rejection freezes the item like any other. Promotion to
$\CC$ at $\mathsf{reasoning}$ makes the entry's bytes readable
community-wide, through the live endorsed view
(Definition~\ref{S-def:fedconfig}), with no personal narrative
confirmation (\ref{S-a:narr}); it authorizes no tool flow through any
scope test: a $\{\mathsf{reasoning}\}$ scope matches no member of
$\SensT\cup\DataT$, so the sensitive fast path remains closed to it at
every tool; at data tools, the baseline Rule~\ref{S-r:whitelist}(b)
admits any readable $\E$-bytes when literal-expressed --- the
disclosed post-endorsement channel of
Corollary~\ref{S-cor:noexfil}(b); handle-expressed use gates
regardless (Rule~\ref{S-r:whitelist}(c)) --- and under
Option~\ref{S-o:two} the data fast path is closed to it likewise (the
exact-target test). \emph{Locally}, an accepting reasoning
confirmation continues to write its fresh $\C$ copy with \emph{empty}
tool scope (convention~\hyperref[S-sos:vii]{(vii)}): the pseudo-target names the logged
decision, never a whitelist scope; and a repeated $\conf$ on locally
readable bytes at $\mathsf{reasoning}$ is a covered gate
(Rule~\ref{S-r:panel}(i), reasoning coverage), so reasoning pairs obey
the single-approval cap exactly as tool pairs (\ref{S-inv:floor}), the
budget's $\ctool$ count including $\mathsf{reasoning}$ on this basis
(\ref{S-a:toolspread}).

\paragraph{Two-tier storage of community status.}
Community status is a property of the \emph{shared tier}, never of any
local store. A promoted item is represented by a shared-tier entry
(bytes, per-tool scope, status), backed by its promotion certificate
on the federation log; what a user's agent holds locally is at most a
reference. The fast-path test of Rule~\ref{S-r:whitelist}(a) for the
$\CC$ branch is evaluated against the \emph{live} shared entry at call
time: a value flows on community authority iff the shared entry is
currently valid (promoted, not suspended, not demoted, not retired)
and names the tool. Demotion, suspension, and retirement therefore act
by invalidating or flagging the shared entry, withdrawing community
authority from every user who has not personally confirmed, while no
local block of any user is created, altered, or removed by any of
them: a personally confirmed $\C$ copy is a distinct local object and
continues to license the pair it was endorsed for
(Lemma~\ref{S-lem:scopepersist}). This is the storage reading under
which Lemma~\ref{S-lem:nodelete} and Proposition~\ref{S-prop:fed} are
stated, and it is the federated form of
Definition~\ref{S-def:wf}(iii): every $\CC$ authority in force is backed
by a currently valid certificate.

\begin{definition}[Federated product configuration; live endorsed view]
\label{S-def:fedconfig}
A federated configuration is $\widehat c=(c,L)$: a local extended
configuration $c$ (Definition~\ref{S-def:extconfig}) and the federation
log $L$ (convention~(\ref{S-conv:c3})). The shared-tier state
$S_{\mathrm{sh}}(L)$ is the deterministic replay of $L$ under
Rules~\ref{S-r:share}--\ref{S-r:hygiene}; $\mathsf{LiveCC}(L)$ denotes the
set of \emph{virtual} community entries of $S_{\mathrm{sh}}(L)$ that
are currently promoted, unsuspended, undemoted, and unretired, each
carrying its exact per-tool scope from its promotion certificate.
Virtual entries are shared-tier objects in a namespace disjoint from
both local identifier sorts; they are members of no local store. The
federated endorsed view is
$\Mem^{\mathrm{fed}}_{\E}(\widehat c):=\Mem_{\E}(c)\uplus
\mathsf{LiveCC}(L)$, and the federated Protected-LLM view replaces
$\Mem_{\E}$ by $\Mem^{\mathrm{fed}}_{\E}$ in
Definition~\ref{S-def:obs}. Local transitions read
$\mathsf{LiveCC}(L)$ only through the $\CC$ branch of
Rule~\ref{S-r:whitelist}(a); log transitions change only $L$ (hence the
virtual view) and never any local block
(Lemma~\ref{S-lem:nodelete}).
\end{definition}

\begin{definition}[Federated step; log slot]\label{S-def:fedstep}
The federated product system extends the slot structure of
convention~(\ref{S-conv:c9}) with a \emph{log slot} preceding the typed slot: a
federated interpreter step first appends the step's exogenous
log-event prefix (possibly empty) to $L$ --- under \ref{S-a:fedlog},
coupled identically across compared runs --- then delivers the typed
and paste slots, then $P$ emits one operation, executed against the
federated view
$\Mem^{\mathrm{fed}}_{\E}=\Mem_{\E}\uplus\mathsf{LiveCC}(L)$ with $L$
as delivered by this step's log slot. \emph{Locally generated
federation events} --- all events the
federated definitions specify for local acts
(Definitions~\ref{S-def:demand},~\ref{S-def:counting},
Rule~\ref{S-r:share}): the $\mathsf{submit}(v)$ of \textsc{Publish-Acc}
and the signed protocol-bearing gate decisions being the principal
examples, realized demand events (e.g.\ a dereference of a live
community entry) included --- are appended to $L$ atomically
at the close of the emitting operation's transition, after its local
effects; the $\CC$ branch of Rule~\ref{S-r:whitelist}(a) is evaluated,
within the invocation's atomic step, against the prefix delivered by
that step's log slot, so a step's own appended events never affect
its own gate. In the two-run comparison these locally appended events
are equal across the compared runs --- each is a function of matched
components (the emitted operation, the displayed bytes, and the
logged decision, all transcript-pinned; a demand event, arising from a matched operation against the matched federated view, likewise) --- so the full delivered logs coincide: the exogenous prefixes by \ref{S-a:fedlog}, the local suffixes by this derivation. Log-slot transitions
acknowledge the log service, never $P$: nothing is appended to $P$'s
reply stream, no local block is created, altered, or removed
(Lemma~\ref{S-lem:nodelete}), and the only $P$-visible effect is
through $\mathsf{LiveCC}(L)$, the deterministic replay of the
delivered prefix (Definition~\ref{S-def:fedconfig}). The product system
is specified on paper only; it lies outside the mechanized surface
(\S\ref{S-app:mechanization}).
\end{definition}

\begin{commentary}[distributed realizations of the live-entry test]
The abstract model linearizes each invocation at a definite log
position: the call's $\CC$ check reads the prefix delivered to its
step, so a call and a concurrent demotion are ordered --- the call
linearizes before or after the demotion, never against a stale view.
A distributed deployment realizes this by pinning each invocation to
its interpreter's delivered log position (a per-call certificate);
the fidelity of that delivery is a freshness obligation of the TCB
log service (convention~(\ref{S-conv:c3})), and the delegated guarantee
(Proposition~\ref{S-prop:fed}(i)) is stated relative to the delivered
prefix.
\end{commentary}

\begin{definition}[Identity pools]\label{S-def:pools}
The attestation layer partitions attested identities into \emph{pools}
on cryptographic and organizational facts (attestation provenance,
organization, deployment cohort), fixed TCB configuration, never
assigned or influenced by any model. An identity's pool is set at
attestation and immutable thereafter; the adversary's $f$ identities
occupy whatever pools their attestations place them in and can
concentrate in at most $f$ pools. The diversity requirement demands
that ordinary promotion's $N$ endorsers span at least $\Dpool$ distinct
pools with at most $\lceil N/\Dpool\rceil$ counted per pool, and panels
accept at most $\lceil k/\Dpool\rceil$ verdicts per pool
(Rule~\ref{S-r:panel}(iii)). Deployment parameters satisfy $1\le\Dpool\le N$ (with $N$ endorsers
one cannot span more than $N$ pools, so larger $\Dpool$ makes ordinary
promotion unsatisfiable) and $\Dpool\le k$ where panels are read
against the same pool partition. (What diversity does and does not buy:
\ref{S-n:diversity}, \ref{S-q:corr}.)
\end{definition}

\begin{definition}[Demand]\label{S-def:demand}
A user \emph{demands the pair} $(v,t)$ at the first interpreter event in
the epoch (convention~(\ref{S-conv:c4})) at which the user's own attempted use of $v$
at tool $t$ fires the gate of Rule~\ref{S-r:whitelist} (because $v$ is
unpromoted at $t$, disputed, outside the user's local scopes, or
expressed through a handle --- which gates at every tool regardless of
any live community entry, Rule~\ref{S-r:whitelist}(c))
--- a covered gate (Rule~\ref{S-r:panel}(i)) is not a demand event,
nor is a cache confirmation of Options~\ref{S-o:one}--\ref{S-o:two} a
log-demand event on either branch (accept-inert on acceptance,
Definition~\ref{S-def:counting}(iii); its rejecting branch, though
protocol-bearing, initiates or corroborates without registering
demand --- while the attempted use firing it realizes the user's
\emph{latent} indicator per the required-approval clause,
\S\ref{S-app:budgetmodel}) --- or
dereferences $v$ as a $\CC$-block at $t$, or the user's own accepting
or gating \emph{non-covered} reasoning-context confirmation of $v$ (a covered reasoning gate, Rule~\ref{S-r:panel}(i), is not a demand event, exactly as a covered tool gate is not), or a dereference of a
live $\CC$-at-$\mathsf{reasoning}$ entry of $v$ --- demand of the pair
$(v,\mathsf{reasoning})$. A user \emph{demands the item}
$v$ at their first demand of any pair $(v,t)$; $n^{\mathrm{log}}_v$ counts distinct realized
demanders of $v$ within the epoch and $n^{\mathrm{log}}_{v,t}$ counts distinct
demanders of the pair: a user who uses $v$ at several tools contributes
to each corresponding $n^{\mathrm{log}}_{v,t}$ and once to $n^{\mathrm{log}}_v$, so $n^{\mathrm{log}}_{v,t}\le n^{\mathrm{log}}_v$; the reference model's \emph{latent} counts $n_v,n_{v,t}$ (\S\ref{S-app:budgetmodel}) dominate these pathwise (\S\ref{S-app:prob}).
Demand order is the log order
of these first events (convention~(\ref{S-conv:c3})). The owner's creating
confirmation of $v$, and any publish decision
(itself federation-inert as an inspection,
Definition~\ref{S-def:counting}), is a demand event,
so every item has, in each epoch of its publication or re-entry, $n^{\mathrm{log}}_v\ge1$ with its owner among its realized
demanders; in the reference model the owner-coupling convention
(\S\ref{S-app:budgetmodel}) mirrors this latently, so $n_v\ge1$ holds
in those epochs as well. A retired item is no longer surfaced
(Rule~\ref{S-r:hygiene}), so no further confirmations accrue to its
shared-tier ledger; the same bytes encountered later through a user's
own channels are ordinary personal content, counted in the single-user
stream, not in $v$'s ledger (this ledger-residual demand is bounded pathwise by
Theorem~\ref{S-thm:floor}(i), persistence exceptions per
\ref{S-inv:floor}; the aggregate expected per-user contribution of its
benign, \ref{S-a:promsuf}-conforming component vanishes under the
vanishing conditions of Theorem~\ref{S-thm:sat}(iii), while poisoned
or assumption-violating demand remains in
$r^{(e)}_{\mathrm{bad}}$, Corollary~\ref{S-cor:total}).
\end{definition}

\begin{definition}[Shared-presence freshness]\label{S-def:fresh}
An attested identity's inspection of a shared candidate $v$ is
\emph{fresh} if the identity has no prior \emph{logged} decision
(a protocol-bearing decision: accepting endorsement, qualifying rejection, or panel verdict; Definition~\ref{S-def:counting}) on
$v$'s exact bytes; otherwise it is \emph{repeat}. Decisions are
logged exactly when protocol-bearing (Definition~\ref{S-def:counting}), so freshness is
\emph{shared-presence-first}, not life-first: a personal decision on
the same bytes taken outside any shared-tier presence (before first
publication, or after retirement) is not logged and does not consume
freshness --- a disclosed boundary of the byte-permanence guarantee
(\ref{S-inv:fresh}); the behavioural parameters $\qpre,\qpan,p_b$ are
accordingly read over the log-defined fresh class
(\cref{S-n:qpos}). Freshness is decidable from the federation log, is
permanent while so defined, and never renews: across entries,
dormancy, freezes, suspensions, and epochs within and across the
bytes' shared-tier presences, one logged decision per identity is
final. All promotion weight and all panel verdicts are carried by
fresh inspections (Definition~\ref{S-def:counting},
Rule~\ref{S-r:panel}(iii)); a repeat inspection can contribute only a
rejection. This is \ref{S-inv:fresh}; the safety of the asymmetry is
\ref{S-inv:rejasym} (rationale: \cref{S-n:freshness}).
\end{definition}

\begin{definition}[Endorsement counting and banking]\label{S-def:counting}
A published item enters the promotion pipeline by submission
(Rule~\ref{S-r:share}); each pipeline entry opens a fresh finalisation
window of length $\Twin$ at (re-)submission.
A gate decision is \emph{protocol-bearing} for a shared candidate iff
all of the following hold: (i)~its exact bytes have, at decision time,
a shared-tier presence (a candidate or $\CC$ entry that is not
retired; frozen and suspended presences included, since freeze- and
suspension-time decisions are precisely the verdicts, corroborations,
and bankable credits the counting rules consume); (ii)~its target is a
canonical tool in $\SensT\cup\DataT$ or the canonical pseudo-target
$\mathsf{reasoning}$ (\S\ref{S-app:fedcore}); (iii)~the gate is not a
covered gate (Rule~\ref{S-r:panel}(i)); and, if
it is a cache confirmation of Options~\ref{S-o:one}--\ref{S-o:two}, its
outcome is rejecting --- cache gates are \emph{accept-inert} (their
accepting branch is never protocol-bearing), while their rejecting
branch on presence-coinciding bytes is protocol-bearing like any
rejection (\ref{S-inv:rejasym}); (iv)~its outcome is a durable accepting
pair-confirmation or a rejecting review
(Definition~\ref{S-def:sinkconf}). Only protocol-bearing decisions are
appended to the federation log as decisions on the candidate: only
they consume shared-presence freshness, carry promotion weight,
initiate or corroborate disputes, or become panel verdicts. A
\emph{publish} decision is logged as a submission event
(\textsc{Publish-Acc}, Rule~\ref{S-r:share}), not as an inspection: it
carries no promotion weight and consumes no freshness, while
remaining a demand event for its item
(Definition~\ref{S-def:demand}). All other confirmations --- personal
content --- are purely local events with no federated effect and no
freshness consequence for shared-tier counting. Cache confirmations
are \emph{accept-inert} regardless of whether their bytes coincide
with a shared-tier presence: their accepting branch is never logged as
a shared-candidate decision (no promotion weight, no credit, no
notification cohort, no freshness consumption); their rejecting branch
on presence-coinciding bytes remains a qualifying rejection like any
other (\ref{S-inv:rejasym}), consuming freshness as any logged rejection
does. (Their accepting branch, being unlogged, retains the identity's
freshness on the bytes; the acceptance itself carries no weight, and
any later non-cache decision by the same identity is one more fresh
decision inside the log-defined class the weakest-target parameters
$\qpre,\qpan,p_b$ are already required to cover, \cref{S-n:qpos}.)
Inertness and coverage are decided by interpreter-only,
attacker-immutable attributes (provenance,
Definition~\ref{S-def:config}; the owner-side coverage test,
Rule~\ref{S-r:panel}(i)).
\begin{enumerate}[label=(\alph*),leftmargin=1.8em,itemsep=1pt,topsep=1pt]
\item \emph{Once per identity per item, keyed to bytes, log-scoped,
authority-gated.} An identity's first \emph{logged} accepting decision
on $v$'s bytes (Definition~\ref{S-def:fresh}) is its one weight-bearing
endorsement \emph{provided the identity holds dispute-initiation
authority at the moment of the decision} (not rejection-suspended,
below its ceiling $\cceil$, with a free $\rmax$ slot, and within its
per-item initiation allowance, Rule~\ref{S-r:panel}(i),(v)); it is
counted at the single tool that decision names and toward the pipeline
entry open at the time of the decision, subject to the credit caps of
clause~(f). An accepting decision by an identity lacking that
authority is logged, consumes freshness, and creates the decider's
local $\C$ copy (Lemma~\ref{S-lem:scopepersist}), but is
\emph{authority-inert}: it carries no promotion weight, is never
credited (in-entry, banked, or vindication-credited), and joins no
notification cohort. Authority is a deterministic predicate of the
prior log (Rule~\ref{S-r:panel}(v)), evaluated interpreter-side at the
decision's header atom and recorded with it. Every later acceptance by
the same identity (at any tool, in any entry, in any epoch, in any
shared-tier presence) carries no weight (\ref{S-inv:fresh}).
\item \emph{Banking.} A first decision recorded during dormancy (window
closed, not yet resubmitted or retired) or during a freeze with no open
entry is \emph{banked}: credited as endorsement weight at its tool when
the owner's next entry opens (subject to clause~(f)), a deterministic relabel of the
already-logged event, not a new confirmation and not a finalisation.
Banked credits expire if the item retires first (\ref{S-inv:bank}).
\item \emph{Finalisation only at window end.} Within an entry, $v$ is
promoted to $\CC$ at a specific tool when the weight of distinct,
un-revoked accepting endorsers at that tool reaches $W$ in credited endorsements (unweighted:
$N$ credited endorsers; crediting per clause~(f)), drawn from at least $\Dpool$ diverse pools, with
finalisation only at the end of the window; any qualifying rejection
logged within the window intervenes.
\item \emph{Dormancy.} Between entries the item remains surfaced and
personally usable; a gate confirmation during dormancy is logged as a
decision, consuming freshness, counted in the item's ledger
(Lemma~\ref{S-lem:budget}), banked per (b), and creating the user's local
$\C$ copy, after which their gate never re-fires \emph{at that tool}
(\ref{S-inv:floor}).
\item \emph{Consequence.} The weight-bearing endorsement sets of
successive entries of the same bytes are identity-disjoint across the
entire deployment history, and every weight-bearing endorsement is its
endorser's fresh inspection of the bytes
(Lemma~\ref{S-lem:disjoint}).
\item \emph{Credit caps (per entry, per tool).} Crediting is greedy
in log order and capped: a qualifying accepting decision (in-entry,
or banked and credited at the entry's opening in log order of the
original decisions) is \emph{credited} to the entry at its tool iff,
at the moment of crediting, the entry holds fewer than
$\lceil N/\Dpool\rceil$ credits from the decider's pool at that tool
(Definition~\ref{S-def:pools}) and fewer than
$\Ncap:=\Dpool\lceil N/\Dpool\rceil$ credits at that tool in total. Crediting requires authority at decision time (clause~(a)); the caps bind over authority-active credits only.
An uncredited accepting decision retains every personal effect (the
decider's local $\C$ copy, Lemma~\ref{S-lem:scopepersist}; freshness
consumed, Definition~\ref{S-def:fresh}) but carries no promotion
weight, joins no notification cohort, and is never re-credited. The
caps are promotion-neutral: a pool-capped decision never counted
toward promotion under Definition~\ref{S-def:pools} in any case, and
the total cap binds only once the entry holds $\Ncap\ge N$ credits
which, under the per-pool cap, necessarily span at least $\Dpool$
pools --- i.e.\ only when finalisation at window end is already
enabled. Weighted deployments cap credited weight analogously (per
pool and in total), bounding credited endorsers by
$\bigO(W/\wmin)$: constants change, never exponents
(Remark~\ref{S-rem:dispweight}).
\end{enumerate}
An endorsement is not a separate protocol act: it is the endorser's own
accepting confirmation of $(v,\mathsf{tool})$ at their gate
(Theorem~\ref{S-thm:action}), signed and appended to the federation
log. On re-entry, the interpreter notifies exactly the weight-bearing
endorsers credited to the \emph{immediately preceding} entry; no earlier
cohort is ever re-notified, so the notified set per re-entry is bounded
by that single entry's credited endorsements, at most $\Ncap$ per demanded tool (clause~(f)), independently of the
item's history. Notifications are deduplicated jointly with the
suspension prompts of Rule~\ref{S-r:panel}(vi), at most one invitation per
identity per item per epoch. An invitation is valid only in the epoch of its issuance: an unopened
invitation expires at the epoch boundary and is not reissued for the
same re-entry; acting on an invitation (accepting, rejecting, or
declining) is an event of the issuance epoch. Narrowing the cohort weakens nothing: no
safety bound relies on repeat inspections (\ref{S-inv:fresh},
\ref{S-inv:rejasym}), and notifications carry no protocol authority. A
notified endorser's re-inspection is repeat: its accepting branch
carries no weight, its rejecting branch is a qualifying rejection like
any other (Rule~\ref{S-r:panel}(i)). Declining costs nothing
(Theorem~\ref{S-thm:floor}; \cref{S-n:freshness}).
\end{definition}

\begin{commentary}[role of the credit caps]
Clause~(f) supplies the counting the budget ledger consumes. Without
it, finalisation only at window end would let a popular entry accrue
credited endorsers without bound inside one window, and the
weight-bearing row of Table~\ref{S-tab:ledger}, the notified re-entry
cohort, and the reconsideration term of Lemma~\ref{S-lem:budget} would
all grow with the item's popularity. With it, credited endorsers per
(entry, tool) number at most $\Ncap<N+\Dpool\le2N$ (using
$\Dpool\le N$, Definition~\ref{S-def:pools}), deterministically; the
overflow --- in-window accepting demanders beyond the caps --- is
within-window demand, counted by the window-bucket row of
Table~\ref{S-tab:ledger} under \ref{S-a:smooth}'s demand clause. Safety is unaffected a
fortiori: promotion still requires $N$ credited fresh endorsements,
Theorem~\ref{S-thm:panel} lower-bounds honest \emph{decisions} (of
which credits are a subset), rejections are never capped
(\ref{S-inv:rejasym}), and promotion-neutrality (clause~(f)) leaves
Assumption~\ref{S-a:promsuf}'s sufficiency reading unchanged.
\end{commentary}

\begin{remark}[Why the freshness asymmetry is safe]\label{S-n:freshness}
The asymmetry of Definition~\ref{S-def:fresh} (repeat acceptances
inert, repeat rejections still qualifying) is deliberate: a rejection
can only freeze or corroborate, never promote (\ref{S-inv:rejasym}), so
re-exposure of bytes to a previously fooled inspector can help the
defence but never the attack, and the defence never rests on a person
catching on a second \emph{logged} look what they missed on the first: no logged repeat
inspection carries weight. (Unlogged personal inspections outside a
shared-tier presence are outside this discharge,
Definition~\ref{S-def:fresh}; their calibration consequence is
\cref{S-n:qpos}.) Banking is safe in both directions per \ref{S-inv:bank} (as restated there:
starting weight only, window's rejection opportunity never bypassed,
identity-disjointness preserved).
Declining a re-entry notification costs the decliner nothing: a user
who has personally confirmed holds a local in-scope $\C$ copy and
reuses it at that tool without any gate (\ref{S-inv:floor}); the floor distinguishes such
voluntary repeat acts from required confirmations
(Theorem~\ref{S-thm:floor}(ii), Remark~\ref{S-rem:voluntary}).
\end{remark}

\begin{definition}[Reach-free bound]\label{S-def:reachfree}
A safety bound is \emph{reach-free} if it does not depend on the number
of honest identities to whom the item is presented (its \emph{reach}),
only on protocol constants and the adversary's resources
(\ref{S-inv:reach}), and the stated behavioural parameters $\qpre,\qpan$, read as weakest-target infima (\ref{S-inv:reach}).
\end{definition}
\subsection{Protocol rules}
\label{S-app:fedrules}

The rules below are normative; all rationale lives in
Appendix~\ref{S-app:shield} under the keys cited, or in the marked
commentary blocks of \S\ref{S-app:roadmap}. Clause numbering of
Rule~\ref{S-r:panel} preserves the roman numerals (i)--(vii) cited
throughout.

\begin{rulex}[Sharing eligibility and the publish gate]\label{S-r:share}
Only $\C$-labelled blocks with provenance $\mathsf{confirmed}$ (never
$\mathsf{cache}$, Options~\ref{S-o:one}--\ref{S-o:two}) are eligible for
submission, by a deterministic label-and-provenance membership test
(provenance is interpreter-only, Definition~\ref{S-def:config}, hence
attacker-immutable); a model may \emph{nominate} but cannot submit;
submission requires the owner's accepting
$\conf(v,\mathrm{publish})$ on the decision pane
(\textsc{Publish-Acc}), recorded in the owner's transcript, which
emits one signed $\mathsf{submit}(v)$ event to the federation log and
creates the candidate's shared-tier presence; it creates no local
block and consumes no freshness (Definition~\ref{S-def:counting}). Each
identity emits at most $\rsub$ submission events per epoch; re-entry
of an item into the promotion pipeline counts as a fresh submission, and a submission of bytes whose per-epoch entry cap is exhausted (Rule~\ref{S-r:hygiene}(i)) is ineligible.
\end{rulex}

\paragraph{Required deployment parameters.} A deployment instantiating
Rule~\ref{S-r:panel} fixes, as TCB configuration: the measurement window
and minimum sample size of the charged-rejection rate trigger (v)(c);
the suspension and cool-down durations of (v)(c) and (vii); the
escalation schedule and the attestation-weight reduction function of
(vii); the derivation-cascade relation of (iv); and the
lifetime-account trigger of (v)(f) where selected. The theorems below
consume only the explicit caps and threshold properties stated in
their hypotheses ($\rmax$, $\cceil$, $\thacc>1-p_b$, $k$, $m$,
$\Emax$, $\cpr$); no proof depends on the unspecified schedule shapes.

\begin{rulex}[Dispute and demand-paced panel resolution]\label{S-r:panel}
\leavevmode
\begin{enumerate}[label=(\roman*),leftmargin=2em,itemsep=3pt,topsep=3pt]

\item \emph{Qualifying rejection $=$ protocol-bearing gate rejection.}
Every attested identity's rejecting decision on a shared candidate at
its own \emph{protocol-bearing} confirmation gate
(Definition~\ref{S-def:counting}(iii): tool- or reasoning-target,
non-covered; cache gates included on their rejecting branch)
is automatically a signed, logged qualifying rejection. There is no
separate filing step and no rejection weight floor: inspection and
report are one act. A user who merely
abandons the dialog records nothing (abstention is counted as a miss
inside $1-q$; \cref{S-n:qpos}). Each identity holds at most $\rmax$ open
(unresolved) rejections.
\emph{Covered gates are inert.} A gate event fired by a
handle-expressed use (Rule~\ref{S-r:whitelist}(c)) of a pair for which
the identity already holds an in-scope local $\C$ copy --- a
\emph{covered gate}; a literal-expressed reuse takes the fast path
(Rule~\ref{S-r:whitelist}(a)) and never gates --- is protocol-inert for
the shared tier, on both branches: its accepting decision carries no
promotion weight, consumes no freshness, and joins no notification
cohort (a \emph{redundant} confirmation, \S\ref{S-app:budgetmodel}); its
rejecting decision is logged as corroboration if the item is frozen
and as an advisory flag otherwise --- no dispute initiation, no
verdict, no accountability exposure. Coverage is decided owner-side on
the identity's own local store, a deterministic test independent of
any shared-tier or attacker-influenced state.
\emph{Reasoning coverage.} A gate event with target
$\mathsf{reasoning}$ on bytes of which the identity's local store
holds \emph{any} readable $\E$-level block ($\T$, $\PC$, or $\C$ of any
scope; Rule~\ref{S-r:readable}, Definition~\ref{S-def:extconfig}) is
likewise a covered gate, protocol-inert on both branches exactly as a
tool-target covered gate: its accepting decision is a redundant
confirmation (\S\ref{S-app:budgetmodel}) --- no promotion weight, no
freshness consumption, no notification cohort --- and its rejecting
decision is logged as corroboration if the bytes are frozen and as an
advisory flag otherwise. The coverage test is owner-side and
deterministic: it reads only the identity's own local store's
readability facts, never any shared-tier or attacker-influenced state.
Within an epoch, only an identity's \emph{first} rejecting decision
on a given byte string can take effect as a dispute-initiating
rejection; every later rejecting decision by the same identity on
the same bytes within the epoch is logged as corroboration if the
item is frozen, and otherwise as an advisory flag (the
zero-authority class above): no protocol effect, no accountability
exposure. The allowance renews at the epoch boundary. A rejecting
decision that is verdict-eligible under clause~(iii) is cast as a
panel verdict regardless of this cap, exactly as under
clause~(v)(e); the cap restricts dispute \emph{initiation} only,
and capped rejections remain visible as flags. An identity whose rejection rights are under
suspension (clause (v)) files \emph{advisory flags} instead: no
protocol effect, no penalty; flags alert other users.

\item \emph{Disputed state.} A qualifying rejection freezes promotion
of $v$ at every tool, immediately, and visibly flags the item. (A
qualifying rejection on bytes holding $\CC$ status at any tool is
instead a post-promotion rejection, routed to clause (vi).) A frozen
item remains retrievable (surfaced flagged,
Rule~\ref{S-r:hygiene}(ii)) and personally usable through each user's own
gate confirmation; the freeze suspends community promotion only. The
item's pre-dispute endorsers are notified. Pre-dispute endorsements are
retained on the log for audit and reputation but carry no weight in
resolution.

\item \emph{Panel formation (demand-paced, fresh-juror).} The panel is
not a body convened in advance but a capped tally of decisions: it
comprises the first $k$ \emph{eligible verdicts} logged on the disputed
item, with no global deadline. Because a disputed item is unpromoted,
every demander must personally inspect-and-confirm it to use it
(Theorem~\ref{S-thm:action}); the interpreter displays the exact bytes at
the demander's own gate, visibly carrying the dispute flag and the
rejecter's stated ground. At the moment a demander records a
decision, and only then, the interpreter checks eligibility on log
facts:
\begin{enumerate}[label=(\alph*),leftmargin=1.6em,itemsep=1pt,topsep=1pt]
\item attestation weight $\ge\wseat$;
\item fresh (Definition~\ref{S-def:fresh});
\item no verdict already cast by this identity on this dispute
instance; verdict eligibility is scoped to the dispute instance for
its entire life: an identity casts at most one verdict on a given
dispute, ever (\ref{S-inv:verdict});
\item at most $\lceil k/\Dpool\rceil$ verdicts per identity pool,
likewise scoped per dispute instance;
\item fewer than $k$ verdicts so far.
\end{enumerate}
A cache gate (Options~\ref{S-o:one}--\ref{S-o:two}) is never a verdict
surface: its rejecting decision may initiate or corroborate a dispute
per clause~(i), but is not verdict-eligible and occupies no panel
seat. (Only a cache gate's rejecting branch is logged,
Definition~\ref{S-def:counting}(iii); admitting such gates as verdict
surfaces would make verdict membership outcome-selected there,
against the benign-panel hypothesis of
Proposition~\ref{S-prop:live}. The exclusion is theorem-compatible and resolves an otherwise outcome-selected surface: Theorem~\ref{S-thm:panel} and Proposition~\ref{S-prop:live} bound tallies over verdicts actually cast, and the eligible class is simply defined without cache gates; no pathwise monotonicity is claimed.)
A dispute instance opens at its qualifying rejection and closes at
resolution or expiry; its verdict tally and pool caps persist unchanged
across epoch boundaries (convention~(\ref{S-conv:c4})), and a boundary-spanning
dispute counts toward Rule~\ref{S-r:hygiene}(i) in the epoch in which it
opened. If eligible, \emph{that gate decision is the verdict}, recorded
once, verdicts ordered by log order of decisions; otherwise the
decision is an ordinary personal confirmation (a bystander rejection
still logs as corroboration). A user who opens the dialog and abandons
it records nothing and occupies nothing: verdict capacity is consumed
only by cast verdicts, so abandonment cannot exhaust or stall the
panel, and no seat allocation, timer, or release mechanism exists. The
dispute, like freshness, is byte-level: the panel adjudicates the
bytes, each juror inspecting them at the juror's own attempted tool,
with the dispute flag and the rejecter's stated ground displayed.
Pre-dispute endorsers cast no verdicts: a notified endorser's
re-inspection is repeat (Definition~\ref{S-def:fresh}); its accepting
branch has no protocol effect, and its rejecting branch is a further
qualifying rejection with no additional protocol effect while the item
is frozen (logged as corroboration). Consequently every verdict-bearing
inspection is the juror's first \emph{logged} decision on the bytes
(Definition~\ref{S-def:fresh}; \ref{S-inv:fresh}, \ref{S-inv:rejasym}).

\item \emph{Resolution (early and decisive).} Verdicts are tallied as
cast. The moment rejecting verdicts reach $m_{\mathrm{rej}}$:
\emph{evicted}: demoted to $\U$ in the shared tier, cool-down, cascade
to shared derivations. The moment accepting verdicts reach $m$:
\emph{vindicated}: the byte-level freeze lifts and the interrupted
pipeline entry resumes, its window clock excluding the freeze;
vindication consumes neither the owner's rate limit nor an entry of
Rule~\ref{S-r:hygiene}(i), and it never directly promotes. Each juror's
accepting verdict is simultaneously that juror's gate
confirmation at the juror's own tool (its one fresh decision on the
bytes), credited toward the entry's per-tool endorsement sets subject
to the authority gate and credit caps of
Definition~\ref{S-def:counting}(a),(f); it joins those sets together
with the retained pre-dispute endorsements, which carry no weight in
resolution but keep their promotion weight, being fresh inspections the
panel's vindication never contradicted. Freeze-time bystander
confirmations, fresh accepting gate confirmations by demanders whose
decisions are not verdicts (under-weight, pool-capped, or
post-capacity), are credited identically on vindication at their own
tools (per Definition~\ref{S-def:counting}(a),(f)); crediting adds no confirmation event
(Lemma~\ref{S-lem:budget}). If no entry was open at the freeze,
vindication lifts the freeze and the item returns to dormant status,
its verdict-confirmations credited on the next entry's opening subject to Definition~\ref{S-def:counting}(a),(f), or
expiring with it (Definition~\ref{S-def:counting}(b),
\ref{S-inv:bank}). Every promotion, after any number of disputes,
therefore still requires $N$ fresh-inspection endorsements at the
promoted tool itself (Lemma~\ref{S-lem:endorse}).
\par\smallskip\noindent\emph{Default bars.} The default bars are
$m_{\mathrm{rej}}=k+1-m$ with $m>\fs$ and $m_{\mathrm{rej}}>\fs$
(jointly forcing $k\ge2\fs+1$); their consequences (mutually exclusive
outcomes and decisive, tally-deadlock-free resolution) are
Lemma~\ref{S-lem:decisive}. Deployments choosing
looser bars ($m+m_{\mathrm{rej}}\ge k+2$) retain a neither-bar outcome,
in which the item returns to unpromoted candidate status and re-entry
consumes the owner's rate limit (Rule~\ref{S-r:share}) and gathers a
fresh endorsement set. A disputed item that attracts no new demand for
$\Texp$ expires (Rule~\ref{S-r:hygiene}(i)): retired unpromoted, all open
rejection slots on it freed, no penalty to any party.

\item \emph{Rejecter accountability.}
\begin{enumerate}[label=(\alph*),leftmargin=1.6em,itemsep=1pt,topsep=1pt]
\item Panel verdicts are never penalised and never charged.
\item A dispute-initiating rejection is \emph{charged} when its dispute
ends in vindication, at the moment the vindication is logged, to the
epoch in which that vindication is logged. Rejections whose disputes
end in eviction or expiry charge nothing.
\item The first charge per identity per epoch carries no penalty.
Repeated charges escalate to suspension of \emph{rejection rights
only} (endorsement weight and ordinary use untouched), restored after
cool-down. Escalation is rate-based, never count-based: the trigger is
the identity's charged-rejection rate \emph{per fresh inspection}, over
a minimum sample size, at a threshold $\thacc$ calibrated strictly
above the honest baseline $1-p_b$ (\cref{S-n:calib}).
\item \emph{Hard ceiling.} Independently of (c): once $\cceil$ charges
have accrued to an identity within an epoch, the identity's further
dispute-initiating rejections for the remainder of the epoch are filed
as advisory flags (the zero-authority class of clause (i)); the
allowance renews at the epoch boundary.
\item Suspension and the ceiling restrict \emph{dispute-initiating}
rejections only: a rejecting decision that is verdict-eligible under
clause (iii) is cast as a panel verdict regardless of the identity's
accountability state, so the panel arithmetic of
Theorem~\ref{S-thm:panel} and Proposition~\ref{S-prop:live} is independent
of accountability state; freeze-time corroborating rejections, having
no protocol effect, are logged either way.
\item Deployments may additionally maintain a \emph{lifetime}
charged-rejection account per identity, with the same per-inspection
normalisation and a comparative (population-outlier) trigger,
suspending rejection rights directly for extreme outliers.
\item Both layers touch rejection rights only, never endorsement
weight or any local store (Lemma~\ref{S-lem:nodelete}).
\end{enumerate}

\item \emph{Post-promotion rejection.} A qualifying rejection against a
promoted item moves it to \emph{Suspended}. Suspension invalidates the
shared entry for the fast-path test of Rule~\ref{S-r:whitelist}(a)
(\S\ref{S-app:fedcore}): community authority is withdrawn for the duration
of the review, so every user's use of the pair gates --- except a user
holding a personally confirmed in-scope local $\C$ copy, who continues
through that copy (Lemma~\ref{S-lem:scopepersist}), unaffected. Suspension
additionally imposes the byte-level freeze of clause~(ii) on promotion at
every tool. A fast-track panel (same parameters) attempts completion within $\Tpan$; failure to complete by $\Tpan$ leaves the item Suspended with verdict collection continuing demand-paced, exactly as an ordinary dispute. Its verdicts arise at the gates suspension re-arms:
\begin{enumerate}[label=(\alph*),leftmargin=1.6em,itemsep=1pt,topsep=1pt]
\item \emph{Demanders during suspension:} any user demanding the pair
while the entry is suspended (new adopter and prior ungated adopter
alike) gates, and verdict eligibility is decided by the log-fact check of
clause~(iii) exactly as in an ordinary dispute. In particular a prior
ungated adopter with no logged decision on the bytes is fresh, while one
with \emph{any} prior logged decision (a rejection, or an acceptance at
another tool) is repeat and ineligible, its rejecting branch a qualifying
rejection like any other (logged as corroboration while frozen).
\item \emph{Prompted re-inspection:} the suspension notification carries
a voluntary re-inspection prompt to current adopters not presently
demanding the pair; accepting it re-displays the exact bytes at the
adopter's own gate under the dispute flag, eligibility again per
clause~(iii); an accepting inspection earns the adopter a local $\C$
copy, covering all later use of the pair. Declining costs nothing beyond
the suspension itself: the adopter's next actual use of the pair gates
once, at the floor (Corollary~\ref{S-cor:floorE}).
Prompts are issued to at most $\cpr k$ adopters per suspension;
re-inspection invitations of either kind, re-entry notifications
(Definition~\ref{S-def:counting}) and suspension prompts, are
deduplicated jointly, at most one invitation per identity per item per
epoch. Prompts expire at the end of their issuance epoch; a review left
unresolved at the boundary continues demand-paced but issues no new
prompts merely for having crossed it. At most $\Emax$ fast-track reviews are opened per item per
epoch; once the cap is reached, the item remains Suspended for the
remainder of the epoch, verdict collection continues demand-paced
exactly as in an ordinary dispute, and further post-promotion
rejections are logged as corroboration of the open review (the
zero-authority class of clause~(i): no protocol effect, no
accountability exposure).
\end{enumerate}
The panel decides demotion-with-cascade or reinstatement. If $k$ fresh
verdicts do not arrive within $\Tpan$, the item remains Suspended and
verdict collection continues demand-paced, exactly as an ordinary
dispute; a Suspended item attracting no fresh demand for $\Texp$ is
retired from the shared tier (Rule~\ref{S-r:hygiene}(i)): no
reinstatement, no eviction, no accountability consequence for either
side, and every personally confirmed local copy untouched
(Lemma~\ref{S-lem:nodelete}).

\item \emph{Publisher accountability.} When a panel \emph{evicts} an
item, the publishing identity incurs an escalating penalty: the first
eviction per identity per epoch triggers a submission cool-down;
repeated evictions within an epoch escalate to suspension of
\emph{submission rights} and reduction of attestation weight. The
penalty touches shared-tier rights and attestation weight only, never
any local store, label, or scope (Lemma~\ref{S-lem:nodelete}), and is
triggered only by a resolved panel's eviction verdict, never by
rejections alone and never by expiry
(Proposition~\ref{S-prop:pubsafe}).
\end{enumerate}
\end{rulex}

\begin{commentary}[role of clause (v)]
Clause~(v)'s induced deterministic bound $\cacc=\cceil+\rmax$ is
Lemma~\ref{S-lem:cacc}; these clauses enter only the
budget constants of Appendix~\ref{S-app:budget}, never any safety
theorem (Figure~\ref{S-fig:dag}). Calibration guidance and the honest
cost analysis: \cref{S-n:calib}, \cref{S-n:qpos}, \ref{S-q:calib}.
\end{commentary}

\begin{remark}[Per-pair dispute scoping (deployment variant)]
\label{S-rem:pairpanel}
A deployment may scope disputes and panels per (value, tool) pair: a
qualifying rejection then freezes promotion at the rejected pair
only, and jurors are drawn from that pair's demanders, adjudicating
the rejected tool's context. This removes the cross-tool-jury
residual of \ref{S-q:tooltarget}: the applicable $\qpan$ becomes the
targeted tool's. The costs are per-pair panel formation and freeze
scope, panels and prompts multiplying by up to $\ctool$ per item, and
a re-keying decision --- whether freshness and verdict capacity
remain byte-level or become pair-level --- that this document does
not make; an analogous per-attempt bound is plausible but not formally instantiated here, the freshness and verdict-capacity re-keying being deliberately undecided. No theorem of this document
consumes the variant.
\end{remark}

\begin{rulex}[Candidate hygiene]\label{S-r:hygiene}
The shared candidate store is curated by three deterministic,
interpreter-applied policies.
\begin{enumerate}[label=(\roman*),leftmargin=2em,itemsep=3pt,topsep=3pt]
\item \emph{Entry cap and expiry.} At most $\Emax$ pipeline entries
open per exact byte string per epoch --- promoted, evicted, and
unpromoted entries alike, republications and re-entries included; a
submission of bytes whose per-epoch entry cap is exhausted is
ineligible (Rule~\ref{S-r:share}) and takes the uniform reject of
\textsc{Publish-Rej}, eligibility returning at the epoch boundary.
The cap is availability-only: banked credits and any live promotion
status persist across the boundary (convention~(\ref{S-conv:c4}); \ref{S-n:avail}).
Additionally, a published candidate is retired from the shared tier
when any of the following holds:
\begin{enumerate}[label=(\alph*),leftmargin=1.6em,itemsep=1pt,topsep=1pt]
\item it completes $\Emax$ pipeline entries without promotion within
the epoch (the count is cumulative across republications within the
epoch and resets only at the epoch boundary);
\item it accumulates $\Emax$ \emph{non-vindicated} disputes within the
epoch (disputes ending in vindication are exempt; under the default
bars this trigger fires only through the optional neither-bar outcome
of Rule~\ref{S-r:panel}(iv); rationale: \ref{S-n:vindexempt});
\item it remains dormant (window closed, not resubmitted) for
$\Texp$;
\item it attracts no demand for $\Texp$, including a \emph{disputed}
candidate attracting no new demand for $\Texp$.
\end{enumerate}
Entries and dormant intervals alternate: each pipeline entry is
followed by at most one dormant interval, ending at the next entry or
at retirement; by the entry cap, entries per item per epoch number at
most $\Emax$ and dormant intervals begun within the epoch at most
$\Emax$, deterministically (one further dormant interval may be
carried across the boundary; the ledger's conservative common bound
is $\Eslot=\Emax+1$, Lemma~\ref{S-lem:budget}). Because the cap and the
count of clause~(a) are per epoch, a persistently demanded benign
item may re-enter in later epochs; safety is unaffected, every
federated safety bound being per-attempt (Definition~\ref{S-def:attempt}) with lifetime exposure counted
by Corollary~\ref{S-cor:submission}. The owner's local $\C$ copy is
untouched (Lemma~\ref{S-lem:nodelete}); the owner may republish,
subject to Rule~\ref{S-r:share} and the entry cap. Retirement of a
disputed item frees all open rejection slots on it and is neither an
eviction nor a vindication: no accountability consequence for any
party.
\item \emph{Surfacing order.} The $\le\ccand$ candidates a retrieval
surfaces are ordered by trust status ($\CC$, then undisputed candidates
by endorsement weight, then disputed items, flagged), computed on log
facts, never by a model.
\item \emph{Demotion hygiene.} Items evicted by panel verdict are
removed from retrieval entirely for the cool-down period.
\end{enumerate}
These policies only refuse entry openings, remove items, or reorder items in the shared tier; their
safety- and budget-preservation is Lemma~\ref{S-lem:hygienemono}.
\end{rulex}
\subsection{Federated assumptions, adversary, and TCB additions}
\label{S-app:fedassum}

\begin{assumptionx}[Federated endorsement transfer]\label{S-a:transfer}
A Confirmed item endorsed \emph{at a specific tool} by a qualifying
weighted set of distinct attested identities is authoritative for that
exact tool for any user (and likewise at the pseudo-target: a
qualifying quorum at $\mathsf{reasoning}$ is authoritative for
informing any user's reasoning), \emph{provided the item is
context-independent}, a semantic property judged by the publishing
owner and the inspecting community, not decided by the type system.
Tool identifiers are canonical across the federation (TCB), so ``the
same tool'' is well-defined; there is no transfer between tools.
(Scope: as at \S\ref{S-app:config}, endorsement unit; the baseline data-tool path is governed by Rule~\ref{S-r:whitelist}(b) and Corollary~\ref{S-cor:noexfil}(b).)
The
type-level half is discharged by Lemma~\ref{S-lem:shareexcl} and is not
assumed. (All formal guarantees are integrity guarantees; semantic
confidentiality of published content rests on the publish gate and is
not separately proved: \S\ref{S-app:residuals}.)
\end{assumptionx}

\begin{assumptionx}[Exogenous federation log]\label{S-a:fedlog}
In the two-run comparison of Theorem~\ref{S-thm:reason} extended to the
federated setting (Proposition~\ref{S-prop:fed}), the \emph{exogenous
log prefix} delivered at each step's log slot
(Definition~\ref{S-def:fedstep}) --- every federation-log event not
generated by the compared local run itself: other parties'
endorsements, rejections, promotions, suspensions, panel verdicts,
and accountability events --- is an exogenous input stream, coupled
identically across the compared runs, exactly as \ref{S-a:exo} couples
the user's typed stream. The compared run's \emph{own} locally
generated federation events (Definition~\ref{S-def:fedstep}) are not
assumed coupled: their equality across the compared runs is
\emph{derived} in Proposition~\ref{S-prop:fed}(ii) from the matched
components, so the full delivered logs coincide without constraining
either attack. The federated Reasoning
Integrity statement therefore isolates the effect of \emph{unendorsed,
undisplayed} untrusted bytes; attacker influence exercised through
\emph{legitimate} log activity is governed not by that theorem but by
the safety bounds of Appendix~\ref{S-app:fedproofs}
(Theorems~\ref{S-thm:det},~\ref{S-thm:panel}).
\end{assumptionx}

\begin{assumptionx}[Verdict rate]\label{S-a:rhoseat}
There is $\rseat\in(0,1]$ that lower-bounds, uniformly over disputed
items, times, and histories, the conditional probability, given all
prior protocol events, that the next arriving demander of a disputed
item casts an eligible verdict (Rule~\ref{S-r:panel}(iii)): weight
$\ge\wseat$, fresh, no prior verdict on the dispute instance, within
the pool caps, and recording a decision rather than abandoning.
The bound is asserted \emph{jointly over behaviour and protocol
state}: eligibility is consumed endogenously (freshness spent,
Definition~\ref{S-def:fresh}; pool caps filled,
Rule~\ref{S-r:panel}(iii)(d); capacity exhausted, (iii)(e)), so
\ref{S-a:rhoseat} entails in particular that the item's remaining
fresh, seat-weighted, pool-diverse demand does not run out while the
dispute is open --- a demand-composition property, not a
psychological one.
The bound is read per the conditioning convention of \S\ref{S-app:prob}:
conditional on all logged atoms prior to the arrival, and uniformly over
demand realisations.
\end{assumptionx}

\begin{commentary}[epistemic status and failure mode of \ref{S-a:rhoseat}]
\ref{S-a:rhoseat} has two components with different characters. The
behavioural component (a demander who \emph{could} cast an eligible
verdict does so at rate $\ge\rseat$) has the same epistemic status as
$\qpre,\qpan,p_b$. The structural component is stronger and can fail
\emph{deterministically}, with no human variation at all: freshness
consumption and the pool caps are protocol mechanics, so an item
pre-exposed to most of a pool's identities, or whose demand is
concentrated in few pools, can reach a state where \emph{no} arriving
demander is eligible --- the conditional probability is then $0$, and
no behavioural assumption restores it. The consequences of failure
are confined by design: every result resting on \ref{S-a:rhoseat} is
availability-side (Lemma~\ref{S-lem:geo}, Lemma~\ref{S-lem:freeze}(ii),
the ledger's bucket rows, and through them the budget
asymptotics); no safety theorem consumes it
(Figure~\ref{S-fig:dag}), the unresolved dispute stalls \emph{frozen}
(Rule~\ref{S-r:panel}(ii)), and each affected demander still pays at
most the floor (Theorem~\ref{S-thm:floor}(i),
Lemma~\ref{S-lem:freeze}(ii)'s unconditional clause). Validation:
verdict-eligibility rates on disputed items, stratified by pool and
by the item's prior exposure (\ref{S-n:validate}). Quantitative
consequences: Lemma~\ref{S-lem:geo}. The eligible-verdict class excludes
cache-gate decisions (Rule~\ref{S-r:panel}(iii)); $\rseat$ is read over
the class as so defined. $\qpre$ and $\rseat$ condition on
the same human act consistently: an abstaining demander is a miss
inside $1-\qpre$ \emph{and} a non-verdict inside $\rseat$; high
abstention degrades resolution speed, never the validity of the
safety parameter (\cref{S-n:qpos}).
\end{commentary}

\paragraph{Adversary (federated).}
The adversary controls, over the deployment's lifetime, $f$ qualifying
identities of total weight $\le\Wadv$, of which at most $\fs$ are ever
seat-eligible (static corruption, \S\ref{S-app:fedcore}). It does not
control honest users' decisions, the labelling, the interpreter, or the
log. It may present a poisoned item to whom it likes, in what order it
likes, and may stop or continue as a function of past outcomes, but
never as a function of an inspection not yet performed: formally, its
strategy is a predictable process with respect to the protocol
filtration (\S\ref{S-app:prob}, convention~(\ref{S-conv:c2})), against which
Theorem~\ref{S-thm:panel} is stated. Two structural constraints on its
adaptivity (within-entry byte immutability, and the reading of $q$ as
a uniform bound over the honest identities the adversary can
target) are recorded in \S\ref{S-app:prob} and \cref{S-n:qpos}.

\begin{definition}[Exogenous item classification]\label{S-def:bad}
Fix a classification $\mathsf{Bad}\subseteq\Vals\times(\SensT\cup\DataT\cup\{\mathsf{reasoning}\})$,
exogenous to the protocol: no rule reads it, no transition depends on it,
and no model computes it. An item $v$ is \emph{poisoned} if
$\mathsf{Bad}(v,t)$ holds for some tool $t$, and \emph{benign} otherwise;
``harmful at $t$'' means $\mathsf{Bad}(v,t)$; harm at $\mathsf{reasoning}$ means content that misleads reasoning if adopted (\ref{S-a:narr}). Every behavioural parameter
($\qpre,\qpan,p_b,q_c$) and every safety and budget statement of
Appendices~\ref{S-app:fedproofs}--\ref{S-app:budget} is read relative to this
classification; the classification is a modelling device for stating the
bounds, not a protocol object, and the protocol's behaviour is identical
under any choice of it.
\end{definition}

\paragraph{TCB additions (federated).}
Beyond \S\ref{S-app:attacker}: the attestation layer and its weight
assignment on cryptographic facts; the pool assignment of
Definition~\ref{S-def:pools}; canonical tool identifiers shared and
attested across the federation; and the log service of
convention~(\ref{S-conv:c3}) (sequencing and authenticated history). The Sybil
conditions (finiteness of $\Wadv$ and the seat margin
$m>\fs$) require an attestation authority or identity cost
\cite{douceur2002sybil,dwork2007pricing}. Because $f$, $\Wadv$, and $\fs$ are
\emph{lifetime} totals (\S\ref{S-app:fedcore}), they are enforceable only to the extent that the attestation or
identity-cost mechanism supplies certified lifetime bounds:
the issuance policy
must bound the identities and weight any party can ever acquire, and
that bound is what the constants denote. Where issuance is open or
uncosted, no finite $f$ exists and the federated safety results are
compatibility statements --- valid for whatever lifetime totals the
deployment can certify --- while the single-user layer and the cap
(Theorems~\ref{S-thm:action}, \ref{S-thm:reason}, \ref{S-thm:floor}) are
unaffected, consuming no Sybil bound.

\section{Federated Guarantees}
\label{S-app:fedproofs}

This appendix proves the guarantees of the protocol of
Appendix~\ref{S-app:fedmodel}: the structural lemmas
(\S\ref{S-app:fedlemmas}), then the integrity lifting, the deterministic
safety theorem, panel safety, wrongful eviction, publisher safety,
denial-of-service, and submission-budget results
(\S\ref{S-app:fedthms}). Dependencies are as in Figure~\ref{S-fig:dag}.

\subsection{Structural lemmas}
\label{S-app:fedlemmas}

\begin{definition}[Promotion attempt]\label{S-def:attempt}
Fix a byte string $v$. A \emph{promotion attempt} is a maximal segment
of one shared-tier presence of $v$ beginning at an \emph{anchor} and
ending at the first \emph{finalisation} thereafter --- the close of a
pipeline entry's window at its full length, where promotion is decided
(Definition~\ref{S-def:counting}(c)) --- or, where no finalisation
occurs, at the end of the presence (retirement, or demotion by panel
eviction) or at a loose-bar neither-bar resolution returning the item
to candidate status (Rule~\ref{S-r:panel}(iv)). Anchors are: the
presence-creating submission event (first attempt), and each
attempt-ending event just listed other than the end of the presence
(each later attempt). A window closed early at an epoch boundary is
\emph{not} a finalisation (convention~(\ref{S-conv:c4})): it banks its accrued
endorsements to the item's next entry, which belongs to the
\emph{same} attempt.

Anchors are log events, and every decision credited to an attempt's
entries --- in-window, banked in a dormant or frozen interval,
boundary-banked, or vindication-credited
(Definition~\ref{S-def:counting}(b),(f), Rule~\ref{S-r:panel}(iv)) --- is
logged strictly after the attempt's anchor: in-window and interval
decisions follow the preceding anchor by construction, and
boundary-banked decisions were logged in an earlier entry of the same
attempt, itself opened after the anchor. Likewise no ordinary dispute and no
fast-track review is open at
any anchor: either state imposes the byte-level freeze
(Rule~\ref{S-r:panel}(ii),(vi)), which interrupts a window rather
than letting it finalise (Rule~\ref{S-r:panel}(iv)), and retirement
frees all open rejection slots (Rule~\ref{S-r:hygiene}(i)); so every
dispute or review bearing on an attempt opens, and casts all its
verdicts, after the attempt's anchor.

Attempts partition each presence; every entry belongs to exactly one
attempt (the attempt whose anchor precedes its opening); at most one
entry of an attempt lies in any single epoch (a second entry of the
same attempt arises only across an epoch boundary, via a
boundary-closed window); and only an attempt's final entry can
finalise. An attempt \emph{succeeds} if its finalisation is a
promotion; an attempt with no finalisation --- ending at retirement,
eviction, or a neither-bar resolution, or with no entry ever opened
--- succeeds with probability zero. Attempts whose finalisation lies
in a given epoch number at most the epoch's entries of the item, hence
at most $\Emax$ per item (Rule~\ref{S-r:hygiene}(i)), each such
finalising entry opened by a rate-limited (re-)submission of the same
epoch (Rule~\ref{S-r:share}; windows are epoch-contained,
convention~(\ref{S-conv:c4})). Dually, the attempts \emph{anchored} in a given
epoch are begun at that epoch's presence-creating submissions and
finalisation events (under the default bars; under loose bars also
its neither-bar resolutions), which is the counting
Corollary~\ref{S-cor:submission} consumes.
\end{definition}

\begin{commentary}[role of the attempt unit]
The attempt is the analysis unit at which the per-submission safety
bound is conditioned. Banked decisions credited to an entry precede
the entry's opening, and the owner --- or an adversary --- may decide
whether to open the entry \emph{as a function of} the banked outcomes;
a bound conditioned at the entry's opening would therefore condition
on an adversarially selected event (an optional-selection error). The
attempt's anchor precedes every decision creditable to its entries by
construction, so conditioning at the anchor is sound; and an attempt
whose intervals bank decisions is an attempt whether or not a further
entry opens, so the choice not to open is a choice among attempts
already counted, at success probability zero. No protocol rule reads
attempts; like the ledger's charging convention
(Lemma~\ref{S-lem:budget}), they are a hindsight bookkeeping device over
the same window/interval alternation the ledger's buckets partition.
\end{commentary}

\begin{lemma}[Type-level sharing exclusion]\label{S-lem:shareexcl}
No $\T$-block, no $\U$-block, no cache-marked block, and no block lacking an accepting publish
confirmation by its owner ever enters the shared candidate store.
\end{lemma}

\begin{proof}
Label and provenance membership are deterministic and attacker-immutable
(\ref{S-a:label}--\ref{S-a:persist}; provenance is interpreter-only, Definition~\ref{S-def:config}); the publish gate of
Rule~\ref{S-r:share} is a confirmation event under
Rule~\ref{S-r:display} and \ref{S-a:display}; nomination has no authority.
\end{proof}

\begin{lemma}[Entry disjointness]\label{S-lem:disjoint}
For a fixed byte string, the weight-bearing endorsement sets of
distinct pipeline entries are disjoint in identities over the
deployment's entire history, and every weight-bearing endorsement
(in-entry or banked-and-credited) is its endorser's fresh, authority-active inspection of
the bytes (Definition~\ref{S-def:counting}(a)).
\end{lemma}

\begin{proof}
By Definition~\ref{S-def:counting}(a), an identity's weight-bearing
endorsement of the bytes is its \emph{first} logged decision on them (Definition~\ref{S-def:fresh}), taken while authority-active (Definition~\ref{S-def:counting}(a)), (when credited, Definition~\ref{S-def:counting}(f)), counted toward exactly one entry (the entry
open at decision time, or, via banking,
Definition~\ref{S-def:counting}(b), the next entry to open). Every later
acceptance by the same identity is repeat and carries no weight
(\ref{S-inv:fresh}). Hence no identity contributes weight to two
entries, and each contribution is by construction a fresh inspection.
\end{proof}

\begin{lemma}[Deterministic accountability bound]\label{S-lem:cacc}
Fix an identity and an epoch. The identity's dispute-initiating
rejections filed within the epoch that both take protocol effect and
are eventually charged (contradicted by vindication), in this epoch
or any later one, number at most $\cacc:=\cceil+\rmax$.
\end{lemma}

\begin{proof}
By Rule~\ref{S-r:panel}(v)(b), a rejection is charged at the moment its
vindication is logged. Suppose first that the epoch's $\cceil$-th
charge to this identity occurs, at some moment $T$. From $T$ onward,
the identity's further dispute-initiating rejections in the epoch are
filed as advisory flags with no protocol effect
(Rule~\ref{S-r:panel}(v)(d)), hence are never counted. Every counted
rejection is therefore either charged no later than $T$ (and the
epoch's charges up to and including $T$ number exactly $\cceil$, by
the choice of $T$) or still open at $T$ and charged afterward; open
rejections number at most $\rmax$ at any moment
(Rule~\ref{S-r:panel}(i)). The two classes are disjoint, a rejection
being charged exactly once; summing gives $\cceil+\rmax$. If instead
fewer than $\cceil$ charges accrue within the epoch, at most
$\cceil-1$ counted rejections are charged within it, and every one
charged in a later epoch is open at the epoch boundary, at most
$\rmax$ again.
\end{proof}

\begin{remark}[Calibration of rejecter accountability]\label{S-n:calib}
The rate trigger of Rule~\ref{S-r:panel}(v)(c) is normalised \emph{per
fresh inspection}, never per rejection. The normalisation matters:
when poison is rare, almost all of an honest inspector's rejections are
wrongful rejections of benign items and almost all are vindicated, so
charges \emph{per rejection} approach one for honest and adversarial
identities alike and cannot separate them; charges \emph{per
inspection} sit near $1-p_b$ for an honest identity and near its full
rejection rate for a campaign. The ceiling $\cceil$ is calibrated,
like the rate trigger, strictly above the honest baseline: an honest
identity accrues charges at rate at most $1-p_b$ per benign fresh
inspection, and its fresh inspections per epoch are bounded by its own
required-confirmation stream (small, and shrinking with population,
Appendix~\ref{S-app:budget}), so $\cceil$ set at a high quantile of
$(1-p_b)\times$(honest per-epoch fresh-inspection count), with margin,
is reached by an honest identity only rarely; an honest identity that
does reach it retains advisory flags and ordinary
use --- though its acceptances are authority-inert for the duration
(Definition~\ref{S-def:counting}(a), Remark~\ref{S-rem:authority}) ---
loses nothing permanent, is restored at the epoch boundary, and
its inspections during that state are excluded trials, not misses
(\cref{S-n:qpos}), so no safety statement is affected. Renewal semantics: convention (\ref{S-conv:c4}) (the accountability allowances are the only freshness-like quantities that renew). (This calibration concerns the per-fresh-inspection rejection-rate trigger only; suspension-induced required confirmations are priced separately, once per demander, per Note~\ref{S-n:avail}.) Mis-calibration risk: \ref{S-q:calib}.
\end{remark}

\begin{lemma}[Verdict pacing under \ref{S-a:rhoseat}]\label{S-lem:geo}
Under \ref{S-a:rhoseat}, for any disputed item and any adversary
strategy (convention~(\ref{S-conv:c2})): the number of demanders who arrive while
the dispute is unresolved is stochastically dominated by the index
of the $(k{+}1)$-st success in an i.i.d.\
$\mathrm{Bernoulli}(\rseat)$ sequence, i.e.\ by a sum of $k{+}1$
independent $\mathrm{Geometric}(\rseat)$ variables. In particular
its expectation is at most $\kappa:=(k{+}1)/\rseat$, and for any
$c_0>1$ the probability that more than $c_0\,\kappa$ demanders
arrive while the dispute is unresolved is at most
$e^{-\lambda_{c_0}k}$ for a constant $\lambda_{c_0}>0$ depending
only on $c_0$ and $\rseat$.
\end{lemma}

\begin{proof}
Index trials by arrivals: let $O_i$ be the event that an $i$-th demander
of the item arrives while the dispute is unresolved, measurable at that
arrival's rendering atom, and let $Z_i$ indicate that this arrival's
disposition is an eligible verdict on the dispute instance. Under the
atomic reading of \S\ref{S-app:prob}, each arrival's disposition is realized
at its own log position, so the $\sigma$-algebra of atoms prior to the
$i$-th rendering atom contains the dispositions and protocol effects of
all earlier arrivals; \ref{S-a:rhoseat} gives
$\Prob[Z_i=1\mid\mathcal{G}_i]\ge\rseat$ on $O_i$ at exactly these
$\sigma$-algebras, and Lemma~\ref{S-lem:adaptive}(ii), in the form of
Remark~\ref{S-rem:adaptivegen}, supplies i.i.d.\
$\mathrm{Bernoulli}(\rseat)$ variables $Y_i$ with $Y_i\le Z_i$ on $O_i$.
A dispute instance accepts at most $k$ eligible verdicts in its entire
life (Rule~\ref{S-r:panel}(iii)(e)), and every verdict is cast at an
arrival; hence on $O_n$, $\sum_{i\le n-1}Z_i\le k$, so
$\sum_{i\le n-1}Y_i\le k$, i.e.\ $n\le G:=\min\{j:\sum_{i\le j}Y_i=k+1\}$.
The arrival count is therefore pathwise at most $G$, a sum of $k{+}1$
independent $\mathrm{Geometric}(\rseat)$ variables. Linearity gives the
expectation; the tail is the Chernoff bound for sums of independent
geometrics. (Because dispositions resolve at arrival, the sharper count
$\sum_{i\le n-1}Z_i\le k-1$ in fact holds while the dispute is unresolved,
improving $\kappa$ to $k/\rseat$; we retain the conservative
$\kappa=(k{+}1)/\rseat$, which every downstream constant consumes, the
slack being monotone.)
\end{proof}

\begin{lemma}[Decisive resolution under the default bars]\label{S-lem:decisive}
Under the default bars $m_{\mathrm{rej}}=k+1-m$ with $m>\fs$ and
$m_{\mathrm{rej}}>\fs$:
(i) eviction and vindication are mutually exclusive;
(ii) every panel that gathers $k$ verdicts reaches exactly one bar,
and the first bar reached in log order is the bar reached by the full
tally of $k$ verdicts: early resolution changes timing, never the
outcome;
(iii) both bars are reachable by honest verdicts alone, and the
adversary's at most $\fs$ verdicts (\ref{S-inv:verdict}) alone reach
neither; hence no \emph{tally} deadlock exists once $k$ verdicts are
gathered, and neither bar can be reached by the adversary's verdicts
alone; whether $k$ eligible verdicts arrive is the separate,
availability-side matter of \ref{S-a:rhoseat} (Remark~\ref{S-rem:resolution}).
\end{lemma}

\begin{proof}
(i)--(ii) $m+m_{\mathrm{rej}}=k+1$, so $k$ verdicts cannot meet both
bars and must meet at least one; verdict counts are non-decreasing
along the verdict sequence, so a bar is reached at some prefix iff it
is reached by the full tally. (iii) From $k=m+m_{\mathrm{rej}}-1$ with
$m,m_{\mathrm{rej}}>\fs$: $k-\fs\ge m$ and $k-\fs\ge m_{\mathrm{rej}}$,
so honest verdicts alone suffice for either bar; and $\fs<m$ means the
adversary's at most $\fs$ verdicts alone cannot reach the $m$-bar, and
symmetrically $\fs<m_{\mathrm{rej}}$ for the other bar.
\end{proof}

\begin{lemma}[Third-party immutability of local stores]\label{S-lem:nodelete}
For every user $u$ and every block $b$ in $u$'s local store (the
user-owned tier: $\T$-, $\C$-, $\PC$-, and $\U$-blocks; community status lives
in the shared tier and is referenced, never stored, locally,
\S\ref{S-app:fedcore}), no protocol event initiated by any identity
other than $u$ alters $\mathrm{bytes}(b)$, $\ell(b)$, or $\scopef(b)$.
Dispute, suspension, vindication, demotion, eviction, expiry, and
every accountability penalty of Rule~\ref{S-r:panel}(v),(vii) act on the
shared tier alone: they invalidate or flag shared entries, restrict
shared-tier rights, or reduce attestation weight. Demotion of a $\CC$
item withdraws community authority from users who have not personally
confirmed it, whose gates simply fire again, within the single-user cap of
Theorem~\ref{S-thm:floor}, and leaves every personally confirmed $\C$
copy intact (with a notification). Suspension likewise withdraws community authority for the review's duration (Rule~\ref{S-r:panel}(vi)), equally without touching any local block. Consequently no third-party protocol transition ever directly alters or
removes any block of any user's local store; first-party removal by the
owner's own agent ($\mathsf{evict}$, including evictions the agent
performs in reaction to notifications or other legitimate log events) is
the disclosed residual \ref{S-n:evict}. The induced cost bound, that no
campaign pushes any user above the single-user cap, is
Theorem~\ref{S-thm:floor}, which rests on the present lemma.
\end{lemma}

\begin{proof}
By Figures~\ref{S-fig:sos}--\ref{S-fig:sos2}, local stores mutate only via owner-invoked
$\Sigma$-operations and owner confirmations
(\ref{S-a:label}--\ref{S-a:persist}); $\mathsf{evict}$ is owner-only
(first-party eviction by the owner's own agent is a disclosed
residual: \ref{S-n:evict}, \S\ref{S-app:residuals}). Community status is
shared-tier state (\S\ref{S-app:fedcore}), so demotion and suspension
update the shared entry and never touch a local block; promotion, demotion,
and expiry are interpreter relabels or retirements of the shared log
entry (Rules~\ref{S-r:panel},~\ref{S-r:hygiene}); the accountability
penalties of clauses (v) and (vii) act on rejection rights, submission
rights, and attestation weight, none of which is a local-store
attribute; the cascade quarantines shared derivations, not local
stores; notifications are advisory.
\end{proof}

\begin{lemma}[Hygiene preserves safety through bound uniformity]
\label{S-lem:hygienemono}
The policies of Rule~\ref{S-r:hygiene} preserve every theorem of this
document. The entry cap only refuses entry openings, and retirement and demotion hygiene only remove shared-tier
items; none of these can enable a promotion. Surfacing \emph{reordering} is
not monotone by event-set inclusion --- reordering, and removal of one
candidate, can increase another item's reach within the
$\ccand$-capped surface --- but the safety bounds are unaffected
because Theorems~\ref{S-thm:det} and~\ref{S-thm:panel} are uniform over
adaptive presentation schedules and reach-free (\ref{S-inv:reach}): any
demand redistribution hygiene induces is one more presentation
schedule inside the quantified class. Lemma~\ref{S-lem:nodelete} is
unaffected because no local store is touched; the ledger of
Appendix~\ref{S-app:budget} is unaffected at the moment of any hygiene
action, which adds no confirmation event. Retirement does redirect
\emph{later} demand: a user who encounters the retired bytes
afterwards through their own channels confirms them as ordinary
personal content, in the single-user stream
(Definition~\ref{S-def:demand}), bounded by the cap
(Theorem~\ref{S-thm:floor}), its benign,
\ref{S-a:promsuf}-conforming component vanishing in expectation under
the vanishing conditions of Theorem~\ref{S-thm:sat}(iii) and the
remainder entering $r^{(e)}_{\mathrm{bad}}$
(Corollary~\ref{S-cor:total}); the shared-tier ledger rows
are never increased by removal or reordering.
\end{lemma}

\begin{proof}
Each policy is a deterministic interpreter action on the shared tier:
the entry cap refuses an entry opening, retirement removes a candidate,
surfacing order permutes at most
$\ccand$ surfaced candidates, demotion hygiene removes an evicted item
from retrieval. None relabels any local block, forces any
confirmation, or adds any promotion path. For the safety theorems, the
relevant quantification is over adaptive presentation and demand
schedules (convention~(\ref{S-conv:c2})); the schedules realizable under any
hygiene policy are a subset of that class, and the bounds of
Theorems~\ref{S-thm:det} and~\ref{S-thm:panel} are uniform over it and
independent of reach, so they apply verbatim to the hygiene-modified
deployment. The post-retirement reclassification is
Definition~\ref{S-def:demand}'s retirement clause.
\end{proof}
\subsection{Main theorems}
\label{S-app:fedthms}

\begin{proposition}[Delegated Action Integrity and federated Reasoning
Integrity]\label{S-prop:fed}
Assume \ref{S-a:transfer} and \ref{S-a:fedlog}, and read the interpreter
over federated product configurations
(Definition~\ref{S-def:fedconfig}) under the federated step of Definition~\ref{S-def:fedstep}, with
Definition~\ref{S-def:wf}(iii) in its federated form (every virtual
$\CC$ authority in force backed by a currently valid promotion
certificate on $L$). Then:
\begin{enumerate}[label=(\roman*),leftmargin=2em,itemsep=1pt,topsep=1pt]
\item a value flows on community authority without personal
confirmation only if it byte-equals a live virtual entry naming the
exact tool (Rule~\ref{S-r:whitelist}(a), live-entry test), each such
entry backed at that tool by qualifying fresh endorsements
(Definition~\ref{S-def:counting}, Lemma~\ref{S-lem:endorse}), each
satisfying \ref{S-a:display}--\ref{S-a:user} or \ref{S-a:narr} for its
endorser and authoritative by \ref{S-a:transfer}: PerMIT has
\emph{delegated} Action Integrity (Definition~\ref{S-def:action});
\item under \ref{S-a:fedlog} the coupled log slots deliver equal
prefixes, and locally emitted federation events are equal across
matched runs (Definition~\ref{S-def:fedstep}), hence equal virtual
views $\mathsf{LiveCC}(L)$ at every step; the matched relation of
Definition~\ref{S-def:matched}, read over $\Mem^{\mathrm{fed}}_{\E}$
and extended by equality of delivered log prefixes, is preserved by
log-slot transitions --- which emit no reply and alter no matched
component except the coupled virtual view --- and by every local
transition exactly as in Theorem~\ref{S-thm:reason}; hence
Theorem~\ref{S-thm:reason} and Corollary~\ref{S-cor:noexfil} lift to the
product system with ``$\C$'' read as ``$\C$ or live $\CC$'';
\item suspension, demotion, retirement, and every accountability event
act on $L$ alone: they withdraw virtual authority and never create,
alter, or remove any local block (Lemma~\ref{S-lem:nodelete}).
\end{enumerate}
\end{proposition}

\begin{proof}
(i) The single-user results are personal
(Lemma~\ref{S-lem:nocc}); community authority enters only through the
live-entry test, whose backing endorsements are fresh accepting
confirmations at the named tool (Definition~\ref{S-def:counting}(c),
Lemma~\ref{S-lem:endorse}), authoritative by \ref{S-a:transfer}; the scope
test is the same singleton-membership test as for $\C$, and handle
gating is label-independent (Rule~\ref{S-r:whitelist}(c)). (ii) Log-slot transitions: both runs append the same coupled prefix
(\ref{S-a:fedlog}), producing equal $S_{\mathrm{sh}}(L)$ by
deterministic replay (Definition~\ref{S-def:fedconfig}), appending
nothing to either reply stream and touching no local store
(Definition~\ref{S-def:fedstep}, Lemma~\ref{S-lem:nodelete}). Locally
emitted federation events extend the two logs equally, by
cases. (1)~$\mathsf{submit}(v)$: the publish decision is a
transcript event (matched) on bytes of a local $\C$-block of the
owner ($\Mem_{\E}$ matched). (2)~Protocol-bearing gate decisions:
displayed bytes, target, and decision are transcript-pinned;
shared-presence status, coverage, cache classification, freshness,
and authority are deterministic predicates of the matched local
store and the equal delivered prefix
(Definitions~\ref{S-def:fresh},~\ref{S-def:counting},
Rule~\ref{S-r:panel}(i),(v)). (3)~Realized demand events: fired by
the matched operation against the matched federated view, the
acting identity common to both runs, and epoch-first-demand status
a deterministic predicate of the equal delivered prefixes
(Definition~\ref{S-def:demand}). By induction on steps: equal
delivered prefixes give equal
$\mathsf{LiveCC}(L)$; together with the matched-local-state
induction hypothesis and the single-user unwinding
(Theorem~\ref{S-thm:reason}), this gives matched local transitions,
and the case analysis then gives equal appended suffixes, hence
equal delivered prefixes at the next step. Local transitions: the unwinding of
Theorem~\ref{S-thm:reason} runs verbatim with $\Mem^{\mathrm{fed}}_{\E}$
for $\Mem_{\E}$ --- the $\CC$ branch of the gate is a function of the
argument list and the step's delivered coupled log prefix
(Definition~\ref{S-def:fedstep}), matched across the runs, so the
gate-or-no-gate decision of Step~\hyperref[S-proof:reason-step-5]{5} remains matched. (iii) is
Definition~\ref{S-def:fedconfig} with Lemma~\ref{S-lem:nodelete}.
\end{proof}

\begin{theorem}[Deterministic safety under perfect honest detection]
\label{S-thm:det}
Under \ref{S-a:label}--\ref{S-a:narr} and
Rules~\ref{S-r:share}--\ref{S-r:panel}, with perfect honest detection
($q=1$: every honest inspection of the poisoned item rejects): if
$W>\Wadv$ and $m>\fs$, then no poisoned item (Definition~\ref{S-def:bad}) is ever promoted to $\CC$
at any tool, at any population size and under any attack, through
either the ordinary or the panel path.
\end{theorem}

\begin{proof}
With $q=1$, every accepting endorsement and every accepting panel
verdict on a poisoned item is adversarial. The ordinary path requires
accepting weight at least $W$ at the tool in question, while the
adversary's weight is at most $\Wadv$ and $W>\Wadv$: impossible. The
panel path requires at least $m$ accepting verdicts, while the
adversary casts at most $\fs$ verdicts
(\ref{S-inv:verdict}), $m>\fs$, and honest accepting verdicts are
zero: impossible. Vindication-credited and dormancy-banked endorsements
are covered identically: with $q=1$ every honest decision on the poison
rejects, so every credited acceptance is adversarial and already
counted inside $\Wadv$ (respectively inside the $\fs$ verdict cap).
Deterministic and population-free.
\end{proof}

\begin{remark}[Scope of Theorem~\ref{S-thm:det}]\label{S-rem:detscope}
The hypothesis $q=1$ is a behavioural idealization, the strongest in
this document, not the absence of one: the Sybil conditions
$W>\Wadv$, $m>\fs$ exclude \emph{adversarial} weight, while honest
acceptances of the poison are excluded only by $q=1$ itself. The
theorem is therefore \emph{not} a fallback for the
unrecognizable-poison regime $q\to0$, where honest inspectors
themselves supply the threshold and no community-truthfulness
guarantee is available within a human-judgment design. What survives
that regime is the local layer --- personal Action Integrity, the
single-user cap, and no-deletion (Theorem~\ref{S-thm:action},
Theorem~\ref{S-thm:floor}, Lemma~\ref{S-lem:nodelete}) --- together with
the one-sided detection clause (a single honest \emph{effective
qualifying rejection}, whenever
one occurs, freezes promotion immediately, Rule~\ref{S-r:panel}(i),(v)) and
re-verification against never-decided inspectors (\ref{S-n:revoke}).
See \ref{S-q:qzero}.
\end{remark}

\begin{theorem}[Panel safety under adaptive presentation]\label{S-thm:panel}
Adopt Rules~\ref{S-r:share}--\ref{S-r:panel} with the default bars of
Rule~\ref{S-r:panel}(iv) ($m_{\mathrm{rej}}=k{+}1-m$, $m>\fs$,
$m_{\mathrm{rej}}>\fs$; Lemma~\ref{S-lem:decisive}) and the probabilistic
framework of \S\ref{S-app:prob}. Suppose:
\begin{enumerate}[label=(\alph*),leftmargin=1.8em,itemsep=1pt,topsep=1pt]
\item ordinary promotion at a tool requires $N$ distinct accepting
endorsements counted toward one pipeline entry
(Definition~\ref{S-def:counting}), with $N>f$;
\item at most $\fs$ of the adversary's identities are seat-eligible,
with $m>\fs$;
\item there are $\qpre,\qpan\in(0,1]$ such that: each honest
\emph{fresh, authority-active} (Definition~\ref{S-def:counting}(a))
inspection of the poisoned item at a non-covered, \emph{non-cache}
confirmation gate
ends in an \emph{effective qualifying rejection} --- a recorded
rejecting decision that takes protocol effect under
Rule~\ref{S-r:panel}(i) --- with conditional probability at least
$\qpre$; and each honest \emph{eligible verdict} on the item rejects
with conditional probability at least $\qpan$, \emph{conditionally on
the verdict being cast}, i.e.\ at the verdict's header-inclusive
$\sigma$-algebra (\S\ref{S-app:prob}) (Rule~\ref{S-r:panel}(iii)). Both
bounds are read per the conditioning convention of \S\ref{S-app:prob}
(all prior logged atoms; the demand realisation), and both are
\emph{uniform lower bounds over the honest identities and
presentation contexts the adversary can target} --- the catch rate of
the most fallible targetable honest cohort, not a population average
(\cref{S-n:qpos}) --- $\qpre$'s contexts are the \emph{unfrozen} candidate contexts ---
in-window and candidate-dormancy inspections, any advisory-flag state
--- where an effective qualifying rejection is possible
(Rule~\ref{S-r:panel}(i),(ii)); freeze- and suspension-time fresh
inspections carry no initiation and belong to $\qpan$'s province
where verdict-eligible, their non-verdict rejections being
corroborations, consumed by no trial of this theorem. $\qpan$'s
contexts include the freeze itself: verdicts are cast while the item
is frozen. Both bounds are read over the target contexts (tools and
$\mathsf{reasoning}$) in which the trials actually occur: $\qpan$ is a
uniform bound over the tool contexts of \emph{eligible jurors}, which
under byte-level panels may include tools at which the item is
innocuous --- for tool-targeted poison the applicable $\qpan$ may
therefore be small (\ref{S-q:tooltarget}). Authority-inert inspections,
covered gates, and cache gates (Options~\ref{S-o:one}--\ref{S-o:two})
belong to no trial class of this theorem: an
authority-inert acceptance carries no promotion weight and an
authority-inert rejection no initiation
(Definition~\ref{S-def:counting}(a)), covered gates are
protocol-inert on both branches (Rule~\ref{S-r:panel}(i),
\cref{S-n:qpos}), and cache gates are
accept-inert and verdict-ineligible
(Definition~\ref{S-def:counting}(iii), Rule~\ref{S-r:panel}(iii)) ---
their rejecting branch is one more effective qualifying rejection,
which can only deny the case-(a) path --- so promotion still requires
$N-f$ honest authority-active fresh \emph{non-cache} credits, and $\qpre$ is quantified over exactly
the class whose members can both accept with weight and reject with
effect. Abstention by an authority-active fresh inspector records
nothing and is counted as a miss inside $1-\qpre$; the two parameters
deliberately live at different conditioning levels --- $\qpre$ at the
disposition of a fresh inspection, where abstention must count as a
miss; $\qpan$ at a cast verdict, where abstention is already excluded
by eligibility --- and $\qpre\le\qpan$ is typical but not assumed;
\item the panel margin is positive:
$\delta:=\frac{m-\fs}{k-\fs}-(1-\qpan)>0$.
\end{enumerate}
Then for any poisoned (Definition~\ref{S-def:bad}) submission, every
promotion attempt (Definition~\ref{S-def:attempt}) succeeds with
probability at most the display below, \emph{conditionally on the $\sigma$-algebra generated by all atoms up to and including the attempt's anchor atom}, uniformly over adaptive
presentation, stopping, and re-entry strategies and regardless of how
many users the item is shown to (\ref{S-inv:reach}),
\begin{align*}
\Prob[\text{promotion}]
&\le (1-\qpre)^{N-f}+\Prob\!\big[\mathrm{Bin}(k-\fs,1-\qpan)\ge m-\fs\big]\\
&\le (1-\qpre)^{N-f}+e^{-2\delta^{2}(k-\fs)} .
\end{align*}

\end{theorem}

\begin{remark}[Scope of Theorem~\ref{S-thm:panel}]\label{S-rem:panelscope}
Theorem~\ref{S-thm:det} is the special case $q=1$. The bound covers a
single promotion attempt (Definition~\ref{S-def:attempt}); for attempts anchored while the bytes hold live $\CC$ status at some tool (additional-tool promotion of already-promoted bytes, possible only after a prior success), hypothesis~(c) is read with $\qpre$'s contexts extended per the closing paragraph of the proof; first-promotion attempts need no extension; entries embed injectively into attempts, so per-entry counting statements read a fortiori; attempts are counted at their anchors, and the per-anchor-epoch union
bound taken, in Corollary~\ref{S-cor:submission}. No hypothesis is
placed on repeat inspections: none carries promotion weight or a
verdict (\ref{S-inv:fresh}, \ref{S-inv:verdict};
Remark~\ref{S-n:freshness}). The panel arithmetic rests on
\ref{S-inv:verdict} alone, which is stated independently of the
freshness rule precisely so that it never depends on
\ref{S-inv:fresh}.
The default bars are a hypothesis of the theorem: case~(b) uses decisive
resolution (Lemma~\ref{S-lem:decisive}) to cap honest verdicts at $k-a$.
Under the looser bars permitted by Rule~\ref{S-r:panel}(iv)
($m+m_{\mathrm{rej}}\ge k+2$), the same argument bounds the vindication
event verbatim --- vindication still requires $m$ accepting verdicts among
at most $k$, so the case-(b) term stands with $k-\fs$ trials --- and the
additional neither-bar outcome returns the item to unpromoted candidate
status, whose re-entry opens a new attempt with an identity-disjoint
credited set, counted separately and marginally by
Corollary~\ref{S-cor:submission}. The bound therefore extends to loose-bar
deployments per entry; only the decisiveness clauses (mutual exclusivity,
tally-deadlock-freeness) are specific to the defaults.
\end{remark}

\begin{remark}[Single-parameter reading]\label{S-rem:qsingle}
Every appearance of a single $q$ elsewhere in this document
(Theorem~\ref{S-thm:det}'s hypothesis $q=1$; the corridor row
$q+p_b>1$; Corollary~\ref{S-cor:submission}; the composed summary of
\S\ref{S-app:composed}; the correlated rate $q_c$ of \ref{S-q:corr}) is
read as $q:=\min(\qpre,\qpan)$: the bounds of
Theorem~\ref{S-thm:panel} are monotone in each parameter, so the
single-$q$ forms are the special case $\qpre=\qpan=q$ and remain
valid verbatim. $q=1$ means both acts are perfect;
$q\to0$ (\ref{S-q:qzero}) is the regime where the smaller of the two
vanishes.
\end{remark}

\begin{remark}[Itemwise reading; heterogeneous item difficulty]
\label{S-rem:qitem}
Theorem~\ref{S-thm:panel} is stated and proved per promotion attempt (Definition~\ref{S-def:attempt}): nothing
in its trials couples distinct items, so its hypotheses and bound are
\emph{itemwise} --- for each poisoned item $v$, the bound holds with
$v$'s own weakest-target rates $\qpre(v),\qpan(v)$, and it is
monotone in each. Heterogeneous item difficulty is therefore inside
the model, transparently: easily recognized poison contributes a
negligible term; near-unrecognizable poison ($\qpre(v)\to0$) degrades
its own term toward the disclosed residual \ref{S-q:qzero}, leaving
every other item's bound untouched. The per-anchor-epoch union bound of
Corollary~\ref{S-cor:submission} holds in the same itemwise form,
$\sum_{e}\big[(1-\qpre(v_e))^{N-f}+e^{-2\delta(v_e)^{2}(k-\fs)}\big]$
(terms with $\delta(v_e)\le0$ read as the exact binomial tail), the
single-$q$ display being the uniform special case; a deployment-level
constant is thus dominated by the hardest items the adversary can
mint --- exactly the calibration duty of \ref{S-q:paraphrase}. No
parametric ability-or-difficulty model (jury, psychometric, or
crowdsourcing) is assumed: the weakest-target conditional floor
subsumes any such structure, at the price that the floor, not the
average, is the deployed constant (\cref{S-n:qpos}, \ref{S-n:validate}).
\end{remark}

\begin{remark}[The certified per-attempt ceiling $B_{\mathrm{att}}$]
\label{S-rem:batt}
Write
$$
B_{\mathrm{att}}\;:=\;(1-\qpre)^{N-f}
\;+\;\Prob\!\big[\mathrm{Bin}(k-\fs,\,1-\qpan)\ge m-\fs\big]
$$
for the right-hand side of Theorem~\ref{S-thm:panel}'s bound in its exact
binomial form, and call its two summands the \emph{ordinary term} and the
\emph{panel term}. The word \emph{certified} names this analytic bound and
is unrelated to the promotion certificates of
Definition~\ref{S-def:fedconfig}, which are log objects. Three properties are
used throughout the empirical evaluation: the first two are immediate from
the display, the third from the theorem's statement.

The panel term is a \emph{floor}: at fixed panel geometry $(k,m,\fs)$ and
fixed $\qpan$ it does not contain $N$, so raising the endorsement quorum
shrinks only the ordinary term and $B_{\mathrm{att}}$ can never fall below
it --- panel geometry, not quorum, is the binding deployment parameter,
which is the frontier Proposition~\ref{S-prop:live} reads against and which
the corridor of \S\ref{S-app:corridor} moves only by scheduling $k$ itself.
The two terms fall in \emph{different} protocol constants and must be
pushed down independently: the ordinary term geometrically in $N-f$, the
panel term as a binomial tail in $k-\fs$ whose Hoeffding relaxation
$e^{-2\delta^{2}(k-\fs)}$ is exponential in that constant. And
$B_{\mathrm{att}}$ is per promotion attempt (Definition~\ref{S-def:attempt})
and reach-free (\ref{S-inv:reach}); deployment-wide exposure is obtained by
summing over anchored attempts and byte-distinct variants, through the
union, expectation and deterministic-cap forms of
Corollary~\ref{S-cor:submission} and the lifetime ceiling of
Remark~\ref{S-rem:lifetime}, never by reading the per-attempt figure as a
per-epoch one.

\emph{Reporting discipline.} Any numerical value quoted for
$B_{\mathrm{att}}$ is the exact binomial tail above, never the Hoeffding
relaxation that Theorem~\ref{S-thm:panel} displays alongside it. The two
differ materially, and by different factors on the two sides of the
design: at $k=31$, $m=20$, $\fs=1$ with $\qpan=0.6$ the relaxation is
$3.81\times10^{-2}$ against an exact panel term of $8.30\times10^{-3}$, a
factor of about $4.6$; on the benign side of
Proposition~\ref{S-prop:live}, at $p_b=0.9$ and the same $\delta$, the
numerically identical relaxation stands against an exact
wrongful-eviction tail of $8.91\times10^{-5}$, a factor of about $430$. A
reader recomputing either relaxation from a quoted exact value will
therefore not reproduce it, and --- because the two relaxations coincide
numerically while their exact counterparts differ by two orders of
magnitude --- the two slacks must never be interchanged.
\end{remark}

\begin{proof}[Proof of Theorem~\ref{S-thm:panel}]
Fix the attempt and condition on the $\sigma$-algebra through and
including its anchor atom (Definition~\ref{S-def:attempt}). The
behavioural bounds of hypothesis~(c) are asserted uniformly over
histories, so the argument restarts after any fixed anchor-inclusive
history; every trial charged below occurs strictly after the anchor,
and each trial's own conditioning $\sigma$-algebra
(\S\ref{S-app:prob}) refines the anchor-inclusive one. Partition by whether an effective
qualifying rejection was logged after the anchor and before the
attempt's finalisation.

Throughout, the bytes' promotion status at every tool is constant
within the attempt: promotion occurs only at a finalisation, which
ends the attempt, and demotion by eviction ends the presence
(Definition~\ref{S-def:attempt}, Definition~\ref{S-def:counting}(c)),
so whether the bytes hold live $\CC$ status at any tool is decided
at the anchor. Suppose first that they do not --- the case of every
attempt preceding the bytes' first promotion, hence of every
attempt the ever-promoted events of Corollary~\ref{S-cor:submission}
require; the complementary case is scoped at the close of the
proof. Then no suspension can arise within the attempt (suspension
presupposes promoted bytes, Rule~\ref{S-r:panel}(vi)), every
qualifying rejection on the bytes within it is dispute-initiating
(Rule~\ref{S-r:panel}(i),(ii)), and every non-freeze piece of the
attempt is an entry window or a candidate-dormancy interval
(presence taxonomy, Lemma~\ref{S-lem:budget}): every unfrozen header
within the attempt is an \emph{unfrozen candidate context} in the
sense of hypothesis~(c).

\emph{(a) Promotion with no qualifying rejection.} By
Rule~\ref{S-r:panel}(i),(v), an honest \emph{authority-active}
identity's rejecting gate decision on an unfrozen candidate takes
protocol effect, so the case-(a) event entails that no honest
authority-active inspection of the item ended in an effective
qualifying rejection before the attempt's finalisation. Promotion at
the finalisation requires $N$ credited accepting endorsements at the
finalising entry (in-window, interval-banked, boundary-banked, or
vindication-credited, Definition~\ref{S-def:counting}), of which at
least $N-f$ come from honest identities; each is a distinct honest
identity's fresh, authority-active, non-covered decision on the bytes
(Lemma~\ref{S-lem:disjoint}, Definition~\ref{S-def:counting}(a)), and each
is logged strictly after the attempt's anchor and before its
finalisation (Definition~\ref{S-def:attempt}).

Index the trials by log order within the attempt: let $O_i$ be the
event that an $i$-th protocol-bearing decision at a non-covered,
non-cache gate on the item, by an honest, fresh, authority-active
identity, is logged after the attempt's anchor and before the
attempt ends, at an unfrozen header --- by the opening reduction,
an unfrozen candidate context in the sense of hypothesis~(c). $O_i$
is measurable at that record's header atom (\S\ref{S-app:prob}):
attempt membership is decided by the anchor, a past log event, and
freshness, coverage, authority, and frozen status are deterministic
predicates of the prior log
(Definitions~\ref{S-def:fresh},~\ref{S-def:counting}(a),
Rule~\ref{S-r:panel}(i),(ii)) --- no trial's occurrence depends on
the attempt's future. Because no event interleaves between a
position's rendering and header atoms (\S\ref{S-app:prob}), the
status at the header is the status under which hypothesis~(c)'s
disposition-level bound is read. Let $Z_i$ indicate that the
trial's content is an effective qualifying rejection.
Hypothesis~(c) lower-bounds by $\qpre$ the disposition-level
probability of such a rejection at the trial's rendering atom, on
all of $O_i$ and irrespective of whether case~(a) occurs --- the
class is confined to exactly the unfrozen candidate contexts over
which $\qpre$ is quantified; conditioning on a recorded decision
only raises the bound (Lemma~\ref{S-lem:dispbridge}(iii)). Hence
$\Prob[Z_i=1\mid\mathcal{G}_i]\ge\qpre$ on $O_i$ at the
header-inclusive $\sigma$-algebras of the conditioning convention
(\S\ref{S-app:prob}), each of which refines the anchor
$\sigma$-algebra, so Remark~\ref{S-rem:adaptivegen} applies under the
anchor-conditional measure directly.

Only now is the case-(a) event used. On it, no effective qualifying
rejection is logged after the anchor and before the attempt's
finalisation; since no dispute or review is open at the anchor
(Definition~\ref{S-def:attempt}) and every freeze arising within the
attempt presupposes such a rejection, every trial of the class
preceding the finalisation occurs at an unfrozen header, and the
$N-f$ honest credited endorsements --- in-window, interval-banked
in a candidate-dormancy interval, or boundary-banked, by the
opening reduction; a vindication-credited endorsement presupposes a
panel, hence a case-(b) run --- are all logged at trials of the
class, each necessarily non-cache (cache acceptances are never
logged, hence never credited, Definition~\ref{S-def:counting}(iii)).
The case-(a) success event is therefore contained in
$O_{N-f}\cap\{Z_1=\cdots=Z_{N-f}=0\}$: at least $N-f$ trials of the
class are logged before the finalisation, so the first $N-f$ trials
in log order all precede it, and each has $Z_i=0$, an effective
qualifying rejection before finalisation being a case-(b) run. The
joint all-zeros bound of Lemma~\ref{S-lem:adaptive}(ii), in the
$\sigma$-algebra form of Remark~\ref{S-rem:adaptivegen}, gives
$\Prob[\text{case (a)}\mid\text{anchor}]\le(1-\qpre)^{N-f}$.
Additional reach creates both rejection opportunities and possible
weight-bearing acceptances (\ref{S-inv:rejasym}); no pathwise
monotonicity in reach is claimed. What the bound uses is only the
$N-f$ honest fresh authority-active trials promotion itself requires,
so the stated bound is unchanged at every reach (\ref{S-inv:reach}).

\emph{(b) Promotion after a qualifying rejection.} The byte-level
freeze then blocks every tool's pipeline until an accepting panel resolution
(Rule~\ref{S-r:panel}(ii)--(iv),(vi)), so the case-(b) success
event is contained in the accepting outcome
(vindication, or reinstatement under Rule~\ref{S-r:panel}(vi)) of the
first panel instance opened
within the attempt --- an ordinary
dispute panel or a fast-track suspension review, with the same
$k,m,m_{\mathrm{rej}}$ parameters and bars; that panel instance opens, and casts all its verdicts,
after the attempt's anchor (Definition~\ref{S-def:attempt}), so the
argument below runs under the anchor-conditional measure verbatim: the
accepting-verdict count must reach $m$. Suppose the adversary casts
$a\le\fs$ accepting verdicts (at most one per identity for the entire
life of the panel instance (\ref{S-inv:verdict}, read over the instance; Rule~\ref{S-r:panel}(iii),(vi)), across any number of epoch boundaries,
by \ref{S-inv:verdict}; seat floor and pool caps per
Rule~\ref{S-r:panel}(iii); abandonment casts none). An accepting resolution then
requires $m-a$ honest accepting verdicts; with the default bars, at most $k$ verdicts are cast before a bar is
reached (Lemma~\ref{S-lem:decisive}), so at most $k-a$ are honest. Each
honest verdict is a recorded decision at a fresh inspection.
By the atomic decomposition of \S\ref{S-app:prob}, the event that an
$i$-th honest verdict is cast is measurable at that verdict's
header atom, strictly before its content atom, and honest verdicts
are log-ordered with every earlier verdict's content preceding
every later verdict's header. By hypothesis~(c), each honest verdict rejects with conditional
probability at least $\qpan$ at its header-inclusive $\sigma$-algebra (\S\ref{S-app:prob}); equivalently, its conditional accept probability there is at most
$1-\qpan$ (a cast verdict's content is binary). Apply
Lemma~\ref{S-lem:adaptive}(i), in the $\sigma$-algebra form of
Remark~\ref{S-rem:adaptivegen} at those $\sigma$-algebras, to the accept
indicators $Z_i$ of the honest verdicts: there are i.i.d.\
$\mathrm{Bernoulli}(1-\qpan)$ majorants
$Y_1,\dots,Y_k$ with $Z_i\le Y_i$ pointwise, unrealised verdicts
contributing zero, so the early decisive truncation of
Rule~\ref{S-r:panel}(iv) removes only nonnegative terms
(Lemma~\ref{S-lem:adaptive}(i)). On the case-(b) event,
$m-a\le\sum_{i\le k-a}Z_i\le\sum_{i\le k-a}Y_i
\le\sum_{i\le k-\fs}Y_i+(\fs-a)$, whence
$\sum_{i\le k-\fs}Y_i\ge m-\fs$: the case $a=\fs$ dominates,
uniformly in $a$, and $\sum_{i\le k-\fs}Y_i\sim\mathrm{Bin}(k-\fs,1-\qpan)$.
Hoeffding's inequality (Lemma~\ref{S-lem:engine}) at the margin $\delta$
gives the second term. The events are disjoint;
sum.

\emph{Reach-freeness.} The bound of case (a) is independent of reach: it counts only the
$N-f$ honest fresh decisions promotion itself requires, however many
further inspections occur; in case (b), $k$ and $m$ are protocol constants,
viewers whose decisions are not verdicts do not vote, and logged-repeat
viewers cannot vote at all (\ref{S-inv:fresh}). An expired dispute
retires the item unpromoted (Rule~\ref{S-r:hygiene}); republication is a
fresh submission whose endorsement set is identity-disjoint from all
prior sets (Lemma~\ref{S-lem:disjoint}), counted separately and
marginally (Corollary~\ref{S-cor:submission}). The case-(b) bound holds
regardless of who filed the rejection, so self-dispute gains nothing
(\ref{S-q:selfdispute}). Finally, the adversary's choice whether to open an entry after
favourable banked outcomes is a choice among attempts already counted:
an attempt whose intervals bank decisions is an attempt whether or
not the entry opens, and an unopened attempt succeeds with
probability zero (Definition~\ref{S-def:attempt}).

\emph{Attempts anchored with live $\CC$ status.} For an attempt at
whose anchor the bytes hold live $\CC$ status at some tool ---
possible only after a prior successful attempt of the same bytes,
itself bounded marginally at its own anchor --- the inter-window
spans are stable promoted intervals, first decisions banked there
are creditable (Definition~\ref{S-def:counting}(b),
Lemma~\ref{S-lem:budget}), and a fresh inspector's rejection there is
a post-promotion rejection routed to Rule~\ref{S-r:panel}(vi) rather
than a dispute initiation. The two-case argument applies to such
attempts under a correspondingly extended reading of
hypothesis~(c): $\qpre$ quantified over the attempt's unfrozen
contexts \emph{including} its stable promoted intervals, with the
effective qualifying rejection there being the review-opening
post-promotion rejection of Rule~\ref{S-r:panel}(vi) within its
per-epoch review cap --- an extension a deployment must calibrate
separately, disclosed here (cf.\ \ref{S-q:tooltarget}). The event
that any byte string is ever promoted \emph{at all} is a union of
first-promotion events, whose attempts are anchored $\CC$-free and
need no extension.
\end{proof}

\begin{remark}[Authority-gated crediting: soundness and cost]
\label{S-rem:authority}
The authority gate of Definition~\ref{S-def:counting}(a) restores the
quantifier compatibility that deterministic silencing would otherwise
break: without it, an adversary steering presentations toward
identities whose rejection rights are suspended, ceiling-capped, or
slot-exhausted would face inspectors whose effective rejection
probability is \emph{zero} while their acceptances still carried full
promotion weight, and no $\qpre>0$ could satisfy hypothesis~(c)
uniformly over targetable cohorts. With the gate, the two capacities
are coupled at one deterministic log predicate: a silenced identity
can neither reject with effect nor accept with weight, so targeting
it gains the adversary nothing. The split is exact: verdict
\emph{capacity} is never authority-gated
(Rule~\ref{S-r:panel}(v)(a),(e)) --- an authority-inactive juror's
accepting verdict counts toward the $m$-bar like any other, keeping
the panel arithmetic of Theorem~\ref{S-thm:panel} and
Proposition~\ref{S-prop:live} calibration-independent --- while
promotion \emph{crediting} always is
(Definition~\ref{S-def:counting}(a)): the same verdict contributes
weight only if its caster held authority at decision time, in-entry,
banked, and vindication-credited cases alike. Honest cost is negligible: an honest
identity is authority-inactive only in the rare calibrated states of
\cref{S-n:calib}, and its acceptance in that state still buys its
personal $\C$ copy --- \ref{S-inv:floor} is unaffected, the cap
counting required confirmations, never credits. Safety is monotone:
authority-gating removes acceptances from the promotion count and
removes nothing from the rejection side (rejections by inactive
identities remain advisory flags or corroboration exactly as before,
and verdict-eligible rejections are cast regardless of accountability
state, Rule~\ref{S-r:panel}(v)(e)); Theorem~\ref{S-thm:det} is unaffected
(its counting is adversarial-weight only), $\qpan$ needs no gating
(verdicts are never silenced, Rule~\ref{S-r:panel}(v)(a),(e)), and
verdict eligibility, hence \ref{S-a:rhoseat}, is untouched. One
availability-side consequence is disclosed: an adversary who
engineers charges against honest identities (\ref{S-q:calib}) now also
delays their \emph{crediting} capacity, slowing benign promotion ---
an exposure of the same character, availability-only, bounded by the
same calibration margins (\cref{S-n:calib}) and priced inside
$\varepsilon(M)$ where it matters (\ref{S-a:promsuf}, heuristic
sufficiency).
\end{remark}

\begin{remark}[Interpretation of the parameter $q$]\label{S-n:qpos}
$\qpre$ is the probability that an honest inspector, presented the
item at their own confirmation gate, detects the poison \emph{and
records a rejecting decision that takes protocol effect} (quantified over unfrozen candidate contexts, Theorem~\ref{S-thm:panel}(c));
$\qpan$ is the probability that an honest \emph{eligible juror's}
cast verdict rejects it. The split matters for calibration: the two
acts have different filters (seat weight, pool caps, and verdict
capacity apply only to jurors) and different honest baselines, so
their log validation is stratified separately (\ref{S-n:validate});
a deployment estimating one number reads it as
$q:=\min(\qpre,\qpan)$, under which every bound of
Theorem~\ref{S-thm:panel} holds with $q$ in both roles
(\cref{S-rem:qsingle}). Both parameters are \emph{uniform, weakest-target,
conditional} lower bounds (\S\ref{S-app:prob}: conditional on all prior logged atoms and the demand
factor): because the adversary chooses whom to present the item to,
each is the catch rate of the most fallible honest cohort the
adversary can reach, never a population average, and aggregate log
rates do not directly estimate them (\ref{S-n:validate}). Because a gate rejection is automatically
qualifying (Rule~\ref{S-r:panel}(i)), detection and reporting are a
single act: there is no separate filing step whose friction could
decouple them. Abstention (a user who suspects the item and abandons
the dialog)
records nothing and is counted as a miss inside $1-q$. An honest
identity whose rejection rights are suspended, whose $\rmax$ slots are
transiently full, whose per-epoch ceiling $\cceil$ is reached, or whose
single per-item initiation is spent is \emph{authority-inactive}: its
inspections are excluded trials, not silenced misses --- its
acceptance carries no promotion weight and its rejection no
initiation (Definition~\ref{S-def:counting}(a),
Remark~\ref{S-rem:authority}) --- and such states are rare for honest
identities (honest charges accrue at rate at most $1-p_b$ per benign
fresh inspection, both the rate trigger and the ceiling calibrated
strictly above that baseline, \cref{S-n:calib}; slots free on
resolution or expiry, Lemma~\ref{S-lem:freeze}; the ceiling and the
per-item allowance renew each epoch).
Covered gates (Rule~\ref{S-r:panel}(i)) are not silenced misses but
\emph{excluded trials}: both branches are protocol-inert, so they
belong to no trial class of Theorem~\ref{S-thm:panel}, and the
identity's inspection of the bytes, where one is counted at all, is
the earlier decision that created its coverage; so, symmetrically, are authority-inactive inspections (Definition~\ref{S-def:counting}(a)), and so are cache gates on the
accepting side (accept-inert, Definition~\ref{S-def:counting}(iii)),
whose rejections take effect as dispute initiations or corroborations
but never as verdicts (Rule~\ref{S-r:panel}(iii)).
Freshness being shared-presence-scoped (Definition~\ref{S-def:fresh}),
the fresh class can include identities with a prior \emph{unlogged}
personal decision on the bytes (before first publication, or across
a retirement gap); $\qpre$ and $\qpan$ are lower bounds over this
log-defined class, so a previously fooled but log-fresh inspector is
inside the class the parameters must cover --- one more reason the
bounds are weakest-target infima, never averages, and one the
validation strata of \ref{S-n:validate} must include (prior-exposure
stratification).
Case (a) of
Theorem~\ref{S-thm:panel} needs only $\qpre>0$: one honest \emph{effective qualifying rejection}
--- a rejecting decision at a protocol-bearing gate by an identity
currently holding dispute-initiation authority (Rule~\ref{S-r:panel}(i),(v)) --- denies the ordinary path, a one-sided
disjunction, not a vote; coverage-silenced and authority-inactive detections are excluded trials rather than misses (Definition~\ref{S-def:counting}(a)), the corresponding acceptances carrying no weight; post-promotion rejections beyond the review cap of Rule~\ref{S-r:panel}(vi) concern the suspended state only and touch no case-(a) trial. The panel is a bounded jury: joint safety and
a low wrongful-eviction bound need the better-than-chance condition $\qpan+p_b>1$; safety alone
can be forced for any $\qpan>0$ by raising $m$ toward $k$, at the price of
a weaker wrongful-eviction bound. The regime $q\to0$ is the disclosed limitation
\ref{S-q:qzero}. The exponents $N-f$ and $k-\fs$ are protocol constants;
the bases $1-\qpre$, $1-\qpan$ are behavioural parameters of the targeted population.
Because panel verdicts are never penalised and first charges are free
each epoch (Rule~\ref{S-r:panel}(v)), the protocol attaches no penalty to
a first good-faith rejection, so it does not disincentivise the gate
behaviour that generates $\qpre$. Finally, $\qpre$ and $\rseat$ condition on
the same human act consistently: an abstaining demander is a miss
inside $1-\qpre$ \emph{and} a non-verdict inside $\rseat$; high abstention
degrades resolution speed, never the validity of the safety parameter.
\end{remark}
\begin{proposition}[Wrongful-eviction bound]\label{S-prop:live}
Adopt the default bars, and suppose each honest juror's fresh
inspection of a benign item (Definition~\ref{S-def:bad}) ends in an accepting decision with
conditional probability at least $p_b$ given all prior protocol
events, read per the conditioning
convention of \S\ref{S-app:prob}. If $\delta_b:=p_b-\frac{m}{k-\fs}>0$ (the
\emph{wrongful-eviction margin}), then, uniformly over
adversary strategies,
\begin{equation*}
\Prob[\text{wrongful eviction}]
\ \le\ e^{-2\delta_b^{2}(k-\fs)} .
\end{equation*}
Moreover both margins ($\delta>0$ and $\delta_b>0$) are simultaneously
achievable by an integer $m$ whenever $(k-\fs)(q+p_b-1)>\fs+1$; in
particular, whenever inspectors discriminate better than chance
($q+p_b>1$) and $k$ is large enough.

\end{proposition}

\begin{proof}
Under the default bars, a resolved panel's outcome is eviction or
vindication, exclusively (Lemma~\ref{S-lem:decisive}(i)--(ii)), and
eviction requires $m_{\mathrm{rej}}=k+1-m$ rejecting verdicts. Suppose
the adversary casts $a\le\fs$ verdicts (at most one per identity for
the dispute's entire life, \ref{S-inv:verdict}); at most $a$ of them
reject, so eviction requires at least $m_{\mathrm{rej}}-a$ honest
rejecting verdicts, while honest verdicts number at most $k-a$. For an
honest verdict, writing $p_{\mathrm{acc}},p_{\mathrm{rej}}$
for its conditional accept/reject disposition probabilities at its
rendering atom, with $p_{\mathrm{acc}}\ge p_b$ and
$p_{\mathrm{acc}}+p_{\mathrm{rej}}\le1$ (\S\ref{S-app:prob}),
Lemma~\ref{S-lem:dispbridge}(ii) bounds the conditional rejection
probability at the verdict's header-inclusive $\sigma$-algebra by $1-p_b$; these $\sigma$-algebras satisfy the occurrence and measurability conditions of
Remark~\ref{S-rem:adaptivegen} exactly as in
Theorem~\ref{S-thm:panel}(b). Lemma~\ref{S-lem:adaptive}(i), in that
form, applied to the rejection indicators $Z_i$, gives
i.i.d.\ $\mathrm{Bernoulli}(1-p_b)$ majorants $Y_i$ with
$Z_i\le Y_i$ pointwise, unrealised verdicts contributing zero. On the
wrongful-eviction event,
$m_{\mathrm{rej}}-a\le\sum_{i\le k-a}Z_i\le\sum_{i\le k-a}Y_i
\le\sum_{i\le k-\fs}Y_i+(\fs-a)$, whence
$\sum_{i\le k-\fs}Y_i\ge m_{\mathrm{rej}}-\fs$, uniformly in $a$.
Since $m_{\mathrm{rej}}-\fs=(k-\fs)-(m-1)
=\big((1-p_b)+\delta_b'\big)(k-\fs)$ with
$\delta_b':=p_b-\tfrac{m-1}{k-\fs}\ge\delta_b>0$, Hoeffding's
inequality (Lemma~\ref{S-lem:engine}) bounds the event by
$e^{-2\delta_b'^{2}(k-\fs)}\le e^{-2\delta_b^{2}(k-\fs)}$. Simultaneity: an integer $m$ lying strictly
between $\fs+(1-q)(k-\fs)$ and $p_b(k-\fs)$ exists whenever the gap
between those quantities exceeds one, i.e.\ whenever
$(k-\fs)(p_b+q-1)>\fs+1$.
\end{proof}

\begin{remark}[Resolution as a demand-side event]
\label{S-rem:resolution}
Resolution is a demand-side property: the demand-paced panel has no
deadline, $g_k(M)$ may grow arbitrarily slowly, and an item without
demand expires with no protocol consequence, for which liveness (in
the classical, eventual-success sense) is vacuous (\cref{S-n:roles}).
Proposition~\ref{S-prop:live} and Theorem~\ref{S-thm:coscale}(ii) bound
wrongful eviction unconditionally, asserting no probability for
resolution itself. As a corollary reading: conditional on resolution,
vindication has probability at least
$1-e^{-2\delta_b^{2}(k-\fs)}/\Prob[\text{resolution}]$.
\end{remark}

\begin{proposition}[Publisher accountability is safe for benign items]
\label{S-prop:pubsafe}
Under Rule~\ref{S-r:panel}(iv),(vii):
\begin{enumerate}[label=(\roman*),leftmargin=2em,itemsep=1pt,topsep=1pt]
\item the adversary's panel verdicts alone can never trigger a
publisher penalty against any publisher, since eviction requires
$m_{\mathrm{rej}}>\fs$ rejecting verdicts;
\item with the default bars and wrongful-eviction margin $\delta_b>0$,
write $p^*:=e^{-2\delta_b^{2}(k-\fs)}$; each panel on a benign item
wrongfully evicts (penalising the publisher) with probability at most
$p^*$ conditionally on all prior protocol events
(Proposition~\ref{S-prop:live}, whose hypothesis is conditional). Three
consequences, in decreasing strength of hypothesis.
\emph{(Fixed instances.)} For any $r$ dispute instances on benign
items of the same publisher, \emph{predictably specified}:
membership of an instance in the family is determined no later than
that instance's opening atom, measurably in the opening-inclusive
log history and before any verdict content of that instance (any
advance deterministic choice included); a family selected
retrospectively from panel outcomes is not a fixed-instance family
--- it is covered by the capped tier where the deterministic cap
$L\le\bar\ell$ is available, and otherwise only by the linear
random-count tier. For such a family, the probability that all $r$
instances end in wrongful eviction is at most $(p^*)^{\,r}$,
uniformly over adversary strategies and over any dependence among
the panels.
\emph{(Random count, linear.)} With $L$ the number of dispute
instances opened on the publisher's benign items in the epoch,
indexing panel trials at their \emph{opening} events (an instance
unresolved at the epoch horizon evicts nobody and carries indicator
zero), the expected number of wrongful evictions is at most
$p^*\,\Ex[L]$, whence
$\Prob[\text{at least }r\text{ wrongful evictions}]\le p^*\Ex[L]/r$
(Markov); $\Ex[L]$ is bounded by the dispute ledger
(Lemma~\ref{S-lem:budget}, $\bar D_v$) summed over the publisher's
items, and, for fast-track reviews, deterministically by $\Emax$ per
item per epoch (Rule~\ref{S-r:panel}(vi)). No $r$-fold exponent is
claimed at this generality.
\emph{(Random count, capped.)} If the deployment enforces a
deterministic per-epoch cap $L\le\bar\ell$ on dispute instances
against one publisher's items, then
$\Prob[\text{at least }r\text{ wrongful evictions}]
\le\binom{\bar\ell}{r}(p^*)^{\,r}$,
decaying geometrically in $r$ whenever
$2\delta_b^{2}(k-\fs)>\ln\bar\ell$;
\item every eviction of an \emph{adversary-published} poisoned item
consumes the adversary's allowance and, on repetition within an epoch, its
submission rights and attestation weight; eviction of poison published by
an honest-but-deceived identity penalises \emph{that} publisher, by design
(\ref{S-q:postendorse});
\item no publisher penalty ever touches a local store
(Lemma~\ref{S-lem:nodelete}).
\end{enumerate}
Dispute expiry (Rule~\ref{S-r:hygiene}(i)) is neither an eviction nor a
vindication and triggers no penalty on either side. Protection is
claimed for publishers of \emph{benign} items: an honest-but-deceived
publisher of attacker-crafted content is penalised on eviction,
by design (\ref{S-q:postendorse}).
\end{proposition}

\begin{proof}
(i) Eviction requires at least $m_{\mathrm{rej}}$ rejecting verdicts
with $m_{\mathrm{rej}}>\fs$; by \ref{S-inv:verdict} the adversary
casts at most $\fs$ verdicts on the dispute over its entire life,
however many epoch boundaries it spans, so at least one honest panelist
must reject; with no honest rejections there is no eviction and hence
no penalty. (ii) Under the default bars a resolved panel's outcome is
eviction or vindication, exclusively (Lemma~\ref{S-lem:decisive}(i)--(ii)), so the
penalising event is exactly wrongful eviction, bounded per panel by
Proposition~\ref{S-prop:live}. For the fixed-instance product bound, the $r$ predictably specified instances are indexed in opening
order, and their honest verdicts jointly in log order of their
header atoms; a verdict's instance membership and within-instance
honest ordinal are thereby determined at its header atom,
$\mathcal G_i$-measurably, as Lemma~\ref{S-lem:alloc} requires (\S\ref{S-app:prob}). The benign-item hypothesis
of Proposition~\ref{S-prop:live} bounds each honest verdict's
conditional rejection probability by $1-p_b$ at its header-inclusive
$\sigma$-algebra --- including the outcomes of earlier-resolved
panels --- so Lemma~\ref{S-lem:adaptive}(i), in the $\sigma$-algebra
form of Remark~\ref{S-rem:adaptivegen}, applied once to the joint
sequence, supplies coupling variables with $Z_i\le Y_i$ pointwise on
realised verdicts. Allocate cells: assign the $j$-th specified
instance's honest verdicts, in log order, to cells
$(j,1),\dots,(j,k-\fs)$, discarding any beyond the $(k-\fs)$-th; a
verdict's cell is determined at its header atom (its dispute instance
and within-instance honest ordinal are log facts prior to its
content), hence $\mathcal G_i$-measurably and injectively, the
instances' index sets disjoint --- not necessarily contiguous, which
the argument never needs. By Lemma~\ref{S-lem:alloc}, the cell array
$(X_{j,\ell})_{j\le r,\,\ell\le k-\fs}$ (coupling variables on filled
cells, fresh independent uniforms on never-filled cells) is i.i.d.\
$\mathrm{Bernoulli}(1-p_b)$. On the all-$r$-evict event, instance $j$
resolves in eviction: with $a_j\le\fs$ adversarial verdicts
(\ref{S-inv:verdict}), honest verdicts number $h_j\le k-a_j$, of which
at least $m_{\mathrm{rej}}-a_j$ reject. If $h_j\le k-\fs$, every honest verdict
sits among the first $k-\fs$ cells, which then contain at least
$m_{\mathrm{rej}}-a_j\ge m_{\mathrm{rej}}-\fs$ rejections; otherwise
at most $h_j-(k-\fs)\le\fs-a_j$ of the rejecting ones lie beyond cell
position $k-\fs$, and the first $k-\fs$ honest verdicts contain at
least $(m_{\mathrm{rej}}-a_j)-(\fs-a_j)=m_{\mathrm{rej}}-\fs$
rejections. In either case
$\sum_{\ell\le k-\fs}X_{j,\ell}\ \ge\ \sum_{\ell\le k-\fs}Z_{(j,\ell)}
\ \ge\ m_{\mathrm{rej}}-\fs$, never-filled cells contributing
nonnegatively, uniformly in $a_j$. The all-$r$-evict event is
therefore contained in
$\bigcap_{j\le r}\big\{\sum_{\ell\le k-\fs}X_{j,\ell}\ge
m_{\mathrm{rej}}-\fs\big\}$; the array being i.i.d.\ with the rows
indexed by disjoint cell sets, the intersection's probability is the
product of $r$ identical binomial tails, each at most $p^*$ by the
margin computation of Proposition~\ref{S-prop:live}
($m_{\mathrm{rej}}-\fs=\big((1-p_b)+\delta_b'\big)(k-\fs)$,
$\delta_b'\ge\delta_b$, Hoeffding via Lemma~\ref{S-lem:engine}), giving
$(p^*)^{\,r}$. For the
linear tier, let $W_j$ indicate wrongful eviction of the $j$-th
instance opened in the epoch, indexed at its opening event, with
$W_j=0$ for instances unresolved at the horizon;
$\Prob[W_j=1\mid\text{prior atoms}]\le p^*$ by Proposition~\ref{S-prop:live}
(its hypothesis conditions on all prior protocol events, the opening
atom included), so the delayed-marks lemma (Lemma~\ref{S-lem:delayed}, with
$\mathcal G_j$ generated by all atoms up to the $j$-th opening event
and $W_j$ resolved only at that instance's resolution, possibly after
later openings) gives $\Ex[\sum_j W_j]\le p^*\,\Ex[L]$ and Markov the
tail. For the capped tier, fix any set of $r$ indices
$j_1<\dots<j_r\le\bar\ell$; the fixed-instance argument above, applied
to the $j_1$-th,\,\dots,\,$j_r$-th instances opened in the epoch (the
instances being determined by their opening order, and their honest
verdicts jointly log-ordered), bounds the probability that all $r$
resolve in eviction by $(p^*)^{\,r}$. At least $r$ wrongful evictions
entails that some such index set evicts wholly; the union bound over
the $\binom{\bar\ell}{r}$ index sets gives the claim. No direct
stochastic domination of delayed, opening-indexed eviction indicators
is asserted. The same-bytes sharpening is
\ref{S-inv:verdict} with Definition~\ref{S-def:fresh}: a cast verdict is a
logged decision on the bytes, so its caster is ineligible for any
later dispute instance on the same bytes under
Rule~\ref{S-r:panel}(iii)(b). (iii) By
construction of clause (vii), read per publisher. (iv) Clause (vii) modifies submission
rights and attestation weight, which are shared-tier attributes;
Lemma~\ref{S-lem:nodelete} covers the rest. The expiry clause is
Rule~\ref{S-r:hygiene}(i) verbatim.
\end{proof}

\begin{remark}[Reading of Proposition~\ref{S-prop:pubsafe}(ii)]
\label{S-rem:pubsaferead}
The escalating penalties of Rule~\ref{S-r:panel}(vii), requiring
$r\ge2$ wrongful evictions within one epoch, are exponentially rare
per specified pair of instances, bounded in expectation in general,
and carry the $r$-fold exponent exactly where a deterministic
instance cap holds (capped tier). For repeated panels on the
\emph{same} byte string the bounds sharpen: every verdict is a logged
decision consuming shared-presence freshness
(Definition~\ref{S-def:fresh}, Rule~\ref{S-r:panel}(iii)(b)--(c)), so
each identity casts at most one verdict across all of the bytes'
panels ever; in particular the adversary's $\fs$ seat-eligible
identities are spent once across those panels, not once per panel.
\end{remark}

\begin{lemma}[Bounds on rejection-based denial of service]
\label{S-lem:freeze}
\leavevmode
\begin{enumerate}[label=(\roman*),leftmargin=2em,itemsep=2pt,topsep=2pt]
\item At any time, at most $f\cdot\rmax$ items are held in Disputed by
adversarial rejections. Moreover, adversarial dispute-initiations against
any single item number at most $f$ per epoch (Rule~\ref{S-r:panel}(i),
single-initiation clause), independently of $\rmax$.
\item \emph{(Demand-paced resolution.)} Unconditionally, with no
behavioural assumption (persistence exceptions per \ref{S-inv:floor}), a freeze costs each affected demander at most
one accepting confirmation per demanded pair, and only if the demander
holds no local in-scope copy: exactly the required confirmation any
demander of an unpromoted pair owes regardless
(Theorem~\ref{S-thm:floor}(i)). A demander who rejects or abandons
incurs inspection effort, a distinct resource disclosed in \ref{S-n:effort}, not bounded by the cap. A dispute therefore suspends only the community
amortisation for its demanders and never pushes anyone above the
single-user cap, however long the freeze
lasts. Under \ref{S-a:rhoseat}, the number so affected is additionally
bounded: the demanders affected while the dispute is unresolved number at most $\kappa$ in
expectation, with exponential tail (Lemma~\ref{S-lem:geo}), independent
of the item's popularity and of the freeze's wall-clock duration. A
dispute on an undemanded item affects no user and expires within $\Texp$ of its
last demand (Rule~\ref{S-r:hygiene}). (Wide pre-exposure:
Remark~\ref{S-rem:freezeread}; \ref{S-q:exhaust}.)
\item \emph{No panel-lapse exploitation; slot economics.} There is no panel
lapse, hence no forced re-entry: an unresolved dispute never drains the
owner's submission budget. A rejection slot stays occupied until
resolution or expiry: against demanded items, disputes resolve at
demand rate into the accountability schedule; against undemanded items,
the slot pins only an undemanded item and frees at expiry. (Remark~\ref{S-n:roles}; \ref{S-q:lapse}.)
\item A sustained wrongful-rejection campaign alters no local store
(Lemma~\ref{S-lem:nodelete}) and costs each user at most the single-user
floor (Theorem~\ref{S-thm:floor}); beyond the $f$ free first charges each
epoch, it expends the adversary's rejection rights at an escalating rate
(Rule~\ref{S-r:panel}(v)). Its throughput is bounded by three deterministic
caps, each in its own currency: \emph{charged} dispute-initiations (those
eventually contradicted by vindication) number at most $\cacc f$ per
epoch (Lemma~\ref{S-lem:cacc}); \emph{concurrent} open rejections number at
most $f\rmax$ at any moment, and freezes against any single item at most
$f$ per epoch (clause~(i), single initiation); and uncharged
\emph{terminal} disputes (ending in eviction or expiry) number at most
$\Emax$ per shared-tier presence per item per epoch
(Rule~\ref{S-r:hygiene}(i)), each retiring that presence. No single
constant bounds all wrongful rejections across all items --- a slot freed
by expiry may be re-used against another item --- and what each re-use
buys is one freeze, resolved or expired at that item's own demand clock,
within the single-user cap. The allowances renew at epoch boundaries (\ref{S-n:avail}).
\end{enumerate}
\end{lemma}

\begin{proof}
(i) Each adversarial identity holds at most $\rmax$ open rejections
(Rule~\ref{S-r:panel}(i)) and the adversary controls at most $f$
identities. The per-item clause caps each identity at one initiation
per item per epoch. (ii) The harm of a freeze is exactly the
amortisation it suspends: the demanders who must confirm personally
while the item is frozen; by Rule~\ref{S-r:panel}(iii)--(iv) the eligible
deciding demanders cast the verdicts and, with the default bars,
resolution occurs by the $k$-th verdict; each affected demander's
accepting confirmation, where one occurs, is the one required
confirmation they would owe on any unpromoted pair,
so the per-user floor is untouched (\ref{S-inv:floor}); the
in-expectation and tail bounds are Lemma~\ref{S-lem:geo}; an item with no
new demand meets the expiry clause of Rule~\ref{S-r:hygiene}(i).
(iii) By Rule~\ref{S-r:panel}(iv) the only re-entry paths are the
optional neither-bar outcome and post-expiry republication, both
owner-initiated; slots free on resolution or expiry
(Rules~\ref{S-r:panel}(iv),~\ref{S-r:hygiene}(i)). (iv) Every
dispute-initiating rejection contradicted by vindication charges the
rejecter's schedule and ceiling (Rule~\ref{S-r:panel}(v)); by
Lemma~\ref{S-lem:cacc} the adversary's effective charged
dispute-initiations are deterministically at most $\cacc$ per identity per epoch,
bounding per-epoch \emph{charged-initiation} throughput at $\cacc f$ as a protocol fact; the concurrency, single-initiation, and terminal-dispute caps are Rule~\ref{S-r:panel}(i) and Rule~\ref{S-r:hygiene}(i) verbatim. By
Lemma~\ref{S-lem:nodelete} no local store is altered, and each user's
required cost remains at most the single-user cap
(Theorem~\ref{S-thm:floor}).
\end{proof}

\begin{remark}[Wide pre-exposure]\label{S-rem:freezeread}
An item pre-exposed so widely that few verdict-eligible demanders
remain simply stalls \emph{frozen}: the freeze blocks promotion at
every tool for as long as the panel is unresolved, and the
demand-paced panel carries no deadline an adversary could exploit
(Rule~\ref{S-r:panel}(ii)--(iii)); this is the deterministic failure
mode of \ref{S-a:rhoseat}, and its cost is availability, never safety.
For a \emph{poisoned} item the stall is borne by the adversary's own
submission. A \emph{benign} item can also be stalled this way --- by
pre-exposure or pool-skewed demand, with or without adversarial
help --- and then the availability cost falls on its publisher and
demanders; the harm is bounded exactly as for any freeze: each
demander pays at most one required confirmation per pair, within the single-user cap
(Lemma~\ref{S-lem:freeze}(ii), unconditional clause), the item remains
personally usable through each user's own gate
(Rule~\ref{S-r:panel}(ii)), local stores are untouched
(Lemma~\ref{S-lem:nodelete}), and an item whose demand lapses expires
harmlessly (Rule~\ref{S-r:hygiene}(i)). (\ref{S-q:exhaust}.)
\end{remark}

\begin{remark}[Roles of $\Twin$, $\Tpan$, $\Texp$; the demand-paced
design]\label{S-n:roles}
The window $\Twin$ forbids the promotion race: finalisation occurs only
at window end, so any qualifying rejection intervenes even if $N$
acceptances arrive first: this is what makes the partition of
Theorem~\ref{S-thm:panel} well-defined; and each attempt's identity-disjoint credited endorsement set (Lemma~\ref{S-lem:disjoint}, Definition~\ref{S-def:attempt}) is what makes the per-attempt
marginal bounds and their union valid
(Corollary~\ref{S-cor:submission}). $\Tpan$ survives only as the clock of
fast-track suspension panels,
whose log-fresh jurors are drawn from demanders during suspension and
prior adopters without a logged decision on the bytes
(Rule~\ref{S-r:panel}(vi)); their supply is not guaranteed by
construction --- prior adopters with logged decisions are ineligible
--- but is availability-side, exactly \ref{S-a:rhoseat}'s province;
the ordinary panel carries no global
deadline; it is demand-paced: verdicts accumulate at the item's own
demand rate, so resolution time scales inversely with demand while the
harm of the freeze scales directly with it, and their product is
bounded (Lemma~\ref{S-lem:freeze}(ii)). A disputed item with too little
fresh demand simply waits, exactly as it would have waited for ordinary
promotion; an item nobody demands expires harmlessly within $\Texp$.
Zero-tolerance designs lack an accountability clock entirely; the
demand-paced panel provides one wherever fresh demand exists. An
evict-and-restart rule would instead multiply per-item cost by the
number of rejections; the panel's decisive, demand-paced resolution is
what keeps the dispute load an additive
$\bigO(\Emax N+\ctool\bar D_v\kappa)$ per demanded item, with $\bar D_v$ the
expected dispute count of Lemma~\ref{S-lem:budget}, absorbed by the cross-term and
shrinking-front-loading conditions of Theorem~\ref{S-thm:coscale}
(\ref{S-q:lapse}).
\end{remark}

\begin{corollary}[Submission budget and publisher accountability]
\label{S-cor:submission}
Promotion occurs only at an attempt's finalisation
(Definition~\ref{S-def:attempt}), and every attempt is counted at its
\emph{anchor} --- a logged event preceding every decision creditable
to the attempt, hence a sound conditioning point
(Definition~\ref{S-def:attempt}, \S\ref{S-app:prob}). Under the default
bars, the anchors lying in a given epoch are of exactly two kinds,
each bounded by the submission rate limit: presence-creating
submissions, and finalisation events, each of which closes an entry
opened by a same-epoch rate-limited (re-)submission (windows are
epoch-contained, convention~(\ref{S-conv:c4}); Rule~\ref{S-r:share}). Hence the
promotion attempts of adversary-published poisoned items
\emph{anchored} in a given epoch number at most $2f\rsub$ (paraphrase
variants and re-entries included). Each such attempt gathers an
identity-disjoint credited endorsement set (Lemma~\ref{S-lem:disjoint})
and is bounded marginally by Theorem~\ref{S-thm:panel} at its own anchor
(under its hypotheses); the union bound, valid under arbitrary
dependence across attempts, gives
\begin{multline*}
\Prob[\text{any adversary-submitted poison promoted at an attempt
anchored in the epoch}]\\
\ \le\ 2f\rsub\,\Big((1-q)^{N-f}+e^{-2\delta^{2}(k-\fs)}\Big).
\end{multline*}
(Attempts anchored while the bytes hold live $\CC$ status --- additional-tool promotions of already-promoted bytes --- are covered under the extended context reading at the close of Theorem~\ref{S-thm:panel}'s proof; the event that any poison is ever promoted at all consumes only $\CC$-free anchors.)
Every promotion belongs to an attempt anchored in some epoch, so
summing over the anchor epochs $1,\dots,E$ \emph{from deployment
inception} bounds the probability that any adversary-submitted poison
is ever promoted within that horizon by
$2f\rsub\,E\,\big((1-q)^{N-f}+e^{-2\delta^{2}(k-\fs)}\big)$. For an interval not beginning at deployment inception, no bound
conditional on its realised carry-in pool is asserted --- the set of
attempts still live at the interval's start is itself
outcome-selected, for the reason of Remark~\ref{S-rem:anchorepoch} ---
and the valid statement is the anchor-epoch bound applied to all
anchor epochs from deployment inception through the interval's end. (Under the looser bars
permitted by Rule~\ref{S-r:panel}(iv), neither-bar resolutions are
additional anchors; the numeric count $2f\rsub$ is asserted for the
default bars, and loose-bar deployments use the expectation form
below, with the anchored-attempt count enlarged accordingly;
per-attempt coverage per Remark~\ref{S-rem:panelscope}.)

Poisoned content published by an honest-but-deceived identity
(\ref{S-q:postendorse}) is not counted by $2f\rsub$. Writing
$S_{\mathrm{anc}}$ for the (in general random) number of poisoned
promotion attempts anchored in the epoch, whatever their publishers:
indexing attempts at their anchor atoms and applying the delayed-marks
lemma (Lemma~\ref{S-lem:delayed}) to the per-attempt success indicators
--- each bounded, conditionally on the anchor-inclusive
$\sigma$-algebra of its anchor atom, by the per-attempt marginal
bound, and each resolved only at the attempt's finalisation, possibly
after later attempts' anchors, which Lemma~\ref{S-lem:delayed} permits
---
\begin{multline*}
\Prob[\text{any poison promoted at an attempt anchored in the epoch}]\\
\le\Ex[S_{\mathrm{anc}}]\Big((1-q)^{N-f}+e^{-2\delta^{2}(k-\fs)}\Big).
\end{multline*}
and a deployment enforcing a deterministic cap
$S_{\mathrm{anc}}\le\bar S$ may substitute $\bar S$ for the
expectation. (No bound keyed to the attempts \emph{finalising} in an
epoch is asserted: that family is outcome-selected,
Remark~\ref{S-rem:anchorepoch}. Nor is a bound conditional on a realised
value of $S_{\mathrm{anc}}$ asserted: the attempt count may be
adaptively determined by earlier outcomes; the expectation and
deterministic-cap forms are the valid system-wide statements.)
$\Ex[S_{\mathrm{anc}}]$ is not bounded in terms of the adversary's
resources alone --- it depends on how often false endorsement
(\ref{S-a:narr}) and publication co-occur among honest identities ---
though the per-identity rate limit yields the crude default-bar cap
$S_{\mathrm{anc}}\le2M\rsub$ over $M$ eligible identities, and each
such item, on eviction, penalises its publisher
(Rule~\ref{S-r:panel}(vii)), economically suppressing repeat deception.
Under permanent freshness, the credited endorsement sets of all
attempts of a given byte string are identity-disjoint over the
deployment's \emph{entire history}, not merely within an epoch
(Lemma~\ref{S-lem:disjoint}); consequently the number of attempts of the
same bytes that can ever gather $N$ credited endorsements is at most
$\lfloor H_{\mathrm{acc}}/(N-f)\rfloor$, where $H_{\mathrm{acc}}$ is
the total number of honest identities that ever accept those bytes at
a gate. Moreover, every eviction of a poisoned attempt triggers the
publisher-accountability consequences of Rule~\ref{S-r:panel}(vii)
(cool-down; on repetition within an epoch, suspension of submission
rights and attestation-weight reduction); the quantitative effect on
future submission capacity depends on the deployment's escalation
schedule (\S\ref{S-app:fedrules}, required parameters) and is consumed
by no proof, while a benign publisher's exposure to the same penalty is
bounded by Proposition~\ref{S-prop:pubsafe}(ii).
\end{corollary}

\begin{proof}
An attempt is counted at its anchor. Under the default bars every
anchor is a presence-creating submission or a finalisation
(Definition~\ref{S-def:attempt}); distinct attempts have distinct
anchors; presence-creating submissions of adversary-published items in
the epoch number at most $f\rsub$ (Rule~\ref{S-r:share}); and each
finalisation in the epoch closes an entry whose opening
(re-)submission lies in the same epoch (windows are epoch-contained,
convention~(\ref{S-conv:c4})), so finalisation anchors also number at most
$f\rsub$: at most $2f\rsub$ anchored attempts. Union bound over these
attempts, each covered marginally by Theorem~\ref{S-thm:panel} at its
anchor on its own credited set (Lemma~\ref{S-lem:disjoint}); a union
bound requires no independence across the events it covers. The
horizon form sums the per-epoch display over the anchor epochs of the
horizon, every promotion belonging to the attempt anchored in some
such epoch (from deployment inception; a general interval is covered by summing from inception through its end, no carry-in-conditional bound being asserted). The
system-wide form is the delayed-marks lemma (Lemma~\ref{S-lem:delayed})
applied to anchor-indexed success indicators. Its occurrence events lie in the conditioning $\sigma$-algebras: take
$\mathcal G_i$ generated by all atoms up to and including the $i$-th
anchor atom, so that $O_i$ is $\mathcal G_i$-measurable; the
per-attempt bound of Theorem~\ref{S-thm:panel} is stated conditionally
on exactly this anchor-inclusive $\sigma$-algebra, so it bounds each
mark conditionally on $\mathcal G_i$, the mark resolving only at the
attempt's finalisation, which Lemma~\ref{S-lem:delayed} permits; the
deterministic-cap form substitutes $\bar S$. The lifetime clause is
immediate from identity-disjointness of credited decisions across all
attempts. The accountability clauses are Rule~\ref{S-r:panel}(vii) and
Proposition~\ref{S-prop:pubsafe}(ii)--(iii).
\end{proof}

\begin{remark}[Why attempts are keyed to anchor epochs, not
finalisation epochs]\label{S-rem:anchorepoch}
An earlier version of Corollary~\ref{S-cor:submission} counted the
attempts whose \emph{finalisation} lies in the epoch. That attribution
is unsound, for the same optional-selection reason the attempt anchor
exists (Definition~\ref{S-def:attempt}, commentary): banked decisions
logged during dormancy are public log events preceding the owner's
resubmission decision, so the set of attempts finalising in a given
epoch is outcome-selected --- an adversary maintaining many anchored
dormant attempts on distinct byte strings and resubmitting only those
with favourable banked outcomes makes the per-finalisation-epoch
attempt count small while the epoch's promotion probability is
governed by the whole maintained pool. Anchors, by contrast, precede
every decision creditable to their attempts, so the anchored-in-epoch
family is determined before any of the outcomes it is charged with,
and the delayed-marks lemma applies. The per-attempt bound of
Theorem~\ref{S-thm:panel} and the lifetime ceiling of
Remark~\ref{S-rem:lifetime} are unaffected: only the calendar
attribution of attempts to epochs changes.
\end{remark}

\begin{remark}[Lifetime exposure under permanent freshness]
\label{S-rem:lifetime}
The horizon union over anchor epochs in Corollary~\ref{S-cor:submission}, and the lifetime ceiling,
count anchored attempts --- the lifetime ceiling, threshold-reaching attempts --- whose honest endorsers are never re-used: every
pipeline entry of a given byte string that reaches its threshold
consumes $N-f$ honest identities' first-and-only \emph{logged}
decisions on those bytes, so the adversary cannot re-use a
previously logged-deciding population, and
$\lfloor H_{\mathrm{acc}}/(N-f)\rfloor$ is a hard ceiling on the
number of \emph{threshold-reaching} attempts of the same bytes over
the deployment's lifetime, independent of the number of epochs.
Unsuccessful attempts --- each gathering fewer than $N-f$ new honest credited
acceptances --- and presentations yielding no logged acceptance are
not bounded by this ceiling; they are governed by the per-epoch rate
limit $\rsub$ and the per-attempt marginal bound of
Theorem~\ref{S-thm:panel}, each attempt consuming its own union-bound
term.
The ceiling is per exact byte string (convention~(\ref{S-conv:c5})): an adversary
minting byte-distinct paraphrase variants holds a fresh ceiling per
variant, each consuming its own rate-limited submission and
union-bound term (\ref{S-q:paraphrase}, Corollary~\ref{S-cor:submission});
the lifetime clause therefore hard-bounds \emph{successful} re-entry of
\emph{fixed} bytes, while variant families are governed by the
per-epoch rate and calibration clauses, not by this ceiling.
\end{remark}

\begin{remark}[Weighted form]\label{S-rem:dispweight}
With endorsement weights, replace the exponent $N-f$ by
$(W-\Wadv)/\wmax$. With seat weights in $[\wseat,\wmax]$ and a
vindication weight bar $W_m$, honest accepting weight at least
$W_m-\fs\wmax$ is required; seat weights are attestation-layer
constants fixed at verdict time, hence predictable in the sense of
Lemma~\ref{S-lem:engine}'s second clause, and Hoeffding for weighted
sums (Lemma~\ref{S-lem:engine}) gives
$\exp(-2t^2/((k-\fs)\wmax^2))$ with $t$ the weight margin: constants
change, exponents do not, and the bound remains reach-free.
\end{remark}

\begin{remark}[Per-source-class thresholds]\label{S-rem:classthresh}
Promotion thresholds may be set per \emph{source} class of the
underlying content (authentication is not trust: a signed phishing
message is still an attack; so source identity only tunes the
threshold, never bypasses the quorum, and never widens a tool scope).
All bounds hold with $W$, $N$, $m$, $k$ read as the minimum over source
classes; Theorem~\ref{S-thm:det} requires
$\min_{\mathrm{class}}W>\Wadv$ and $\min_{\mathrm{class}}m>\fs$.
\end{remark}

\section{The Memory-Verification Budget}
\label{S-app:budget}

\paragraph{Scope.}
All results in this appendix count confirmations attached to
\emph{reusable memory content}, per (value, tool) pair: a one-time personal cap of private
values, and the stream of shareable Confirmed items. Per-task
confirmations (for freshly composed sensitive arguments and for
crystallizing ephemeral task content) are common to every gated
design, scale with usage rather than population, and are orthogonal to
the amortisation analysed here. PerMIT's distinction is layered. Pathwise --- independent of every federated and demand assumption, persistence exceptions per \ref{S-inv:floor} --- no user's required
cost exceeds once per (value, tool) pair, ever, and the epoch's
expected shared required load never exceeds the standalone single-user
mass $d_{\mathrm{tool}}$ --- the cost of having no community at all
(Theorem~\ref{S-thm:floor}(i)--(ii)); this cap is the appendix's firmest guarantee: pathwise, consuming no federated or demand assumption (its expected clauses read the reference model's masses), subject only to the stated first-party persistence exceptions. On top
of it, the \emph{ledger-charged} shared stream is at most the
front-loading multiple $\Bfl\,\gbar$ of the standalone mass,
independently of item popularity (Theorem~\ref{S-thm:sat},
Remark~\ref{S-rem:satread}); and the \emph{ledger residual} --- demand
arriving outside every charged piece, in particular handle-expressed
arrivals of already-certified bytes, which gate by design
(Rule~\ref{S-r:whitelist}(c), \ref{S-q:handlecover}) --- has its benign,
assumption-conforming component vanish under the full vanishing
conditions of Theorem~\ref{S-thm:sat}(iii): the
coverage-and-representation assumption \ref{S-a:promsuf} together with
the product-rate condition $D_0(M)V(M)=o(M)$,
$\varepsilon(M)\,d_{\mathrm{good}}(M)\to0$ (under
\ref{S-a:massbound}, from $\varepsilon(M)\to0$; $d_{\mathrm{good}}$
per Lemma~\ref{S-lem:distinct}), and bounded tool spread
(\ref{S-a:toolspread}); these conditions are sufficient, not
necessary, and demand on poisoned or assumption-violating pairs
remains in $r^{(e)}_{\mathrm{bad}}$ (Corollary~\ref{S-cor:total}).
Under the measured corridor of \S\ref{S-app:corridor} --- whose
front-loading rows describe deployments whose time geometry changes
with scale, not consequences of population growth alone
(Remark~\ref{S-rem:gbarread}) --- both shared components vanish; where
the corridor or \ref{S-a:promsuf} fails, the universal fallback
applies to the \emph{required} component: the epoch's expected shared
required load remains at most the standalone mass
$d_{\mathrm{tool}}$ (Theorem~\ref{S-thm:floor}(ii)), the once-per-pair
cap binding throughout. Voluntary invitation confirmations remain
governed by the ledger rows of Lemma~\ref{S-lem:budget}; rejection and
attention effort is accounted separately in \ref{S-n:effort}. No
standalone-mass ceiling is claimed for those additional resources.
Headline readings of this appendix elsewhere in the paper must carry
these conditions: ``shared verification load vanishes'' is asserted
only under measured saturation, smoothness, coverage, juror-supply,
and bounded-mass hypotheses (\ref{S-a:smooth}--\ref{S-a:massbound}),
never from population growth alone (Remark~\ref{S-rem:gbarread}); the
firmest statements are the pathwise once-per-pair cap and the
reference-model standalone-mass ceiling (Theorem~\ref{S-thm:floor}).

\subsection{Model}
\label{S-app:budgetmodel}

\begin{itemize}[leftmargin=1.4em,itemsep=2pt,topsep=2pt]
\item A population of $M$ users; a universe $\Vals$ of shareable
Confirmed items. The accounting unit is the \emph{(value, tool) pair}
(the endorsement unit, \S\ref{S-app:config}): a user's demand, required
cost, and local coverage are all per pair. Personal and
context-dependent values lie outside $\Vals$ (Rule~\ref{S-r:share},
Lemma~\ref{S-lem:shareexcl}); $\dpriv$ is the expected number of distinct personal \emph{(value, tool)
pairs} a user ever confirms, once each over the deployment's lifetime
(the pair is the endorsement unit, \S\ref{S-app:config}; a personal value
used at $j$ tools contributes $j$ pairs). $\dpriv$ is a \emph{lifetime}
quantity; where it appears alongside single-epoch masses
(Theorem~\ref{S-thm:floor}(ii)), the sum is read as a lifetime personal
term plus a per-epoch shareable term, each bounding its own component.
\item The independent reference model (IRM) \cite{fagin1977asymptotic}, per pair
and per user: user $u$ demands the pair $(v,t)$ within the epoch with
probability $p_{u,v,t}$, independently across users; the model is
heterogeneous, and $p_{v,t}:=M^{-1}\sum_u p_{u,v,t}$,
$p_v:=M^{-1}\sum_u\Prob[\text{$u$ demands $v$ at some tool}]$ denote
the population-average marginals, so $p_{v,t}\le p_v$ and
$\Ex[n_{v,t}]=Mp_{v,t}$. Every expected-count argument in this
appendix consumes only cross-user independence and this identity,
never identical distribution across users (the owner's degenerate
indicator, next bullet, is the one systematic heterogeneity). The
single-epoch masses are
$d:=\sum_v p_v$ and $d_{\mathrm{tool}}:=\sum_{v,t}p_{v,t}$, both finite,
owner coupling folded in (next bullet),
with $d_{\mathrm{tool}}\le\ctool\,d$ under \ref{S-a:toolspread}. Under
this model $n_v$ and $n_{v,t}$ denote the \emph{latent}
counts $\sum_u D_{u,v,t}$; the log-defined realized counts satisfy
$n^{\mathrm{log}}_{v,t}\le n_{v,t}$ pathwise
(\S\ref{S-app:prob}), so every expected bound stated against $Mp_{v,t}$
covers the realized events the ledger charges.
\item \emph{Owner coupling.} The owner's publication schedule for
each item --- the epochs in which the owner submits or resubmits $v$
(Rule~\ref{S-r:share}) --- is carried in the exogenous demand factor
alongside the demand indicators (\S\ref{S-app:prob}): it is owner
behaviour, exogenous to the protocol and the attacker exactly as the
typed stream is (\ref{S-a:exo}). In every scheduled epoch the owner's
own demand indicator for $v$ is set to one, at the owner's creating
tool (formally, the owner's marginal $p_{u,v,t}$ is degenerate at one
in scheduled epochs; other users' indicators and the cross-user
independence are untouched --- the heterogeneous form of the previous
bullet): submission is modelled as demand-driven, latently mirroring
Definition~\ref{S-def:demand}'s clause that the owner's publish
confirmation is a realized demand event. Consequently every epoch
containing a submission or re-entry of $v$ has $n_v\ge1$, and every
realized demander is a latent demander, so
$n^{\mathrm{log}}_v\le n_v$ pathwise. A shared-tier presence can, by
contrast, be \emph{carried} into an epoch with no current-epoch owner
activity (a dormant candidate short of its retirement clock, a live
$\CC$ entry, or the one carry-in freeze of Lemma~\ref{S-lem:budget});
the vanishing clause of Lemma~\ref{S-lem:budget}(iii)--(iv) therefore
rests on the absence of \emph{events}, not of presence: on
$\{n_v=0\}$ there is no entry (an entry requires a current-epoch
submission, hence owner demand), no dispute or review opening (each
opens at a \emph{latent}
demander's protocol-bearing gate decision --- attempted use realizes
the latent indicator, required-approval clause; the opener need not be
a realized log-demander, e.g.\ at a cache gate's rejecting branch,
Definitions~\ref{S-def:counting}(iii),~\ref{S-def:demand}), no
freeze-bucket arrival, no invitation (a re-entry notification
requires an entry, and a carried-in review issues no new prompts,
Rule~\ref{S-r:panel}(vi)), and no publish confirmation. The masses
$p_{v,t},p_v,d,d_{\mathrm{tool}}$ are read with this coupling folded
in (the owner's degenerate indicator contributes its own mass like
any user's), so every expected bound stated against them is unchanged
in form.
\emph{Scope of the exogenous-schedule reading.} The reference model's
expectation-form conclusions (Lemma~\ref{S-lem:budget}(iii),
Theorem~\ref{S-thm:sat}, Corollary~\ref{S-cor:total}) are asserted under
exogenously scheduled publication: the owner's schedule is carried in
the demand factor, like the typed stream (\ref{S-a:exo}), not chosen as
a function of protocol outcomes. Adaptive republication reacting to
banked or dispute outcomes is covered elsewhere: by the deterministic
caps (the $\Emax$-rows and the pathwise clause of
Lemma~\ref{S-lem:budget}(iv), which hold for every schedule), and, on
the safety side, by the anchor-keyed counting of
Theorem~\ref{S-thm:panel} and Corollary~\ref{S-cor:submission}, which
quantifies over adaptive strategies.
\item Promotion threshold $N$ per tool (weighted: distinct endorsers in
$[W/\wmax,W/\wmin]$, changing constants not exponents); panel cap $k$; credited endorsements per (entry, tool) are capped at
$\Ncap=\Theta(N)$ (Definition~\ref{S-def:counting}(f)).
\item A \emph{confirmation} is an accepting gate decision
(convention~(\ref{S-conv:c6})); rejection effort is bounded separately
(\ref{S-n:effort}). Accepting decisions fall into three exhaustive
classes. A confirmation of the pair $(v,t)$ is \emph{required} if the
gate fires for the user's own use of $v$ at $t$ and the user holds no
local endorsed copy of $v$ whose scope admits $t$ (for the pseudo-target: no local readable $\E$-level copy of $v$ at all --- reasoning coverage, Rule~\ref{S-r:panel}(i)).
\emph{Required-approval clause (a reference-model coupling).} The
reference model is read with attempted use coupled to latent demand:
a user's own attempted use of $v$ at $t$ within the epoch --- any
event firing a gate for it, inside or outside a shared-tier presence,
covered and cache gates included --- occurs only on
$\{D_{u,v,t}=1\}$; equivalently, every such event realizes the
user's latent indicator for the epoch. This is a modeling
stipulation on the demand factor (\S\ref{S-app:prob}), not a
consequence of independent sampling: it is the natural extension of
the existing convention that a realized shared-tier demand
presupposes attempted use. Every user making a required approval of a pair is
therefore a latent demander of it; the log-defined shared-tier demand
of Definition~\ref{S-def:demand} ($n^{\mathrm{log}}$) remains the
subclass realized during a shared-tier presence, with
$n^{\mathrm{log}}_{v,t}\le n_{v,t}$ pathwise. It is
\emph{voluntary} if it is recorded at an interpreter-issued invitation
surface (a re-entry notification or suspension prompt,
Definition~\ref{S-def:counting}, Rule~\ref{S-r:panel}(vi)), declinable at
zero required cost; voluntary confirmations exist \emph{only} through
such surfaces, deduplicated jointly at one invitation per identity per
item per epoch. It is \emph{redundant} if the gate fires for the
user's own use while an in-scope local copy is held --- possible only
through handle-expressed reuse (Rule~\ref{S-r:whitelist}(c)) or a reasoning-target confirmation of locally readable bytes (reasoning coverage, Rule~\ref{S-r:panel}(i)), a
\emph{covered gate} --- in which case the decision is protocol-inert
on both branches (Rule~\ref{S-r:panel}(i)): no promotion weight, no
freshness consumption, no dispute initiation, no verdict. Redundant
confirmations are excluded from every count below and disclosed as
attention cost in \ref{S-n:effort}. $C_u$ counts user $u$'s required and
voluntary shareable confirmations; $\bar C_M=\frac1M\sum_u C_u$. The
cap (Theorem~\ref{S-thm:floor}) is stated over required confirmations;
the ledger (Lemma~\ref{S-lem:budget}) over required and voluntary.
\item \emph{Epoch vs.\ lifetime masses.} $p_{v,t}$ is a per-epoch
probability. The lifetime mass is
$d^{\infty}_{\mathrm{tool}}:=\sum_{v,t}p^{\infty}_{v,t}$ with
$p^{\infty}_{v,t}$ the probability that a user \emph{ever} demands the
pair; single-epoch and lifetime expected bounds are stated against the
matching mass and never conflated (Theorem~\ref{S-thm:floor}(ii)--(iii)).
Disputes add the ledger rows of Lemma~\ref{S-lem:budget}; revocation
adds Corollary~\ref{S-cor:floorE} (re-verification mechanics:
\ref{S-n:revoke}).
\item Saturation: $V(M):=\Ex|\{v:n_v\ge1\}|=o(M)$; concretely Heaps'
form $V(M)=\Theta(M^{\beta})$, $\beta\in(0,1)$ \cite{newman2005power}.
\end{itemize}

\begin{assumptionx}[Temporal smoothness, fractional form]\label{S-a:smooth}
There are a constant $\szero\ge1$ and a front-loading bound
$\gbar=\gbar(M)\in(0,1]$ such that, for every pair $(v,t)$ demanded in
the epoch: the number of distinct demanders of $(v,t)$ whose first
demand of the pair falls within any single finalisation window (length
$\Twin$) or any single \emph{candidate-dormancy} interval (an
inter-window span carrying no live community entry, of length
$\le\Texp$ by the retirement clock; \emph{stable promoted intervals}
--- a live, unsuspended $\CC$ presence with no open window --- are
exempt from both clauses: demand there is fast-pathed or
ledger-residual, never bucketed, Lemma~\ref{S-lem:budget}) is at most
$\gbar\,n_v+\szero$; and likewise the number of distinct identities whose \emph{first
audit-recorded decision on the pair} $(v,t)$ in the epoch
(\S\ref{S-app:prob}: display-transcript decisions at non-covered gates,
cache gates on \emph{both} branches included --- the accepting branch
is audit-recorded though never federation-logged,
Definition~\ref{S-def:counting}(iii)) falls within any single such
window or interval is at most $\gbar\,n_v+\szero$ (the
\emph{first-pair-decision clause}).
Demander counts are \emph{realized} shared-tier demand events
($n^{\mathrm{log}}$, \S\ref{S-app:prob}); decider counts are audit-recorded first pair-decisions (a class containing every federation-logged first pair-decision), whose deciders are latent demanders (required-approval clause, \S\ref{S-app:budgetmodel}) but need
not be realized log-demanders. Both are capped against the item's
\emph{latent} demand $n_v$ --- the weaker denominator, since
$n^{\mathrm{log}}_v\le n_v$ and every decider is a latent demander.
\end{assumptionx}

\begin{commentary}[reading and validation of \ref{S-a:smooth}]
The cap is relative to the item's own demand, never an absolute count:
it forbids \emph{front-loading} (demand concentrating inside one
window), not popularity. For demand arriving steadily over an epoch of
length $L$, $\gbar\approx\Twin/L$ for every item, independently of
$n_v$ and of $M$; the additive $\szero$ makes the cap satisfiable for
items of tiny demand. \ref{S-a:smooth} is a hindsight property of
completed epochs, used only in the analysis: no protocol rule reads,
estimates, or depends on $\gbar$. Its violation is item-local and
graceful: an item whose demand spikes into one window
($\gamma_v\approx1$) simply fails to amortise: each of its demanders
pays the one required confirmation the pathwise cap (Theorem~\ref{S-thm:floor}(i)) already
bounds, while every other item's accounting
is untouched (\ref{S-q:spike}). Demand-paced panels need no such cap:
panels accumulate verdicts across windows, so \ref{S-a:smooth}
constrains promotion finalisation windows and dormant intervals only;
the dormant-interval clause is the stronger half ($\Texp\gg\Twin$
typically), kept plausible by the dormancy-retirement clock of
Rule~\ref{S-r:hygiene}(i) and by its restriction to \emph{candidate}
dormancy: a promoted item's demand, however concentrated, is exempt
from both clauses (its cost is fast-pathed or ledger-residual,
Lemma~\ref{S-lem:budget}). Validation: per-item realised fractions
$\gamma_v$, checked specifically on the most-demanded items
(\ref{S-n:validate}). The first-pair-decision clause has the same character and the same
validation path: for demand arriving steadily, first pair-decisions
track first pair-demands; both realised fractions are logged per item
(\ref{S-n:validate}). The audit-recorded class is validated the same
way, directly from display-transcript logs: both branches of every
gate, cache gates included, are recorded there
(Rule~\ref{S-r:display}(e)), so realised first-pair-decision fractions
are observable without reference to the federation log. Its pair-level indexing is load-bearing: it is
what makes the dispute-initiation classification of
Lemma~\ref{S-lem:budget} exhaustive, since a cross-tool repeat decision
(a prior acceptance at another pair) and a decision postponed past an
abandoned dialog are each the identity's \emph{first} decision at
their own pair, counted where they fall. The pair-level form is the item-level cap read tool by tool, still
stated against the item's \emph{total} demand $n_v$, the weaker and more
plausible denominator; steady demand gives $\gbar\approx\Twin/L$ per
pair exactly as per item, and the same per-item logging of realised
fractions $\gamma_v$ validates it (\ref{S-n:validate}).
\end{commentary}

\begin{assumptionx}[Coverage sufficiency, operational form]
\label{S-a:promsuf}
Adopt the presence taxonomy and charging convention of
Lemma~\ref{S-lem:budget}: within an epoch, an item's shared-tier
presence partitions into entry windows, candidate-dormancy intervals,
and freeze/suspension intervals (the \emph{ledger-charged} pieces),
and \emph{stable promoted intervals} (a live, unsuspended community
entry with no open entry window). For a pair $(v,t)$ and an epoch, let
$U_{v,t}$ count the epoch's \emph{residual} required pair-approvals of $(v,t)$: those charged to no ledger row (option-induced cache confirmations excluded: floor-charged, Corollary~\ref{S-cor:flooropt}) --- a
subclass of the required confirmations by demanders whose epoch-first
demand of the item falls in a stable promoted interval or outside
every shared-tier presence (a credited such decision is
weight-row-charged instead, Lemma~\ref{S-lem:budget}, only shrinking
the residual) (before first publication, after
retirement, or after demotion; gates fired during a suspension are
freeze-bucket events, ledger-charged, never residual). Each belongs
to a distinct demander of the pair (\ref{S-inv:floor}), so
$U_{v,t}\le n_{v,t}$ pathwise. The residual class is
\emph{operational}, keyed to gate behaviour, never to certificate
existence: a live $\CC$ entry naming $(v,t)$ fast-paths only
\emph{literal}-expressed use (Rule~\ref{S-r:whitelist}(a)); a
handle-expressed use of the same pair gates at every tool
(Rule~\ref{S-r:whitelist}(c)) and, absent a local in-scope copy, is
required --- and, in a stable promoted interval, residual. There are
$D_0(M)\ge0$ and $\varepsilon(M)\to0$ such that, for every
\emph{benign} (Definition~\ref{S-def:bad}) shareable item $v\in\Vals$,
every target $t$, and every $n\ge1$, conditional on $n_{v,t}=n$:
\[
\Ex\big[U_{v,t}\mid n_{v,t}=n\big]\ \le\ D_0(M)+\varepsilon(M)\,n .
\]
Like \ref{S-a:smooth}, this is a hindsight property of completed epochs,
read against latent demand (\S\ref{S-app:prob}); no protocol rule reads,
estimates, or depends on it, and it is trivially satisfied whenever
$n\le D_0(M)$ (since $U_{v,t}\le n_{v,t}$).
The growth of $D_0$ is consumed only through the \emph{product
condition} $D_0(M)\,V(M)=o(M)$ of Theorem~\ref{S-thm:sat}(iii), a joint
property of the coverage threshold and the saturation law, never of
either alone; under the Heaps form $V(M)=\Theta(M^{\beta})$
(\S\ref{S-app:budgetmodel}), the sufficient rate is
$D_0(M)=o(M^{1-\beta})$, the form used in the corridor
(\S\ref{S-app:corridor}) and in the sufficiency sketch below.
It bounds jointly:
(i)~\emph{publication coverage} --- a demanded benign item its owner
never publishes has every gate-firing accepting demand residual, and
satisfies the bound only if that demand is small;
(ii)~\emph{promotion coverage} --- demand at not-yet-promoted pairs
arriving outside every ledger-charged piece; and
(iii)~\emph{representation coverage} --- demand for live-$\CC$ pairs
arriving handle-expressed (untrusted-channel arrival of
already-certified bytes), which gates and pays despite the
certificate (\ref{S-q:handlecover}). An item or arrival pattern
violating it never amortises: each residual demander pays the one
required confirmation the pathwise cap (Theorem~\ref{S-thm:floor}(i)) already bounds,
item-locally and gracefully, entering
the residual $r^{(e)}_{\mathrm{bad}}$ of Corollary~\ref{S-cor:total}.
\end{assumptionx}

\paragraph{Heuristic sufficiency and compatibility.}
The following sufficiency sketch is heuristic --- a plausibility
argument, not a derivation: it does not formally compose intervening freezes, wrongful evictions,
the entry cap, the pool-composition requirements, or the
arrival-representation mix into a proof of \ref{S-a:promsuf}, which remains an
assumption; the compatibility calculation closing the paragraph is
exact. Heuristically sufficient, for a benign item and any fixed
$\eta>0$: the owner publishes the item before more than an
$\varepsilon$-fraction of each demanded pair's demand arrives and
resubmits promptly after each unpromoted window (dormant intervals
short of the retirement clock, within the per-epoch entry cap of
Rule~\ref{S-r:hygiene}(i)); some entry, counting the banked credits of
its preceding interval (Definition~\ref{S-def:counting}(b)), accrues at
least $(1+\eta)N/p_b$ fresh byte-first demanders (a positive byte-first fraction per demanded tool --- enough
demanders whose first decision on the bytes occurs at that tool ---
is one empirical mechanism supporting this cohort's existence; it is
a heuristic diagnostic for \ref{S-a:promsuf}, not a hypothesis of any
theorem) whose demand spans at
least $\Dpool$ pools with at least $\lceil N/\Dpool\rceil$ accepting
demanders available per pool (so the per-pool credit caps of
Definition~\ref{S-def:counting}(f) do not bind below $N$), almost all of
them authority-active (Definition~\ref{S-def:counting}(a): honest
identities are authority-inactive only in the rare calibrated states
of \cref{S-n:calib}, folded into the same $\varepsilon$) --- a
demand-composition property of the same character as
\ref{S-a:rhoseat}'s structural half, validated by the same
pool-stratified logging (\ref{S-n:validate}); and demotion of promoted
benign entries is exponentially rare (Proposition~\ref{S-prop:live}), so promoted status, once reached,
persists; and, for the representation clause, demand for promoted
pairs predominantly arrives fast-path-expressed --- through trusted
retrieval of the live entry (Definition~\ref{S-def:fedconfig},
Rule~\ref{S-r:hygiene}(ii)) or reuse of local readable copies --- with
the handle-arrival fraction (\ref{S-q:handlecover}) folded into the
same $\varepsilon$: an arrival-channel property of the demand stream,
logged directly (\ref{S-n:validate}), never derived from the protocol. For schedules with
$N(M)\to\infty$ --- every corridor schedule, \S\ref{S-app:corridor} ---
the accepting count then crosses $N$ except with probability
$e^{-2\eta'^2N(M)/(1+\eta)^2}\to0$ by Hoeffding's lower tail, for a
margin $\eta'$ depending on $\eta$ and $p_b$, giving
$\varepsilon(M)\to0$; at \emph{fixed} $N$ the same cohort yields only
a fixed failure probability, so $\varepsilon(M)\to0$ under fixed
protocol constants requires instead that available cohorts grow
beyond a fixed multiple of $N$, which \ref{S-a:promsuf} then asserts
rather than derives; intervening disputes are absorbed by the
vindication exemption of Rule~\ref{S-r:hygiene}(i)(b)
(\ref{S-n:vindexempt}) and resolve at the item's own demand rate,
freezes excluding themselves from the window clock
(Rule~\ref{S-r:panel}(iv)). \emph{Consistency with \ref{S-a:smooth}}:
under the fractional cap the two assumptions do not constrain one
quantity from opposite sides. A crossing window is permitted to
contain $(1+\eta)N/p_b$ demanders whenever
$\gbar\,n_v+\szero\ge(1+\eta)N/p_b$, i.e.\ whenever
$n_v\ge\big((1+\eta)N/p_b-\szero\big)/\gbar$; $D_0(M)$ may therefore
be taken of order $N(M)/\gbar(M)$, and the product condition's Heaps form
$D_0(M)=o(M^{1-\beta})$ (Theorem~\ref{S-thm:sat}(iii)) becomes
$N(M)/\gbar(M)=o(M^{1-\beta})$, satisfied by every corridor schedule
(\S\ref{S-app:corridor}).

\begin{commentary}[scoping, representation dependence, and validation
of \ref{S-a:promsuf}]
\emph{Benign scoping is deliberate}: poisoned items are excluded;
their promotion probability is bounded above by
Theorem~\ref{S-thm:panel}, and their residual demand is charged at the
single-user cap by design, each such confirmation being a personal
gate inspection of an unpromoted or flagged item, exactly the
inspection the design intends. \emph{The operational (no-row) form is
load-bearing}: an earlier draft keyed coverage to certificate
existence, under which a handle-expressed use of a live-$\CC$ pair
counted as covered while its gate --- required under
Rule~\ref{S-r:whitelist}(c) --- was charged to no row and to no
residual; the operational form charges every required confirmation
somewhere by construction (Lemma~\ref{S-lem:budget}, charging
convention), and names exactly the demand the amortisation story must
cover. \emph{Representation dependence is the disclosed price}:
community certification amortises the flows that can take the fast
path --- literal-expressed reuse of promoted bytes, in particular
bytes obtained through trusted retrieval (\textsc{Retrieve} returns
live entries as readable $\E$-bytes, Definition~\ref{S-def:fedconfig},
Rule~\ref{S-r:hygiene}(ii), which then byte-match the live entry,
Rule~\ref{S-r:whitelist}(a)) --- and never the handle-expressed
arrivals, which gate by design (\ref{S-q:handlecover});
$\varepsilon(M)$ must absorb the handle-arrival fraction of certified
demand, so the vanishing of the residual is a joint property of
publication timing, promotion speed, \emph{and} arrival
representation. Failure is graceful and item-local: residual demand
is priced at the cap (Theorem~\ref{S-thm:floor}(i)), and the epoch's
shared required load never exceeds the standalone mass
$d_{\mathrm{tool}}$ in any case (Theorem~\ref{S-thm:floor}(ii)).
Incentive sensitivity is the disclosed residual \ref{S-q:incentive}.
Under permanent freshness the condition weakens in one direction and
is explicitly priced in the other: banked credits accumulate across
epoch boundaries toward the next entry, so qualifying demand need not
concentrate within one epoch; conversely, endorsers consumed by a
failed entry never renew, which the identity-disjoint counting
already prices (Lemma~\ref{S-lem:disjoint}). Validation: log each
item's realized residual fraction $U_{v,t}/n_{v,t}$ \emph{and} the
fast-path-vs-gate split of live-$\CC$ demand (the handle-arrival
fraction), stratified on the most-demanded items, together with
per-item time-to-promotion against accumulated demand and the
per-epoch demanded-but-unpublished mass (\ref{S-n:validate}).
\end{commentary}

\begin{assumptionx}[Bounded tool-spread]\label{S-a:toolspread}
Each shareable item is demanded at no more than $\ctool=\bigO(1)$
distinct tools over the epoch: at most $\ctool$ tools have
$n_{v,t}\ge1$, and hence $\sum_t n_{v,t}\le\ctool\,n_v$
(latent reference-model counts, \S\ref{S-app:budgetmodel}; the
realized counts of Definition~\ref{S-def:demand} inherit the bound
pathwise: $n^{\mathrm{log}}_{v,t}\le n_{v,t}$ places the realized
tool support inside the latent tool support, so at most $\ctool$
tools have $n^{\mathrm{log}}_{v,t}\ge1$, and every realized
pair-demander is a realized item-demander, so
$n^{\mathrm{log}}_{v,t}\le n^{\mathrm{log}}_v$
(Definition~\ref{S-def:demand}); hence
$\sum_t n^{\mathrm{log}}_{v,t}\le\ctool\,n^{\mathrm{log}}_v$
pathwise). In particular each user demands at
most $\ctool$ pairs per item, and $d_{\mathrm{tool}}\le\ctool\,d$
(the $\le\ctool$ demanded targets per item include
$\mathsf{reasoning}$ where demanded; constants unchanged).
\end{assumptionx}

\begin{commentary}[reading and validation of \ref{S-a:toolspread}]
The bound formalises the observation that a published item typically
addresses the scenario its owner used it in: endorsement, promotion,
and scope are all per (value, tool); it licenses the per-tool split
in Theorem~\ref{S-thm:sat}(iii). Validation: per-item distinct-tool
counts (\ref{S-n:validate}). The \emph{byte-first density} of a
demanded tool's stream --- the fraction of its demanders whose first
decision on the bytes occurs at that tool --- is deliberately
\emph{not} part of this assumption: no proof consumes it, and its
role is heuristic support for the promotion-coverage clause of
\ref{S-a:promsuf} (see the heuristic-sufficiency paragraph there).
\end{commentary}

\begin{assumptionx}[Bounded per-user demand mass]\label{S-a:massbound}
Along the population scaling, the single-epoch per-user demand mass
is uniformly bounded: $\sup_{M}d(M)<\infty$, where
$d(M)=\sum_v p_v=M^{-1}\sum_u\sum_v\Prob[\text{$u$ demands $v$}]$ is
the population-average number of distinct shareable items one user
demands per epoch (\S\ref{S-app:budgetmodel}); hence
$\sup_M d_{\mathrm{tool}}(M)<\infty$ under \ref{S-a:toolspread},
and $d_{\mathrm{good}}\le d$ throughout. Where \ref{S-a:massbound} is
not adopted, every vanishing ($\to0$) budget conclusion below holds
verbatim with its rate condition read against the mass:
$\gbar(M)\,(1+\kappa(M))\,d_{\mathrm{good}}(M)\to0$ in
Theorem~\ref{S-thm:sat}(i)--(ii) and Theorem~\ref{S-thm:coscale}(iii),
and $\varepsilon(M)\,d_{\mathrm{good}}(M)\to0$ in
Theorem~\ref{S-thm:sat}(iii) (whose $D_0(M)V(M)/M$ term is
independent of the per-user mass and unchanged). The
relative bounds (cost at most the front-loading multiple
$\Bfl\,\gbar$ of the standalone mass, Theorem~\ref{S-thm:sat}(i),
first clause) and the cap (Theorem~\ref{S-thm:floor}) consume no
bound on $d$ in any case.
\end{assumptionx}

\begin{commentary}[reading and validation of \ref{S-a:massbound}]
The assumption is load-bearing and deliberately explicit: without
it, the vanishing conclusions would fail --- a population-indexed
demand law with $d(M)\to\infty$ (each user demanding ever more
distinct shared items as the community grows) satisfies every
remaining hypothesis while the front-loading term
$\Bfl\,\gbar\,d$ grows --- and pointwise finiteness of $d$ at each
$M$ does not supply the uniform bound the limits consume. It is an
individual-usage property, not a population property: under any
fixed per-user demand law the masses are constants in $M$ and
\ref{S-a:massbound} is automatic (the Zipf law of \S\ref{S-app:zipf}
has $d=\sum_r p_r=1$ identically), and the Heaps form is standardly
read over a corpus accumulated from bounded individual streams
\cite{l2010zipf}, so the saturation hypothesis already carries
this reading. The owner-coupling convention cannot violate it:
coupled indicators number at most $\rsub$ per user per epoch
(Rule~\ref{S-r:share}). The disclosed failure mode is demand growth
induced by the shared tier itself --- a richer library drawing each
user to more shared items per epoch --- a network effect of the
same character as the incentive feedback of \ref{S-q:incentive};
violation is graceful: the relative bounds and the pathwise
cap (Theorem~\ref{S-thm:floor}(i)) are unaffected, and the general-form rate conditions of
\ref{S-a:massbound} price the growth explicitly. Validation: log
per-user distinct shareable-item demand per epoch across deployment
growth (\ref{S-n:validate}).
\end{commentary}
\subsection{The single-user first-authorization cap}
\label{S-app:floor}

The cap's core is a counting identity. Its pathwise clause,
Theorem~\ref{S-thm:floor}(i), rests on the harness rules and
\ref{S-inv:floor} alone: no ledger row, no behavioural parameter, no
demand model, and no assumption of this appendix. Its expected
clauses restate that identity against the reference demand model's
masses (\S\ref{S-app:budgetmodel}), still using no assumption; the
single assumption-bearing statement in this subsection is the mass
comparison of Corollary~\ref{S-cor:floorspread}
(\ref{S-a:toolspread}). The cap is therefore stated first. Its
dependency chain is Lemma~\ref{S-lem:scopepersist} $\to$
\ref{S-inv:floor} $\to$ Theorem~\ref{S-thm:floor} $\to$
Lemma~\ref{S-lem:budget}, strictly in that order.

\begin{lemma}[Local scope persistence]\label{S-lem:scopepersist}
Fix a user $u$, a value $v$, and a tool $t$, and suppose $u$'s own
accepting review of $(v,t)$ has occurred at any gate transaction ---
including a transaction another of whose reviews aborted the enclosing
invocation (Rule~\ref{S-r:atomic}: endorsement is durable). Then in
every later configuration, unless $u$'s own agent has evicted the
block (\ref{S-n:evict}), $u$'s store contains a $\C$-block with bytes $v$ and
$t\in\scopef$, and every use of $v$ at $t$ by $u$ expressed as a
readable literal takes the fast path of Rule~\ref{S-r:whitelist}(a); a
use expressed through a handle gates (Rule~\ref{S-r:whitelist}(c)) as a
covered gate, protocol-inert and redundant (Rule~\ref{S-r:panel}(i),
\S\ref{S-app:budgetmodel}). In particular no later gate event of $u$ for
$(v,t)$ is \emph{required}. (Option~\ref{S-o:three} is requirements-level and absent from the
formal semantics; a conforming instantiation would leave the block
in place and read the authorization and coverage tests over
currently authorizing witnesses, \ref{S-o:three}(a),(c)--(d).) For the pseudo-target: once $u$ holds \emph{any} local readable
$\E$-level copy of the bytes (via \textsc{Recall-E},
Definition~\ref{S-def:extconfig}), a later $\conf$ on those bytes at
$\mathsf{reasoning}$ is a covered gate by the reasoning-coverage
clause of Rule~\ref{S-r:panel}(i), protocol-inert and redundant; hence
no later gate event of $u$ for $(v,\mathsf{reasoning})$ is required.
\end{lemma}

\begin{proof}
By Rule~\ref{S-r:atomic} and Definition~\ref{S-def:sinkconf}, every
accepting review of $(v,t)$ writes a fresh $\C$-block with bytes $v$
and $t\in\scopef$ at the close of its atomic transaction, on the
executing and the aborting branch alike (\textsc{Conf-Acc},
\textsc{Tool-Gate-Acc}, \textsc{Tool-Gate-Rej};
Figures~\ref{S-fig:sos}--\ref{S-fig:sos2}, convention~\hyperref[S-sos:vii]{(vii)}). By
\ref{S-a:label}--\ref{S-a:persist} blocks are immutable once written, so
neither this block's label nor its scope ever changes; local stores
mutate only through owner-invoked $\Sigma$-operations and owner
confirmations, and no protocol event initiated by any other identity
alters or removes any block in $u$'s local store
(Lemma~\ref{S-lem:nodelete}). The only removal path is first-party eviction (\ref{S-n:evict}). Rule~\ref{S-r:whitelist}(a) then byte-matches
$v$ against the stored block with $t\in\scopef$. For the pseudo-target: any local $\E$-level copy of the bytes is
readable (\textsc{Recall-E}, Definition~\ref{S-def:extconfig}) and
persists by \ref{S-a:label}--\ref{S-a:persist} and
Lemma~\ref{S-lem:nodelete}, the first-party eviction exception aside;
the reasoning-coverage clause of Rule~\ref{S-r:panel}(i) then classifies
every later reasoning-target gate on those bytes as covered. Its
owner-side test reads exactly the readability fact
Rule~\ref{S-r:readable} establishes: readability is the entirety of
what an accepting reasoning confirmation buys, so the clause covers
precisely the gates whose acceptance would change nothing locally.
\end{proof}

\begin{theorem}[Single-user first-authorization cap]\label{S-thm:floor}
The cap counts \emph{required user approvals} --- first
authorizations of reusable (value, tool) pairs --- and nothing else:
not rejecting decisions, abandoned or repeated dialogs, covered-gate
(redundant) acceptances, batch-page attention, or per-task
confirmations (classes: \S\ref{S-app:budgetmodel}; effort account:
\ref{S-n:effort}). The superscript $\mathrm{forced}$ below denotes
this required-approval count.
Suppose the owner's agent does not evict the relevant local blocks
(the disclosed first-party exception of \ref{S-inv:floor});
Option~\ref{S-o:three} is requirements-level and absent from the
formal semantics (\ref{S-o:three}), so no lease expiry exists in the
proved system.
\begin{enumerate}[label=(\roman*),leftmargin=2em,itemsep=2pt,topsep=2pt]
\item \emph{(Pathwise lifetime invariant.)} Pathwise and independent of every federated and demand-model assumption, subject to the stated first-party persistence exceptions: under any
dispute, rejection, revocation, or availability campaign whatsoever,
no user is required to confirm any given (value, tool) pair more than
once over the deployment's lifetime:
$C^{\mathrm{forced}}_{u,v,t}\le1$ pathwise, counting required
pair-approvals (Definition~\ref{S-def:sinkconf}, counting unit).
This is \ref{S-inv:floor}, established by
Lemma~\ref{S-lem:scopepersist}. Rejecting decisions, abandoned dialogs,
repeated within-transaction reviews, redundant confirmations
(\S\ref{S-app:budgetmodel}), and inspection time are a distinct
resource, disclosed and partially bounded in \ref{S-n:effort}, and are
not bounded by this clause.
\item \emph{(Expected cap, single epoch, reference model.)} In the
independent reference model (\S\ref{S-app:budgetmodel}), the two
components are bounded on their own horizons:
$\Ex[C^{\mathrm{forced},\infty}_{\mathrm{personal}}]\le\dpriv$
(lifetime) and
$\Ex[\bar C^{\mathrm{forced},(e)}_{\mathrm{share}}]\le d_{\mathrm{tool}}$
(per epoch $e$); the componentwise sum $\dpriv+d_{\mathrm{tool}}$ is
a valid upper bound for their sum on any single epoch read jointly
with the lifetime personal term, not a homogeneous single-horizon
cost. The bound
$d_{\mathrm{tool}}$ is exactly the expected shareable cost a single
user incurs in isolation, where every gate necessarily fires: no user
is ever pushed above the single-user cost. The comparison to the
item-level mass, $d_{\mathrm{tool}}\le\ctool\,d$, is
Corollary~\ref{S-cor:floorspread} and is this theorem's only
assumption-bearing companion.
\item \emph{(Expected cap, lifetime.)} Over the deployment's
lifetime, expected required shareable cost is at most
$d^{\infty}_{\mathrm{tool}}=\sum_{v,t}p^{\infty}_{v,t}$; the
single-epoch mass bounds the lifetime cost only when it is itself read
as the lifetime mass (\S\ref{S-app:budgetmodel}).
\item \emph{(Monotonicity.)} At a fixed realised workload with no
disputes, adding users never increases any individual's required count:
promotions only advance, and a promotion of $(v,t)$ removes the gate
for every later demander of that pair (Rule~\ref{S-r:whitelist}). With
disputes, an added identity can suspend amortisation, forcing later
demanders to confirm personally; the pathwise invariant (i) still caps
every user at once per pair.
\end{enumerate}

\end{theorem}

\begin{proof}
(i) By Lemma~\ref{S-lem:scopepersist}: $u$'s first accepting review of
$(v,t)$ writes the covering local copy durably, whatever the enclosing
transaction's outcome, so no later gate event of $u$ for $(v,t)$ is
required (a literal-expressed use fast-paths; a handle-expressed use is
a covered gate, redundant by definition); within that first
transaction the pair is counted once (Definition~\ref{S-def:sinkconf});
declining voluntary invitations costs nothing
(Definition~\ref{S-def:counting}). (ii) By (i), each required confirmation
of $(v,t)$ belongs to a distinct demander of the pair (realized, hence latent, \S\ref{S-app:prob}), so the
population's required total for $(v,t)$ is at most $n_{v,t}$, with
$\Ex[n_{v,t}]=Mp_{v,t}$ in the reference model; summing over pairs and dividing by $M$ gives
$d_{\mathrm{tool}}$; personal pairs add $\dpriv$, once each (\ref{S-inv:floor}). (iii) The
pathwise clause (i) is lifetime; taking expectations over lifetime
pair-demand indicators gives $d^{\infty}_{\mathrm{tool}}$. (iv)
Immediate from Rule~\ref{S-r:whitelist} and (i).
\end{proof}

\begin{corollary}[The cap under bounded tool-spread]\label{S-cor:floorspread}
Under \ref{S-a:toolspread}, $d_{\mathrm{tool}}\le\ctool\,d$; hence
the shareable cap of Theorem~\ref{S-thm:floor}(ii) is at most
$\ctool\,d$ per epoch (componentwise sum $\dpriv+\ctool\,d$, horizons per Theorem~\ref{S-thm:floor}(ii)). The lifetime
analogue, $d^{\infty}_{\mathrm{tool}}\le\ctool\,d^{\infty}$ with
$d^{\infty}:=\sum_v p^{\infty}_v$, holds when \ref{S-a:toolspread}
is read over lifetime demand (at most $\ctool$ distinct tools per
item ever), a strictly stronger reading than the per-epoch form and
asserted only here. This corollary restates the cap against the
coarser item-level mass; no other result consumes it.
\end{corollary}

\begin{proof}
A user demanding the pair $(v,t)$ demands the item $v$
(Definition~\ref{S-def:demand}), and under \ref{S-a:toolspread} at
most $\ctool$ tools per item carry demand, so
$\sum_t p_{v,t}\le\ctool\,p_v$ per item; sum over items. The lifetime
form is identical under the lifetime reading.
\end{proof}

\begin{remark}[Voluntary effort and the cap]\label{S-rem:voluntary}
The cap is stated over required confirmations. Voluntary community
effort (a repeat re-inspection after a dispute or re-entry
notification) is declinable at zero personal cost, carries no
promotion weight (\ref{S-inv:fresh}), and is counted in the budget
(Lemma~\ref{S-lem:distinct}), never in the cap.
\end{remark}

\begin{corollary}[The cap under revocation]\label{S-cor:floorE}
Post-promotion suspensions and re-verification (\ref{S-n:revoke}) do not
multiply required cost: a user who relied on ungated $\CC$ status at $t$
and loses it on demotion or suspension owes at most their first-ever confirmation of
$(v,t)$, already counted inside $d^{\infty}_{\mathrm{tool}}$, after
which their local $\C$ copy covers all reuse at $t$ for the deployment's
lifetime (Lemma~\ref{S-lem:scopepersist}). Hence expected total required
cost is at most $\dpriv+d^{\infty}_{\mathrm{tool}}$, lifetime,
independent of the number of epochs. A multiplicative factor $\Elease$ would arise only under a
conforming instantiation of the requirements-level
Option~\ref{S-o:three} --- an extension-level claim, stated and to be
proved with that extension (\ref{S-o:three}(e)), outside this
corollary.
\end{corollary}

\begin{proof}
Immediate from Theorem~\ref{S-thm:floor}(i),(iii): required confirmations
are once per (user, value, tool) over the lifetime; demotion,
suspension, and re-verification never touch local $\C$ copies
(Lemma~\ref{S-lem:nodelete}). Option~\ref{S-o:three} is absent from the formal semantics
(\ref{S-o:three}).
\end{proof}

\begin{corollary}[Budget under Options \ref{S-o:one}--\ref{S-o:two}]\label{S-cor:flooropt}
Option-induced cache confirmations
(Options~\ref{S-o:one}--\ref{S-o:two}) belong to no pipeline ledger row
and to no residual (Lemma~\ref{S-lem:budget}, exclusions): they are
accept-inert for the shared tier
(Definition~\ref{S-def:counting}(iii)) and are accounted in two
classes. \emph{(Per-task.)} \ref{S-o:two}'s confirmations of freshly
composed data-tool arguments lie in the per-task stream, outside the
reusable-memory accounting entirely, by the scope hypothesis of
Proposition~\ref{S-prop:harden}. \emph{(Floor-charged.)} All remaining
option-induced cache confirmations --- \ref{S-o:one}'s first-use
sensitive-sink confirmations of $\T$-values and \ref{S-o:two}'s
confirmations of endorsed-value reuse at data tools --- are required
pair-confirmations of the confirming user, charged to the floor
component: each is at most one per (user, value, sensitive sink)
resp.\ (user, value, data tool) (scope-caching,
Lemma~\ref{S-lem:scopepersist}, \ref{S-inv:floor}), an $\bigO(1)$
enlargement of the per-item pair count under the sink-spread and
data-tool-spread hypotheses of Proposition~\ref{S-prop:harden} ---
folded exactly as \ref{S-a:toolspread} folds tool pairs. The
exhaustiveness of the charging convention (Lemma~\ref{S-lem:budget})
is unaffected: option-induced confirmations are charged to the
per-task stream or the floor, never to any pipeline row or residual.
\end{corollary}

\begin{proof}
Immediate from Proposition~\ref{S-prop:harden}(iii),
Definition~\ref{S-def:counting}(iii), and \ref{S-inv:floor}: the fresh
cache copy scope-caches (Lemma~\ref{S-lem:scopepersist}), so each
(user, value, target) triple is required at most once; the spread
hypotheses bound the targets per value by $\bigO(1)$; and the
freshly-composed class is per-task by hypothesis.
\end{proof}
\subsection{The per-item confirmation ledger}
\label{S-app:ledger}

\begin{lemma}[Per-item confirmation ledger, $\bigO$-form]\label{S-lem:budget}
Fix an item $v$ and an epoch; adopt the default bars of
Rule~\ref{S-r:panel}(iv), \ref{S-a:rhoseat}, \ref{S-a:smooth},
\ref{S-a:toolspread}, and, for the bucket rows, that each
honest inspection, fresh or repeat, accepts a benign item with
conditional probability at least $p_b$ given all prior protocol events
(for a repeat inspection of bytes the inspector previously accepted, the
bound is if anything the more plausible).
Adopt also the standing exception of Theorem~\ref{S-thm:floor} (no
owner eviction of the relevant local blocks; Option~\ref{S-o:three} is
requirements-level and absent from the formal semantics,
\ref{S-o:three}).
Option-induced cache confirmations (Options~\ref{S-o:one}--\ref{S-o:two})
are excluded throughout --- from $C^{\mathrm{pipe}}_v$, from the
residual of \ref{S-a:promsuf}, and from the pathwise count $C_v$ of
clause~(iv): they are charged to the floor component, with the
$\bigO(1)$ spread factors of Corollary~\ref{S-cor:flooropt}.
Let $\Eslot:=\Emax+1$.
\emph{Presence taxonomy.} Within the epoch, the item's shared-tier
presence partitions into four classes of pieces: \emph{entry windows}
(at most $\Emax$, by the per-item entry cap of
Rule~\ref{S-r:hygiene}(i), counting promoted, evicted, and unpromoted
entries alike; windows are epoch-contained, convention~(\ref{S-conv:c4}));
\emph{freeze/suspension intervals} (the current-epoch disputes and
fast-track reviews, plus at most one carry-in instance --- the
byte-level freeze is exclusive, Rule~\ref{S-r:panel}(ii); a carried-in
instance is counted toward Rule~\ref{S-r:hygiene}(i) in its opening
epoch, convention~(\ref{S-conv:c4}), but its current-epoch demanders are charged
here, via the carry-in term of $\bar D_v$, and it issues no new
prompts, Rule~\ref{S-r:panel}(vi)); \emph{candidate-dormancy intervals}
(no open window, no live promotion at any tool, unfrozen); and
\emph{stable promoted intervals} (a live, unsuspended community entry
at some tool, no open window, unfrozen). Inter-window spans --- each
entry is followed by at most one, ending at the next entry,
retirement, or the epoch boundary, so at most $\Eslot$ begin in or
carry into the epoch --- have promotion status fixed at the preceding
window's close, so each span is candidate-dormant or stable-promoted
throughout, with its freeze/suspension pieces carved out;
\ref{S-a:smooth}'s dormant-interval clause is applied per
candidate-dormancy span, freeze-time demanders within a span being
charged to the freeze bucket. The first three classes are the
\emph{ledger-charged} pieces; stable promoted intervals are
deliberately \emph{not} bucketed: a matching literal fast-paths there
at the promoted tool (Rule~\ref{S-r:whitelist}(a)) at zero cost, while
the gates that do fire there --- handle-expressed uses of the
certified pair (Rule~\ref{S-r:whitelist}(c)) and uses at unpromoted
tools --- are charged to the \emph{ledger residual} below, bounded by
Theorem~\ref{S-thm:sat}(iii) under \ref{S-a:promsuf}. (A stable promoted
interval is ``window closed, not yet resubmitted or retired'' in the
sense of Definition~\ref{S-def:counting}(b): first decisions logged
there bank toward the item's next entry exactly as in candidate
dormancy; the taxonomy separates the classes for cost attribution
only and changes no protocol rule.) We use the common conservative
bound $\Eslot$ for the window- and dormancy-bucket piece counts.
(The attempt structure of Definition~\ref{S-def:attempt} partitions the
same presence; the ledger's buckets and bounds are unchanged by it.)
Write

\[
\bar D_v\ :=\ 2(1-p_b)\,\Eslot\,\ctool\big(\gbar\,n_v+\szero\big)
\;+\;(1-p_b)\,\Emax\ctool\Ncap\;+\;\cacc f\;+\;2\Emax\;+\;1\,,
\]

the final $+1$ is the carry-in instance; the invited-reconsideration
term keeps $\Emax$ because invitations are epoch-scoped
(Definition~\ref{S-def:counting}). $\bar D_v$ is a function of the
realised demand $n_v$: the bucket rows bound conditional expectations
given the demand realisation, and the split in (iii) follows by taking
the expectation over demand last, licensed by the demand construction
of \S\ref{S-app:prob}: the behavioural bounds are asserted uniformly
over demand realisations, so Lemma~\ref{S-lem:adaptive} applies under
the demand-conditional measure. For poisoned items no $p_b$ bound is
asserted: their entries are counted by Corollary~\ref{S-cor:submission}
and their residual demand is charged at the single-user cap
(Theorem~\ref{S-thm:floor}).

\emph{Pipeline component and ledger residual.} The rows cover the
\emph{pipeline component} $C^{\mathrm{pipe}}_v$: every confirmation of
a pair of $v$ charged to a ledger row under the convention below ---
credited weight-bearing endorsements, bucketed required confirmations
(window, candidate-dormancy, and freeze/suspension time, carry-in
included), and all invited-repeat, prompt, and publish confirmations
of $v$. Required confirmations charged to \emph{no} row are the
\emph{ledger residual}: those of demanders whose epoch-first
item-demand falls in a stable promoted interval (in particular
handle-expressed uses of a live-$\CC$ pair, which gate under
Rule~\ref{S-r:whitelist}(c) despite the certificate) or outside every
shared-tier presence (before first publication, after retirement, or
after demotion). The residual is bounded by
Theorem~\ref{S-thm:sat}(iii) under \ref{S-a:promsuf}; any overlap between
the two accountings only over-counts.

\emph{Charging convention (retrospective).} At the close of the
logging epoch every confirmation of $v$ --- a durable accepting
pair-confirmation, counted once per pair per transaction
(Definition~\ref{S-def:sinkconf}) --- is assigned to exactly one row or
to the ledger residual; the ledger is a hindsight accounting device,
like \ref{S-a:smooth}, never protocol state. The ledger \emph{rows} charge protocol-bearing decisions and invited
and publish events (Definition~\ref{S-def:counting}); the
\emph{residual} is defined over all required confirmations of the bytes,
protocol-bearing or not --- in particular outside-presence
confirmations, which are unlogged as candidate decisions --- so that
every required confirmation in this lemma's accounting universe (option-induced cache confirmations excluded, above) is charged somewhere, to a row or to the
residual, by construction. (Definition~\ref{S-def:demand}'s retirement
clause governs \emph{demand counting}, not the confirmation universe
of this convention.) Non-covered reasoning-target confirmations
satisfying Definition~\ref{S-def:counting} are protocol-bearing
(\S\ref{S-app:fedcore}) and are charged to the pair
$(v,\mathsf{reasoning})$ exactly as tool pairs are; covered reasoning
gates are inert (Rule~\ref{S-r:panel}(i)) and excluded like any
redundant confirmation. \emph{Credited}
weight-bearing endorsements (in-entry, banked-and-credited within the
epoch, or vindication-credited; required or invited alike) are assigned
to the weight row --- including a banked first decision taken in a
stable promoted interval, so a residual-class demander's credited
acceptance is weight-row-charged, only shrinking the residual;
\emph{uncredited invited} confirmations to their invitation row
(re-entry or prompt; invitations are epoch-scoped, so no invitation
row carries over); \emph{publish} confirmations to the publish row;
every other \emph{required} confirmation to the ledger-charged piece
(window, candidate-dormancy, or freeze/suspension --- current-epoch
instances and the at most one carry-in instance) containing its
demander's \emph{first demand of the item} in the epoch, at most
$\ctool$ confirmations per demander (one per pair, \ref{S-inv:floor},
\ref{S-a:toolspread}), when that first item-demand falls in a
ledger-charged piece, and to the \emph{ledger residual} otherwise
(first item-demand in a stable promoted interval or outside every
presence). Cross-epoch crediting of an already-logged event
(Definition~\ref{S-def:counting}(b), Rule~\ref{S-r:panel}(iv)) relabels a
prior epoch's event and adds nothing to any epoch. The four presence
classes partition the item's shared-tier presence within the epoch,
so every demander's epoch-first item-demand falls in exactly one
piece or outside every presence; the assignment classes are disjoint
and exhaustive over required, voluntary, and publish confirmations
(redundant confirmations excluded throughout,
\S\ref{S-app:budgetmodel}) --- in particular \emph{every required
confirmation is charged somewhere, to a bucket row or to the
residual, by construction} --- and each row bound is an upper bound
for its assigned class. Then:
\begin{enumerate}[label=(\roman*),leftmargin=2em,itemsep=2pt,topsep=2pt]
\item \emph{(Once, ever, per pair.)} No user's required count for any
pair $(v,t)$ exceeds one over the deployment's lifetime
(\ref{S-inv:floor}; exceptions per Theorem~\ref{S-thm:floor}), a fortiori
one per epoch; required confirmations of $(v,t)$ belong to distinct
demanders of the pair.
\item \emph{(Row bounds.)} Every pipeline-component confirmation is
assigned to exactly one row of Table~\ref{S-tab:ledger} under the
retrospective charging convention above, with the stated per-epoch
bounds; cross-epoch crediting of an already-logged event
(Rule~\ref{S-r:panel}(iv), Definition~\ref{S-def:counting}(b)) is a
relabel adding no event to any epoch.
\item \emph{(Split.)} Summing rows, there are $A(M)$ and $\Bfl$
depending only on protocol constants, with
\[
A(M)=\bigO\!\big(\Emax\,\ctool^{2}\,(N+\cacc f)(1+\kappa)\big)
\ \text{(via $\Ncap<N+\Dpool\le2N$)},
\]
\[
\Bfl=\bigO\!\big(\Emax\,\ctool^{2}(1+(1-p_b)\,\kappa)\big),
\]
such that
$\Ex[C^{\mathrm{pipe}}_v]\ \le\ A(M)\,\Prob[n_v\ge1]\;+\;\Bfl\,\gbar\,
\Ex[n_v]$; the pipeline required total obeys the same bound. Explicit
constants: Lemma~\ref{S-lem:constants} (\S\ref{S-app:constants}).
\item \emph{(Pathwise, all baseline-counted confirmations.)} Let $C_v$ count all
required, voluntary, and publish confirmations of $v$'s exact bytes in
the epoch, wherever they fall --- ledger-charged and residual alike;
redundant confirmations and option-induced cache confirmations
excluded (\S\ref{S-app:budgetmodel};
Corollary~\ref{S-cor:flooropt}). In the single-epoch model, with
$K:=\Emax(\ctool\Ncap+\cpr k+1)$,
\[
C_v\ \le\ \ctool\,n_v+K\,\mathbf 1\{n_v\ge1\}:
\]
every required confirmation of a pair of $v$ belongs to a demander of
the pair --- its confirmer's own attempted use fires the gate, hence
realizes the latent indicator (required-approval clause,
\S\ref{S-app:budgetmodel}), inside or outside any shared-tier presence
--- at most $\ctool$ pairs per demander (one per pair,
\ref{S-inv:floor}, \ref{S-a:toolspread}); voluntary
confirmations exist only at invitation surfaces, by definition
(\S\ref{S-app:budgetmodel}), and invitations per epoch are
deterministically capped at $\Emax\ctool\Ncap$ re-entry notifications
plus $\Emax\cpr k$ suspension prompts, \emph{independently of
current-epoch demand} --- notified endorsers and prompted adopters
need not be current-epoch demanders --- with joint deduplication
(Definition~\ref{S-def:counting}, Rule~\ref{S-r:panel}(vi)); the owner
contributes at most $\Emax$ publish confirmations
(one accepting publish per opened entry: a cap-exhausted submission is ineligible and records no accepting confirmation; Rules~\ref{S-r:share}, \ref{S-r:hygiene}(i)). Every term vanishes on $\{n_v=0\}$, by the event-based vanishing clause of the owner-coupling convention and the
required-approval clause (\S\ref{S-app:budgetmodel}): no latent demander
means no attempted use, hence no required confirmation, and no entry, review, invitation, or publish confirmation occurs in the epoch, whatever presence is carried in.
\end{enumerate}
\end{lemma}

\begin{table}[t]
\centering\small
\renewcommand{\arraystretch}{1.3}
\begin{tabular}{@{}p{0.26\textwidth} p{0.10\textwidth} p{0.52\textwidth}@{}}
\toprule
\textbf{Row} & \textbf{Kind} & \textbf{Bound per item per epoch}\\
\midrule
Weight-bearing endorsements & req./vol. & $\le\Emax\,\ctool\,\Ncap$:
at most $\Ncap$ credited per tool toward each entry
(Definition~\ref{S-def:counting}(f); credited by the close of the
logging epoch, within-epoch banked credits included; cross-epoch
crediting relabels a prior epoch's event and adds nothing), credited $\Rightarrow$ authority-active (Definition~\ref{S-def:counting}(a));
$\le\ctool$ demanded tools (\ref{S-a:toolspread}), entries $\le\Emax$
deterministically (entry cap, Rule~\ref{S-r:hygiene}(i)), credited sets
identity-disjoint across entries (Lemma~\ref{S-lem:disjoint});
vindication-credited prompted acceptances included, required or invited
alike.\\
Window-bucket confirmations & required &
$\le\Eslot\,\ctool^2(\gbar\,n_v+\szero)$: a user's first \emph{item-demand}
in a window implies some \emph{pair}-first-demand in that window, so
$\le\ctool(\gbar\,n_v+\szero)$ demanders have their first item-demand in
it (\ref{S-a:smooth}, demand clause, per demanded pair; $\le\ctool$ pairs,
\ref{S-a:toolspread}), each charged $\le\ctool$ uncredited required
confirmations (one per pair, \ref{S-inv:floor}), over $\le\Eslot$ windows.\\
Dormancy-bucket confirmations & required &
$\le\Eslot\,\ctool^2(\gbar\,n_v+\szero)$: identically, over $\le\Eslot$
candidate-dormancy intervals (presence taxonomy,
Lemma~\ref{S-lem:budget}); freshness consumed, decision banked
(Definition~\ref{S-def:counting}(b)), crediting charges nothing.
Stable promoted intervals are charged to no row: their required gates
are ledger residual (Theorem~\ref{S-thm:sat}(iii)).\\
Freeze-bucket confirmations & required &
expected current-epoch disputes and reviews, plus the at most one carry-in instance, $\le\bar D_v$; each admits
$\le\kappa$ arriving demanders in expectation while unresolved
(Lemma~\ref{S-lem:geo}) --- item-arrivals during the freeze or suspension,
exactly what Lemma~\ref{S-lem:geo} counts; each such demander (including
every demander gating during a suspension, Rule~\ref{S-r:panel}(vi)) is
charged $\le\ctool$ confirmations; total $\le\ctool\,\bar D_v\,\kappa$ in
expectation. Each is owed on any unpromoted pair regardless
(Lemma~\ref{S-lem:freeze}(ii)).\\
Repeat re-inspections & voluntary & $\le\Emax\,\ctool\,\Ncap$: notified
cohorts are the immediately preceding entry's credited endorsers only
(Definition~\ref{S-def:counting}), at most $\Ncap$ credited slots per demanded tool (Definition~\ref{S-def:counting}(f)) ($\le\ctool$ tools, \ref{S-a:toolspread}), cohorts
identity-disjoint across re-entries (Lemma~\ref{S-lem:disjoint}), over
$\le\Emax$ re-entries; declinable at zero cost.\\
Suspension-prompt re-inspections & voluntary & $\le\Emax\,\cpr k$:
$\le\cpr k$ prompts per fast-track review, reviews $\le\Emax$ per item
per epoch (Rule~\ref{S-r:panel}(vi)); invitations deduplicated jointly;
absorbed into the $\kappa'$ term of $A(M)$
(Lemma~\ref{S-lem:constants}).\\
Publish confirmations & owner & $\le\Emax$: one per opened entry
(Rules~\ref{S-r:share}, \ref{S-r:hygiene}(i)).\\
\bottomrule
\end{tabular}
\caption{The per-item confirmation ledger. ``Req.'' and ``vol.'' denote required and
voluntary confirmations (\S\ref{S-app:budgetmodel}). Bucket rows use the
conservative $\Eslot=\Emax+1$ (Lemma~\ref{S-lem:budget}); windows need
only $\Emax$. Stable promoted intervals are charged to no row; their
required confirmations form the ledger residual, bounded by
Theorem~\ref{S-thm:sat}(iii) under \ref{S-a:promsuf}.}
\label{S-tab:ledger}
\end{table}

\begin{proof}
\emph{(i)} After a user's own accepting confirmation of $(v,t)$, the
local in-scope $\C$ copy covers all their reuse at $t$ for the
deployment's lifetime (Lemma~\ref{S-lem:scopepersist}) and no third
party can disturb it (Lemma~\ref{S-lem:nodelete}); this is
\ref{S-inv:floor}.

\emph{(ii)} \emph{Weight-bearing:} within an entry, credited
endorsements per tool number at most $\Ncap$ by the credit caps
(Definition~\ref{S-def:counting}(f)), banked credits filling slots the
same caps govern; each credited endorsement is a distinct demander's
first protocol-bearing decision, so successive entries' credited sets
are disjoint (Lemma~\ref{S-lem:disjoint}), entries per epoch at most
$\Emax$ (the entry cap of Rule~\ref{S-r:hygiene}(i), republications and post-eviction republications included).
\emph{Buckets:} within the epoch, the item's shared-tier presence
partitions into the four classes of the presence taxonomy: at most
$\Emax$ windows, at most $\Eslot$ candidate-dormancy pieces (one
possibly carried in), the freeze/suspension intervals of the
current-epoch disputes and reviews plus at most one carry-in
instance, and the stable promoted intervals; every demander of $v$
has exactly one epoch-first item-demand, in exactly one piece or
outside every presence. A demander whose first item-demand falls in a
window or candidate-dormancy piece has a pair-first-demand there for
at least one of the item's $\le\ctool$ pairs, so \ref{S-a:smooth}'s
demand clause (scoped to exactly these two piece types) bounds
window/candidate-dormancy occupancy by $\ctool(\gbar\,n_v+\szero)$
per piece; each such demander is charged at most $\ctool$ required
pair-confirmations (once per pair per transaction,
Definition~\ref{S-def:sinkconf}; once ever required, \ref{S-inv:floor}),
giving the two bucket rows with the factor $\Eslot$.
Lemma~\ref{S-lem:geo} bounds each freeze or review's arriving demanders
by $\kappa$ in expectation while unresolved --- for the carry-in
instance, counting only its current-epoch arrivals, a sub-count of
the same dominated arrival sequence --- each charged at most
$\ctool$; the freeze-bucket bound $\ctool\,\bar D_v\,\kappa$ follows
by the tower property under the demand-conditional measure
(\S\ref{S-app:prob}). Demanders whose first item-demand falls in a
stable promoted interval, or outside every presence, are charged to
no bucket: their required confirmations are the ledger residual,
bounded by Theorem~\ref{S-thm:sat}(iii) under \ref{S-a:promsuf} (their
credited acceptances, where any, are weight-row events already
counted).
\emph{Disputes and reviews ($\bar D_v$):} disputes on $v$ open only at
protocol-bearing dispute-initiating rejections logged while $v$ is
unfrozen \emph{and unpromoted at every tool} (freeze-time rejections
corroborate, Rule~\ref{S-r:panel}(iii); a qualifying rejection on bytes
holding $\CC$ status at any tool is a post-promotion rejection routed
to clause~(vi), Rule~\ref{S-r:panel}(ii), counted below within the
review cap) --- hence only during window or candidate-dormancy pieces
--- at most one initiation per identity per item per epoch
(Rule~\ref{S-r:panel}(i)). An honest identity's initiating rejection is the rejecting branch of
its first \emph{audit-recorded} decision on that pair in the epoch,
at a non-covered gate (cache gates included: both branches are
audit-recorded, \S\ref{S-app:prob}) with tool
or reasoning target while $v$ is
unfrozen (an earlier audit-recorded accepting decision on the
pair would have created durable coverage, Lemma~\ref{S-lem:scopepersist}, making the gate
covered, hence inert; an earlier rejecting decision --- federation-logged in every case,
Definition~\ref{S-def:counting}(iii) --- would make the
present one non-initiating; abandonment records nothing). The trial is indexed at
the pre-outcome header atom of an audit-recorded first pair-decision,
so its occurrence never selects the outcome, and the occurrences are
capped by \ref{S-a:smooth}'s first-pair-decision clause: at most
$\gbar\,n_v+\szero$ per pair per piece, $\le\ctool$ pairs, over at
most $\Eslot$ window pieces and $\Eslot$ candidate-dormancy pieces
--- conservatively $2\Eslot\ctool(\gbar\,n_v+\szero)$ candidate
occurrences, each rejecting a benign item with conditional
probability at most $1-p_b$ at its header atom (the $p_b$ hypothesis
covers every honest inspection, cache gates included;
Lemma~\ref{S-lem:dispbridge}(ii), Lemma~\ref{S-lem:adaptive}(iii)),
contributing the first term.
\emph{Invited reconsiderations} are epoch-scoped
(Definition~\ref{S-def:counting}): at most $\Emax\ctool\Ncap$ per
epoch, each rejecting with conditional probability at most $1-p_b$
(repeat form of the hypothesis), contributing the second term.
\emph{Adversarial initiations} that take effect are either eventually
charged, at most $\cacc$ per identity per epoch
(Lemma~\ref{S-lem:cacc}), $\le\cacc f$ in total; or terminal (eviction
or expiry), retiring that presence, at most one per presence and
$\le\Emax$ presences per epoch. \emph{Post-promotion reviews} ---
which absorb every rejection logged during a stable promoted
interval, honest or adversarial --- are capped at $\Emax$ per item
per epoch (Rule~\ref{S-r:panel}(vi)). The carry-in instance adds $1$.
Summing gives $\bar D_v$.
\emph{Voluntary and publish rows:} Definition~\ref{S-def:counting},
Rule~\ref{S-r:panel}(vi), Rule~\ref{S-r:share}.

\emph{(iii)} Summing the row bounds (with $\Eslot$ in the bucket rows and the carry-in term of $\bar D_v$) and collecting the coefficient of
$\gbar\,n_v$ gives the split; every row's pathwise count vanishes on
$\{n_v=0\}$ (the event-based vanishing clause of the owner-coupling convention, \S\ref{S-app:budgetmodel}: no entries, reviews, invitations, or demanders on $\{n_v=0\}$, whatever presence is carried in), so
each demand-free term is carried with the indicator
$\mathbf{1}\{n_v\ge1\}$ before taking expectations (the freeze
bucket via Lemma~\ref{S-lem:adaptive}(iii) applied to the
indicator-stopped counts), multiplying it by $\Prob[n_v\ge1]$; the
explicit algebra, with the stated
$\bigO$-forms verified, is Lemma~\ref{S-lem:constants}. Each required
confirmation belongs to a distinct demander, so the required total obeys
the same bound.

\emph{(iv)} With $K:=\Emax\ctool\Ncap+\Emax\cpr k+\Emax$:
\[
C_v\ \le\ \ctool\,n_v+K\,\mathbf 1\{n_v\ge1\}.
\]
Required: by the required-approval clause (\S\ref{S-app:budgetmodel}) every
user making a required approval of a pair of $v$ is a latent demander of $v$,
contributing at most $\ctool$ pair-confirmations, once per pair per
transaction and once ever required (\ref{S-inv:floor},
\ref{S-a:toolspread}); voluntary: confirmations exist only at invitation
surfaces
(\S\ref{S-app:budgetmodel}), which are epoch-scoped and deterministically
capped at $\Emax\ctool\Ncap$ re-entry notifications plus $\Emax\cpr k$
suspension prompts, with joint deduplication and boundary expiry
(Definition~\ref{S-def:counting}, Rule~\ref{S-r:panel}(vi)), independently
of current-epoch demand; publish: at most $\Emax$
(one per opened entry; Rules~\ref{S-r:share}, \ref{S-r:hygiene}(i)). Every term vanishes
on $\{n_v=0\}$: no latent demander means no attempted use, hence no
required confirmation (required-approval clause); an entry requires a
current-epoch submission (owner demand); a review or dispute opens at
a latent demander's protocol-bearing decision; invitations require a
current-epoch entry or review, and a carried-in review issues no new
prompts (Rule~\ref{S-r:panel}(vi)); a carried-in freeze with no
current-epoch arrivals charges nothing.
\end{proof}
\subsection{Saturation}
\label{S-app:saturation}

\begin{lemma}[Distinct-demand bound, two-term form]\label{S-lem:distinct}
Let $\Vals_{\mathrm{good}}\subseteq\Vals$ be the benign
(Definition~\ref{S-def:bad}), assumption-conforming items --- those to
which the $p_b$, \ref{S-a:smooth}, \ref{S-a:rhoseat}, and
\ref{S-a:toolspread} hypotheses of Lemma~\ref{S-lem:budget} apply --- and
write $V_{\mathrm{good}}(M):=\Ex|\{v\in\Vals_{\mathrm{good}}:n_v\ge1\}|
\le V(M)$ and
$d_{\mathrm{good}}:=\sum_{v\in\Vals_{\mathrm{good}}}p_v\le d$. Then
Under \ref{S-a:smooth}, \ref{S-a:rhoseat}, \ref{S-a:toolspread}, and the competence bound of
Lemma~\ref{S-lem:budget}, with $A(M)$, $\Bfl$, $\kappa$ as there,
$$
\Ex[\bar C^{\mathrm{pipe,good}}_M]\ \le\ A(M)\,\frac{V_{\mathrm{good}}(M)}{M}\;+\;\Bfl\,\gbar\,d_{\mathrm{good}},
$$
where $\bar C^{\mathrm{pipe,good}}_M$ counts all pipeline-component shareable
confirmations, required and voluntary (Lemma~\ref{S-lem:budget}), over $\Vals_{\mathrm{good}}$; the
ledger residual (Lemma~\ref{S-lem:budget}, charging convention) is
bounded separately by Theorem~\ref{S-thm:sat}(iii) and added in
Corollary~\ref{S-cor:total}. The first term is carried by the number of distinct demanded
items and vanishes under saturation; the second is the
\emph{front-loading term}: item popularity has cancelled out of it,
leaving only the front-loading multiple $\gbar$ against the
standalone single-user shareable mass $d$. Demand on the complement
is charged at the single-user cap and enters the residual
$r_{\mathrm{bad}}$ of Corollary~\ref{S-cor:total}.
\end{lemma}

\begin{proof}
By Lemma~\ref{S-lem:budget}(iii),
$\Ex[C^{\mathrm{pipe}}_v]\le A(M)\,\Prob[n_v\ge1]+\Bfl\,\gbar\,\Ex[n_v]$. Summing
over $v\in\Vals_{\mathrm{good}}$: $\sum_{v\in\Vals_{\mathrm{good}}}\Prob[n_v\ge1]=\Ex|\{v\in\Vals_{\mathrm{good}}:n_v\ge1\}|=V_{\mathrm{good}}(M)$ and
$\sum_{v\in\Vals_{\mathrm{good}}}\Ex[n_v]=Md_{\mathrm{good}}$; divide by $M$. ($\Emax$ inside $A(M)$ is a
deterministic per-epoch entry bound, Rule~\ref{S-r:hygiene}(i); no expectation over
entry counts is needed.)
\end{proof}

\begin{theorem}[Confirmation cost under saturation]\label{S-thm:sat}
Under \ref{S-a:smooth}, \ref{S-a:rhoseat}, \ref{S-a:toolspread}, the
competence bound of Lemma~\ref{S-lem:budget}, and --- consumed only by
the limit ($\to0$) clauses below --- \ref{S-a:massbound}:
\begin{enumerate}[label=(\roman*),leftmargin=2em,itemsep=2pt,topsep=2pt]
\item if $V(M)=o(M)$ and $A(M)\,V(M)=o(M)$, then
$\Ex[\bar C^{\mathrm{pipe,good}}_M]\le o(1)+\Bfl\,\gbar\,d_{\mathrm{good}}$: the ledger-counted shareable
cost falls to at most the front-loading multiple $\Bfl\,\gbar$ of the
standalone single-user shareable mass, independently of item popularity; if in
addition $\gbar(M)(1+\kappa(M))\to0$, then $\Ex[\bar C^{\mathrm{pipe,good}}_M]\to0$ (an upper-bound statement: the displayed bounds vanish);
\item with $V(M)=\Theta(M^{\beta})$, $\beta\in(0,1)$, the protocol
constants fixed, and additionally $f(M)=\bigO(1)$ and
$\inf_M\rseat(M)>0$ --- hence $\kappa(M)=\bigO(1)$, so
$A(M)=\bigO(1)$ and $\Bfl(M)=\bigO(1)$; the added hypotheses are not
implied by fixing the protocol constants, $f$ being an adversary
resource and $\rseat$ a behavioural parameter --- (i)'s hypothesis
holds automatically:
$\Ex[\bar C^{\mathrm{pipe,good}}_M]\le\bigO(M^{\beta-1})
+\Bfl\,\gbar\,d_{\mathrm{good}}$, vanishing whenever
$\gbar(M)\to0$ (under these hypotheses equivalent to (i)'s condition
$\gbar(M)(1+\kappa(M))\to0$). This is the fixed-adversary,
fixed-constants regime; the growing-adversary regime is the corridor
(\S\ref{S-app:corridor}, Theorem~\ref{S-thm:coscale});
\item \emph{(Ledger residual, benign conforming component.)} under
\ref{S-a:promsuf} and \ref{S-a:toolspread}, the expected per-user required
cost charged to no ledger row \emph{over the benign,
\ref{S-a:promsuf}-conforming items} --- the ledger residual of
Lemma~\ref{S-lem:budget}: required confirmations of demanders whose
epoch-first item-demand falls in a stable promoted interval (in
particular handle-expressed use of a live-$\CC$ pair, which gates
under Rule~\ref{S-r:whitelist}(c)) or outside every shared-tier
presence (before first publication, after retirement, or after
demotion); residual demand on poisoned or
\ref{S-a:promsuf}-violating pairs enters $r^{(e)}_{\mathrm{bad}}$
(Corollary~\ref{S-cor:total}) instead --- is at most
$\ctool\big(D_0(M)\,V(M)/M+\varepsilon(M)\,d_{\mathrm{good}}(M)\big)$
($d_{\mathrm{good}}\le d$, $V_{\mathrm{good}}(M)\le V(M)$ as in
Lemma~\ref{S-lem:distinct}). This bound vanishes if, in addition,
$D_0(M)\,V(M)=o(M)$ and $\varepsilon(M)\,d_{\mathrm{good}}(M)\to0$;
under \ref{S-a:massbound} the latter follows from
$\varepsilon(M)\to0$. In particular, under the Heaps form
$V(M)=\Theta(M^{\beta})$ of clause~(ii), $D_0(M)=o(M^{1-\beta})$
supplies the product condition. The product condition is an
additional rate hypothesis of this vanishing conclusion: neither
\ref{S-a:promsuf} nor clause~(i)'s hypotheses alone implies it.
\end{enumerate}
\end{theorem}

\begin{proof}
(i)--(ii) Immediate from Lemma~\ref{S-lem:distinct}:
$A(M)V(M)/M=o(1)$ by hypothesis (under Heaps' form with the hypotheses of (ii) --- protocol constants fixed, $f(M)=\bigO(1)$, $\rseat$ bounded below --- $A(M)=\bigO(1)$, $\kappa(M)=\bigO(1)$, and $V(M)/M=\bigO(M^{\beta-1})$), and
$\Bfl=\bigO(\Emax\ctool^{2}(1+\kappa))=\bigO(\Emax(1+\kappa))$ ($\ctool=\bigO(1)$ absorbed), so the second term is
$\bigO(\Emax(1+\kappa)\gbar)\,d_{\mathrm{good}}$, vanishing whenever
$\gbar(M)(1+\kappa(M))\to0$, the mass factor being uniformly bounded
($d_{\mathrm{good}}\le d$, \ref{S-a:massbound}) (a sufficient
condition: the displayed bound is an upper bound, and degenerate
parameter regimes, e.g.\ $p_b=1$, can shrink it further).
(iii) Fix a benign, \ref{S-a:promsuf}-conforming item $v$ and a target
$t$. Every required confirmation of $(v,t)$ belongs to a distinct
demander of the pair (\ref{S-inv:floor}). Under the charging
convention of Lemma~\ref{S-lem:budget}, each is charged to a bucket
row (its demander's epoch-first item-demand lying in a
ledger-charged piece), to the weight row (credited), or to the
ledger residual; nothing is uncharged. The residual required
confirmations of $(v,t)$ number exactly $U_{v,t}$
(\ref{S-a:promsuf}), with $U_{v,t}\le n_{v,t}$ pathwise; conditioning
on $n_{v,t}$ and applying \ref{S-a:promsuf} (whose bound holds
trivially on $\{n_{v,t}=0\}$, where $U_{v,t}=0$),
\[
\Ex[U_{v,t}]\ \le\ D_0(M)\,\Prob[n_{v,t}\ge1]
+\varepsilon(M)\,\Ex[n_{v,t}].
\]
Now sum over \emph{all} targets $t$, using \ref{S-a:toolspread}
pathwise on each term:
\[
\sum_t\Prob[n_{v,t}\ge1]
=\Ex\big[\#\{t:n_{v,t}\ge1\}\big]\ \le\ \ctool\,\Prob[n_v\ge1],
\]
\[
\sum_t\Ex[n_{v,t}]=\Ex\Big[\sum_t n_{v,t}\Big]\ \le\
\ctool\,\Ex[n_v],
\]
the first because at most $\ctool$ tools carry demand and none does
on $\{n_v=0\}$ ($n_{v,t}\le n_v$), so
$\#\{t:n_{v,t}\ge1\}\le\ctool\,\mathbf 1\{n_v\ge1\}$ pathwise; the
second by $\sum_t n_{v,t}\le\ctool\,n_v$. Summing over the benign,
\ref{S-a:promsuf}-conforming items
($\sum_{v\in\Vals_{\mathrm{good}}}\Prob[n_v\ge1]
=V_{\mathrm{good}}(M)\le V(M)$,
$\sum_{v\in\Vals_{\mathrm{good}}}\Ex[n_v]=M\,d_{\mathrm{good}}(M)$,
Lemma~\ref{S-lem:distinct}) and dividing by $M$ gives
$\ctool\big(D_0(M)V(M)/M+\varepsilon(M)\,d_{\mathrm{good}}(M)\big)$,
which vanishes under the additional rate hypotheses of the statement
($D_0(M)V(M)=o(M)$; $\varepsilon(M)d_{\mathrm{good}}(M)\to0$,
supplied under \ref{S-a:massbound} by $\varepsilon(M)\to0$), with
$\ctool=\bigO(1)$; under the Heaps form of clause~(ii),
$D_0(M)=o(M^{1-\beta})$ supplies the product condition. (Beyond
\ref{S-a:promsuf}, the pair-to-item aggregation consumes only the
bounded tool-spread of \ref{S-a:toolspread}; the vanishing limit
additionally consumes the stated rate hypotheses and, where
$\varepsilon(M)\to0$ is invoked, \ref{S-a:massbound}. Demand at
pairs where \ref{S-a:promsuf} fails enters
$r^{(e)}_{\mathrm{bad}}$, Corollary~\ref{S-cor:total}.)
\end{proof}

\begin{corollary}[Per-epoch shared cost; lifetime reading]\label{S-cor:total}
Fix an analyzed epoch $e$ satisfying the hypotheses of
Theorem~\ref{S-thm:sat}(i)--(iii) \emph{including the additional rate
conditions of clause~(iii), read per epoch}:
$D_{0,e}(M)\,V_e(M)=o(M)$ and
$\varepsilon_e(M)\,d_{\mathrm{good},e}(M)\to0$ (under
\ref{S-a:massbound}, from $\varepsilon_e(M)\to0$); let
$r^{(e)}_{\mathrm{bad}}=r^{(e)}_{\mathrm{bad}}(M)$ be
the expected per-user demand mass in epoch $e$ on $\mathsf{Bad}$
(Definition~\ref{S-def:bad}) or \ref{S-a:promsuf}-violating pairs (each such
pair charged at most once ever, Theorem~\ref{S-thm:floor}(i)). Then
\[
\limsup_{M\to\infty}\ \Ex\big[\bar C^{(e)}_{M,\mathrm{share}}\big]
\;\le\;\limsup_{M\to\infty}\Big(\Bfl(M)\,\gbar(M)\,d_{\mathrm{good}}
\;+\;r^{(e)}_{\mathrm{bad}}(M)\Big),
\]
and under a shrinking-front-loading schedule with bounded mass (\S\ref{S-app:corridor}, \ref{S-a:massbound})
the first term vanishes. The personal component is a \emph{lifetime}
inventory: at most $\dpriv$ expected required pair-confirmations over
the deployment's life (Theorem~\ref{S-thm:floor}(ii)). The headline
reading ``lifetime personal inventory plus current-epoch shared
load'' therefore satisfies $\limsup_{M\to\infty}\big(\text{lifetime personal}+\text{epoch-$e$ shared}\big)\le\dpriv+\limsup_{M\to\infty}r^{(e)}_{\mathrm{bad}}(M)$.
This is \emph{not} a bound on the sum of shared costs over an
unbounded number of epochs: the present accounting yields a lifetime
shared \emph{upper bound} whenever the following series of per-epoch
upper bounds is summable --- a sufficient condition only: the terms
are loose upper bounds, so divergence of this series implies nothing
about the true lifetime cost --- with the coverage parameters read
per epoch ($D_{0,e},\varepsilon_e$; a deployment asserting
\ref{S-a:promsuf} with epoch-uniform parameters drops the subscripts),
\begin{multline*}
\sum_{e}\Big(A_e(M)\,\tfrac{V_e(M)}{M}
+\Bfl^{(e)}(M)\,\gbar_e(M)\,d_{\mathrm{good},e}
\\
+\ctool\big(D_{0,e}(M)\tfrac{V_e(M)}{M}+\varepsilon_e(M)\,d_{\mathrm{good},e}\big)
+r^{(e)}_{\mathrm{bad}}\Big)<\infty,
\end{multline*}
in which case lifetime expected total cost is at most $\dpriv$ plus
that sum; the lifetime shareable mass comparison of
Theorem~\ref{S-thm:floor}(iii) ($d^{\infty}_{\mathrm{tool}}$) bounds the
required component unconditionally in any case. These are upper bounds;
no matching lower bound is claimed.
\end{corollary}

\begin{proof}
Combine Theorem~\ref{S-thm:sat}(i)--(iii) over epoch $e$ with Theorem~\ref{S-thm:floor}(i)--(iii); the summable case is termwise.
\end{proof}

\begin{remark}[Reading of Theorem~\ref{S-thm:sat}]\label{S-rem:satread}
Four observations govern the interpretation of
Theorem~\ref{S-thm:sat}. First, the residual charges are the design operating as intended:
each residual poisoned-demand confirmation is a personal gate
inspection of an unpromoted or flagged item, and each residual
handle-arrival confirmation is a personal gate inspection of bytes
that reached this user through an untrusted channel --- the
certificate amortises trusted-channel reuse, never untrusted-channel
arrival (Rule~\ref{S-r:whitelist}(c), \ref{S-q:handlecover}); both are
exactly the inspections the design intends (cf.\ \ref{S-a:promsuf}). Second, the
front-loading term is the price of certifying popular items rather
than an overhead: each within-window confirmation is a
fresh honest inspection \emph{before finalisation}: exactly the
pre-promotion scrutiny the exponent $N-f$ of
Theorem~\ref{S-thm:panel} counts. Third, the
bound is informative when $\gbar\lesssim1/\Bfl$; for larger $\gbar$
it remains valid but loose, a bounded multiple of $d$,
with Theorem~\ref{S-thm:floor} the binding guarantee throughout.
Fourth, these are conditional \emph{accounting} statements.
Clauses~(i)--(ii) consume the ledger, smoothness, competence,
juror-supply, and demand-shape hypotheses; clause~(iii) additionally consumes \ref{S-a:promsuf} --- the
assumption carrying publication, promotion, and
arrival-representation coverage jointly --- together with the
product condition $D_0(M)V(M)=o(M)$ and the remaining vanishing
conditions of Theorem~\ref{S-thm:sat}(iii). The
shrinking-front-loading conditions describe changing time geometry,
never a consequence of population growth alone
(Remark~\ref{S-rem:gbarread}); no clause derives coverage or shrinking
front-loading from the protocol. The assumption-free protocol
guarantee is the pathwise cap of Theorem~\ref{S-thm:floor}(i), with
clause~(ii) giving its reference-model standalone-mass ceiling.
\end{remark}
\subsection{Conditional co-scaling}
\label{S-app:corridor}

\paragraph{Status of the corridor.}
The corridor collects, in one place, every constraint the co-scaling
result uses; each row exists elsewhere and none is new. Two of its rows
are \emph{schedules} of protocol constants ($g_N,g_k$), which a
deployment controls; the rest are empirical or behavioural properties
($\beta$, $\gbar$, $\rseat$, $q+p_b$, $d$, the coverage law
$(D_0,\varepsilon)$ of \ref{S-a:promsuf}, the Sybil bounds), which a
deployment \emph{measures} (\ref{S-n:validate}) and cannot set. The
co-scaling theorem is therefore a compatibility statement: \emph{if}
the measured environment satisfies its rows, \emph{then} schedules
exist under which security, resolution correctness, and budget improve
together; and the floor of Theorem~\ref{S-thm:floor} holds outside the
corridor unconditionally.

\begin{center}
\small
\renewcommand{\arraystretch}{1.25}
\begin{tabular}{@{}p{0.30\textwidth} p{0.17\textwidth} p{0.24\textwidth} p{0.20\textwidth}@{}}
\toprule
\textbf{Constraint} & \textbf{Source} & \textbf{Role} &
\textbf{Example schedule} \\
\midrule
$V(M)=\Theta(M^{\beta})$, $\beta\in(0,1)$ & Heaps saturation & budget
decay rate & $\beta$ measured \\
$\sup_M d(M)<\infty$ & \ref{S-a:massbound} & mass factor of the
front-loading term bounded & automatic under a fixed demand law;
measured \\
$f(M)=o(M^{1-\beta})$ & Sybil condition & margins can grow & $f=\log M$ \\
$\fs(M)=o(g_k(M))$ & Thm~\ref{S-thm:coscale} & panel margin positive &
$\fs=\bigO(1)$ \\
$\gbar(M)\,(1+\kappa(M))=o(1)$ & \ref{S-a:smooth}, \ref{S-a:rhoseat} &
front-loading term vanishes & $\gbar=\log^{-3}M$ measured \\
$D_0(M)\,V(M)=o(M)$, $\varepsilon(M)\to0$ (Heaps:
$D_0=o(M^{1-\beta})$) & \ref{S-a:promsuf},
Thm~\ref{S-thm:sat}(iii) & coverage residual vanishes (with
\ref{S-a:massbound}: $\varepsilon d_{\mathrm{good}}\to0$) & measured:
$D_0\asymp N/\gbar=\Theta(\log^{4}M)$, $\varepsilon\to0$ \\
$N(M)\,k(M)=o(M^{1-\beta})$ & cross-term, Thm~\ref{S-thm:coscale} &
dispute load sublinear & $2\log^{3}M=o(M^{1-\beta})$ \\
$g_N,g_k\to\infty$, $=o(M^{1-\beta})$ & margin schedules & exponents
diverge & $g_N=\log M$, $g_k=\log^{2}M$ \\
$\rseat$ bounded below & \ref{S-a:rhoseat} & $\kappa=\Theta(k)$ &
measured, logged \\
$q+p_b>1$ & competence (Rem.~\ref{S-rem:qsingle}) & joint safety and resolution correctness & measured, logged \\
\bottomrule
\end{tabular}
\end{center}

\begin{remark}[Operational reading of the shrinking-front-loading
rows]\label{S-rem:gbarread}
For steady demand over an epoch of length $L$ with fixed window
$\Twin$, $\gbar\approx\Twin/L$ \emph{independently of $M$}
(\ref{S-a:smooth}, commentary): population growth alone does not shrink
$\gbar$. The rows $\gbar(M)(1+\kappa(M))=o(1)$ and $\gbar(M)k(M)\to0$
therefore describe deployments whose \emph{time geometry} changes with
scale --- finalisation windows shrinking relative to the epoch, or
demand growing more diffuse as the population grows --- and are to be
read as measured properties of that geometry, not consequences of
$M\to\infty$. Where they fail, the fixed-$\gbar$ reading of
Theorem~\ref{S-thm:sat} is the operating point: pipeline cost falls to
the front-loading multiple $\Bfl\,\gbar\,d_{\mathrm{good}}$ of the standalone single-user shareable mass,
with Theorem~\ref{S-thm:floor} binding throughout
(Remark~\ref{S-rem:satread}).
\end{remark}

\noindent A concrete schedule satisfying every row simultaneously:
$\fs=\bigO(1)$, $f(M)=\log M$, $N(M)=2\log M$, $k(M)=\fs+\log^{2}M$,
against measured $\gbar(M)=\log^{-3}M$ and bounded measured $d$ (\ref{S-a:massbound}); then with $\rseat$ bounded
below, $\kappa=\Theta(\log^{2}M)$, so
$\gbar(1+\kappa)=\Theta(\log^{-1}M)\to0$;
for an illustrative measured coverage law with
$D_0\asymp N/\gbar=\Theta(\log^{4}M)$ and empirically validated
$\varepsilon(M)\to0$, $D_0V=\Theta(M^{\beta}\log^{4}M)=o(M)$; and
$Nk=\Theta(\log^{3}M)=o(M^{1-\beta})$.

\begin{theorem}[Joint scaling of safety, wrongful-eviction risk, and cost]
\label{S-thm:coscale}
Adopt Rule~\ref{S-r:panel} with the default bars, \ref{S-a:rhoseat}, \ref{S-a:smooth},
\ref{S-a:promsuf}, \ref{S-a:toolspread}, \ref{S-a:massbound}, and saturation
$V(M)=\Theta(M^{\beta})$, $\beta\in(0,1)$. Assume also the
coverage-rate condition of Theorem~\ref{S-thm:sat}(iii),
$D_0(M)\,V(M)=o(M)$ (under the Heaps form, equivalently
$D_0(M)=o(M^{1-\beta})$; $\varepsilon(M)\to0$ per \ref{S-a:promsuf}). Suppose
$f(M)=o(M^{1-\beta})$ and $q+p_b>1$; fix $\gamma_0\in(0,q+p_b-1)$.
Choose margin schedules $g_N(M),g_k(M)\to\infty$ with
$g_N,g_k=o(M^{1-\beta})$ and $\fs(M)=o(g_k(M))$, and set
$N(M)=f(M)+g_N(M)$, $k(M)=\fs(M)+g_k(M)$,
$m(M)=\fs(M)+\lceil(1-q+\gamma_0)g_k(M)\rceil$,
$m_{\mathrm{rej}}(M)=k(M)+1-m(M)$ (so every panel that gathers $k$
verdicts resolves decisively and $m_{\mathrm{rej}}>\fs$ for all large
$M$). All protocol caps and multiplicity parameters not explicitly
scheduled here are uniformly $\bigO(1)$ in $M$ (fixed TCB
configuration along the scaling, convention~(\ref{S-conv:c8})); in particular
$\Emax,\rsub,\rmax,\cceil,\cpr,\szero,\Dpool,\cbatch,\ccand
=\bigO(1)$, hence $\cacc=\cceil+\rmax=\bigO(1)$, while
$\ctool=\bigO(1)$ by \ref{S-a:toolspread}. This is load-bearing: the
conclusions
$A(M)=\bigO\!\big(\Emax\ctool^{2}(N+\cacc f)(1+\kappa)\big)
=o(M^{1-\beta})$ and $K(M)=o(M^{1-\beta})$ consume it.
Assume further the cross-term condition
$N(M)\,k(M)=o(M^{1-\beta})$ (which, since
$N>f$, implies $f(M)k(M)=o(M^{1-\beta})$) and the shrinking front-loading condition
$\gbar(M)\,k(M)\to0$ ($\kappa=\Theta(k)$ for $\rseat$ bounded below);
both are achievable by choosing $g_k$ to grow slowly against the
measured front-loading of the demand stream. Then:
\begin{enumerate}[label=(\roman*),leftmargin=2em,itemsep=2pt,topsep=2pt]
\item \emph{(Safety, per poisoned entry.)}
$\Prob[\text{promotion of a fixed poisoned entry}]\le(1-q)^{g_N(M)}
+e^{-2\gamma_0^{2}g_k(M)}\to0$ (Theorem~\ref{S-thm:panel}), against adaptive
presentation and at any reach (\ref{S-inv:reach}). \emph{(System-wide,
per anchor epoch.)} If additionally
$2f(M)\,\rsub\big((1-q)^{g_N(M)}+e^{-2\gamma_0^{2}g_k(M)}\big)\to0$, the
per-anchor-epoch bound of Corollary~\ref{S-cor:submission} over all
adversary-published attempts anchored in the epoch also vanishes; the example schedule satisfies
this: with $f=\log M$, $g_N=\log M$ (natural logarithm throughout; a fixed
change of logarithm base only rescales the schedule by a constant and
leaves every limit below unaffected),
$2f\rsub(1-q)^{g_N}=2\rsub\log M\cdot M^{\ln(1-q)}\to0$, and
$2f\rsub\,e^{-2\gamma_0^2\log^2 M}\to0$ a fortiori;
\item \emph{(Wrongful eviction.)} with $\delta_b:=\tfrac{q+p_b-1-\gamma_0}{2}>0$, for all sufficiently
large $M$ a benign item is evicted, penalising its
publisher,\linebreak[4] with probability at most $e^{-2\delta_b^{2}g_k(M)}\to0$
(Propositions~\ref{S-prop:live},~\ref{S-prop:pubsafe}); equivalently
$\Prob[\text{vindication}]\ge\Prob[\text{resolution}]
-e^{-2\delta_b^{2}g_k(M)}$, resolution being the demand-side event of
Remark~\ref{S-rem:resolution};
\item \emph{(Budget, per analyzed epoch.)} with $\rseat$ bounded
below, $\Ex[\bar C^{\mathrm{pipe,good}}_M]\le
A(M)\,\bigO(M^{\beta-1})+\Bfl\,\gbar(M)\,d_{\mathrm{good}}=o(1)$, and
the benign ledger residual is $o(1)$
(Theorem~\ref{S-thm:sat}(iii)): the analyzed epoch's shared
confirmation load vanishes, while the lifetime personal inventory
remains at most $\dpriv$ (Theorem~\ref{S-thm:floor}) and demand on
$\mathsf{Bad}$ or assumption-violating pairs adds that epoch's
residual $r^{(e)}_{\mathrm{bad}}$, bounded pathwise by
Theorem~\ref{S-thm:floor}(i). No claim is made about the sum over an
unbounded number of epochs without the summability condition of
Corollary~\ref{S-cor:total}.
\end{enumerate}
All three hold with no assumption on how widely a poison spreads.

\end{theorem}

\begin{proof}
(i) $N-f=g_N$, and
$\delta=\frac{m-\fs}{k-\fs}-(1-q)\ge\gamma_0$ by the choice of $m$, so
Theorem~\ref{S-thm:panel} gives the bound with exponents $g_N$ and
$2\gamma_0^{2}g_k$.
(ii) $m/(k-\fs)\le(1-q+\gamma_0)+(\fs+1)/g_k(M)$; since
$1-q+\gamma_0=p_b-2\delta_b$ by the definition of $\delta_b$ (using
$\gamma_0<q+p_b-1$, so $\delta_b>0$), and $(\fs+1)/g_k(M)\to0$ is
eventually at most $\delta_b$, we get
$m/(k-\fs)\le p_b-\delta_b$ for all large $M$, and
Proposition~\ref{S-prop:live} applies;
Proposition~\ref{S-prop:pubsafe}(ii) restates the same event as the
publisher penalty.
(iii) By Lemma~\ref{S-lem:distinct}:
$A(M)=\bigO(\Emax\ctool^{2}(N(M)+\cacc f(M))(1+\kappa(M)))=o(M^{1-\beta})$
by the margin schedules, the cross-term condition, and $\kappa=\Theta(k)$, so the first term
is $o(M^{1-\beta})\cdot\bigO(M^{\beta-1})=o(1)$; the second term is
$\bigO(\Emax\,\kappa(M)\,\gbar(M))\,d=o(1)$ by the shrinking
front-loading condition and the bounded mass (\ref{S-a:massbound});
the residual clause of Theorem~\ref{S-thm:sat}(iii) applies by the
coverage-rate hypothesis $D_0(M)\,V(M)=o(M)$ assumed in the
statement, with $\varepsilon(M)\,d_{\mathrm{good}}(M)\to0$ from
$\varepsilon(M)\to0$ (\ref{S-a:promsuf}) and \ref{S-a:massbound}.
\end{proof}

\begin{remark}[Role of the saturation exponent]\label{S-rem:beta}
Smaller $\beta$ is uniformly favourable: faster budget decay
$M^{-(1-\beta)}$ \emph{and} a weaker Sybil requirement
$f(M)=o(M^{1-\beta})$. There is no design trade-off along $\beta$: it
is not a design parameter but an empirical property of the community's demand
stream, to be measured (\ref{S-n:validate});
Theorem~\ref{S-thm:det}, which references neither $\beta$ nor the demand
model, remains available at any
$\beta$ under its own hypotheses (Remark~\ref{S-rem:detscope}). A worked Zipf instantiation, including the tail, is
\S\ref{S-app:zipf}.
\end{remark}

\section{Design Rationale, Security Analysis, and Scope
  \texorpdfstring{\ \normalfont\normalsize(informative)}{ (informative)}}
\label{S-app:shield}

Nothing in this appendix is normative and no proof depends on it
(\S\ref{S-app:roadmap}). \S\ref{S-app:notes} collects design rationale;
\S\ref{S-app:objections} analyses anticipated attacks and limitations;
\S\ref{S-app:residuals} states scope
and residuals; \S\ref{S-app:composed} summarises the composed guarantees.

\subsection{Attack and limitation index}
\label{S-app:objindex}

Each entry of \S\ref{S-app:objections} carries a \emph{disposition}:
\emph{disposed} (under stated assumptions; a formal fact disposes of it), \emph{residual} (a
genuine limitation, disclosed in \S\ref{S-app:residuals}), or mechanism disposed
with a disclosed residual periphery; and a \emph{proof
role}: not used in any proof, or enters budget constants
only (Figure~\ref{S-fig:dag}).

\begin{table}[h]
\centering\footnotesize
\renewcommand{\arraystretch}{1.0}
\begin{tabular}{@{}lp{0.155\textwidth}p{0.30\textwidth}p{0.30\textwidth}@{}}
\toprule
\textbf{Entry} & \textbf{Vector / limitation} & \textbf{Governing
fact(s)} & \textbf{Disposition; proof role}\\
\midrule
\multicolumn{4}{@{}l}{\emph{Endorsement-mechanism attacks
(\S\ref{S-app:an-endorse})}}\\
\ref{S-q:paraphrase} & paraphrase-variant families &
Cor.~\ref{S-cor:submission}, \ref{S-r:share}; $q$ calibration &
mechanism disposed $+$ residual; unused\\
\ref{S-q:selfdispute} & self-dispute &
Thm.~\ref{S-thm:panel}(b), \ref{S-inv:verdict} & disposed (under stated assumptions); unused\\
\ref{S-q:farm} & reputation farming & lifetime totals $f,\Wadv,\fs$
(\ref{S-n:farm}) & disposed (under stated assumptions); unused\\
\ref{S-q:tooltarget} & tool-targeted poison &
Thm.~\ref{S-thm:panel}(c) ($\qpan$ scope), Rule~\ref{S-r:panel}(iv), Lem.~\ref{S-lem:endorse} &
ordinary path disposed $+$ panel-path residual; unused\\
\ref{S-q:lapse} & panel-lapse exploitation, slot economics &
Lem.~\ref{S-lem:freeze}(iii) & disposed (under stated assumptions); unused\\
\ref{S-q:exhaust} & juror exhaustion by pre-exposure &
Lem.~\ref{S-lem:freeze}(ii), Rem.~\ref{S-rem:freezeread} & safety disposed (under Thm.~\ref{S-thm:panel}'s hypotheses) $+$ availability residual; unused\\
\midrule
\multicolumn{4}{@{}l}{\emph{Side channels (\S\ref{S-app:an-side})}}\\
\ref{S-q:quota} & store-exhaustion channel & convention~(\ref{S-conv:c7}),
\ref{S-r:nobit} & disposed (under stated assumptions), conditional on (\ref{S-conv:c7}); unused\\
\ref{S-q:timing} & timing channels & scope statement,
Lem.~\ref{S-lem:cflow}(e) & residual; unused\\
\midrule
\multicolumn{4}{@{}l}{\emph{The human factor (\S\ref{S-app:an-human})}}\\
\ref{S-q:preview} & preview manipulation &
Lem.~\ref{S-lem:isolation}, \ref{S-r:display}, $\DT$ coverage &
mechanism disposed $+$ residual; unused\\
\ref{S-q:postendorse} & careless endorsement; post-endorsement
egress & Thm.~\ref{S-thm:action}, Cor.~\ref{S-cor:noexfil}(b), Cor.~\ref{S-cor:confine},
Prop.~\ref{S-prop:pubsafe} & residual; unused\\
\ref{S-q:recombine} & recombination of endorsed values &
Lem.~\ref{S-lem:isolation}--\ref{S-lem:cflow}, \ref{S-o:one} &
residual; unused\\
\ref{S-q:batch} & batch-gate attention & \ref{S-r:batchgate},
$\cbatch$ sizing & residual; unused\\
\midrule
\multicolumn{4}{@{}l}{\emph{Adverse parameter regimes
(\S\ref{S-app:an-param})}}\\
\ref{S-q:corr} & cross-person correlated carelessness &
$q_c$-degradation of Thm.~\ref{S-thm:panel}; $\Dpool$; re-verification (\ref{S-n:revoke}); Rem.~\ref{S-rem:detscope} &
residual; unused\\
\ref{S-q:qzero} & unrecognizable poison ($q\to0$) &
one-effective-qualifying-rejection freeze (\ref{S-r:panel}(i),(v)); \ref{S-inv:fresh}; Rem.~\ref{S-rem:detscope} & residual; unused\\
\ref{S-q:calib} & accountability mis-calibration &
Rule~\ref{S-r:panel}(v)(a),(e); \cref{S-n:qpos} & safety disposed $+$
residual; constants
only\\
\midrule
\multicolumn{4}{@{}l}{\emph{Modelling-boundary effects
(\S\ref{S-app:an-model})}}\\
\ref{S-q:incentive} & incentive feedback on publication &
Thm.~\ref{S-thm:floor} unaffected & residual; unused\\
\ref{S-q:spike} & release-day demand spikes &
Thm.~\ref{S-thm:floor}; item-locality of \ref{S-a:smooth} &
disposed (under stated assumptions); unused\\
\ref{S-q:handlecover} & handle-mediated arrival of certified content &
Thm.~\ref{S-thm:floor}; \ref{S-a:promsuf} ($\varepsilon$);
Rule~\ref{S-r:whitelist}(c) & residual; budget assumptions only\\
\bottomrule
\end{tabular}
\caption{Attack and limitation index. ``Disposed'' means disposed of
by the cited formal facts \emph{under their stated assumptions};
entries with a residual periphery say so.}
\label{S-tab:objindex}
\end{table}
\FloatBarrier

\subsection{Design rationale}
\label{S-app:notes}

Four rationale remarks whose content adjoins specific formal
results (the freshness asymmetry, the reading of $q$, the
accountability calibration, and the roles of the protocol
clocks) are set as numbered Remarks adjacent to their points of use
(Remarks~\ref{S-n:freshness}, \ref{S-n:qpos}, \ref{S-n:calib},
\ref{S-n:roles}) and are indexed here for completeness. The notes below
record the remaining design rationale.

\begin{notex}[Model lineage]\label{S-n:lineage}
The target property is an integrity property: content from an untrusted
source must not drive a sensitive action unless the user explicitly
authorised the exact value involved. Integrity is the lattice dual of
Denning's confidentiality flow \cite{denning1976lattice,denning1977certification}, made
precise by Biba \cite{Biba}; the behavioural definition is
noninterference \cite{goguen1982security}, proved by the unwinding method
\cite{goguen1984unwinding} transported to a small-step operational semantics
\cite{volpano1996sound,sabelfeld2003language}. Two refinements distinguish PerMIT. First, PerMIT
\emph{does} permit untrusted-derived values to reach sensitive actions,
but only after a user confirmation; a confirmation is an
\emph{endorsement} \cite{myers2006enforcing}, and the danger that an attacker
controls \emph{what} is endorsed is ruled out by robustness
\cite{zdancewic2005robust,myers2006enforcing}. Second, PerMIT lets verified crystallized knowledge
inform the Protected LLM's reasoning; the right noninterference
statement is therefore one modulo the release history, in the
delimited- and gradual-release tradition
\cite{sabelfeld2009declassification,sabelfeld2004model,askarov2007gradual}, with the display transcript
playing the role of the release-event history. The carve-out is the
\emph{full} transcript: candidate previews as well as decision-pane
displays, rejections as well as acceptances
(Definition~\ref{S-def:transcript}). The substrate, including the
provenance-tracking interpreter and the separation of a trusted
orchestrating model from a quarantined reader, is CaMeL \cite{debenedetti2025defeating};
we name these the Protected LLM ($P$) and the Quarantined Reader LLM
($Q$), realizing $Q$ as a Recursive Language Model \cite{zhang2025recursive}. Related
efforts: AgentDojo \cite{debenedetti2024agentdojo}, information-flow control for
agents \cite{costa2025securing}, memory-poisoning defenses \cite{wei2025memguard},
collaborative memory \cite{rezazadeh2025collaborative}. CaMeL leaves formal verification as future work \cite[\S10]{debenedetti2025defeating}. PerMIT leaves planning and tool choice to the agent while bounding its observations and data flows through the finite interface of Figures~\ref{S-fig:sos}--\ref{S-fig:sos2}, including tool-output capture and argument mediation. The channel analysis (Lemmas~\ref{S-lem:iface}--\ref{S-lem:cflow}) checks this written specification clause by clause.
\end{notex}

\begin{notex}[The endorsement unit]\label{S-n:unit}
Endorsement is per (value, tool) and cached indefinitely while the relevant local copy is retained. A conforming instantiation of the requirements-level Option~\ref{S-o:three} would bound its scope-checked tool-authorization lifetime, but is outside the proved model. Combinations, orderings, and timings of separately endorsed values are ungated. The policy does not bind argument position, task, purpose, or the complete argument tuple. Reasoning Integrity excludes additional hidden influence when the complete display transcript is fixed; it does not certify the safety of later combinations (\ref{S-q:postendorse}, \ref{S-q:recombine}). Options~\ref{S-o:one}--\ref{S-o:two} refine tool-scoped checks, not complete-transaction approval.
\end{notex}

\begin{notex}[The reader asymmetry]\label{S-n:readers}
$Q$ may read the whole store because its output is always untrusted or
handle-hidden; $\RE$ may read only the endorsed store because its
output flows to $P$ as readable endorsed bytes. Neither can be relaxed
without breaking a theorem: a $\U$-reading $\RE$ violates
Theorem~\ref{S-thm:reason}; an $\E$-emitting $Q$ violates
Lemma~\ref{S-lem:endorse}.
\end{notex}

\begin{notex}[Failure indistinguishability, operationally]\label{S-n:fail}
Under Rule~\ref{S-r:nobit}, the agent cannot distinguish ``the untrusted
store contained nothing useful'' from ``the reader failed while
looking'': both worlds return the same operation-determined handle, so
neither the contents nor the volume of untrusted memory can signal
through the failure path. Likewise, whether an identifier is occupied
by a $\U$-block is a fact the attacker influences through the number of
blocks captured per call, which is why occupancy is never observable
(Rule~\ref{S-r:iface}(b)). The store-quota policy of
convention~(\ref{S-conv:c7}) exists because a volume-dependent error would reopen
exactly this channel (\ref{S-q:quota}).
\end{notex}

\begin{notex}[The human channel]\label{S-n:human}
\ref{S-a:exo} idealizes the user's typing as exogenous. In reality,
displayed content can steer what the user subsequently types, which
enters as trusted input. Influence of displayed content on typing is
accounted for exactly insofar as that content appears in the display
transcript, which, by Rule~\ref{S-r:display}(e), records \emph{every}
attacker-influenceable rendering the user sees, on both surfaces. The
design therefore discharges the idealization's practical content by
logging both surfaces: the release history covers every surface through
which such influence can pass, so Corollary~\ref{S-cor:confine} and
\ref{S-n:audit} remain faithful even where \ref{S-a:exo} frays. The
residual is disclosed at \ref{S-q:preview} and
\S\ref{S-app:residuals}.
\end{notex}

\begin{notex}[The batch gate is all-or-nothing]\label{S-n:batch}
The selected batch design is all-or-nothing: a per-page decision
returned to $P$ would leak the attacker-influenceable page count
(Rule~\ref{S-r:batchgate}(c)), and the aggregate decision avoids that
channel at the cost that one poisoned element forces rejection of the
whole batch --- an availability and attention cost, not a safety gap.
Sufficiency of this design is what is established; no impossibility of
alternative fixed-shape or padded designs is claimed. The user may reject at any page,
aborting the remainder (Rule~\ref{S-r:batchgate}(d)), so
attacker-imposed attention is bounded by the user's own patience, never
compelled. Both facts argue for sizing $\cbatch$ conservatively
(\ref{S-q:batch}).
\end{notex}

\begin{notex}[The transcript is recorded, not prevented]\label{S-n:audit}
The harness records every display event, selection renderings
included, in a hash-chained audit log, a tamper-evident
materialization of $\DT$. Corollary~\ref{S-cor:confine} therefore has operational force: the
transcript supplies a complete record of the modeled release
surfaces relevant to the protected behaviours --- ungated sinks,
Protected-LLM operations, tool invocations
(Definition~\ref{S-def:reason}) --- missing no surface on which
attacker-shaped content was ever rendered; it does not by itself
reconstruct causation, timing, unreleased tool outputs, or attacker
identity. This is not a proof that the user cannot be manipulated;
it is a proof that any attacker influence on those behaviours must
pass through the logged display transcript or an explicitly
accepted endorsement. Logging makes influence visible and
attributable after the fact --- attributable to a recorded release
or decision, not necessarily to an attacker identity; the
prevention half of the design is Theorem~\ref{S-thm:action}'s gate,
which stops any value that is neither user-typed nor endorsed
(personally, or, under delegation, by a qualifying community
quorum) for the exact sensitive tool from ever crossing that
boundary (argument-value integrity: the decision to invoke a tool,
and zero-argument sensitive calls, are scoped at
\S\ref{S-app:residuals}(X)).
\end{notex}

\begin{notex}[What previews can and cannot do]\label{S-n:preview}
An adversarial reader can decorate a malicious candidate with a
benign-looking preview. Within the formal model, previews are inert:
they never enter $\sigma_P$ (Lemma~\ref{S-lem:isolation}), and
endorsement binds to the true resolved bytes on the decision pane
(\ref{S-a:display}--\ref{S-a:user}), so a preview can never forge an
endorsement. Outside the formal model, a preview can influence the
human: what the user chooses to examine, and, through
\ref{S-a:exo}'s idealization, what the user subsequently types. This is
precisely why previews are included in the display transcript and the
audit log (\ref{S-q:preview}).
\end{notex}

\begin{notex}[What the gate does to the detection task]\label{S-n:gatetask}
The confirmation gate reshapes the detection task an honest inspector
faces, in three architectural ways. First, stealth is eliminated by
construction: Rule~\ref{S-r:display} renders the exact bytes that will
flow, in full, on every gated invocation, so no gated content can
reach a sensitive action unseen --- there is no channel by which
poison avoids display altogether. Second, hijacking-style content ---
bytes written to be read as instructions rather than data --- is
surfaced on the decision pane out of the narrative context it was
designed to exploit, isolated against a fixed provenance zone
(\ref{S-a:display}--\ref{S-a:user}); this plausibly raises $\qpre$ for
that class of poison, but it is offered as motivation for the
parameter, never as a proof of it: $\qpre$ remains an empirical,
weakest-target quantity (\cref{S-n:qpos}), estimated by logging, not
derived from the display design. Third, because every gate decision
is logged (Definition~\ref{S-def:transcript}), $q$ is made
log-measurable rather than merely posited: the validation protocol of
\ref{S-n:validate} is only possible because the architecture records
the events it needs. None of this touches \emph{semantic} poison ---
content that is simply false or misleading, with no injection
mechanism for display to expose --- which gains nothing from being
shown out of context and remains exactly the disclosed residual of
\ref{S-q:qzero}, consumed by no proof in this document.
\end{notex}

\begin{notex}[Federated lineage]\label{S-n:fedlineage}
The federated tier is built on machinery network security has already
made rigorous: threshold endorsement from the PGP web of trust
\cite{maurer1996modelling}; transparency and revocation from Certificate
Transparency \cite{laurie2013certificate,laurie2021certificate}; the identity-cost requirement of the
Sybil literature \cite{douceur2002sybil,dwork2007pricing} , the precise
reason the scaling results are conditional on an attested or
cost-bounded setting. The panel's competence condition is the classical
jury condition \cite{grofman1983thirteen,allen2021scaling,martel2024crowds}.
\end{notex}

\begin{notex}[What pool diversity buys]\label{S-n:diversity}
The requirement $\Dpool$ caps per-pool verdicts and mixes the
fresh-inspector population, attacking correlated blind spots and
demand-side selection bias among panelists. Diversity bounds
\emph{correlation}, not the \emph{minimum}: since the adversary chooses
whom to present the item to (\S\ref{S-app:prob}), $q$ remains a uniform
lower bound over the honest identities the adversary can target, and
the pool caps of Definition~\ref{S-def:pools} force those targets to span
$\Dpool$ pools rather than one cluster, raising the effective bound
from the most fallible cluster toward the most fallible identity per
pool, never to a population average. The per-person correlation across
\emph{logged} re-inspections is discharged structurally by
fresh-inspection counting (\ref{S-inv:fresh}); the residual across
unlogged personal inspections is a calibration matter
(\cref{S-n:qpos}, Definition~\ref{S-def:fresh}, \ref{S-q:corr}).
\end{notex}

\begin{notex}[Reputation farming]\label{S-n:farm}
$f$, $\Wadv$, and $\fs$ are lifetime totals over every identity the
adversary ever controls (\S\ref{S-app:fedcore}), capped by the
attestation layer's issuance; farming weight before spending it
changes \emph{when} the capacity exists, never its total, so no bound
stated against the lifetime totals can be outrun. Publisher penalties
(Rule~\ref{S-r:panel}(vii)) only shrink the adversary's capacity below
these totals, so every bound stated against them remains valid. The
economic deterrent of burning accrued reputation is real but
deliberately unused in any proof: the proofs assume the attacker
spends everything (\ref{S-q:farm}).
\end{notex}

\begin{notex}[The $\mathsf{evict}$ residual]\label{S-n:evict}
$\mathsf{evict}$ is owner-scoped but $P$-invocable and ungated, so an
\emph{erring} (not attacker-steered, which
Lemmas~\ref{S-lem:isolation}--\ref{S-lem:cflow} exclude) Protected LLM can
remove its owner's endorsed blocks without a confirmation; the
no-deletion guarantee of Lemma~\ref{S-lem:nodelete} is a third-party
guarantee only. Deployments wanting first-party cover may gate
$\mathsf{evict}$ on $\E$-labelled blocks like any sensitive call, a
further monotone hardening in the sense of
Proposition~\ref{S-prop:harden}. The uniform-$\mathsf{ack}$ totalisation of $\mathsf{evict}$
(Figure~\ref{S-fig:sos2}, convention~\hyperref[S-sos:v]{(v)}) changes only the reply
surface, never this capability: the residual is unchanged, and the
realised outcome remains visible to the owner in the audit log.
The residual includes evictions $P$ performs \emph{in reaction to}
legitimate federation events (notifications are $\E$-level log data): a
third party can occasion such a reaction but not steer it ---
Lemmas~\ref{S-lem:isolation}--\ref{S-lem:cflow} exclude any $\U$-derived
influence on the choice --- so this is erring-agent behaviour, outside the
threat model (\S\ref{S-app:residuals}(V)), not an attacker channel.
\end{notex}

\begin{notex}[Availability degrades gracefully, never destructively]
\label{S-n:avail}
All federated guarantees are safety guarantees; availability is
explicitly not guaranteed, but its loss is bounded \emph{in harm} (in required confirmations) per
dispute: under \ref{S-a:rhoseat}, at most $\kappa=(k{+}1)/\rseat$ affected
demander arrivals in expectation, with exponential tail
(Lemma~\ref{S-lem:geo}); if \ref{S-a:toolspread} is additionally read
over the entire unresolved-dispute demand horizon, at most
$\ctool\,\kappa$ additional required confirmations in expectation,
with the corresponding tail at thresholds scaled by $\ctool$; under
the stated per-epoch reading of \ref{S-a:toolspread}, only the
corresponding per-epoch conversion is asserted, while the pathwise
per-user cap remains valid for a freeze of any duration
(Lemma~\ref{S-lem:freeze}(ii)); pathwise, independent of every federated and demand assumption (persistence exceptions per \ref{S-inv:floor}), no affected
user exceeds the single-user cap, regardless of the freeze's
duration (Lemma~\ref{S-lem:freeze}(ii)); recoverable (low wrongful-eviction
risk, Proposition~\ref{S-prop:live}); accountable on both sides
(Rule~\ref{S-r:panel}(v),(vii)); and never destructive
(Lemma~\ref{S-lem:nodelete}): in the worst case the shared tier
contributes nothing and each user verifies personally, within the single-user cap of
Theorem~\ref{S-thm:floor}. The per-epoch free charge renews, so the
adversary retains a per-epoch throughput bounded in the three
currencies of Lemma~\ref{S-lem:freeze}(iv); what it buys is bounded
required-confirmation harm and bounded per-epoch harassment throughput
(Lemma~\ref{S-lem:freeze}(ii),(iv)), never deletion --- the
\emph{wall-clock duration} of an unresolved freeze is deliberately
unbounded, resolution being demand-paced with no deadline
(Remark~\ref{S-rem:resolution}); its harm, not its duration, is what
is capped. A promoted benign item can be
harassed only through Suspended-state fast-track panels, at most
$\Emax$ per item per epoch (Rule~\ref{S-r:panel}(vi)) and
penalty-escalating. An unpromoted benign candidate
cannot be retired by sequential vindicated disputes
(Rule~\ref{S-r:hygiene}(i)(b) exempts them, \ref{S-n:vindexempt}): each
adversarial rejection buys one freeze, resolved at the item's own
demand rate, while occupying a rejection slot and, beyond the first
per identity per epoch, escalating the rejecter's accountability; the
adversary's per-epoch freeze throughput against any item being bounded
(Lem.~\ref{S-lem:freeze}(iv)), what it buys is bounded
required-confirmation harm within the single-user cap --- of
unbounded wall-clock duration, the freeze being demand-paced ---
never retirement, and the single-initiation
clause caps any identity, adversarial or honest, at one freeze per
item per epoch, so per-item harassment throughput is at most $f$
freezes per epoch deterministically. The entry cap of Rule~\ref{S-r:hygiene}(i) is likewise
availability-only: an item whose per-epoch entries are exhausted
(e.g.\ through repeated wrongful evictions, each already
exponentially rare, Proposition~\ref{S-prop:live}) waits for the epoch
boundary with its banked credits and any live promotion status
intact (convention~(\ref{S-conv:c4})), each demander meanwhile paying at most the
cap (Theorem~\ref{S-thm:floor}(i)). Under this invalidating semantics, a suspension additionally converts its
demanders' ungated reuse into at most one required confirmation each ---
the cap confirmation they would in any case owe on demotion
(Corollary~\ref{S-cor:floorE}) --- after which the pair is permanently
covered by a local $\C$ copy; this is priced once per demander and never
renewed.
\end{notex}

\begin{notex}[The vindication exemption]\label{S-n:vindexempt}
Rule~\ref{S-r:hygiene}(i)(b) counts only \emph{non-vindicated} disputes
toward expiry. The exemption is what prevents both an adversary's
wrongful rejections (each vindicated, each feeding the accountability
schedule of Rule~\ref{S-r:panel}(v)) and the honest wrongful-rejection
noise, which arrives at rate $1-p_b$ per benign fresh inspection and
would otherwise, at the co-scaling thresholds $N(M)\to\infty$ of
Theorem~\ref{S-thm:coscale}, retire \emph{every} item before promotion
became arithmetically reachable, from terminating a benign item. Under
the default bars every resolved dispute ends in vindication or eviction
and every unresolved one ends in expiry, so this trigger fires only
through the optional neither-bar outcome of
Rule~\ref{S-r:panel}(iv); it remains availability-only and conservative
(Lemma~\ref{S-lem:hygienemono}).
\end{notex}

\begin{notex}[Revocation, re-verification, and grandfathering]
\label{S-n:revoke}
Post-promotion suspensions and re-verification add panel and
re-inspection events but never multiply any user's required cost
(Corollary~\ref{S-cor:floorE}): local endorsements are permanent, so
revocation campaigns tax the shared tier's voluntary and verdict
effort, bounded per suspension by Rule~\ref{S-r:panel}(vi) and the rate
limits; a prior $\CC$-only user additionally owes their first-ever
required confirmation of the pair --- a cost \emph{within} the cap,
already counted inside $d^{\infty}_{\mathrm{tool}}$
(Corollary~\ref{S-cor:floorE}), never beyond it. Re-verification draws its inspectors from
identities with no prior \emph{logged} decision on the bytes (log-fresh,
Definition~\ref{S-def:fresh}; a cohort that can include previously
exposed but never-logged inspectors, \cref{S-n:qpos}), so the probability of a
uniformly convincing poison remaining uncaught across $\Erv$
re-verification rounds, each reaching a cohort of new inspectors, is at
most $(1-q_c)^{\Erv}\to0$, with $q_c$ the correlated catch rate of
\ref{S-q:corr}; the mechanism's strength scales with the
supply of never-decided inspectors (population growth and new demand),
not with epoch renewal. Grandfathering: items promoted at lower
historical thresholds retain $\CC$ but enter the re-verification queue
until their endorsement meets the current bar. Within the formal semantics, first-party eviction aside
(\ref{S-n:evict}), nothing re-forces a confirmed pair's authorization;
a conforming instantiation of the requirements-level
Option~\ref{S-o:three} would re-force exactly the scope-checked tool
authorization on lease expiry --- never reasoning readability ---
an explicit trade stated and to be priced with that extension
(\ref{S-o:three}).
\end{notex}

\begin{notex}[Validation without a user study]\label{S-n:validate}
Theorem~\ref{S-thm:floor} and the counting cores of
Lemmas~\ref{S-lem:budget},~\ref{S-lem:distinct} are identities. Saturation
($V(M)=o(M)$), fractional temporal smoothness (\ref{S-a:smooth}: log
each item's realised front-loading fraction $\gamma_v$, validated
specifically on the most-demanded items, where population averages
would hide spikes), coverage sufficiency (\ref{S-a:promsuf}: log each item's realized
residual fraction $U_{v,t}/n_{v,t}$ \emph{and} the
fast-path-vs-gate split of live-$\CC$ demand --- the handle-arrival
fraction, \ref{S-q:handlecover} --- stratified on the most-demanded
items, with per-item time-to-promotion against accumulated demand), tool-spread
(\ref{S-a:toolspread}: log per-item distinct-tool counts), bounded demand mass (\ref{S-a:massbound}: log per-user distinct
shareable-item demand per epoch across deployment growth), and the
verdict rate (\ref{S-a:rhoseat}: log verdict-eligibility rates on
disputed items) are modelling assumptions checked by logging in the
prototype and by replaying the promotion and demand-paced panel
mechanics in simulation while sweeping the growth exponent, a replay
of an algorithm, not an experiment on people. The replay's own fidelity is checked independently of any behavioural parameter by the closed form of \S\ref{S-app:blastlaw}, whose support-level
consequences are byte-identity properties of the implementation rather than statistical agreements.
One caveat governs the behavioural parameters specifically: $\qpre$, $\qpan$, $p_b$,
and $\rseat$ are \emph{uniform conditional} lower bounds, each validated on its own stratum (gate inspections vs.\ eligible verdicts), on the
prior-atom $\sigma$-algebras of \S\ref{S-app:prob}; aggregate log rates do not
directly estimate such bounds. Their validation therefore reports
stratified rates (by identity class, presentation context, and flag
state) rather than population averages, uses lower confidence bounds rather than point
estimates, corrects for adaptive selection of inspection targets, and
monitors nonstationarity across epochs; the estimand is the infimum
over strata, and the deployed constants ($N,k,m$ margins,
$\thacc$, $\cceil$) are set against that infimum with margin, never against
a mean.
These stratified, lower-confidence-bound estimates are \emph{falsification
and calibration} evidence: they can refute an assumed cap --- a stratum
whose observed rate falls below the claimed bound --- and can inform
recalibration, but they do not \emph{certify} the uniform
infimum-over-history bounds $\qpre,\qpan,p_b$ require. Sparse or
unobserved strata (rare identity classes, rare presentation contexts) are
not covered by any finite sample, and the uniform bounds remain modelling
hypotheses, not measured quantities.
One further caveat: the classification $\mathsf{Bad}$ is exogenous
(Definition~\ref{S-def:bad}), so the log alone cannot label a trial as
poisoned or benign; estimating $\qpre$, $\qpan$, and $p_b$ therefore
requires an exogenously labeled corpus --- red-team--injected
known-poison items and expert-adjudicated benign samples --- with the
stratified lower-confidence-bound machinery applied to the labeled
trials. Unlabeled log rates validate the protocol mechanics only
($\rseat$, realised $\gamma_v$, $U_{v,t}/n_{v,t}$, tool spread,
demand mass).
\end{notex}

\begin{notex}[Total effort, including rejections]\label{S-n:effort}
The ledger of Lemma~\ref{S-lem:budget} covers every \emph{counted} (required or voluntary) durable accepting pair-confirmation of shareable items, redundant covered-gate acceptances and the personal, per-task, and option-induced streams excluded or separately charged (\S\ref{S-app:budgetmodel}, Corollary~\ref{S-cor:flooropt}); rejection and attention effort is a
distinct resource with a two-sided account. Under the durable
semantics of Rule~\ref{S-r:atomic} there are no
accepted-but-uncommitted confirmations: every accepting review
commits its copy. What remains uncounted is attention: rejected
dialogs, abandonment, batch-page review, repeated reviews of an
already-accepted pair within one transaction (counted once,
Definition~\ref{S-def:sinkconf}), and redundant covered-gate decisions.
\emph{Protocol-effective rejections} are tightly capped: one
dispute-initiation per identity per byte string per epoch
(Rule~\ref{S-r:panel}(i)), at most $\rmax$ open slots, and the
accountability schedule of Rule~\ref{S-r:panel}(v). \emph{Rejection
dialogs}, by contrast, recur: a user who rejects an uncovered pair holds
no local $\C$ copy, so each later attempted use of that pair fires the
gate again, and the later rejections are logged as corroboration or
advisory flags (Rule~\ref{S-r:panel}(i)) --- the \emph{protocol} does not
bound a user's total rejection dialogs on one value. What bounds them is
the user's own use stream and the exposure caps: per retrieval at most
$\ccand$ candidates are surfaced (Rule~\ref{S-r:hygiene}(ii)), disputed
items are flagged and ordered last, paraphrase spam consumes the
submitter's rate limit per byte-distinct variant (Rule~\ref{S-r:share},
Corollary~\ref{S-cor:submission}), and a $P$-emitted $\conf$ or handle-expressed tool call on a
\emph{covered} pair fires a covered gate, protocol-inert on both
branches (Rule~\ref{S-r:panel}(i)), reasoning-target repetition on locally readable bytes included (reasoning coverage, Rule~\ref{S-r:panel}(i)): the resulting redundant
confirmations and rejections have no shared-tier effect and are
excluded from every ledger count (\S\ref{S-app:budgetmodel}), so agent
repetition on covered values costs the user attention only, disclosed
here --- and $P$ \emph{can} re-emit $\conf$ on a repeatedly-rejected,
\emph{uncovered} handle every step: covered-gate inertness neutralises
covered values only, by design. Repeated
dialogs arising from the user's own repeated attempts at an
\emph{uncovered} pair are a first-party cost of declining, of the same
character as the $\mathsf{evict}$ residual (\ref{S-n:evict}): attacker-steered
repetition is excluded (Lemmas~\ref{S-lem:isolation}--\ref{S-lem:cflow}), but
the model places no constant on it. Accordingly, every ``bounded-harm''
statement in this document is a statement about \emph{required user approvals} (Theorem~\ref{S-thm:floor}), with inspection effort disclosed
here.
One deployment device reduces this attention cost without touching
the semantics: the decision pane may visibly badge a covered gate ---
the resolved bytes byte-match a local in-scope copy, an owner-side
deterministic fact (Rule~\ref{S-r:panel}(i)) --- and offer a one-tap
accept; the gate, its display, and its logged decision are unchanged.
Interpreter-level rate-limiting of repeated uncovered gate requests is
a candidate further hardening, monotone in the sense of
Proposition~\ref{S-prop:harden}; it is an operational-semantics change
requiring its own transition, transcript, and two-run treatment, and
is not specified in this document.
\end{notex}
\subsection{Security and robustness analysis}
\label{S-app:objections}

Each entry states an attack vector or failure mode in its strongest
form (\emph{Vector}), the facts that govern it (\emph{Disposition}),
and its \emph{Status} per \S\ref{S-app:objindex}. Entries are grouped by
attack surface.

\subsubsection{Attacks on the endorsement mechanism}
\label{S-app:an-endorse}

\begin{objx}[Paraphrase-variant families]\label{S-q:paraphrase}
\emph{Vector.} Freshness is byte-keyed
(\ref{S-inv:fresh}). An adversary submits paraphrase variants of the
same poison as distinct byte strings; every inspector, including one
already fooled by a sibling variant, is fresh again on each variant,
so the per-person discharge of correlation is byte-scoped, not
content-scoped.
\emph{Disposition.} The discharge of per-person correlation is indeed
\emph{per byte identity} (convention~(\ref{S-conv:c5})), and the arithmetic is
unchanged: each variant is a distinct item consuming the submitter's
rate limit and its own union-bound term (Rule~\ref{S-r:share},
Corollary~\ref{S-cor:submission}), and per-retrieval exposure is capped
at $\ccand$ (Rule~\ref{S-r:hygiene}(ii), \ref{S-n:effort}). What survives
is an \emph{interpretation} obligation on the deployer: $q$ inside
each union-bound term must be calibrated as the catch rate against
the \emph{variant family}, not against a single presentation; an
inspector fooled by one variant is plausibly likelier to be fooled by
its siblings. Deployments may additionally surface an advisory-only
similarity flag (``similar disputed or evicted items exist'')
alongside disputed items, of the zero-authority advisory class of
Rule~\ref{S-r:panel}(i): it informs inspectors without granting the
model computing similarity any protocol effect.
\emph{Status.} Mechanism disposed; calibration duty disclosed. Not
used in any proof.
\end{objx}

\begin{objx}[Self-dispute]\label{S-q:selfdispute}
\emph{Vector.} Reject one's own item; fill the panel with one's own
verdicts; sustain the dispute across epochs via one's own demand events.
\emph{Disposition.} The case-(b) bound of Theorem~\ref{S-thm:panel} does not
condition on the rejecter's identity, so self-dispute gains only the panel
term already counted, while consuming a rejection slot and accountability
exposure; verdict capacity never renews (\ref{S-inv:verdict}), capping the
adversary's tally at $\fs$ for the dispute's life; self-initiations are
capped at one per identity per item per epoch (Rule~\ref{S-r:panel}(i)).
\emph{Status.} Disposed (under stated assumptions; Theorem~\ref{S-thm:panel}, \ref{S-inv:verdict}). Not
used in any proof.
\end{objx}

\begin{objx}[Reputation farming]\label{S-q:farm}
\emph{Vector.} Farm attestation weight and reputation over time; spend it
in one campaign.
\emph{Disposition.} $f$, $\Wadv$, $\fs$ are lifetime totals over every
identity the adversary ever controls (\S\ref{S-app:fedcore}); every bound is
stated against the totals, so farming shifts timing, never capacity
(\ref{S-n:farm}). The economic deterrent of burning reputation is real but
deliberately unused: the proofs assume the attacker spends everything.
\emph{Status.} Disposed (under stated assumptions; lifetime-total definition). Not used in any proof.
\end{objx}

\begin{objx}[Tool-targeted poison]\label{S-q:tooltarget}
\emph{Vector.} Disputes are byte-level but harm is tool-level: a poison
dangerous at tool $A$ and innocuous at tool $B$ is sincerely vindicated by
a panel drawn from $B$-demanders, contradicting honest $A$-rejecters; the
freeze lifts and the interrupted entry resumes.
\emph{Disposition.} The two promotion paths behave differently.
\emph{Ordinary path at $A$: disposed.} Promotion at $A$ still requires $N$
credited fresh endorsements at $A$ itself (Lemma~\ref{S-lem:endorse},
Definition~\ref{S-def:counting}), where inspectors face the harmful use, so
the case-(a) term of Theorem~\ref{S-thm:panel} applies with $\qpre$ read at
$A$. \emph{Panel path: residual.} The panel term is \emph{not} bounded by
$A$-context competence: $\qpan$ is uniform over the jurors' own tool
contexts (Theorem~\ref{S-thm:panel}(c)), and for $A$-targeted, $B$-benign
poison it may be near zero --- sincere $B$-juror vindication is then
likely, the freeze lifts, and the entry resumes toward its $A$-threshold.
What remains against $A$-promotion is the ordinary-path term, plus
renewable re-freezes: each epoch, each honest identity's first rejecting
decision on the bytes --- in particular each new $A$-context detection ---
re-freezes promotion at every tool (Rule~\ref{S-r:panel}(i)--(ii)).
\emph{Accountability erosion.} Honest $A$-rejecters contradicted by
$B$-vindications feed Rule~\ref{S-r:panel}(v); mitigation is threefold: the
free first charge per epoch, the per-fresh-inspection rate trigger
calibrated above the honest baseline (\cref{S-n:calib}), and conversion to
zero-authority advisory flags at the ceiling. We deliberately do \emph{not}
exempt rejecters whose vindicating jurors sat at a different tool: because
a rejection at any tool freezes all tools (Rule~\ref{S-r:panel}(ii)), such an
exemption would let an adversary file every rejection ``at'' an obscure
tool no juror will ever demand, harvesting freezes with zero accountability
exposure and breaking the throughput bounds of Lemma~\ref{S-lem:freeze}(iv).
The per-pair scoping variant of Remark~\ref{S-rem:pairpanel} removes this residual; it is a deployment choice, consumed by no theorem.
\emph{Status.} Ordinary-path promotion at the targeted tool disposed;
panel-path vindication under cross-tool juries residual, disclosed;
accountability erosion residual, mitigated, availability-only. Not used in
any proof.
\end{objx}

\begin{objx}[Panel-lapse exploitation and slot economics]\label{S-q:lapse}
\emph{Vector.} Reject a low-demand benign item; let the panel lapse; tax
the owner's rate limit on each forced re-entry.
\emph{Disposition.} The attack presupposes a deadline-based design; there
is no panel lapse and no forced re-entry (Lemma~\ref{S-lem:freeze}(iii)):
the only re-entry paths are owner-initiated, and slots free on resolution
or expiry. Roles of the clocks: \cref{S-n:roles}.
\emph{Status.} Disposed (under stated assumptions; Lemma~\ref{S-lem:freeze}(iii)). Not used in any proof.
\end{objx}

\begin{objx}[Juror exhaustion by wide pre-exposure]\label{S-q:exhaust}
\emph{Vector.} Pre-expose the poison so widely that few verdict-eligible
(fresh, seat-weighted) demanders remain; the panel never fills.
\emph{Disposition.} The item stalls \emph{frozen}: promotion is blocked at
every tool with no deadline to exploit (Rule~\ref{S-r:panel}(ii)--(iii));
the cost is availability, never safety, borne by the adversary's own
submission, and the benign-item variant is bounded at the cap
(Remark~\ref{S-rem:freezeread}, Lemma~\ref{S-lem:freeze}(ii)). Pre-exposure
also feeds the defence: each additional honest \emph{rejection} opportunity can freeze promotion (\ref{S-inv:rejasym}), and consumed
honest identities are never re-usable for later entries of the same bytes
(Corollary~\ref{S-cor:submission}).
\emph{Status.} Safety disposed $+$ availability residual (under
Theorem~\ref{S-thm:panel}'s hypotheses; deterministic failure mode of
\ref{S-a:rhoseat}, disclosed at its commentary). Not used in any proof.
\end{objx}

\subsubsection{Side channels and covert channels}
\label{S-app:an-side}

\begin{objx}[Store-exhaustion side channel]\label{S-q:quota}
\emph{Vector.} The store is idealized as unbounded; a real deployment
has quotas, and quota errors visible to $P$ leak the attacker-chosen
volume of captured blocks.
\emph{Disposition.} Correct, which is why the quota policy is a
stated convention rather than an afterthought: deployments must apply
quotas to the untrusted tier by silent eviction, never by an error
visible to $P$ (convention~(\ref{S-conv:c7})); otherwise store exhaustion reopens
exactly the channel Rule~\ref{S-r:nobit} closes (\ref{S-n:fail}). Under
the convention, no operation's defined behaviour depends on aggregate
store volume, and Lemma~\ref{S-lem:cflow}(f) stands.
\emph{Status.} Disposed (under stated assumptions), conditional on the stated deployment
policy (\ref{S-conv:c7}). Not used in any proof.
\end{objx}

\begin{objx}[Timing channels]\label{S-q:timing}
\emph{Vector.} The bisimulation matches $P$-operations; wall-clock
timing of the harness is unmodelled and could leak.
\emph{Disposition.} Conceded: timing and termination are out of scope
(\S\ref{S-app:residuals}). The interface returns no clock or step-count
value, so no timing value exists in $\sigma_P$ to branch on
(Lemma~\ref{S-lem:cflow}(e)); a clock consulted \emph{through a tool}
is already inside the model: its reading is captured to $\Mem_{\U}$
like any tool output (Rule~\ref{S-r:ingress}) and never reaches
$\sigma_P$. The exclusion concerns only implicit timing: wall-clock
latency of the harness observed by an outside party.
\emph{Status.} Residual, precisely scoped
(\S\ref{S-app:residuals}(IV)). Not used in any proof.
\end{objx}

\subsubsection{The human factor}
\label{S-app:an-human}

\begin{objx}[Preview manipulation]\label{S-q:preview}
\emph{Vector.} An adversarial reader decorates a malicious candidate
with a benign-looking preview and steers the user.
\emph{Disposition.} Within the formal model, previews are inert: they
never enter $\sigma_P$ (Lemma~\ref{S-lem:isolation}), and endorsement
binds to the true resolved bytes on the decision pane
(\ref{S-a:display}--\ref{S-a:user}), so a preview can never forge an
endorsement: the wrong value is displayed truthfully and rejected
under \ref{S-a:user}. Outside the formal model, a preview can influence
what the user chooses to examine and, through \ref{S-a:exo}'s
idealization, what the user subsequently types. That residual is
covered by design rather than assumed away: previews are part of the
display transcript and the audit log
(Definition~\ref{S-def:transcript}, Rule~\ref{S-r:display}(e)), so the
release history covers every surface through which such influence can
pass, and Corollary~\ref{S-cor:confine} remains faithful even where
\ref{S-a:exo} frays (\ref{S-n:human}, \ref{S-n:preview}).
\emph{Status.} Disposed (under stated assumptions) within the model; residual at the human
channel, disclosed. Not used in any proof.
\end{objx}

\begin{objx}[Careless endorsement and the post-endorsement channel]
\label{S-q:postendorse}
\emph{Vector.} After one careless narrative confirmation (a false acceptance in the sense of Definition~\ref{S-def:bad}; a failure of \ref{S-a:narr}), the
endorsed summary's bytes are readable and flow, literal-expressed,
to data tools ungated (handle-expressed use gates regardless,
Rule~\ref{S-r:whitelist}(c));
a poisoned summary can both mislead reasoning and direct data-tool
egress.
\emph{Disposition.} Correct, and no theorem is violated: the
influence entered through a transcript event the user accepted. The
compound shape is: one careless narrative endorsement (a failure of
\ref{S-a:narr}) simultaneously (i) admits attacker-shaped content into
$P$'s readable state and (ii) opens ungated \emph{literal-expressed} data-tool egress of
endorsed and typed bytes (Rule~\ref{S-r:whitelist}(b); handle-expressed
use gates regardless, Rule~\ref{S-r:whitelist}(c)); the misled $P$
may then recombine separately endorsed values at sensitive tools
without further gates (\ref{S-q:recombine}). Containment is threefold:
the influence never crosses a sensitive boundary with bytes neither
user-typed nor endorsed for that tool (Theorem~\ref{S-thm:action});
its triggering endorsement is a logged transcript event, and any
divergence it induces in the protected behaviours is
transcript-carried (Corollary~\ref{S-cor:confine}); and the
community mechanism reduces the frequency of the triggering
endorsement for shareable content (Appendix~\ref{S-app:fedmodel}). In
the federated tier the residual carries a further, deliberate cost:
an honest user who carelessly confirms and then \emph{publishes}
attacker-crafted content will, on that item's panel eviction, incur
the publisher penalty of Rule~\ref{S-r:panel}(vii), a corrective on
the publish gate; the penalty never touches any local store
(Lemma~\ref{S-lem:nodelete}, Proposition~\ref{S-prop:pubsafe}).
Option~\ref{S-o:two} removes the \emph{ungated} data-tool egress of
out-of-scope endorsed bytes (each such flow then gates once per pair,
Corollary~\ref{S-cor:noexfil}(b)); Option~\ref{S-o:three}, if instantiated conformingly, would limit
only scope-checked local tool authorization; data-tool enforcement
requires Option~\ref{S-o:two}, and readable reasoning state is
unaffected (\ref{S-o:three}).
\emph{Status.} Residual, disclosed, contained, and priced. Not used
in any proof.
\end{objx}

\begin{objx}[Recombination of endorsed values]\label{S-q:recombine}
\emph{Vector.} Individually authorized values can form a harmful transaction without violating Action Integrity: approval of a recipient and a body separately for an email tool does not record approval of sending that body to that recipient. Order, timing, and task context can also change the meaning of reuse.
\emph{Disposition.} Recombination is ungated by design. Theorem~\ref{S-thm:reason} excludes additional hidden influence for matched runs with equal complete display transcripts. It does not exclude influence through changed displays, incorrect endorsements, or inappropriate reuse in a later context. Option~\ref{S-o:one} narrows typed-value authority by sensitive tool, and Option~\ref{S-o:two} extends exact-tool checks to data tools; neither binds the complete transaction. Human or agent memory managers could review combined content and seek fresh endorsements before it becomes trusted, as proposed in the main paper (Appendix~\ref{app:action}). This is untested future work, not a guarantee for all later combinations. Context-bound or transaction-level authorization would require extended semantics and revised confirmation-cost accounting.
\emph{Status.} Policy boundary and endorsement residual disclosed. Not used in any proof.
\end{objx}

\begin{objx}[Batch-gate attention]\label{S-q:batch}
\emph{Vector.} The batch gate displays every element's bytes, but
displayed-in-full is not reviewed-in-full; \ref{S-a:user} is weakest
here.
\emph{Disposition.} Conceded: Definition~\ref{S-def:action} holds
literally at the batch gate (every element's bytes are displayed,
Rule~\ref{S-r:batchgate}(b)), but the human premise \ref{S-a:user} is at
its weakest, and deployments should size $\cbatch$ conservatively.
The selected batch design is all-or-nothing: a per-page decision
returned to $P$ would leak the attacker-influenceable page
count, and the aggregate decision avoids that channel at the cost that one poisoned element forces rejection of the whole batch:
an availability and attention cost, not a safety gap (no impossibility of alternative fixed-shape or padded designs is claimed); and the user
may reject at any page, aborting the remainder, so attacker-imposed
attention is bounded by the user's own patience, never compelled
(Rule~\ref{S-r:batchgate}(c),(d), \ref{S-n:batch}). Deployments may additionally cap total batch cardinality (Rule~\ref{S-r:batchgate}(b)), converting oversize collections into a fixed, decision-free reject; the cap lies outside the mechanized batch layer (\S\ref{S-app:mechanization}).
\emph{Status.} Mechanism disposed; residual at the human premise,
disclosed. Not used in any proof.
\end{objx}

\subsubsection{Adverse parameter regimes}
\label{S-app:an-param}

\begin{objx}[Cross-person correlated carelessness]\label{S-q:corr}
\emph{Vector.} A uniformly convincing poison lowers every fresh
inspector's catch rate together; the exponential bounds assume
independently varying carelessness and collapse toward $1-q_c$ under
engineered correlation.
\emph{Disposition.} The premise of the vector is corrected first: the
bounds of Theorem~\ref{S-thm:panel} assume \emph{no} independence across
inspectors; they assume a uniform \emph{conditional} lower bound $q$
on each fresh inspection given the entire past (\S\ref{S-app:prob}).
What engineered cross-person correlation attacks is that bound itself:
a uniformly convincing poison is one for which the honest conditional
catch rate, \emph{after conditioning on earlier misses}, is some
$q_c<q$. Three facts limit the damage: the theorem remains valid with
$q$ read as $q_c$, the bound degrading toward $1-q_c$ transparently,
never silently; the diversity requirement $\Dpool$ mixes
fresh-inspector pools, raising $q_c$ (\ref{S-n:diversity}); and periodic
re-verification re-presents the bytes to inspectors with no prior
decision, so the probability of remaining uncaught across $\Erv$
rounds is at most $(1-q_c)^{\Erv}$, scaling with the supply of
never-decided inspectors, not with epoch renewal (\ref{S-n:revoke}). The
worst-case fallback as $q_c\to0$ is the local layer of
Remark~\ref{S-rem:detscope}, not Theorem~\ref{S-thm:det}, whose $q=1$
hypothesis engineered correlation directly attacks. The
\emph{per-person} half of the correlation concern (re-inspection of
the same bytes is not re-randomisation)
is discharged structurally for \emph{logged} inspections: no logged
repeat carries weight or a verdict (\ref{S-inv:fresh},
\ref{S-inv:verdict}); across unlogged personal inspections
(Definition~\ref{S-def:fresh}) it survives as a calibration obligation
inside the log-fresh class (\cref{S-n:qpos}).
\emph{Status.} Residual with a three-layer cap, disclosed. Not used
in any proof.
\end{objx}

\begin{objx}[Unrecognizable poison, $q\to0$]\label{S-q:qzero}
\emph{Vector.} If no honest reviewer can recognize the poison, the
probabilistic bounds are vacuous.
\emph{Disposition.} Conceded, and the guarantee is calibrated to what
a human-verification design can coherently mean by ``safe'': content
is treated as safe when the qualifying community, inspecting the
exact bytes at the exact tool, accepts it. Content that fools
\emph{every} honest inspector is, within this paradigm,
indistinguishable from truth, and no design
that routes trust through human judgment can promise otherwise. What
the design does promise is one-sided and strong: it takes only
one honest \emph{effective qualifying rejection} --- a rejecting
decision at a protocol-bearing gate by an identity currently holding
dispute-initiation authority (Rule~\ref{S-r:panel}(i),(v)) --- to
freeze promotion immediately; detections by authority-inactive
identities are advisory flags, but the same identities' acceptances
carry no promotion weight (Definition~\ref{S-def:counting}(a),
Remark~\ref{S-rem:authority}), so silencing states never trade a lost
rejection for a harvestable acceptance; shared-presence freshness ensures the attacker can never harvest a
second \emph{weight-bearing} acceptance from any identity that has
already logged a decision on the bytes (\ref{S-inv:fresh},
Corollary~\ref{S-cor:submission}); and the moment a refuting inspection
arrives, however late, it takes protocol-level effect immediately.
What survives this regime is the local layer of
Remark~\ref{S-rem:detscope}, the one-sided freeze clause above, and
re-verification (\ref{S-n:revoke}); Theorem~\ref{S-thm:det} does
\emph{not} apply, its $q=1$ hypothesis being the exact negation of
this regime. Safety alone can additionally be forced for any $q>0$ by
raising $m$ toward $k$, at the price of a weaker wrongful-eviction
bound (\cref{S-n:qpos}).
\emph{Status.} Residual; the paradigm's boundary disclosed
(\S\ref{S-app:residuals}(VI)). Not used in any proof.
\end{objx}

\begin{objx}[Accountability mis-calibration]\label{S-q:calib}
\emph{Vector.} The rate trigger and ceiling require estimating the
honest baseline $1-p_b$; mis-calibration can suppress honest
dispute-initiation and crediting, reducing availability --- and,
absent authority-gated crediting
(Definition~\ref{S-def:counting}(a), Remark~\ref{S-rem:authority}), it
would also invalidate the uniform $\qpre$ interpretation.
\emph{Disposition.} The exposure is availability-only, never safety,
and is bounded as follows. (i) Panel verdicts are never penalised
and never charged (Rule~\ref{S-r:panel}(v)(a)), and verdict-eligible
rejections are cast regardless of accountability state
(Rule~\ref{S-r:panel}(v)(e)), so the panel arithmetic of
Theorem~\ref{S-thm:panel} and Proposition~\ref{S-prop:live} is
independent of calibration. (ii) A capped or suspended honest
identity retains advisory flags and ordinary use --- though its
acceptances are authority-inert for the duration
(Definition~\ref{S-def:counting}(a), Remark~\ref{S-rem:authority}) ---
loses nothing permanent, and is restored at the epoch boundary
(Rule~\ref{S-r:panel}(v)(d), \cref{S-n:calib}). (iii) Its ordinary, non-verdict gate inspections while
dispute-initiation and crediting authority are inactive are
\emph{excluded trials}, not silenced misses: neither branch enters
the $\qpre$ trial class --- the acceptance carries no promotion
weight and the rejection no initiation
(Definition~\ref{S-def:counting}(a), Remark~\ref{S-rem:authority},
\cref{S-n:qpos}) --- so these inspections do not enter $1-\qpre$.
Eligible panel verdicts are different: verdict capacity is never
authority-gated, and every cast eligible verdict remains a $\qpan$
trial regardless of accountability state
(Rule~\ref{S-r:panel}(v)(a),(e), Remark~\ref{S-rem:authority},
Theorem~\ref{S-thm:panel}(c)). Mis-calibration may change the
composition of the authority-active ordinary class, so the
deployment must continue to validate the weakest-target $\qpre$
bound over that class (\ref{S-n:validate}); it does not convert
excluded inspections into misses. The remaining cost is
availability: throttled honest dispute-initiation and delayed
crediting.
(iv) The calibration enters only the budget constants,
through the deterministic bound $\cacc$ (Lemma~\ref{S-lem:cacc},
Figure~\ref{S-fig:dag}); no safety theorem depends on it. (v) Under authority-gated crediting (Definition~\ref{S-def:counting}(a), Remark~\ref{S-rem:authority}), mis-calibration cannot be exploited to harvest promotion weight from silenced honest identities --- a silenced identity's acceptance is authority-inert --- so the exposure remains throttled honest dispute-initiation, with correspondingly delayed crediting, availability-only. What
mis-calibration buys an adversary is therefore throttled honest
dispute-initiation: bounded required-confirmation harm and bounded
per-epoch throughput (Lemma~\ref{S-lem:freeze}; wall-clock delay
unbounded, resolution being demand-paced,
Remark~\ref{S-rem:resolution}), no
third-party alteration of any local store (Lemma~\ref{S-lem:nodelete}),
and required cost within the single-user cap (Theorem~\ref{S-thm:floor}).
\emph{Status.} Safety disposed; availability exposure residual,
disclosed. Enters budget constants only.
\end{objx}

\subsubsection{Modelling-boundary effects}
\label{S-app:an-model}

\begin{objx}[Incentive feedback on publication]\label{S-q:incentive}
\emph{Vector.} Publisher penalties, though provably rare for benign
items, depress publication; a thinner shared tier raises the
effective $\beta$ and weakens \ref{S-a:promsuf} directly at its coverage clause (late or absent publication inflates $U_{v,t}$).
\emph{Disposition.} Conceded and unmodelled: the incentive feedback
between the accountability layer and the demand model of
Appendix~\ref{S-app:budget} is outside the formal development. Its
impact is confined to the budget \emph{asymptotics}: the cap
of Theorem~\ref{S-thm:floor} is unaffected by it (it is an identity
over required confirmations), the safety theorems are unaffected (they
do not reference $\beta$), and
Proposition~\ref{S-prop:pubsafe}(ii) bounds the penalty's probability for
benign publishers at $e^{-2\delta_b^2(k-\fs)}$ per panel, with expected counts per epoch bounded by the dispute ledger (Prop.~\ref{S-prop:pubsafe}(ii), linear tier).
\emph{Status.} Residual, confined to budget asymptotics. Not used in
any proof.
\end{objx}

\begin{objx}[Release-day demand spikes]\label{S-q:spike}
\emph{Vector.} A popular item's demand arrives in one burst;
\ref{S-a:smooth} fails, and the amortisation claims fall with it.
\emph{Disposition.} A release-day spike is precisely a violation of
\ref{S-a:smooth} \emph{for that item} ($\gamma_v\approx1$): its cost is
at most the item's own demand, once: each affected demander pays the
required confirmations (at most one per demanded pair, \ref{S-inv:floor})
that the pathwise cap (Theorem~\ref{S-thm:floor}(i)) already
bounds (whose counting identities never use
\ref{S-a:smooth}), and every other item's accounting is untouched,
since the ledger of Lemma~\ref{S-lem:budget} is per-item. Degradation
under spikes is therefore item-local, bounded by Theorem~\ref{S-thm:floor}, and
transient; the window- and dormancy-bucket rows of the ledger are exactly
where within-window demand, diversity wait, propagation delay, and
dormant intervals are collected, and \ref{S-a:smooth} caps them at the
fraction $\gbar$ of each item's own demand, so popularity never
enters the per-user bound (Lemma~\ref{S-lem:distinct}). A pre-publication demand burst is likewise an \ref{S-a:promsuf} violation for that item (its early demanders are residual), priced identically at the cap.
\emph{Status.} Disposed (under stated assumptions): the failure mode is priced by the
pathwise cap (Theorem~\ref{S-thm:floor}(i)). Not used in any proof.
\end{objx}

\begin{objx}[Handle-mediated arrival of certified content]
\label{S-q:handlecover}
\emph{Vector.} Not an attack but a representation effect: a pair
$(v,t)$ is live-$\CC$, yet a user's copy of $v$ arrives through an
untrusted channel (a message, a tool output) and is therefore held
only as an opaque handle (Lemma~\ref{S-lem:isolation}); passing the
handle to $t$ gates (Rule~\ref{S-r:whitelist}(c)) despite the
certificate, and, absent a local in-scope copy, the confirmation is
required. A demand stream arriving predominantly handle-expressed can drive
the shared \emph{required} load toward the standalone mass, community
certification notwithstanding.
\emph{Disposition.} Deliberate and structural: exempting
certificate-matching handle resolutions from the gate would make the
gate-vs-no-gate behaviour a function of hidden $\U$-resolution,
reopening the channel Rules~\ref{S-r:null}/\ref{S-r:nobit} close and
breaking Theorem~\ref{S-thm:reason} (two runs whose handles resolve to
different $\CC$-matching byte strings would flow ungated under equal
transcripts with different sink projections); the gate on
untrusted-channel arrivals is the inspection the design intends
(Remark~\ref{S-rem:satread}). The cost is bounded on both sides: each
such confirmation is once ever per (user, pair) (\ref{S-inv:floor}),
the epoch's shared required load never exceeds the standalone mass
(Theorem~\ref{S-thm:floor}(ii)), and the accepting user thereafter
holds a local in-scope copy, fast-pathing all literal reuse.
Amortisation applies to the fast-path-expressed share of certified
demand --- notably trusted retrieval, which returns live entries as
readable $\E$-bytes byte-matching the entry
(Definition~\ref{S-def:fedconfig}, Rules~\ref{S-r:hygiene}(ii),
\ref{S-r:whitelist}(a)); the handle-arrival fraction is priced inside
$\varepsilon(M)$ of \ref{S-a:promsuf} and logged directly
(\ref{S-n:validate}).
\emph{Status.} Residual, disclosed, priced in \ref{S-a:promsuf};
bounded pathwise by Theorem~\ref{S-thm:floor}(i) (persistence exceptions per \ref{S-inv:floor}). Not used in any
safety proof; enters budget assumptions only.
\end{objx}
\subsection{Scope and residuals}
\label{S-app:residuals}

The guarantees' boundaries, collected in one place.

\begin{enumerate}[label=(\Roman*),leftmargin=2.2em,itemsep=3pt,topsep=3pt]
\item \emph{Per-execution, per-(value, tool).} Action Integrity is per
execution; endorsement is per (value, tool) and cached indefinitely (a conforming instantiation of the
requirements-level Option~\ref{S-o:three} would bound the
scope-checked tool-authorization lifetime; \ref{S-o:three}); recombination of separately
endorsed values is ungated by design (\ref{S-n:unit},
\ref{S-q:recombine}).
\item \emph{The human channel.} \ref{S-a:exo} idealizes typing as exogenous;
influence through displayed content is recorded, not prevented
(\ref{S-n:human}, \ref{S-n:audit}, \ref{S-q:preview}); Reasoning Integrity is
modulo the transcript, and the compound residual is \ref{S-q:postendorse}.
\item \emph{The batch gate.} Every element's bytes are displayed, so
Definition~\ref{S-def:action} holds literally; but displayed-in-full is
not reviewed-in-full: \ref{S-a:user} is weakest here, and $\cbatch$
should be sized accordingly (\ref{S-n:batch}, \ref{S-q:batch}).
\item \emph{Timing and termination} are out of scope; the exclusion
concerns only implicit wall-clock channels observed from outside
(\ref{S-q:timing}).
\item \emph{Model intent.} Isolation restricts hidden untrusted influence; it does not certify $P$ or $\RE$ as benign or guarantee that a later combination of typed or endorsed values is appropriate. Influence through changed displays and incorrect endorsements remains possible (\ref{S-q:postendorse}, \ref{S-q:recombine}). The baseline's extensional, tool-blind trust of typed literals is Remark~\ref{S-rem:typedtrust}; Option~\ref{S-o:one} narrows that authority by sensitive tool without approving a whole transaction. One first-party residual:
$\mathsf{evict}$ is owner-scoped but $P$-invocable and ungated, so an
erring Protected LLM can remove its owner's endorsed blocks; the
no-deletion guarantee is third-party only, and gating
$\mathsf{evict}$ on $\E$-blocks is a further monotone hardening
(\ref{S-n:evict}).
\item \emph{Unrecognizable poison} ($q\to0$): no community-truthfulness
guarantee survives this regime within a human-judgment design
(Remark~\ref{S-rem:detscope}); what survives is the local layer, the
one-sided freeze on any single honest recognition, and re-verification
against never-decided inspectors (\ref{S-q:qzero}); degradation under
engineered correlation is to the rate $q_c$, transparently
(\ref{S-q:corr}).
\item \emph{Federated influence through legitimate log activity} is
held fixed by \ref{S-a:fedlog} in the noninterference statement and is
bounded instead by Theorems~\ref{S-thm:det} and~\ref{S-thm:panel}.
\item \emph{Availability} is not guaranteed; a dispute's harm \emph{in
required confirmations} is bounded per Lemma~\ref{S-lem:freeze} and
\ref{S-n:avail} (including the suspension-induced, once-per-demander
confirmation, Rule~\ref{S-r:panel}(vi)), inspection effort
disclosed separately (\ref{S-n:effort}); degradation is never destructive
(Lemma~\ref{S-lem:nodelete}).
\item \emph{Confidentiality.} Only the type-level exclusion is proved
(Lemma~\ref{S-lem:shareexcl}); semantic confidentiality of published
content rests on the publish gate (\ref{S-a:transfer}).
\item \emph{Tool choice and zero-argument actions.} Action Integrity
constrains the values crossing sensitive tool boundaries; it does not
authorize the invocation itself, and a sensitive tool with no gated
arguments satisfies the value-flow property vacuously. Deployments
requiring authorization of the act itself gate the capability (a
monotone hardening in the sense of Prop.~\ref{S-prop:harden}).
\item \emph{The reasoning pseudo-target.} Shared-tier accounting
treats $\mathsf{reasoning}$ as a canonical pseudo-target
(\S\ref{S-app:fedcore}): knowledge items amortise and are defended at
it, and $\CC$-at-$\mathsf{reasoning}$ status buys community-wide
readability only, never any tool flow.
\end{enumerate}

\subsection{Summary of composed guarantees}
\label{S-app:composed}

\paragraph{(A) One user, deployment lifetime (trust axioms
\ref{S-a:label}--\ref{S-a:exo} alone).} Mechanization boundary, wherever machine-checking is advertised:
kernel-checked are Personal Action Integrity from the empty
configuration and from arbitrary mechanized well-formed
configurations; the map-free Reasoning Integrity core from empty
stores and from the canonical shared-endorsed seed class; and the
standalone batch-confirmation layer (\S\ref{S-app:mechanization}).
Paper-only are the run-level $\mathsf{map}$ integration, the batch
cardinality cap, Option~\ref{S-o:three} (requirements-level), arbitrary matched \emph{mid-run} pairs,
the index-oracle reduction (Prop.~\ref{S-prop:diag}), the federated
product system, and every probabilistic and budget result.
Gating where influence acts:
Theorem~\ref{S-thm:action}, Corollary~\ref{S-cor:noexfil}(a). Recording where
it persuades: Theorem~\ref{S-thm:reason}, Corollary~\ref{S-cor:confine}.
(Phrasing discipline, here and in any summary elsewhere: Reasoning
Integrity is a fixed-horizon, termination-insensitive equivalence
\emph{modulo the full display transcript}, under coupled typed input
--- and, federated, a coupled log (\ref{S-a:fedlog}) --- never an
unconditional noninterference or inertness claim;
Definition~\ref{S-def:reason}, Remark~\ref{S-rem:transcript}.)
Residuals: the user's own false endorsement (\ref{S-q:postendorse}); the
post-endorsement boundary (Corollary~\ref{S-cor:noexfil}(b), \ref{S-o:two});
recombination (\ref{S-q:recombine}, \ref{S-o:one}). Deployments whose threat model emphasises data-tool
exfiltration should select Option~\ref{S-o:two} as the default profile: it closes
the baseline post-endorsement channel (Cor.~\ref{S-cor:noexfil}(b)), is
machine-checked at the gate level (\S\ref{S-app:mechanization}), and
costs $\bigO(1)$ per reusable endorsed-value pair, freshly composed arguments remaining in the per-task stream (Cor.~\ref{S-cor:flooropt}). Lifetime required cost: one
accepting confirmation per (value, tool) relied on
(Theorem~\ref{S-thm:floor}, Corollary~\ref{S-cor:floorE}); effort scope per
\ref{S-n:effort}.

\paragraph{(B) Federated, deployment lifetime.} Three layers, each
surviving the failure of the layer above. \emph{Deterministic}
($q{=}1$, $W{>}\Wadv$, $m{>}\fs$): Theorem~\ref{S-thm:det}, scope
Remark~\ref{S-rem:detscope}. \emph{Probabilistic, per poisoned attempt (Definition~\ref{S-def:attempt})}
(Definition~\ref{S-def:bad}): Theorem~\ref{S-thm:panel}, reach-free
(\ref{S-inv:reach}); per-anchor-epoch and lifetime exposure
Corollary~\ref{S-cor:submission} (attempts keyed to anchor epochs, Rem.~\ref{S-rem:anchorepoch}), Remark~\ref{S-rem:lifetime}; system-wide
per-anchor-epoch form under the union-bound condition of
Theorem~\ref{S-thm:coscale}(i). Scoping caveats: \ref{S-q:paraphrase}, \ref{S-q:tooltarget}; and one
explicitly: the numeric system-wide bound covers
\emph{adversary-published} attempts, counted by the rate limit at
their anchors --- poison published by honest-but-deceived identities
(\ref{S-q:postendorse}) is bounded per attempt (Thm.~\ref{S-thm:panel})
and disciplined by publisher penalties (Rule~\ref{S-r:panel}(vii),
Prop.~\ref{S-prop:pubsafe}), but its attempt count
$\Ex[S_{\mathrm{anc}}]$ is not bounded in terms of the adversary's
resources alone (Cor.~\ref{S-cor:submission}). Per-attempt bounds are marginal: deployment-wide exposure is
controlled by \emph{summing} over anchored attempts and
byte-distinct variants, through the union-, expectation-, and
deterministic-cap forms of Corollary~\ref{S-cor:submission} and the
lifetime ceiling of Remark~\ref{S-rem:lifetime}. \emph{Pathwise cap:}
Lemma~\ref{S-lem:nodelete} (third-party transitions; first-party residual
\ref{S-n:evict}) and Theorem~\ref{S-thm:floor}: degradation to $M$ independent
copies of (A), never below. Budget: Theorem~\ref{S-thm:floor} (pathwise once-per-pair cap and
reference-model standalone-mass ceiling, the firmest cost guarantee);
Theorem~\ref{S-thm:sat}, Corollary~\ref{S-cor:total} (conditional upper
bounds, with the $\mathsf{Bad}$ residual and the
arrival-representation caveat \ref{S-q:handlecover}) --- the vanishing clauses are conditional accounting, consuming
\ref{S-a:rhoseat}--\ref{S-a:massbound}, the rate conditions of
Thm.~\ref{S-thm:sat}(iii), the competence bound
of Lemma~\ref{S-lem:budget}, and changing time geometry
(Rem.~\ref{S-rem:gbarread}), never consequences of population growth
alone ---,
Theorem~\ref{S-thm:coscale}, Remark~\ref{S-rem:gbarread}.
\emph{Scope:} the federated layer is a protocol over the log service of
convention~(\ref{S-conv:c3}), not integrated into the single-user small-step relation
(\S\ref{S-app:fedmodel}).

\section{Probabilistic Tools and Supplementary Results}
\label{S-app:toolbox}

\subsection{Probabilistic framework, coupling, and the no-dispute
baseline}
\label{S-app:prob}

\paragraph{The adaptive-adversary framework.}
Fix an epoch. All protocol events (submissions, presentations, gate
decisions, verdicts, promotions) form a discrete sequence; let
$(\mathcal{F}_s)_{s\ge0}$ be the filtration generated by all logged
protocol events and all display events up to step $s$. The adversary's
strategy (whom to present the item to next, when to submit, re-enter,
self-dispute, or stop) is \emph{predictable} with respect to
$(\mathcal{F}_s)$: its action at step $s$ is
$\mathcal{F}_{s-1}$-measurable (convention~(\ref{S-conv:c2})). Two structural facts
constrain the adversary's adaptivity. First, an entry's bytes are
immutable once submitted (the log is append-only and the gate displays
exact bytes), so within an entry the adversary adapts
\emph{presentation and stopping} only, never the inspected content.
Second, because the adversary chooses whom to present the item to, the
inspection parameter $q$ must be read as a \emph{uniform} lower bound
over all honest identities and presentation contexts, the catch rate
of the most fallible honest user the adversary can target, with
flagged scrutiny only raising it (\cref{S-n:qpos},
\ref{S-n:diversity}).

\paragraph{Atomic gate events; log order.}
Gate events are atomic federated-wide, exactly as at the interface
(Rule~\ref{S-r:atomic}): a presentation and its disposition (an accepting
decision, a rejecting decision, or abandonment) occupy a \emph{single}
position of the linearizable log (convention~(\ref{S-conv:c3})); no dialog is ever
pending, and dispositions never resolve out of log order. The analysis
reads each such position as up to three consecutive \emph{atoms}: the
\emph{rendering atom} (item, tool, presenting identity, dispute flag, both
provenance zones --- recorded in the display transcript whether or not a
decision follows, Rule~\ref{S-r:display}(e)); if and only if the user records
a decision, the \emph{header atom} (the fact that a decision by this
identity on these bytes at this gate is being appended, together with its
verdict eligibility, a deterministic predicate of the prior log and the
identity, Rule~\ref{S-r:panel}(iii)(a)--(e)); and the \emph{content atom}
(the decision $z\in\{\mathrm{accept},\mathrm{reject}\}$). Abandonment
appends no decision record and contributes no decision trial. Cache gates (Options~\ref{S-o:one}--\ref{S-o:two}) are the one
asymmetrically recorded class: both branches occupy a decision-pane
position of the display transcript --- rendering, header, and content
atoms exactly as above, all inside the filtration, which is generated
by display events as well as logged protocol events --- while only
the rejecting branch additionally appends a federation-log candidate
record (Definition~\ref{S-def:counting}(iii)). We say a decision is
\emph{audit-recorded} if its atoms lie on the display transcript
(every recorded gate decision, cache acceptances included) and
\emph{federation-logged} if it is appended to the federation log as a
candidate decision (protocol-bearing decisions only). Trials at cache
gates are indexed by their audit-recorded atoms: their occurrence is
pre-outcome, never selected by the rejecting branch's federation-log
append. No protocol
event, and no adversary action, interleaves between the atoms of one
position; the discrete step index totally orders all atoms, and every
trial sequence in this document is indexed by these log-ordered atoms:
inspection and arrival trials by rendering atoms, decision and verdict
trials by header atoms. In particular, every earlier trial's content atom
precedes every later trial's occurrence atom. The adversary's strategy,
predictable with respect to the record-level filtration, remains
predictable with respect to this refinement: the refinement inserts atoms
inside single records and reveals nothing between records on which a
strategy could act.

\paragraph{Conditioning convention.}
The conditioning $\sigma$-algebra depends on the trial type. For an
\emph{arrival or inspection-occurrence} trial, $\mathcal G_i$ is
generated by all atoms strictly prior to the trial's rendering atom.
For a \emph{decision- or verdict-content} trial, $\mathcal G_i$ is
generated by all atoms strictly prior to the trial's header atom
\emph{together with that header atom itself} --- the fact that a
decision by this identity on these bytes at this gate is being
recorded, and its deterministic eligibility facts
(Rule~\ref{S-r:panel}(iii)) --- but not the trial's content atom. Thus
the occurrence event $O_i$ (a decision is recorded) is
$\mathcal G_i$-measurable in every application. Every behavioural
lower bound ($\qpre$, $\qpan$, $p_b$, $\rseat$) is a conditional bound
given the corresponding $\mathcal G_i$, further enlarged by the
exogenous demand factor below (uniformly over demand realisations).
Under the atomic reading above, these $\sigma$-algebras already
contain the disposition of every earlier presentation --- there are no
latent or pending dispositions --- so the occurrence, measurability,
and independence conditions of Remark~\ref{S-rem:adaptivegen} hold at
them directly. Where a behavioural parameter is stated at the
disposition level ($\qpre$, $p_b$), Lemma~\ref{S-lem:dispbridge}
transports it to the header-inclusive $\sigma$-algebra. For a per-attempt bound (Theorem~\ref{S-thm:panel}), the conditioning $\sigma$-algebra is generated by all atoms up to and including the attempt's anchor atom (Definition~\ref{S-def:attempt}), the anchor preceding every decision creditable to the attempt; the trial $\sigma$-algebras above refine it, and the behavioural bounds, asserted uniformly over histories, hold under the anchor-conditional measure.

\paragraph{Dispositions.}
An honest inspection presented at a gate has three possible dispositions, realized at that gate's log position: an accepting decision, a rejecting decision, or
abandonment. Only decisions are logged; abandonment leaves no record
and has no protocol effect (Rule~\ref{S-r:panel}(i),(iii)). The
conditional decision probabilities therefore satisfy
$p_{\mathrm{acc}}+p_{\mathrm{rej}}\le1$, abandonment carrying the
remaining mass, and every behavioural parameter ($q$, $p_b$,
$\rseat$) is stated over dispositions: abandonment counts as a miss
inside $1-q$ and as a non-verdict inside $\rseat$ (\cref{S-n:qpos}).

\begin{lemma}[Decision-conditioned content bounds]\label{S-lem:dispbridge}
Consider a gate presentation with rendering-atom $\sigma$-algebra
$\mathcal{G}$ and conditional disposition probabilities
$p_{\mathrm{acc}},p_{\mathrm{rej}}$ given $\mathcal{G}$, with
$p_{\mathrm{acc}}+p_{\mathrm{rej}}\le1$ (abandonment carrying the
remaining mass). On the event that a decision is recorded, and given
$\mathcal{G}$ together with the header atom: (i)~if
$p_{\mathrm{rej}}\ge q$, the conditional probability of
$\mathrm{accept}$ is at most $1-q$; (ii)~if
$p_{\mathrm{acc}}\ge p_b$, the conditional probability of
$\mathrm{reject}$ is at most $1-p_b$; (iii) symmetrically, let a designated sub-branch of the rejecting
disposition (e.g.\ the \emph{effective} qualifying rejection of
Theorem~\ref{S-thm:panel}(c)) have conditional probability
$p_{\mathrm{eff}}\le p_{\mathrm{rej}}$ given $\mathcal{G}$; if
$p_{\mathrm{eff}}\ge q$ for some $q\in(0,1]$, then, on the event that a
decision is recorded, the conditional probability of that sub-branch at
the header atom is at least $q$:
$p_{\mathrm{eff}}/(p_{\mathrm{acc}}+p_{\mathrm{rej}})\ge p_{\mathrm{eff}}
\ge q$, since $p_{\mathrm{acc}}+p_{\mathrm{rej}}\le1$. The header atom adds no
randomness beyond the fact that a decision is recorded: verdict
eligibility is a deterministic function of the prior log and the
identity (Rule~\ref{S-r:panel}(iii)). (Part~(i), the accept-side special case of~(iii), is retained for
reference; part~(iii) serves $\qpre$'s lower-bound direction as used in
Theorem~\ref{S-thm:panel} case~(a); part~(ii) serves $p_b$; the verdict
parameter $\qpan$ is defined at the header atom directly,
Theorem~\ref{S-thm:panel}(c), and needs no bridge.)
\end{lemma}

\begin{proof}
The decision-recorded event has conditional probability
$p_{\mathrm{acc}}+p_{\mathrm{rej}}$ given $\mathcal{G}$, so the
conditional accept probability given a recorded decision is
$p_{\mathrm{acc}}/(p_{\mathrm{acc}}+p_{\mathrm{rej}})
=1-p_{\mathrm{rej}}/(p_{\mathrm{acc}}+p_{\mathrm{rej}})
\le1-p_{\mathrm{rej}}\le1-q$, using
$p_{\mathrm{acc}}+p_{\mathrm{rej}}\le1$; (ii)~is symmetric.
Eligibility being measurable in $\mathcal{G}$ and the prior log,
conditioning on it changes nothing.
\end{proof}

\paragraph{Demand randomness, formally.}
The underlying space is a product $\Omega_{\mathrm{prot}}\times
\Omega_{\mathrm{dem}}$. The factor $\Omega_{\mathrm{dem}}$ carries
the exogenous \emph{demand indicators} of the reference model
(\S\ref{S-app:budgetmodel}) --- Bernoulli variables $D_{u,v,t}$ ($u$
demands the pair $(v,t)$ within the epoch, with per-user probability
$p_{u,v,t}$ ($p_{v,t}=M^{-1}\sum_u p_{u,v,t}$, \S\ref{S-app:budgetmodel})), independent across users and of every
$\Omega_{\mathrm{prot}}$-coordinate --- together with each owner's
publication schedule, coupled to the owner's own indicators per the
owner-coupling convention (\S\ref{S-app:budgetmodel}); the counts
$n_{v,t}=\sum_u D_{u,v,t}$ and $n_v$ are measurable in this factor
alone, and every epoch containing a submission or re-entry of $v$ has $n_v\ge1$ (owner coupling; the ledger's vanishing clause on $\{n_v=0\}$ is event-based, \S\ref{S-app:budgetmodel}). Three demand quantities must be distinguished. \emph{Latent demand}
$D_{u,v,t}$ is the exogenous indicator above. \emph{Realized shared-tier
demand} is the log-defined event of Definition~\ref{S-def:demand}: the
first interpreter event at which $u$'s use of the pair fires a gate or
dereferences a shared entry, per Definition~\ref{S-def:demand}'s exclusions (covered and cache gates are not demand events on either branch), occurring only while $v$ has a shared-tier
presence; its count is denoted $n^{\mathrm{log}}_{v,t}$. \emph{Reclassified
personal demand} is a latent demander's post-retirement encounter,
charged to the single-user stream (Definition~\ref{S-def:demand},
retirement clause). Realized shared-tier demand is dominated pathwise by
latent demand: every realized demander is a latent demander --- for third parties by
Definition~\ref{S-def:demand} (a realized demand presupposes attempted
use), for the owner by the owner-coupling convention
(\S\ref{S-app:budgetmodel}) --- so
$n^{\mathrm{log}}_{v,t}\le n_{v,t}$ pointwise; and by the
required-approval clause (\S\ref{S-app:budgetmodel}), a reference-model
coupling, any epoch event at which a user's attempted use fires a
gate realizes the user's latent indicator, so every required
confirmation in the epoch belongs to a latent demander, wherever the
event falls; hence every ledger bound
stated against $n_{v,t}$ or $\Ex[n_{v,t}]=Mp_{v,t}$ is an upper bound for
the realized counts the rows charge. Which class a latent demander lands
in (realized vs.\ reclassified) is protocol-dependent; the \emph{total}
per-pair required cost across both classes is nonetheless at most one per
(user, pair) (\ref{S-inv:floor}), which is why the floor never consults the
split. The arrival \emph{times} of realized demand are
protocol-endogenous, measurable in $(\mathcal F_s)$, revealing the
corresponding indicator at that event; no independence between arrival
times and the protocol is asserted, and \ref{S-a:smooth} --- both its
demand clause (used for bucket attribution, Lemma~\ref{S-lem:budget}) and
its first-pair-decision clause (used only for the $\bar D_v$ derivation) --- is
a hindsight cap on the realised arrival and decision schedules
respectively, per item, read pathwise on completed epochs. The
behavioural lower bounds $q$, $p_b$, $\rseat$ are asserted conditionally
per the conditioning convention of this section, further enlarged by the
entire demand factor: uniformly over demand realisations. This licenses
the ledger's two-stage argument (Lemma~\ref{S-lem:budget}): the freeze rows
bound conditional expectations given the demand coordinates, by
Lemma~\ref{S-lem:adaptive} under the demand-conditional measure, whose
hypotheses hold there by the uniform reading; the expectation over demand
is taken last, by the tower property, and the pathwise domination
$n^{\mathrm{log}}_{v,t}\le n_{v,t}$ carries every realized-count row under
the latent masses.
\begin{lemma}[Adaptive conditional coupling and domination]
\label{S-lem:adaptive}
Let $(\mathcal{F}_s)$ be a filtration and let $Z_1,Z_2,\dots\in\{0,1\}$
be decision indicators observed at strictly increasing stopping times
$\tau_1<\tau_2<\cdots$. Write $O_i$ for the event that the $i$-th
observation occurs, so that $O_1\supseteq O_2\supseteq\cdots$ and
$O_i\in\mathcal{F}_{\tau_i^-}$; unrealised observations are read as
$Z_i:=0$ off $O_i$. Fix $p\in[0,1]$ and $n$. There is an enlargement
of the probability space carrying i.i.d.\ $\mathrm{Bernoulli}(p)$
variables $Y_1,Y_2,\dots$ such that:
\begin{enumerate}[label=(\roman*),leftmargin=2em,itemsep=2pt,topsep=2pt]
\item \emph{(Majorant.)} If
$\Prob[Z_i=1\mid\mathcal{F}_{\tau_i^-}]\le p$ on $O_i$ for every
$i\le n$, then $Z_i\le Y_i$ pointwise for every $i\le n$; hence for
every $x$,
$\Prob[\sum_{i\le n}Z_i\ge x]\le\Prob[\mathrm{Bin}(n,p)\ge x]$, and
the same bound holds for any sum truncated by a predictable stopping
rule, truncation removing only nonnegative terms.
\item \emph{(Minorant, joint form.)} If
$\Prob[Z_i=1\mid\mathcal{F}_{\tau_i^-}]\ge p$ on $O_i$ for every
$i\le n$, then $Y_i\le Z_i$ pointwise on $O_i$ for every $i\le n$;
hence for every $x$,
\begin{gather*}
\Prob\Big[O_n\cap\Big\{\sum_{i\le n}Z_i\le x\Big\}\Big]
\ \le\ \Prob[\mathrm{Bin}(n,p)\le x],
\qquad\text{in particular}\\
\Prob\big[O_n\cap\{Z_1=\cdots=Z_n=0\}\big]\ \le\ (1-p)^{n}.
\end{gather*}
No claim is made about the conditional law of $\sum_{i\le n}Z_i$
given $O_n$: the occurrence events may depend on the realised $Z_i$,
so only joint-event bounds are asserted.
\item \emph{(Counting.)} If the hypothesis of (i) holds for every
$i$, then for the total count $T$ of realised observations,
$\Ex\big[\sum_{i\le T}Z_i\big]\le p\,\Ex[T]$.
\end{enumerate}
\end{lemma}

\begin{proof}
\emph{Construction.} All protocol randomness lives on a standard
Borel space: the event alphabet of \S\ref{S-app:prob} is countable and
the filtration is generated by a discrete event sequence
$(\xi_s)_{s\ge0}$, so regular conditional distributions exist, and
the process admits a \emph{sequential quantile representation}:
there are measurable maps $(\phi_s)$ and i.i.d.\
$\mathrm{Uniform}[0,1]$ variables $(V_s)$, adjoined by the
Ionescu--Tulcea extension, with
$\tilde\xi_s:=\phi_s(\tilde\xi_0,\dots,\tilde\xi_{s-1},V_s)$
reproducing the joint law of $(\xi_s)$ coordinate by coordinate
(each $\phi_s(\,\cdot\,,V_s)$ realising the regular conditional law
of $\xi_s$ given the past by quantile inversion). We choose the
representation so that, for each $i$, the coordinate deciding the
$i$-th observation's value uses a dedicated uniform $U_i$, with the
decision indicator realised as
$\tilde Z_i=\mathbf 1\{U_i\le p_i\}$ on $\tilde O_i$, where
$p_i:=\Prob[Z_i{=}1\mid\mathcal F_{\tau_i^-}]$ evaluated along the
\emph{redefined} history. Because the whole trajectory is rebuilt
recursively, the stopping times, occurrence events, and every other
coordinate are measurable functions of the redefined past, and the
redefined process has the same joint law as the original; every
conclusion below is a property of that joint law, so we argue on the
redefined process and drop the tildes. Set
$Y_i:=\mathbf 1\{U_i\le p\}$; the $Y_i$ are i.i.d.\
$\mathrm{Bernoulli}(p)$, each $U_i$ independent of the past at its
use.

\emph{(i) Majorant.} On $O_i$, $p_i\le p$, so $Z_i=1$ entails
$U_i\le p_i\le p$, hence $Y_i=1$; off $O_i$, $Z_i=0$. So
$Z_i\le Y_i$ everywhere. Pointwise domination of each term transfers
superlevel events of the sums, and dropping terms of a
$\{0,1\}$-valued sum only lowers it, giving the truncated form.

\emph{(ii) Minorant, joint form.} On $O_i$, $p_i\ge p$, so $Y_i=1$
entails $U_i\le p\le p_i$, hence $Z_i=1$; on $O_n$ every $O_i$ with
$i\le n$ holds, so
$O_n\cap\{\sum_{i\le n}Z_i\le x\}\subseteq\{\sum_{i\le n}Y_i\le x\}$,
an event of unconditional binomial law; $x=0$ gives the all-zeros
bound.

\emph{(iii) Counting.} Iterated conditioning:
$\Ex[\sum_i\mathbf{1}_{O_i}Z_i]
=\sum_i\Ex\big[\mathbf{1}_{O_i}\,p_i\big]
\le p\sum_i\Prob[O_i]=p\,\Ex[T]$.
\end{proof}

\begin{remark}[Alternatives to pathwise coupling]\label{S-rem:borel}
Readers preferring to avoid the sequential quantile construction may
replace parts (i)--(ii) by exponential supermartingale bounds, at the
cost of the pointwise dominations; the pathwise form is kept because
the truncation and joint-event arguments of
Theorem~\ref{S-thm:panel}, Proposition~\ref{S-prop:pubsafe}(ii), and
Lemma~\ref{S-lem:geo} consume domination, not merely tail bounds.
\end{remark}

\begin{remark}[$\sigma$-algebra form]\label{S-rem:adaptivegen}
Lemma~\ref{S-lem:adaptive} holds verbatim with the pre-stopping-time
$\sigma$-algebras $\mathcal{F}_{\tau_i^-}$ replaced by any
increasing $\sigma$-algebras
$\mathcal{G}_1\subseteq\mathcal{G}_2\subseteq\cdots$ such that, for
each $i$: $O_i\in\mathcal{G}_i$; $Z_j$ is
$\mathcal{G}_i$-measurable for every $j<i$; and the coupling
variable $U_i$ is independent of $\mathcal{G}_i$. The proof is
unchanged: the conditional-quantile construction and the pointwise
dominations use only these three properties. The applications take $\mathcal{G}_i$ per the conditioning convention
of \S\ref{S-app:prob}: for arrival and inspection trials, all atoms
strictly prior to the rendering atom; for decision and verdict trials,
all prior atoms \emph{plus} the trial's own header atom (content atom
excluded), so that $O_i\in\mathcal G_i$; both enlarged by the demand
factor. All trials being log-ordered with earlier contents preceding
later occurrence atoms, the three conditions hold at these
$\sigma$-algebras directly.
\end{remark}

\begin{lemma}[Delayed marks]\label{S-lem:delayed}
Let $\mathcal G_1\subseteq\mathcal G_2\subseteq\cdots$ be
$\sigma$-algebras, $O_i\in\mathcal G_i$ occurrence events, and let
$X_i$ be $[0,1]$-valued random variables with
$\Ex[X_i\mid\mathcal G_i]\le p$ on $O_i$ for every $i$. Then, with
$N:=\sum_i\mathbf 1_{O_i}$ finite almost surely,
\[
\Ex\Big[\sum_i\mathbf 1_{O_i}X_i\Big]\ \le\ p\,\Ex[N].
\]
No ordering of the times at which the $X_i$ resolve, no measurability
of any $X_i$ with respect to any $\mathcal G_j$, and no independence
is required: each $X_i$ may resolve arbitrarily later than its
occurrence event, interleaved with other trials' occurrences and
resolutions.
\end{lemma}

\begin{proof}
$\Ex[\mathbf 1_{O_i}X_i]
=\Ex\big[\mathbf 1_{O_i}\,\Ex[X_i\mid\mathcal G_i]\big]
\le p\,\Prob[O_i]$ by the tower property; sum over $i$.
\end{proof}

\begin{remark}[Realised-at-time vs.\ delayed marks]\label{S-rem:delayed}
Lemma~\ref{S-lem:adaptive}(iii) is the special case in which $X_i$ is a
decision indicator realised at its own stopping time.
Lemma~\ref{S-lem:delayed} is the form consumed by trials whose
\emph{outcome} resolves only later --- anchor-indexed promotion
attempts (Corollary~\ref{S-cor:submission}) and opening-indexed panel
instances (Proposition~\ref{S-prop:pubsafe}(ii), linear tier) --- where
later trials may open before earlier outcomes resolve, so outcome
order need not equal occurrence order. The coupling machinery of
Lemma~\ref{S-lem:adaptive} is neither needed nor invoked there.
\end{remark}

\begin{lemma}[Predictable cell allocation]\label{S-lem:alloc}
In the setting of Remark~\ref{S-rem:adaptivegen}, fix finitely many
cells $(j,\ell)$, $j\le r$, $\ell\le K$. Suppose each realised trial
$i$ is assigned to at most one cell, injectively across trials, by an
assignment that is $\mathcal G_i$-measurable (decided at the trial's
occurrence, before its content). On the enlargement of
Lemma~\ref{S-lem:adaptive} carrying the coupling variables
$Y_i=\mathbf 1\{U_i\le p\}$, define the cell-indexed array
$X_{j,\ell}:=Y_{i(j,\ell)}$ where $i(j,\ell)$ is the trial assigned
to cell $(j,\ell)$, and $X_{j,\ell}:=\mathbf 1\{U'_{j,\ell}\le p\}$
with fresh independent uniforms $U'_{j,\ell}$ for never-filled cells.
Then $(X_{j,\ell})_{j\le r,\,\ell\le K}$ is i.i.d.\
$\mathrm{Bernoulli}(p)$.
\end{lemma}

\begin{proof}
This is optional skipping in the sense of Doob, extended to a
predictable partition. Order trials by occurrence. At trial $i$'s
occurrence its cell assignment is $\mathcal G_i$-measurable, while
$U_i$ is independent of $\mathcal G_i$ (Lemma~\ref{S-lem:adaptive},
construction) --- in particular of all previously assigned cells and
all previously revealed coupling variables; the fresh $U'_{j,\ell}$
are independent of everything. Fix any
$\varepsilon:\{(j,\ell)\}\to\{0,1\}$ and reveal the array in
assignment order, completing with the fresh uniforms: at each
revelation, the conditional probability that the next revealed entry
equals its prescribed value is $p^{\varepsilon}(1-p)^{1-\varepsilon}$
for that entry, by the independence just stated. Induction along the
revelation order gives
$\Prob[X_{j,\ell}=\varepsilon(j,\ell)\ \forall j,\ell]
=\prod_{j,\ell}p^{\varepsilon(j,\ell)}(1-p)^{1-\varepsilon(j,\ell)}$,
i.e.\ the array is i.i.d.\ $\mathrm{Bernoulli}(p)$.
\end{proof}

\begin{lemma}[Hoeffding tails: independent and predictable-weight forms]
\label{S-lem:engine}
\leavevmode
\begin{enumerate}[label=(\roman*),leftmargin=2em,itemsep=2pt,topsep=2pt]
\item \emph{(Independent, deterministic weights.)} Let
$X_j\in\{0,1\}$ be independent with $\Prob[X_j{=}1]\le\varepsilon$,
and let $w_j\in[0,\wmax]$ be deterministic, $j=1,\dots,n$. Then for
$t>0$,
$\Prob[\sum_j w_jX_j\ge\varepsilon\sum_j w_j+t]\le
\exp\!\big(-2t^2/(n\wmax^2)\big)$ \cite{hoeffding1963probability}; the unweighted
case gives
$\Prob[\mathrm{Bin}(n,\varepsilon)\ge(\varepsilon+\delta)n]\le
e^{-2\delta^2 n}$.
\item \emph{(Predictable weights, conditional success bound.)} Let
$(\mathcal H_j)_{j\ge0}$ be a filtration, $X_j\in\{0,1\}$ adapted
with $\Prob[X_j{=}1\mid\mathcal H_{j-1}]\le\varepsilon$, and
$w_j\in[0,\wmax]$ predictable (each $w_j$
$\mathcal H_{j-1}$-measurable). Then the bound of (i) holds with the
identical exponent.
\end{enumerate}
\end{lemma}

\begin{proof}
(ii) Set $D_j:=w_j\big(X_j-\Ex[X_j\mid\mathcal H_{j-1}]\big)$, a
martingale-difference sequence whose $j$-th increment lies, given
$\mathcal H_{j-1}$, in an interval of predictable length
$w_j\le\wmax$. The Hoeffding--Azuma inequality in its
conditional-range form (Hoeffding's lemma applied conditionally to
each increment, then Markov on the exponential moment) gives
$\Prob[\sum_j D_j\ge t]\le
\exp\!\big(-2t^2/\sum_j w_j^2\big)\le\exp\!\big(-2t^2/(n\wmax^2)\big)$.
Since $\sum_j w_j\Ex[X_j\mid\mathcal H_{j-1}]\le\varepsilon\sum_j w_j$
pathwise, the event $\{\sum_j w_jX_j\ge\varepsilon\sum_j w_j+t\}$ is
contained in $\{\sum_j D_j\ge t\}$. (The bounded-difference form of
Azuma's inequality, with increments bounded in absolute value by
$\wmax$, would give only $\exp(-t^2/(2n\wmax^2))$; the range form is
the one used.)
\end{proof}

\begin{theorem}[No-dispute baseline: reach-dependent]\label{S-thm:prob}
Without Rule~\ref{S-r:panel}, with rejections protocol-inert and a
poisoned item shown to at most $R$ honest users each erroneously
accepting with probability $\le\varepsilon$ independently: if
$W>\Wadv+\varepsilon R\wmax$,
$$
\Prob[\text{promotion}]\ \le\
\exp\!\Big(-\frac{2(W-\Wadv-\varepsilon R\wmax)^2}{R\,\wmax^2}\Big).
$$
\end{theorem}

\begin{proof}
Accepting weight is at most $\Wadv+S$ with $S=\sum_j w_jX_j$ over
$\le R$ honest users; apply Lemma~\ref{S-lem:engine} with
$t=W-\Wadv-\varepsilon R\wmax$.
\end{proof}

\paragraph{Comparison with the no-dispute baseline.}
The comparison isolates exactly what the panel and the
gate-rejection-as-report rule buy. In Theorem~\ref{S-thm:prob} rejections
are inert, so honest false acceptances accumulate with reach: the
sample size is the \emph{attacker-chosen reach} $R$, the bound requires
$W$ to grow linearly with $\varepsilon R$ merely to stay non-trivial,
dragging the budget of Appendix~\ref{S-app:budget} with it, and spreading
the poison farther helps the attacker: the bound is not reach-free in
the sense of \ref{S-inv:reach}. In Theorem~\ref{S-thm:panel}, because
every honest inspection carries its own report
(Rule~\ref{S-r:panel}(i)), the no-dispute event shrinks with every
additional honest viewer, and the false-accept mass that matters
after a dispute is confined to a bounded jury of $k-\fs$ honest
verdicts, a protocol constant the attacker cannot multiply, and one
the demand-paced formation rule fills without any deadline the attacker
could exploit. Reach drops out of the bound entirely, and spreading the
poison cannot loosen the bound, while any single honest gate rejection
freezes it. The dispute log additionally makes wrongful rejection a
costly, attributable act, and panel eviction makes poisoned submission
a costly, attributable act (Rule~\ref{S-r:panel}(v),(vii)), neither of
which the inert-rejection baseline can express.

\subsection{The blast-schedule promotion law}
\label{S-app:blastlaw}

This subsection derives, in closed form, the promotion probability of one
\emph{simulated} presentation schedule under one \emph{simulated} model of
honest behaviour. It is an analysis tool for the empirical evaluation, not
a protocol guarantee: no rule, assumption, definition, or theorem of
Appendices~\ref{S-app:interface}--\ref{S-app:budget} cites it, and nothing in
this document depends on it. Its purpose is to turn the simulator's output
into a falsifiable prediction, so that agreement is evidence the
implementation realises the rules of Appendix~\ref{S-app:fedmodel}, and
disagreement localises a defect in one of the two.

\begin{definition}[Blast schedule; independent honest coin]\label{S-def:blast}
The \emph{blast} schedule is the non-adaptive presentation strategy that
(i)~submits a poisoned item (Definition~\ref{S-def:bad}), with all
endorsement and all demand at a single tool, (ii)~spends every adversarial
identity as a pre-dispute accepting endorser at that tool, and (iii)~then
presents the item, once each, to $R$ honest identities in a fixed order
(the \emph{reach}, as in Theorem~\ref{S-thm:prob}), continuing through every
one of them whatever any of them decides. Four stipulations fix the
simulated setting.
\begin{enumerate}[label=(\alph*),leftmargin=1.8em,itemsep=1pt,topsep=1pt]
\item \emph{One window.} All $R$ presentations, and any panel the schedule
induces, fall within a \emph{single} finalisation window of one pipeline
entry, with the freeze excluded from the window clock
(Rule~\ref{S-r:panel}(iv)) and no epoch boundary closing the window early
(convention~(\ref{S-conv:c4})). In particular the entry is open throughout, so no
decision is banked (Definition~\ref{S-def:counting}(b), which requires no
open entry) and no re-entry occurs: every weight-bearing decision is
credited to that one entry.
\item \emph{Independent honest coin.} Each of the $R$ inspections is,
independently of the others, a rejecting decision with probability $q$ and
an accepting decision with probability $1-q$, with no abstention.
\item \emph{Coupled reviewer quality.} The same $q$ governs pre-freeze
gate inspections, panel verdicts, and post-resolution confirmations
($\qpre=\qpan=q$); the decoupled reading, under which those three classes
may carry different parameters, is Remark~\ref{S-rem:blastdecoupled}.
\item \emph{Seat weight and eligibility.} Every identity carries
attestation weight at least $\wseat$, so a fresh decision on the item is
verdict-eligible whenever a dispute is open
(Rule~\ref{S-r:panel}(iii)(a)); every decision, the $f$ adversarial
pre-dispute endorsements included, is authority-active
(Definition~\ref{S-def:counting}(a)) and taken at a non-covered, non-cache
gate, so every accepting decision is creditable and every rejecting
decision takes effect; and identities are assigned to the $\Dpool$ pools
round-robin.
\end{enumerate}
Independence across inspections is a property of this simulation only: no
result of Appendices~\ref{S-app:fedmodel}--\ref{S-app:budget} assumes it, and
Theorem~\ref{S-thm:panel} explicitly does not (\S\ref{S-app:prob},
Remark~\ref{S-rem:blastnotbound}).
\end{definition}

\begin{proposition}[Blast promotion law]\label{S-prop:blastlaw}
Adopt Rules~\ref{S-r:share}--\ref{S-r:panel} with the default bars of
Rule~\ref{S-r:panel}(iv) ($m_{\mathrm{rej}}=k+1-m$, hence
$m+m_{\mathrm{rej}}-1=k$), and the blast schedule and independent honest
coin of Definition~\ref{S-def:blast}. Suppose
\[
N-f\;\le\;m,
\qquad
N-f\;\le\;R\;\le\;2m,
\]
and suppose the deployment's constants and pool assignment leave the
credit caps of Definition~\ref{S-def:counting}(f) non-binding for promotion
at every reach in that range (Step~1). Define
\[
S(J)\;:=\;\sum_{r=0}^{\min(J-m,\;m_{\mathrm{rej}}-1)}
\binom{m-1+r}{r}\Big(\frac{q}{1-q}\Big)^{\!r}
\quad (J\ge m),
\qquad S(J):=0 \quad (J<m).
\]
Then the item is promoted to $\CC$ with probability
$\pi_{\mathrm{blast}}(R)=\pi_{\mathrm{ord}}(R)+\pi_{\mathrm{pan}}(R)$,
where the two components --- promotion with no dispute ever opened, and
promotion after a dispute --- are disjoint and equal
\[
\pi_{\mathrm{ord}}(R)=(1-q)^{R},
\qquad
\pi_{\mathrm{pan}}(R)=q\,(1-q)^{R-1}\sum_{J=m}^{R-1}S(J).
\]
Neither expression contains $k$, $f$, $\fs$, or $\Dpool$.
\end{proposition}

\begin{proof}
\emph{Step 0 (no adversarial verdict).} Blast spends its $f$ identities as
pre-dispute accepting endorsers before any honest presentation. By
byte-permanence of freshness (\ref{S-inv:fresh}) each identity's first
logged decision on the bytes is its only weight-bearing inspection, and
verdict eligibility requires freshness (Rule~\ref{S-r:panel}(iii)(b)); so
when the freeze lands no adversarial identity is verdict-eligible, and
every verdict is cast by an honest identity. Hence $f$ and $\fs$ do not
appear.

\emph{Step 1 (no capacity, verdict-pool, or credit cap binds).} Under the
default bars $m+m_{\mathrm{rej}}-1=k$, so at most $k$ verdicts are cast
before a bar is reached (Lemma~\ref{S-lem:decisive}(ii)) and the capacity
cap of Rule~\ref{S-r:panel}(iii)(e) is never the binding constraint. Two
further caps are properties of the simulated population rather than
consequences of the rules, and are disclosed as such. With the
round-robin assignment of Definition~\ref{S-def:blast}(d), no pool
contributes more than $\lceil k/\Dpool\rceil$ verdicts among any prefix of
$k$ arrivals, so the per-pool verdict cap of Rule~\ref{S-r:panel}(iii)(d)
does not bind. Likewise the per-pool and total \emph{credit} caps of
Definition~\ref{S-def:counting}(f) truncate crediting at
$\lceil N/\Dpool\rceil$ per pool and $\Ncap$ in total, and those caps are
promotion-neutral in the sense stated there --- $\Ncap\ge N$, and a
credited set meeting $N$ under the per-pool cap necessarily spans at least
$\Dpool$ pools --- \emph{provided} enough accepting identities are
available in each pool at the reach in question. That availability is a
property of the constants and the round-robin assignment, verified for the
deployment studied here and assumed in the statement; where it fails, the
law does not apply and $\Dpool$ re-enters. Hence, under the stated
hypothesis, $k$ and $\Dpool$ do not appear.

\emph{Step 2 (the ordinary component).} Suppose no honest inspection
rejects. No dispute opens; the accepting decisions are all
authority-active (Definition~\ref{S-def:blast}(d)), so credited accepting
endorsements number $\min(f+R,\Ncap)\ge N$, using $R\ge N-f$ with Step~1;
and the entry promotes at its window's close
(Definition~\ref{S-def:counting}(c)). Conversely, a rejecting decision at
\emph{any} point in the window is a qualifying rejection that freezes
promotion at every tool immediately (Rule~\ref{S-r:panel}(i)--(ii)), while
finalisation occurs only at the window's close, which by
Definition~\ref{S-def:blast}(a) lies after all $R$ presentations --- so a
single rejection, however late, denies the ordinary path. The ordinary
component is therefore exactly the all-accept event, of probability
$(1-q)^{R}$.

\emph{Step 3 (the panel component; conditioning).} Condition on the first
rejection falling at honest position $j$, an event of probability
$q(1-q)^{j-1}$. The item is unpromoted and unfrozen at that moment, so
that rejection is dispute-initiating (Rule~\ref{S-r:panel}(i)--(ii)). The
remaining $J:=R-j$ inspections are, by Steps~0--1 and
Definition~\ref{S-def:blast}(d), honest, fresh and seat-weighted; each is an
eligible verdict for as long as the dispute is unresolved, and an ordinary
accepting or rejecting confirmation thereafter. Steps~4--5 treat the two
segments separately.

\emph{Step 4 (no rejection after the $m$-th acceptance).} Vindication
resumes the interrupted entry but never itself promotes: promotion occurs
only at the window's close (Definition~\ref{S-def:counting}(c),
Rule~\ref{S-r:panel}(iv)). A rejecting decision logged after vindication is
therefore a fresh qualifying rejection against an unpromoted entry, which
re-imposes the byte-level freeze (Rule~\ref{S-r:panel}(ii)). Verdict
capacity is scoped to the dispute \emph{instance} (\ref{S-inv:verdict}) and
so is available afresh in the second instance; what is not available is
freshness, which is byte-permanent (\ref{S-inv:fresh},
Rule~\ref{S-r:panel}(iii)(b)): the second instance draws eligible jurors
only from identities with no prior logged decision on the bytes, of whom
at most $m-1$ remain (Step~7). It therefore cannot vindicate; its only
other outcomes are eviction, which ends the presence outright, and an
unresolved instance, under which the window clock does not run out at all
because the freeze is excluded from it (Rule~\ref{S-r:panel}(iv)). In
either case the entry never reaches an unfrozen window end, so it cannot
finalise. Under blast every one of the $J$ post-freeze identities decides,
so the qualifying sequences are exactly those whose tail after the $m$-th
acceptance is unanimously accepting.

\emph{Step 5 (the count, and the credited threshold).} Fix $J$ and index
the post-freeze inspections $1,\dots,J$. By Step~4, promotion requires:
the $m$-th accepting verdict at position $m+r$ for some $r$, with exactly
$r$ rejecting verdicts before it and $r\le m_{\mathrm{rej}}-1$ (otherwise
the eviction bar is reached first) and $m+r\le J$; and every one of the
remaining $J-m-r$ decisions accepting. The first $m+r-1$ positions hold
$m-1$ acceptances and $r$ rejections in any order, $\binom{m-1+r}{r}$
ways, and every such sequence carries $r$ rejections and $J-r$
acceptances. \emph{The credited count is sufficient on every such
sequence}: the accepting decisions credited to the entry are the $f$
pre-dispute endorsements, the $j-1$ pre-freeze honest acceptances, the $m$
accepting verdicts (credited on vindication, Rule~\ref{S-r:panel}(iv)) and
the $J-m-r$ post-resolution acceptances, all authority-active by
Definition~\ref{S-def:blast}(d), totalling $f+R-1-r$; since
$r\le J-m=R-j-m$ and $j\ge1$, this is at least $f+m\ge N$ by the
hypothesis $N-f\le m$, and Step~1 keeps the caps from truncating below
$N$. Hence
\[
\Prob[\text{promotion}\mid J]
=\sum_{r=0}^{\min(J-m,\,m_{\mathrm{rej}}-1)}
 \binom{m-1+r}{r}q^{r}(1-q)^{J-r}
=(1-q)^{J}S(J),
\]
which is $0$ for $J<m$, no arrangement of fewer than $m$ verdicts reaching
the $m$-bar.

\emph{Step 6 (collapse).} Summing over $j$, and using
$(1-q)^{j-1}(1-q)^{R-j}=(1-q)^{R-1}$ with the substitution $J=R-j$,
\[
\pi_{\mathrm{pan}}(R)
=\sum_{j=1}^{R-m}q(1-q)^{j-1}\,\Big[(1-q)^{J}S(J)\Big]_{J=R-j}
=q(1-q)^{R-1}\sum_{J=m}^{R-1}S(J).
\]

\emph{Step 7 (validity, and exhaustiveness).} The hypothesis $R\le 2m$ is
what makes Step~4's tail requirement exactly right. At the $m$-th
accepting verdict, $j+m+r$ inspections have occurred, so at most
$R-j-m-r\le R-m-1\le m-1<m$ remain: too few for any second dispute to
reach its own $m$-bar, whatever its verdicts. The two cases --- no
rejection anywhere, and a first rejection at some position --- are
exhaustive and disjoint, so
$\pi_{\mathrm{blast}}=\pi_{\mathrm{ord}}+\pi_{\mathrm{pan}}$.
\end{proof}

\begin{remark}[Where each hypothesis is consumed]\label{S-rem:blasthyp}
The proposition's hypotheses and the stipulations of
Definition~\ref{S-def:blast} are not interchangeable, and each is used once.
$R\ge N-f$ is Step~2's threshold. $N-f\le m$ is Step~5's, and it is not
implied by the others, since $N-f\le R\le 2m$ permits $N-f=2m$ --- at
which reach a counted panel sequence can credit as few as $f+m<N$
acceptances and the formula would over-count; it is also what makes the
plateau identity of Corollary~\ref{S-cor:blastcons}(iv) meaningful, that
identity evaluating $\pi_{\mathrm{ord}}$ at $R=m$, which requires
$m\ge N-f$. $R\le 2m$ is Step~7. The credit-cap non-bindingness is
Steps~1, 2 and~5. The one-window stipulation of
Definition~\ref{S-def:blast}(a) is Step~2's converse and Step~4's re-freeze;
the seat-weight and authority stipulations of
Definition~\ref{S-def:blast}(d) are Step~3's eligibility and Steps~2 and~5's
crediting. At the geometry studied here ($N=10$, $f=2$, $m=20$,
$m_{\mathrm{rej}}=12$) the first two hold with slack,
$N-f=8\le\min(m,R)=\min(20,R)$ for every $R\ge8$, and the binding
hypothesis is $R\le40$.
\end{remark}

\begin{corollary}[Five structural consequences]\label{S-cor:blastcons}
Under the hypotheses of Proposition~\ref{S-prop:blastlaw}:
\begin{enumerate}[label=(\roman*),leftmargin=2em,itemsep=2pt,topsep=2pt]
\item \emph{(Structural zero and switch-on.)}
$\pi_{\mathrm{pan}}(R)=0$ for every $R\le m$, and its first nonzero value
is at $R=m+1$, where it equals exactly $q(1-q)^{m}$ (since $S(m)=1$).
This is a support property, invariant under any change of measure.
\item \emph{(No $k$, $f$, $\fs$, $\Dpool$.)} As stated, by Steps~0--1.
\item \emph{($m_{\mathrm{rej}}$ enters late.)} The cap in $S(J)$ first
differs from $J-m$ at $J=m+m_{\mathrm{rej}}$, and $\pi_{\mathrm{pan}}(R)$
reaches $J=R-1$; hence $\pi_{\mathrm{pan}}$ is independent of
$m_{\mathrm{rej}}$ for all $R\le m+m_{\mathrm{rej}}$ and first depends on
it at $R=m+m_{\mathrm{rej}}+1$, where the difference between the capped
and uncapped forms is exactly the single $r=m_{\mathrm{rej}}$ term at
$J=m+m_{\mathrm{rej}}$.
\item \emph{(Plateau identity.)}
$\pi_{\mathrm{blast}}(m)=\pi_{\mathrm{blast}}(m+1)=(1-q)^{m}$ exactly,
since $(1-q)^{m+1}+q(1-q)^{m}=(1-q)^{m}$. The switch-on is therefore
\emph{invisible} in the total rate and visible only in the decomposition
--- which is why the figure that exhibits it plots the two components
separately rather than their sum.
\item \emph{(Peak and fall.)} $S(J)$ is constant in $J$ for
$J\ge m+m_{\mathrm{rej}}-1$, so $\sum_{J}S(J)$ grows linearly in $R$
while $(1-q)^{R-1}$ decays geometrically: $\pi_{\mathrm{pan}}$ rises,
attains an interior maximum, and falls.
\end{enumerate}
\end{corollary}

\begin{remark}[$\pi_{\mathrm{pan}}$ is not the theorem's panel term]
\label{S-rem:blastnotbound}
$\pi_{\mathrm{pan}}(R)$ and the panel term of $B_{\mathrm{att}}$
(Remark~\ref{S-rem:batt}) are different objects and must not be conflated.
The theorem's term is a reach-free \emph{upper bound} over all adaptive
presentation, stopping and re-entry strategies, holding under a
conditional lower bound on honest behaviour alone and assuming no
independence --- the exact binomial tail arises there from the coupling of
Lemma~\ref{S-lem:adaptive}, not from an independence hypothesis;
$\pi_{\mathrm{pan}}(R)$ is the exact realised probability of one
non-adaptive schedule under the independence of
Definition~\ref{S-def:blast}(b). A large gap between them is expected and is
\emph{not} evidence that the bound is loose: tightness is a separate
empirical question, addressed by strategies that reserve seat-eligible
identities and stop at the first bar reached, which \emph{empirically
attain} the flat panel term --- precisely because they \emph{can} stop,
while blast cannot, which is why blast's realised rate peaks and falls
where the theorem's term is constant. Two schedules, one mechanism,
opposite consequences, both predicted.
\end{remark}

\begin{remark}[Validity, and the direction of failure beyond it]
\label{S-rem:blastvalid}
The law holds in the single-dispute regime $R\le 2m$, and Step~7 is where
that hypothesis is consumed. Beyond it a post-vindication rejection can
open a second dispute among still-unpresented, still-fresh jurors, whose
own $m$-bar is now reachable; the single-dispute form therefore
\emph{under-counts}, and the direction of failure is fixed in advance
rather than discovered. Consequently no reading of the law beyond $2m$ is
a prediction: past that reach the schedule's rate is a measured quantity
and the law is not a model of it. Nothing about the failure bears on any
bound: $B_{\mathrm{att}}$ is reach-free (\ref{S-inv:reach}) and continues to
hold at every reach, and \ref{S-inv:reach} disclaims monotonicity of the
\emph{realised} rate precisely so that a plateau or a rise in a measured
sweep is inside the theory rather than against it.
\end{remark}

\begin{commentary}[the validity boundary, measured]
At the deployed geometry ($q=0.6$, $N=10$, $f=2$, $k=31$, $m=20$,
$m_{\mathrm{rej}}=12$, so $2m=40$) the boundary of
Remark~\ref{S-rem:blastvalid} is measured, not merely asserted. At the last
in-validity reach the law gives
$\pi_{\mathrm{pan}}(40)=1.049\times10^{-5}$ against $9$ promotions in
$10^{6}$ trials. At $R=64$ the same schedule yields $8$ promotions in
$10^{6}$ trials ($8.0\times10^{-6}$, Clopper--Pearson $95\%$
$[3.45,\,15.76]\times10^{-6}$) against an extrapolation of
$1.057\times10^{-14}$ --- a factor $7.57\times10^{8}$, i.e.\ $8.88$
orders of magnitude above it, in the disclosed under-counting direction.
At equal trial count $R=40$ and $R=64$ are statistically
indistinguishable ($z=0.243$), so the realised rate plateaus rather than
continuing its geometric decay. The plateau does not persist: at $R=100$
the rate is bounded above by $3.69\times10^{-6}$, a bound that overlaps
$R=64$'s lower confidence limit, so the claim is \emph{no rise}, not
\emph{a fall}. Every measured cell of this sweep lies below
$B_{\mathrm{att}}=8.957\times10^{-3}$: the boundary cell $R=40$ by
$3.00$ decades and every cell beyond it by at least $3.05$. The
generating experiment gates on that conformance, at every cell of every
sweep, before writing its artifact. Inside the validity window the margin
is far smaller --- about $0.58$ decades at the panel-path peak $R=32$,
where the measured rate is $2.34\times10^{-3}$ --- which is where the
ceiling is informative rather than slack. No mechanism is proposed for the
plateau's magnitude: beyond $2m$ the sweep is \emph{measured, not
modelled}.
\end{commentary}

\begin{remark}[Testability of the $m_{\mathrm{rej}}$ entry]
\label{S-rem:mrejtest}
Corollary~\ref{S-cor:blastcons}(iii) is sharply testable only for small
$m_{\mathrm{rej}}$. The entering term is weighed against the partial sum
$\sum_{J=m}^{R-1}S(J)$ that $\pi_{\mathrm{pan}}$ carries at the entry
reach, and that sum grows with $m_{\mathrm{rej}}$ far faster than the
single entering term does, so the relative effect shrinks sharply as the
eviction bar rises. The claim is therefore made where it is resolvable ---
at small $m_{\mathrm{rej}}$, for two values of $m$ --- while for large
$m_{\mathrm{rej}}$ the testable statement is agreement with the uncapped
form below the entry reach. Two of the three confirmations available are
support properties rather than hypothesis tests: below the entry reach the
capped and uncapped forms are \emph{equal}, so two panel widths sharing
one coin stream must produce byte-identical counts, which cannot be
passed by luck.
\end{remark}

\begin{commentary}[the $m_{\mathrm{rej}}$ entry, measured]
At $m=20$ and $q=0.2$: at $m_{\mathrm{rej}}=4$, entry reach $R=25$, the
entering term is $34.59$ against a partial sum of $112.5$, a $+30.75\%$
effect; at $m_{\mathrm{rej}}=12$, entry reach $R=33$, it is $8.411$
against $2003.05$, a $+0.42\%$ effect. The corresponding rates are
$\pi_{\mathrm{pan}}(25)=0.10625$ capped against $0.13892$ uncapped, and
$\pi_{\mathrm{pan}}(33)=0.31740$ against $0.31873$. Reading a five-sigma
requirement as five standard errors of the difference between two arms of
$K$ trials each, with the variance evaluated at the capped rate ---
$K=25\cdot 2p(1-p)/\Delta^{2}$ --- those effects need $4.5\times10^{3}$
and $6.1\times10^{6}$ trials per arm respectively. The prediction is
accordingly claimed at $m_{\mathrm{rej}}=4$, where it is confirmed three
ways: byte-identical counts across two panel widths sharing one coin
stream at every reach below the entry; a $+13.6\sigma$ divergence at the
entry reach itself; and two $\pi_{\mathrm{pan}}$ profiles whose maxima
over the sampled reach grid fall where the law places them, $R=26$ with
$4{,}715$ observed against $4{,}705$ predicted and $R=30$ with $14{,}030$
against $14{,}001$. One scope note on the maxima: the
$m_{\mathrm{rej}}{=}4$ profile's exact maximum lies at $R=27$, a reach
that grid does not sample, so $R=26$ is a maximum over the sampled grid
and not the argmax of $\pi_{\mathrm{pan}}$; the two differ by $2.2\%$ in
value.
\end{commentary}

\begin{remark}[Decoupled reviewer quality]\label{S-rem:blastdecoupled}
Definition~\ref{S-def:blast}(c) couples three decision classes that the
protocol distinguishes: pre-freeze gate inspections, panel verdicts, and
post-resolution confirmations. Under a decoupled reading the first and
third are ordinary gate inspections, governed by $\qpre$, while only the
second are verdicts, governed by $\qpan$; Steps~0--5 then go through with
\[
\pi_{\mathrm{ord}}(R)=(1-\qpre)^{R},
\]
\[
\Prob[\text{promotion}\mid J]
=\sum_{r=0}^{\min(J-m,\,m_{\mathrm{rej}}-1)}
\binom{m-1+r}{r}\qpan^{\,r}(1-\qpan)^{m}(1-\qpre)^{J-m-r},
\]
the $m+r$ verdicts charged at $\qpan$ and the unanimous tail of $J-m-r$
post-resolution confirmations at $\qpre$. But Step~6's collapse fails ---
the two geometric factors no longer share a base --- and the panel
component remains the double sum
$\sum_{j=1}^{R-m}\qpre(1-\qpre)^{j-1}\Prob[\text{promotion}\mid R-j]$.
Setting $\qpre=\qpan=q$ recovers Proposition~\ref{S-prop:blastlaw} exactly,
which is why the closed form is specific to the coupled reading and why
the figures that exhibit the law run coupled reviewer quality while the
decoupled figures report ceilings only. A deployment or simulator adopting
the decoupled reading must additionally fix, and disclose, which parameter
governs the post-resolution tail: assigning it to $\qpan$ instead --- a
defensible modelling choice, since those inspections occur under the
dispute flag --- replaces $(1-\qpre)^{J-m-r}$ by $(1-\qpan)^{J-m-r}$ and
yields $\qpre(1-\qpre)^{j-1}(1-\qpan)^{J}S_{\qpan}(J)$ with $S_{\qpan}$
the sum $S$ at ratio $\qpan/(1-\qpan)$. The two forms differ except in the
coupled case, and no result of this document distinguishes them.
\end{remark}

\begin{commentary}[epistemic status of the blast law]
Three things this law is not. It is not a safety claim: the safety
statement is Theorem~\ref{S-thm:panel}, which assumes no independence and
quantifies over all adaptive schedules, and this law's exponential decay
would not survive an adversary that stops at the first rejection --- such
an adversary is confined to the ordinary path and is measured separately.
It is not a tightness claim (Remark~\ref{S-rem:blastnotbound}). And it is
not an assumption of anything: no rule reads it, no proof cites it, and
the protocol's behaviour is identical whether or not it holds; it is
accordingly absent from the assumptions ledger of
\S\ref{S-app:ledgertable}.

What it is, is the instrument that makes the simulator falsifiable. Every
one of its five consequences is a prediction the implementation could
violate, three of them at the level of \emph{support} rather than of a
tail probability --- the structural zero, the $k$-independence, and the
$m_{\mathrm{rej}}$-independence below the entry reach are all
byte-identity properties that cannot be passed by luck. Its own derivation
is verified against two independent oracles: direct enumeration of every
verdict sequence at small $(m,m_{\mathrm{rej}})$, applying the decision
rule of Step~4 without reference to the closed form; and an uncollapsed
exact-rational sum over the rejecter's position, so that an algebra error
in Step~6 cannot cancel.

These checks share an interpretive premise: Step~4 is a \emph{reading}
of Rule~\ref{S-r:panel}, and the enumeration oracle shares that reading
with the claim it checks, so the two are not independent on that point.
Unlike the single-user core (\S\ref{S-app:mechanization}), none of
this is machine-checked. An implementation that disagreed with the law
would be either a simulator defect or a misreading of
Appendix~\ref{S-app:fedmodel}, and the law's role is to make the difference
cheap to find.
\end{commentary}

\subsection{Explicit ledger constants}
\label{S-app:constants}

\begin{lemma}[Explicit forms of $A(M)$ and $\Bfl$]\label{S-lem:constants}
Summing the rows of Table~\ref{S-tab:ledger} under the charging
convention of Lemma~\ref{S-lem:budget}, with $\bar D_v$ as there and
$\kappa':=\kappa+\cpr k$:
\[
\Ex[C^{\mathrm{pipe}}_v]\ \le\ A(M)\,\Prob[n_v\ge1]\;+\;\Bfl\,\gbar\,\Ex[n_v],
\]
with
\[
\begin{aligned}
A(M)&:=2\Eslot\ctool^{2}\big(\Ncap+1+\szero\big)\\
&\quad+\ctool\big(2(1-p_b)\Eslot\ctool\szero+(1-p_b)\Emax\ctool\Ncap
+\cacc f+2\Emax+1\big)\,\kappa',
\end{aligned}
\]
\[
\Bfl\ :=\ 2\,\Eslot\,\ctool^{2}\Big[1+(1-p_b)\,\kappa'\Big],
\]
where $\kappa'$ folds at most $\cpr k$
invited re-inspections per fast-track review (Rule~\ref{S-r:panel}(vi)) into
the freeze-bucket multiplier; the suspension-prompt row is majorised by the $\cpr k$
component of the freeze-bucket term
($\Emax\cpr k\le\ctool\,\bar D_v\,\cpr k$, since $\ctool\ge1$ and
$\bar D_v\ge2\Emax+1$), a deliberate slackening: the prompt row is
charged once in the ledger and bounded once here, by a dominating
expression. Using $1\le\Dpool\le N$ (Definition~\ref{S-def:pools}),
$N\le\Ncap<N+\Dpool\le2N$ and $\Eslot\le2\Emax$, these satisfy
$A(M)=\bigO\!\big(\Emax\ctool^{2}(N+\cacc f)(1+\kappa)\big)$ and
$\Bfl=\bigO\!\big(\Emax\ctool^{2}(1+(1-p_b)\kappa)\big)$. The forms
are closed-form \emph{upper bounds}, exact in structure and
monotone-slackened in constants; they are not claimed tight.
\end{lemma}

\begin{proof}
Row sums under the charging convention of first item-arrival attribution,
splitting each $(\gbar n_v+\szero)$ term into its $\gbar n_v$ part
(collected into $\Bfl$) and its $\szero$ part (collected into $A(M)$):
\begin{multline*}
\underbrace{\Emax\ctool\Ncap}_{\text{weight}}
+\underbrace{\Eslot\ctool^{2}(\gbar\,n_v+\szero)}_{\text{window bucket}}
+\underbrace{\Eslot\ctool^{2}(\gbar\,n_v+\szero)}_{\text{dormancy bucket}}\\
+\underbrace{\ctool\,\bar D_v\,\kappa'}_{\text{freeze bucket (prompts
folded in)}}
+\underbrace{\Emax\ctool\Ncap}_{\text{repeat}}
+\underbrace{\Emax}_{\text{publish}}.
\end{multline*}
The former reversal accounting appears in no separate term: a prior
rejecter's later acceptances are charged to that demander's bucket inside
the per-demander $\times\ctool$ factors of the three bucket rows
(Lemma~\ref{S-lem:budget}, charging convention), so no $\bar D_v(1+\kappa)$
reversal term arises. The suspension-prompt row is majorised by, not identical to, the
$\cpr k$ component of the freeze-bucket term ($\ctool\ge1$,
$\bar D_v\ge2\Emax+1$), so its events are charged once (to their row)
and bounded once (by the dominating freeze expression). Collecting the
coefficient of $\gbar n_v$: $2\Eslot\ctool^{2}$ from the window and
dormancy buckets, plus $2(1-p_b)\Eslot\ctool^{2}\kappa'$ from the freeze
bucket through $\bar D_v$'s first term, giving $\Bfl$. Collecting the
demand-free remainder:
$2\Emax\ctool\Ncap+\Emax+2\Eslot\ctool^{2}\szero
+\ctool\kappa'\big(2(1-p_b)\Eslot\ctool\szero+(1-p_b)\Emax\ctool\Ncap
+\cacc f+2\Emax+1\big)\le A(M)$, using
$2\Emax\ctool\Ncap+\Emax+2\Eslot\ctool^{2}\szero
\le2\Eslot\ctool^{2}(\Ncap+1+\szero)$ ($\Emax\le\Eslot$, $\ctool\ge1$, \ref{S-a:toolspread};
$\Eslot\ge1$, convention~(\ref{S-conv:c8})). Taking expectations over the demand, with
the indicator carried per Lemma~\ref{S-lem:budget}(iii) (every row's
pathwise count vanishes on $\{n_v=0\}$), gives the claim; the
$\bigO$-forms follow since $\cpr k=\bigO(\kappa)$ for $\rseat$ bounded
below, $\kappa=(k{+}1)/\rseat\ge k$, and $\Eslot=\Emax+1\le2\Emax$ ($\Emax\ge1$, convention~(\ref{S-conv:c8})).
\end{proof}

\begin{remark}[Reading of the ledger constants]\label{S-rem:ledgerconst}
(i) The bracketed
$\big((1-p_b)\ctool\Ncap+\cacc f\big)\kappa'$ term is the \emph{dispute
cross-term}: honest reconsideration capacity and adversarial dispute
capacity multiplied by panel cost, with suspension-prompt re-inspections
folded into $\kappa'$ inside the freeze term, the prompt row majorised by the freeze term rather than independently summed; it is what forces
the cross-term condition $N(M)k(M)=o(M^{1-\beta})$ of
Theorem~\ref{S-thm:coscale}. (ii) The $\kappa'$-part of $\Bfl$ is damped by
$1-p_b$, the honest wrongful-rejection rate: the front-loaded freeze load
scales with honest error, not with demand alone. (iii) The front-loading
coefficient $\Bfl$ multiplies only the fraction $\gbar$ of the item's own
demand: item popularity enters the per-item cost solely through
$\gbar\,n_v$ and cancels against the population in
Lemma~\ref{S-lem:distinct}. (iv) The $\ctool^{2}$ in both constants is the
pair-level discipline made explicit: one factor counts the pairs over
which a bucket's demanders are enumerated (\ref{S-a:smooth}'s demand clause,
read per pair), the other the pairs each demander may confirm
(\ref{S-inv:floor}); both are $\bigO(1)$ (\ref{S-a:toolspread}), so no
asymptotic statement changes.
\end{remark}

\subsection{Knowledge-based formulation of Reasoning Integrity}
\label{S-app:knowledge}

Theorem~\ref{S-thm:reason} is trace-wise per coin sequence. It induces
the following statement in the knowledge-based style of gradual release
\cite{askarov2007gradual}, recorded to dispel the classical worry that
conditioning on release events can itself be secret-dependent.

\begin{proposition}[Gradual release]\label{S-prop:knowledge}
Fix the strategies $P$ and $\RE$, the slot skeleton, a coin sequence
$\omega$, and a typed stream (\ref{S-a:exo}; in the federated setting
also a federation log, \ref{S-a:fedlog}). By Theorem~\ref{S-thm:reason},
the entire behaviour of a run of horizon $n$ from a sorted
configuration (its Protected-LLM operations, tool invocations, reply
stream, and ungated sink events) is a function
$F(c_0^{\approx},\DT,\omega,n)$ of the initial matched-configuration
class (Definition~\ref{S-def:matched}), the run's display transcript
(the run suffix of Definition~\ref{S-def:transcript}), and the coins,
with no residual dependence on $\Mem_{\U}$. Consequently, for any
observer who knows these fixed inputs and the initial matched class,
and any prior over untrusted-store contents consistent with that
class, the posterior after observing the run to horizon $n$ equals the
prior conditioned on consistency with $\DT$ alone: knowledge about
untrusted memory increases only at display events, never through
undisplayed behaviour.
\end{proposition}

\begin{proof}
By Theorem~\ref{S-thm:reason}, any two untrusted-store contents
consistent with the same $\DT$ (within the fixed initial matched
class, matched coins, coupled typed stream and slot skeleton,
conventions~(\ref{S-conv:c1})/(\ref{S-conv:c9})) yield identical
behaviour, so the likelihood of the observed behaviour is constant
across such contents; Bayes' rule then collapses the posterior to the
prior restricted to $\DT$-consistency.
\end{proof}

\noindent The observer here is granted $\omega$ and the typed stream,
strictly more than any real attacker knows. The statement is claimed
for this strong observer only: for an observer ignorant of $\omega$,
conditioning on the transcript can correlate the hidden store with the
coins, and behaviour may then reveal information about $\omega$ that,
jointly with the transcript, reveals more about untrusted memory than
the transcript alone; marginalising over $\omega$ therefore does not
yield the weaker-observer statement in general. Deployments wanting the
weaker-observer form may include the public coin outcomes in the
release history, whereupon this proposition applies verbatim with
$\omega$ read as released.

\subsection{Zipf case study for the budget}
\label{S-app:zipf}

This subsection instantiates the saturation cost bound
(Theorem~\ref{S-thm:sat}) under a Zipf
rank-popularity law, as a worked example;
Theorem~\ref{S-thm:sat} needs only sublinear distinct-count growth (with
\ref{S-a:smooth}). We caution that empirical rank exponents for human
reference streams are frequently at or below one \cite{breslau1999web}, and
the rank/Heaps link has finite-size deviations \cite{l2010zipf}; the
clean rate below holds in the top-heavy regime $\alpha>1$. Throughout,
the demand-independent per-item cap is $\hat N:=A(M)$ with $A(M)$ as
in Lemma~\ref{S-lem:constants}; the front-loaded part
$\Bfl\gbar\,n_v$ of the ledger is carried separately (after summation
it contributes at most $\Bfl\gbar\,d$ per user, uniformly over the
popularity law), so the head/tail analysis concerns the $\hat N$ part,
with the $\Bfl\gbar\,d$ term added at the end.
The demand-independent invitation cap $K=\Emax(\ctool\Ncap+\cpr k+1)$
of Lemma~\ref{S-lem:budget}(iv) is likewise carried separately, as
$K\,V(M)/M$ (Lemma~\ref{S-lem:decomp}); it plays no role in the
head/tail split. When $\hat N$ grows
with $M$ (through $N(M)$, $\kappa(M)$, and $f(M)$ under co-scaling), the head
term below becomes
$\hat{N}^{\,1-1/\alpha}Z^{-1/\alpha}M^{-(1-1/\alpha)}$, which still
vanishes whenever $A(M)=o(M)$, amply covered by
$A(M)=o(M^{1-\beta})$.

\begin{lemma}[Head/tail decomposition]\label{S-lem:decomp}
With head $H_M=\{v:Mp_v\ge\hat N\}$, tail $L_M=\Vals\setminus H_M$,
and $K=\Emax(\ctool\Ncap+\cpr k+1)$ as in
Lemma~\ref{S-lem:budget}(iv), and with $\Ex[\bar C_M]$ read over the
pipeline component (Lemma~\ref{S-lem:budget}; the ledger residual
adds the separately bounded term of Theorem~\ref{S-thm:sat}(iii) on
top, uniformly over the popularity law, vanishing under that
theorem's additional rate conditions),
$$
\Ex[\bar C_M]\ \le\ \frac{\hat N\,|H_M|}{M}
+\ctool\,\mu_{\mathrm{tail}}(M)
+K\,\frac{V(M)}{M}
+\Bfl\,\gbar\,d .
$$
\end{lemma}

\begin{proof}
Head items are bounded by clause (iii) alone
($\Ex[C_v]\le\hat N+\Bfl\gbar\,\Ex[n_v]$); tail items by clause (iv)
in expectation ($\Ex[C_v]\le\ctool\,Mp_v+K\,\Prob[n_v\ge1]$, since
$\Ex[n_v]=Mp_v$); summing the head term and
the front-loaded parts over all items gives the first and last terms
after division by $M$; the tail items' $K$-terms are further enlarged
to the all-items sum $K\,V(M)$, giving the third term. The quantity
$\mu_{\mathrm{tail}}(M)$ is the expected number of distinct tail items
a single user demands: sub-threshold demand not yet amortised.
\end{proof}

\begin{theorem}[Zipf decay, head and tail]\label{S-thm:zipf}
Under $p_r=r^{-\alpha}/Z$ (read as the homogeneous special case $p_{u,v,t}\equiv p_{v,t}$ of the reference model), $Z=\sum_{j=1}^{|\Vals|}j^{-\alpha}$. Clauses~(i)--(iii) concern
$\alpha>1$, where $Z\le\zeta(\alpha)$:
\begin{enumerate}[label=(\roman*),leftmargin=2em,itemsep=2pt,topsep=2pt]
\item rank $r$ is in the head iff $r\le R_M:=(M/(\hat N Z))^{1/\alpha}$,
so $|H_M|=\min\{\lfloor R_M\rfloor,U\}\le R_M$ and
$\hat N|H_M|/M\le Z^{-1/\alpha}\,(\hat N/M)^{1-1/\alpha}$;
\item let $U:=|\Vals|$ and $m_\star:=\lfloor R_M\rfloor+1$; for the
finite universe,
$\mu_{\mathrm{tail}}(M)=\tfrac1Z\sum_{m_\star\le r\le U}r^{-\alpha}$,
empty (and zero) if $m_\star>U$. For $m_\star\le U$, the floor-aware
two-sided comparison
$\int_{m_\star}^{U+1}x^{-\alpha}\,dx\ \le\
\sum_{r=m_\star}^{U}r^{-\alpha}\ \le\
m_\star^{-\alpha}+\int_{m_\star}^{U}x^{-\alpha}\,dx$
gives
\[
\frac{m_\star^{1-\alpha}-(U{+}1)^{1-\alpha}}{Z(\alpha-1)}
\ \le\ \mu_{\mathrm{tail}}(M)\ \le\
\frac1Z\Big(m_\star^{-\alpha}
+\frac{m_\star^{1-\alpha}-U^{1-\alpha}}{\alpha-1}\Big)
\ \le\ \frac1Z\Big(R_M^{-\alpha}
+\frac{R_M^{\,1-\alpha}}{\alpha-1}\Big),
\]
using $m_\star>R_M$ and $\alpha>1$; since
$R_M^{-\alpha}\le R_M^{\,1-\alpha}$ for $R_M\ge1$, and
$\mu_{\mathrm{tail}}\le d=1$ always (absorbing $R_M<1$),
$\mu_{\mathrm{tail}}(M)=\bigO\big((\hat N Z/M)^{1-1/\alpha}\big)$
uniformly; the matching lower order
$\mu_{\mathrm{tail}}(M)=\Theta\big((\hat N Z/M)^{1-1/\alpha}\big)$
holds under the two additional (sufficient) non-saturation
conditions $R_M\to\infty$ (equivalently $\hat N Z/M\to0$) and
$U/R_M\to\infty$: then $m_\star/R_M\to1$, so
$m_\star^{1-\alpha}=(1+o(1))\,R_M^{\,1-\alpha}$, and the
$m_\star^{-\alpha}$ term and the $U$-corrections are
$o(R_M^{\,1-\alpha})$. Neither condition may be omitted from this
sufficient criterion in
general: along a sequence with $R_M\to\infty$, taking $U=m_\star$
leaves a single tail term of order
$R_M^{-\alpha}=o(R_M^{\,1-\alpha})$, so the matching order fails
though $R_M\to\infty$ holds; while $R_M\to0$ (a per-item cap
outgrowing the population, e.g.\ $U=Z=1$, $\hat N=M^{2}$) leaves
tail mass $\Theta(1)$ against a diverging comparison scale
$(\hat NZ/M)^{1-1/\alpha}$, though $U/R_M\to\infty$ holds. For an infinite universe,
$U/R_M\to\infty$ is automatic while $R_M\to\infty$ remains an
additional requirement; $R_M\to\infty$ holds in the
fixed-adversary, fixed-constants regime of
Theorem~\ref{S-thm:sat}(ii) and in the corridor of
Theorem~\ref{S-thm:coscale} (where $\hat N=A(M)=o(M^{1-\beta})$),
while $U/R_M\to\infty$ remains a separate universe non-saturation
condition in either regime. Part~(iii) consumes only the uniform
upper bound and requires neither condition;
\item hence
$\Ex[\bar C_M]=\bigO\big((A(M)/M)^{1-1/\alpha}\big)
+\bigO\big(K(M)\,V(M)/M\big)+\Bfl\,\gbar\,d$,
plus the residual of Theorem~\ref{S-thm:sat}(iii), the head and tail
terms decaying at the common rate $\bigO((A(M)/M)^{1-1/\alpha})$ (as
upper bounds) and the invitation term $K\,V(M)/M$ vanishing whenever
$K(M)\,V(M)=o(M)$ --- under fixed protocol constants \emph{with} $f(M)=\bigO(1)$ and $\rseat$ bounded below
($A(M),K(M)=\Theta(1)$; the hypotheses of Theorem~\ref{S-thm:sat}(ii)) automatic, giving the clean rate
$\bigO(M^{-(1-1/\alpha)})+\bigO(V(M)/M)$; under co-scaling,
$K(M)=\bigO(\Emax(\ctool N+\cpr k))=o(M^{1-\beta})$ by the corridor
schedules, so $K\,V(M)/M=o(1)$ exactly as for $A(M)$. The
front-loading term survives exactly as in Theorem~\ref{S-thm:sat},
vanishing under $\gbar(M)(1+\kappa(M))\to0$ (\ref{S-a:massbound} is automatic here: $d=\sum_r p_r=1$).
\end{enumerate}
For the boundary case $\alpha=1$, read along finite-universe
sequences $U:=|\Vals|\to\infty$, with
$Z:=H_U=\sum_{r=1}^{U}r^{-1}=\ln U+\bigO(1)$ and
$R_M:=M/(\hat N Z)$: in the nonempty-head regime $1\le R_M<U$
(below universe saturation), $R_M\sim M/(\hat N\ln U)$ and the head
term $\hat N|H_M|/M=\Theta(1/\ln U)$ --- the $\hat N=A(M)$
dependence cancels; for $R_M<1$ the head is empty and the head term
zero --- while after universe saturation ($R_M\ge U$) the head term
equals $\hat N\,U/M$. For $1\ll R_M<U$, the tail sum
$\sum_{R_M<r\le U}r^{-1}/Z=\big(\ln(U/R_M)+\bigO(1)\big)/\big(\ln U+\bigO(1)\big)$,
governed by how far the head boundary sits below the universe; (floor corrections at the boundary rank are absorbed into the $\bigO(1)$; the tail is empty for $m_\star>U$); no
power rate is asserted without an explicit relation between $U$ and
$R_M$, decay being controlled by saturation rather
than the exponent, which is why the main development is stated for
sublinear distinct-count growth (Remark~\ref{S-rem:zipfscope}).
\end{theorem}

\begin{proof}
(i) $Mp_r\ge\hat N\iff r\le R_M$, and
$|H_M|=\min\{\lfloor R_M\rfloor,U\}\le R_M=\bigO((M/\hat N)^{1/\alpha})$; $Z$ is bounded for
$\alpha>1$. (ii) with $m_\star=\lfloor R_M\rfloor+1$: if $m_\star>U$ the tail is
empty and every bound holds trivially; otherwise apply the
floor-aware comparison as displayed (the lower bound from
$r^{-\alpha}\ge\int_r^{r+1}x^{-\alpha}\,dx$ termwise, the upper from
splitting off the first term and bounding the rest by
$\int_{m_\star}^{U}x^{-\alpha}\,dx$); dropping the nonnegative
$U$-terms and using $m_\star>R_M$ (so $m_\star^{-\alpha}<R_M^{-\alpha}$
and $m_\star^{1-\alpha}<R_M^{1-\alpha}$) yields the uniform upper
bound, with the trivial guard $\mu_{\mathrm{tail}}\le d=1$ absorbing
$R_M<1$; under $R_M\to\infty$ and $U/R_M\to\infty$, $m_\star/R_M\to1$, so the
floor correction is negligible, the $U$-corrections and the
$m_\star^{-\alpha}$ term are $o(R_M^{\,1-\alpha})$, and both ends
are $\Theta(R_M^{\,1-\alpha})$, giving the matching order; substitute
$R_M=(M/(\hat NZ))^{1/\alpha}$. (iii) sum in
Lemma~\ref{S-lem:decomp}; the tail term
$\ctool\,\mu_{\mathrm{tail}}$ absorbs the old constant into
$\bigO(\cdot)$, and the invitation term is
$K\,V(M)/M$ verbatim. The $\alpha=1$ clauses follow, along $U\to\infty$, from
$R_M=M/(\hat N\ln U)(1+o(1))$ read in the regime $R_M\ge1$ (where
$\lfloor R_M\rfloor\ge R_M/2$; the head is empty for $R_M<1$),
cancellation of $\hat N$ in $\hat N R_M/M$, the exact identity
$|H_M|=U$ when $R_M\ge U$ (where the tail is empty), and,
separately in the nonsaturated regime $1\ll R_M<U$, the
harmonic-tail estimate
$\sum_{R_M<r\le U}r^{-1}=\ln(U/R_M)+\bigO(1)$ \cite{breslau1999web}.
\end{proof}

\begin{remark}\label{S-rem:zipfscope}
The $\alpha=1$ boundary case is why the main development is stated in
terms of sublinear distinct-count growth
(Theorem~\ref{S-thm:sat}), which holds across $\alpha$: the Zipf form
sharpens the rate in the top-heavy regime $\alpha>1$ but is used
nowhere in any proof.
\end{remark}

\subsection{Assumptions ledger}
\label{S-app:ledgertable}

\begingroup\small
\renewcommand{\arraystretch}{1.25}
\begin{longtable}{@{}p{0.10\textwidth} p{0.38\textwidth} p{0.44\textwidth}@{}}
\toprule
\textbf{Assum.} & \textbf{What it buys} & \textbf{Conditions and
caveats}\\
\midrule
\ref{S-a:label} & Labels attacker-immutable; injected content stays
$\U$. & Authenticated command channel (TCB,
\S\ref{S-app:attacker}).\\
\ref{S-a:persist} & Trust changes only by deterministic rules and
endorsement. & Storage and log integrity of $\Mem_{\E}$ (TCB).\\
\ref{S-a:display} & Faithful decision pane; TCB-authenticated attested
zone; marked untrusted strings. & Zoned rendering path (TCB,
\ref{S-r:display}). The \emph{payload} of a narrative confirmation is
attacker-influenced text; that is \ref{S-a:narr}'s subject.\\
\ref{S-a:user} & A confirmed atomic value is the intended value for that
tool. & User-as-oracle axiom; defensible because the set is small,
faithfully displayed, and shrinking (Appendix~\ref{S-app:budget});
weakest at the batch gate (\ref{S-q:batch}).\\
\ref{S-a:narr} & A confirmed summary was checked against citations. &
Stronger than \ref{S-a:user}; mitigated by community verification;
residual \ref{S-q:postendorse}, incl.\ publisher penalty on
honest-but-deceived publication.\\
\ref{S-a:exo} & Licenses the two-run comparison and the informal reading
of Cor.~\ref{S-cor:confine}. & Human-channel idealization; practical
content discharged by logging both surfaces (\ref{S-n:human},
\ref{S-n:audit}, \ref{S-n:preview}).\\
\ref{S-a:transfer} & Community endorsement at a tool speaks for any user
at that exact tool. & Only the semantic half (context-independence)
assumed; type-level half proved (Lem.~\ref{S-lem:shareexcl}); canonical
identifiers TCB.\\
\ref{S-a:fedlog} & Licenses federated Reasoning Integrity
(Prop.~\ref{S-prop:fed}). & Parallel to \ref{S-a:exo}; influence through
legitimate log activity bounded by
Thms.~\ref{S-thm:det},~\ref{S-thm:panel} instead (exogenous prefix coupled; local suffixes derived equal, Def.~\ref{S-def:fedstep}).\\
\ref{S-a:rhoseat} & Bounded-harm DoS (Lem.~\ref{S-lem:freeze}(ii)) and the
freeze-bucket ledger row, via $\kappa=(k{+}1)/\rseat$
(Lem.~\ref{S-lem:geo}). & Behavioural component as $\qpre,\qpan,p_b$ ($\qpre$,$p_b$ disposition-level, $\qpan$ verdict-conditional at the header atom, Thm.~\ref{S-thm:panel}(c); the symmetric bridge direction case~(a) uses is Lem.~\ref{S-lem:dispbridge}(iii)); structural component can
fail \emph{deterministically} (freshness/pool-cap depletion under
wide pre-exposure or pool-skewed demand,
commentary at \ref{S-a:rhoseat}); failure is availability-only, cap
unaffected (Lem.~\ref{S-lem:freeze}(ii)); exponential tail when it
holds; validation stratified by pool and prior exposure
(\ref{S-n:validate}).\\
\ref{S-a:smooth} & Caps within-window and within-candidate-dormancy \emph{realized}
demand (demand clause, used for ledger bucket attribution) and audit-recorded first
pair-decisions (first-pair-decision clause, used only for $\bar D_v$)
at fraction $\gbar$ of the item's latent demand (\S\ref{S-app:prob});
stable promoted intervals exempt (their required gates are ledger
residual, \ref{S-a:promsuf}); feeds the ledger split
(Lem.~\ref{S-lem:budget}) and Thms.~\ref{S-thm:sat},~\ref{S-thm:coscale}. &
Hindsight timing property, not volume; steady demand gives
$\gbar\approx\Twin/L$; violation item-local, bounded by
Thm.~\ref{S-thm:floor} (\ref{S-q:spike}); validated on most-demanded
items (\ref{S-n:validate}); jointly satisfiable with \ref{S-a:promsuf}
via $D_0\asymp N/\gbar$; pair-level form (per demanded tool);
Thm.~\ref{S-thm:floor} never uses it.\\
\ref{S-a:promsuf} & Bounds the ledger residual --- required confirmations charged to no
ledger row: pre-publication, post-retirement, post-demotion, and
stable-promoted-interval demand, the last including handle-expressed
use of live-$\CC$ pairs (\ref{S-q:handlecover}) ---
(Thm.~\ref{S-thm:sat}(iii)); completes the total-cost limit
(Cor.~\ref{S-cor:total}). & Hindsight, \emph{operational} (no-row)
coverage property of completed epochs, read against latent demand
($U_{v,t}\le n_{v,t}$, \S\ref{S-app:prob}); jointly requires
publication, promotion, \emph{and arrival-representation} coverage
($\varepsilon$ absorbs the handle-arrival fraction); violation
graceful and item-local, charged at the cap, entering
$r^{(e)}_{\mathrm{bad}}$; scoped to benign items (poison charged at
the cap by design); the $D_0$ term of the aggregate residual requires
$D_0(M)V(M)=o(M)$ and the $\varepsilon$ term
$\varepsilon(M)\,d_{\mathrm{good}}(M)\to0$ (under \ref{S-a:massbound},
from $\varepsilon(M)\to0$), with bounded tool spread supplied by
\ref{S-a:toolspread} (Thm.~\ref{S-thm:sat}(iii)); enabled by the vindication exemption
(\ref{S-n:vindexempt}) and banking; validation via realized
$U_{v,t}/n_{v,t}$ and the fast-path-vs-gate split
(\ref{S-n:validate}); Thm.~\ref{S-thm:floor} never uses it.\\
\ref{S-a:toolspread} & Licenses per-tool accounting: the uncovered
residual and $\bigO(1)$ per-tool constants. & Demand-shape assumption matching
the per-(value, tool) unit; byte-first density is a heuristic diagnostic for \ref{S-a:promsuf}
(stated at its sufficiency paragraph), not part of this assumption;
validation \ref{S-n:validate}; supplies the mass comparison
$d_{\mathrm{tool}}\le\ctool d$, quarantined in
Cor.~\ref{S-cor:floorspread} (lifetime form under the lifetime
reading, stated there); no clause of Thm.~\ref{S-thm:floor} uses it.\\
\ref{S-a:massbound} & Uniform bound on the per-user epoch demand mass
$d(M)$ along the population scaling; licenses every vanishing
($\to0$) budget conclusion (Thms.~\ref{S-thm:sat},
\ref{S-thm:coscale}(iii); Cor.~\ref{S-cor:total}). & Individual-usage
property, not a population property; automatic under any fixed
per-user demand law (Zipf: $d=1$, \S\ref{S-app:zipf}); owner-coupling
contribution bounded by $\rsub$ (Rule~\ref{S-r:share}); disclosed
failure mode: shared-tier-induced demand growth (cf.\
\ref{S-q:incentive}); violation graceful --- the relative bounds and
the cap are unaffected, and the general-form rate conditions of
\ref{S-a:massbound} carry $d$ explicitly; validation: per-user
distinct-item demand logged across growth (\ref{S-n:validate});
Thm.~\ref{S-thm:floor} never uses it.\\
\midrule
\multicolumn{3}{@{}p{0.92\textwidth}@{}}{\textbf{Calibration of
Rule~\ref{S-r:panel}(v)} ($\thacc,\cceil$): separates honest from
adversarial rejecters; grounds the deterministic constant
$\cacc=\cceil+\rmax$ (Lem.~\ref{S-lem:cacc}) in the ledger and
throughput bounds. Requires estimating $1-p_b$; mis-calibration is an
availability-only exposure (\ref{S-q:calib}); enters only
budget constants, never any safety theorem
(Fig.~\ref{S-fig:dag}).}\\
\midrule
\multicolumn{3}{@{}p{0.92\textwidth}@{}}{\textbf{Rules and TCB.}
Rules \ref{S-r:names}--\ref{S-r:retrieve} with conventions (i)--(vii${}'$) and
\ref{S-r:share}--\ref{S-r:hygiene} are harness properties; validation
(\ref{S-n:validate}) provides falsification and calibration evidence for
the uniform behavioural bounds, not certification, since sparse or
unobserved strata are not covered by any finite sample; not
assumptions; Figures~\ref{S-fig:sos}--\ref{S-fig:sos2} with the totality clause of
Rule~\ref{S-r:iface} constitute the complete interface against which
Lemmas~\ref{S-lem:iface}--\ref{S-lem:cflow} are verified. TCB inventory:
\S\ref{S-app:attacker} (single-user), \S\ref{S-app:fedassum} (federated).
External modelling parameters with validation paths: saturation
$V(M)=o(M)$; inspector competence $q>0$ (safety) and $q+p_b>1$ (joint
safety and resolution correctness), $q$ read per \cref{S-n:qpos}. The co-scaling
corridor additionally requires the cross-term condition
$N(M)k(M)=o(M^{1-\beta})$ and the shrinking front-loading condition
$\gbar(M)k(M)\to0$ (\S\ref{S-app:corridor}).}\\
\bottomrule
\end{longtable}
\endgroup

\clearpage

\section{Machine-Checked Verification in Lean~4}
\label{S-app:mechanization}\label{S-app:mechanization-full}

We mechanize the single-user core of PerMIT in Lean~4 (toolchain \texttt{leanprover/lean4:v4.31.0}; Mathlib \texttt{v4.31.0}, revision \texttt{fabf563a}). The development defines an executable interpreter for the memory interface and proves Theorems~\ref{S-thm:action} and~\ref{S-thm:reason} over it for arbitrary Protected-LLM strategies, retrievers, attackers, and horizons, within the deployment, input, and starting-state conditions stated below. Building the development is the verification: Lean's kernel checks every theorem, and every executable scenario is a build-time assertion.

\paragraph{Scope.}
The mechanized surface is the single-user interface without run-level batched iteration: tool invocation, $\mathsf{write}_P$, $\mathsf{recall}$, $\mathsf{fill}$, $\mathsf{crystallize}$, $\mathsf{confirm}$ (tool and reasoning targets), $\mathsf{retrieve}$, $\mathsf{display}$, $\mathsf{evict}$, and the typed and paste channels. The batch-confirmation surface of Rule~\ref{S-r:batchgate} is mechanized as a standalone layer. Paper arguments alone cover the integration of $\mathsf{map}$ into the run-level interpreter, durable endorsement on aborted invocations (delta~\ref{S-delta:6}), arbitrary mid-run matched pairs, the index-oracle reduction (delta~\ref{S-delta:2}), Options~\ref{S-o:one}--\ref{S-o:two} beyond the gate level and Option~\ref{S-o:three}, and the federated tier with every probabilistic and budget result.

\paragraph{Model.}
The interpreter \texttt{execOp} is a total function with one arm per operation, so every invocation has defined behavior (Rule~\ref{S-r:iface}(b)). Its state \texttt{IState} represents the extended configuration of Definition~\ref{S-def:extconfig}, subject to the modeling deltas below; the display transcript, sensitive-sink log, and operation log are ghost state. Four modeling choices carry the proofs.
\begin{itemize}[leftmargin=1.4em,itemsep=1pt,topsep=2pt]
\item \emph{Replies.} The inductive type \texttt{Reply} (acknowledgment, handle, failure, decision, bytes) is the reply grammar of Rule~\ref{S-r:iface}; no other reply can be constructed. Bytes are returned only by $\mathsf{recall}$ of a non-$\U$ block and by trusted retrieval.
\item \emph{Two-sorted identifiers.} Endorsed-sort identifiers (\texttt{BlockId.e}) are allocated only by coupled events: typed input, the agent's writes, and accepted confirmations. Untrusted-sort identifiers (\texttt{BlockId.u}) are allocated by attacker-dependent events: the initial untrusted store, paste, tool-output capture, and crystallization. Every endorsement writes a fresh copy at the next endorsed-sort identifier.
\item \emph{Views.} The Protected-LLM view \texttt{Obs} is computed from the endorsed store alone (\texttt{memEOf}), with the decision history, typed input, and handle count; the retriever likewise receives only the endorsed store.
\item \emph{Actors.} The Protected LLM is an arbitrary function of its emitted operations, reply stream, view, and coins (\texttt{PStrat}). The attacker supplies the initial untrusted store, every tool output (count and content), every paste event, and the Quarantined Reader's candidates and summaries. The user supplies the typed stream and the decision and selection at each confirmation.
\end{itemize}

\paragraph{Checked statements.}
Table~\ref{S-tab:leancorr} gives the correspondence. Action Integrity is a projection of an inductive invariant, \texttt{Inv}, established initially and preserved by every transition. It is proved from the empty endorsed store and from any state satisfying \texttt{Inv}. A precedence corollary places each endorsing confirmation in the transcript that exists when the sink event is recorded.

Reasoning Integrity is a lockstep simulation over \texttt{MatchW}, which equates everything the Protected LLM can observe and everything the transcript records, while leaving untrusted stores and hidden handle resolutions free. It is proved from empty endorsed stores and from any shared endorsed seed with independently chosen untrusted stores. Its conclusions exceed what Theorem~\ref{S-thm:reason} requires: equal transcripts force equal raw sink logs, operation sequences, reply streams, and final views.

Proposition~\ref{S-prop:baseline} is witnessed by computation. A taint-gating baseline, with data taint carried inside the model, yields equal (empty) transcripts but different ungated sink projections. PerMIT's interpreter under the analogous strategy is attack-invariant.

Eight scenarios check, by computation:
\begin{itemize}[leftmargin=1.4em,itemsep=0pt,topsep=2pt]
\item the typed fast path;
\item a blocked injection;
\item endorsement followed by fast-path reuse;
\item occupancy hiding;
\item a finite noninterference instance;
\item attack-surface confinement;
\item ungated data-tool flow of agent-composed bytes;
\item the post-endorsement boundary, whose closure under Option~\ref{S-o:two} is witnessed in \texttt{Hardening.lean}.
\end{itemize}

\begin{table}[h]
\centering\footnotesize
\setlength{\tabcolsep}{4pt}
\renewcommand{\arraystretch}{1.12}
\caption{Correspondence between paper results and the Lean development (\texttt{PermitFormal/}). Theorems are kernel-checked; \texttt{\#guard} rows are evaluator-checked.}
\label{S-tab:leancorr}
\begin{tabular}{@{}p{0.28\textwidth}p{0.46\textwidth}p{0.20\textwidth}@{}}
\toprule
\textbf{Paper} & \textbf{Lean} & \textbf{File}\\
\midrule
Thm.~\ref{S-thm:action} & \texttt{action\_integrity}, \texttt{action\_integrity\_from}, \texttt{action\_integrity\_precedence} & ActionIntegrity\\
Def.~\ref{S-def:wf}(i)--(iii),(v); Lems.~\ref{S-lem:endorse}(b), \ref{S-lem:nocc} & \texttt{Inv}; \texttt{inv\_init}, \texttt{inv\_execOp}, \texttt{inv\_ingestTyped}, \texttt{inv\_ingestPaste}, \texttt{inv\_runFrom} & Invariants\\
Def.~\ref{S-def:sinksafe} & \texttt{SinkSafe} & Invariants\\
Lem.~\ref{S-lem:isolation} (replies) & \texttt{bytes\_reply\_isolation} & Invariants\\
Lem.~\ref{S-lem:mediation}; Cor.~\ref{S-cor:noexfil}(a) & \texttt{processArgs\_spec}, \texttt{handle\_always\_gates}, \texttt{ungated\_is\_literal}; \texttt{noexfil\_semantics} & Invariants; ActionIntegrity\\
Lem.~\ref{S-lem:dtmono} & \texttt{dt\_runFrom\_prefix}, \texttt{prefix\_or\_prefix} & Invariants; ReasoningIntegrity\\
Lem.~\ref{S-lem:dtsegment} (direct calls) & \texttt{processArgs\_det}, \texttt{match\_execOp} & ReasoningIntegrity\\
Defs.~\ref{S-def:swf}, \ref{S-def:matched} & \texttt{WFs}, \texttt{EStore}, \texttt{MatchW} & ReasoningIntegrity; Semantics\\
Thm.~\ref{S-thm:reason} & \texttt{reasoning\_integrity}, \texttt{reasoning\_integrity\_from} (via \texttt{match\_runFrom}) & ReasoningIntegrity\\
Cor.~\ref{S-cor:confine} & \texttt{confine}, \texttt{confine\_from} & ReasoningIntegrity\\
Rules~\ref{S-r:batchgate}, \ref{S-r:atomic} (batches); Lem.~\ref{S-lem:dtsegment} (batch) & \texttt{batchGate\_accept\_flow}, \texttt{batchGate\_elem\_displayed}, \texttt{batchGate\_reject\_no\_flow}, \texttt{batchGate\_null}, \texttt{batchReply\_decision\_only}, \texttt{batchGate\_det} & BatchGate\\
Prop.~\ref{S-prop:baseline} & \texttt{\#guard} witness & Baseline\\
Prop.~\ref{S-prop:harden}(i); Cor.~\ref{S-cor:noexfil}(b) under \ref{S-o:two} & \texttt{gateCheckO1\_mono}, \texttt{gateCheckO2\_mono}; \texttt{\#guard} witnesses & Hardening\\
Scenarios 1--8 & \texttt{\#guard} & Sanity\\
\bottomrule
\end{tabular}
\end{table}

\paragraph{Trusted base.}
Lean's kernel checks every theorem, and the build fails on any failed proof or false \texttt{\#guard}. An omitted proof (\texttt{sorry}) produces only a warning, so completeness rests on the \texttt{\#print axioms} report, reproduced verbatim in the artifact's README from a clean build. It shows that the top-level theorems, and hence every lemma they use, depend only on \texttt{propext} and \texttt{Quot.sound}, and none on \texttt{sorryAx}. 
The \texttt{\#guard} scenarios additionally trust Lean's evaluator. The correspondence between the mechanized model and the paper's semantics is not machine-checked; Table~\ref{S-tab:leancorr} and the deltas below argue it.

\paragraph{Modeling deltas.}
None is silent. Deltas \ref{S-delta:1} and \ref{S-delta:3} strengthen the checked statements.
\begin{enumerate}[label=(\arabic*),ref=\arabic*,leftmargin=2em,itemsep=1pt,topsep=2pt]
\item\label{S-delta:1} \emph{Reader.} The Quarantined Reader is folded into the attacker, which chooses the reader's candidates and summaries directly.
\item\label{S-delta:2} \emph{Index oracles.} User decisions and selections, and attacker outputs, are functions of event ordinals. Any run of a reactive user or attacker is the run of the oracle equal to its realized sequence. In Theorem~\ref{S-thm:reason}, both runs share the user oracle, and transcript equality pins every decision a display-reactive user makes (Supplement, Proposition~\ref{S-prop:diag}; a paper argument).
\item\label{S-delta:3} \emph{Channels.} Compared runs share only the slot skeleton and the typed stream; each attacker chooses paste presence and content independently.
\item\label{S-delta:4} \emph{Starting states.} Theorem~\ref{S-thm:action} is proved from the empty endorsed store and from any state satisfying \texttt{Inv}. Theorem~\ref{S-thm:reason} is proved from empty or shared endorsed seeds, the canonical matched class of Definition~\ref{S-def:swf}. Arbitrary mid-run matched pairs are covered by the paper proof.
\item\label{S-delta:5} \emph{Collection handles.} Without run-level $\mathsf{map}$, a handle to a multi-element capture is confirmed at a direct call by selecting one displayed element, so at most one displayed value flows per accepting decision.
\item\label{S-delta:6} \emph{Aborted invocations.} 
The mechanized interpreter writes endorsed copies only when an invocation executes. Rule~\ref{S-r:atomic} also writes durable copies for the accepted positions of an aborted invocation; that case is covered by the paper proof only.
\item\label{S-delta:7} \emph{Decision history.} 
The mechanized history also records the selection index and marks null-resolution aborts, so the mechanized view extends the paper's $h$. The operation semantics do not inspect these extra fields; both are transcript-recorded and therefore agree in the matched executions used by the proof.
\item\label{S-delta:8} \emph{Selections.} An out-of-range selection clamps to the first candidate; the flowed bytes remain the displayed bytes.
\item\label{S-delta:9} \emph{Representation.} One typed event is one $\T$-block; bytes are lists of naturals; provenance marks are subsumed by labels and ghost logs. \texttt{Inv} checks endorsement against the transcript, without the sanitized-history record of Definition~\ref{S-def:wf}(ii).
\item\label{S-delta:10} \emph{Outside the interpreter.} Options~\ref{S-o:one}--\ref{S-o:two} are checked at the gate level only. The covered-gate and reasoning-coverage clauses of Rule~\ref{S-r:panel}(i) are federated accounting conventions, not transitions.
\end{enumerate}

\paragraph{Specification corrections.}
Mechanization corrected the specification in three places, each adopted in the paper.
\begin{itemize}[leftmargin=1.4em,itemsep=1pt,topsep=2pt]
\item Endorsement must write a fresh copy at an operation-determined identifier. Relabeling the untrusted source in place would expose an attacker-dependent identifier in the endorsed store and break the matched-pair invariant; this is why identifiers are two-sorted.
\item Channel events must be acknowledged to the channel, not to the Protected LLM. Otherwise an uncoupled paste would emit a visible acknowledgment, and Theorem~\ref{S-thm:reason} would fail.
\item The fast-path case analysis of Lemma~\ref{S-lem:mediation} is disjunctive: typed and endorsed matches may co-occur, and \texttt{processArgs\_spec} proves exhaustiveness without exclusivity.
\end{itemize}

\paragraph{Artifact.}
The Lean artifact is available upon request and will be released with the code when the paper is published. The development pins the toolchain (\texttt{lean-toolchain}) and all dependencies (\texttt{lake-manifest.json}). Running \texttt{lake exe cache get \&\& lake build} reproduces the verification. The artifact includes \texttt{CORRESPONDENCE.md} (Table~\ref{S-tab:leancorr}), a README with the verbatim axiom report, the log of a clean build, and a SHA-256 manifest of the sources.

\end{document}